\documentclass[journal]{IEEEtran}
\usepackage{amsmath} 
\usepackage{amssymb}
\usepackage{graphicx} 
\usepackage{float} 
\usepackage{subcaption} 
\usepackage{indentfirst}
\usepackage{algorithmic}
\usepackage{algorithm}
\usepackage{booktabs} 
\usepackage{multirow}
\usepackage{array} 
\usepackage{caption}
\usepackage[utf8]{inputenc}
\usepackage{makecell} 
\usepackage{bm}
\usepackage{tikz} 
\usepackage{inputenc}\inputencoding{utf8}\DeclareUnicodeCharacter{200B}{}
\usepackage{url}
\usepackage{cite}
\title{Learning High-Quality Initial Noise for Single-View Synthesis with Diffusion Models} 
\author{Zhihao~Zhang,~Xuejun~Yang,~Weihua~Liu,~Mouquan~Shen
	\thanks{(\emph{Corresponding author: Zhihao Zhang.})}
	\thanks{Z. Zhang, X. Yang and M. Shen are with the College of Electrical Engineering and Control Science, Nanjing Tech University, Nanjing 211800, China (e-mail: zhihaozhang94@njtech.edu.cn; 202461206072@njtech.edu.cn;
	shenmouquan@njtech.edu.cn).}
	\thanks{W. Liu is with the Yongjiang Laboratory, Ningbo 315201, China (e-mail: weihua-liu@ylab.ac.cn).}
}
\date{8/25/2020} 
\begin{document} 
    \maketitle 
    \begin{figure*}[htbp]
    	\centering
    	\fontsize{9pt}{12pt}\selectfont  
    	\begin{tabular}{c c c c c c}
    		&
    		\begin{subfigure}[t]{.133\linewidth}
    			\centering
    			\includegraphics[width=\linewidth]{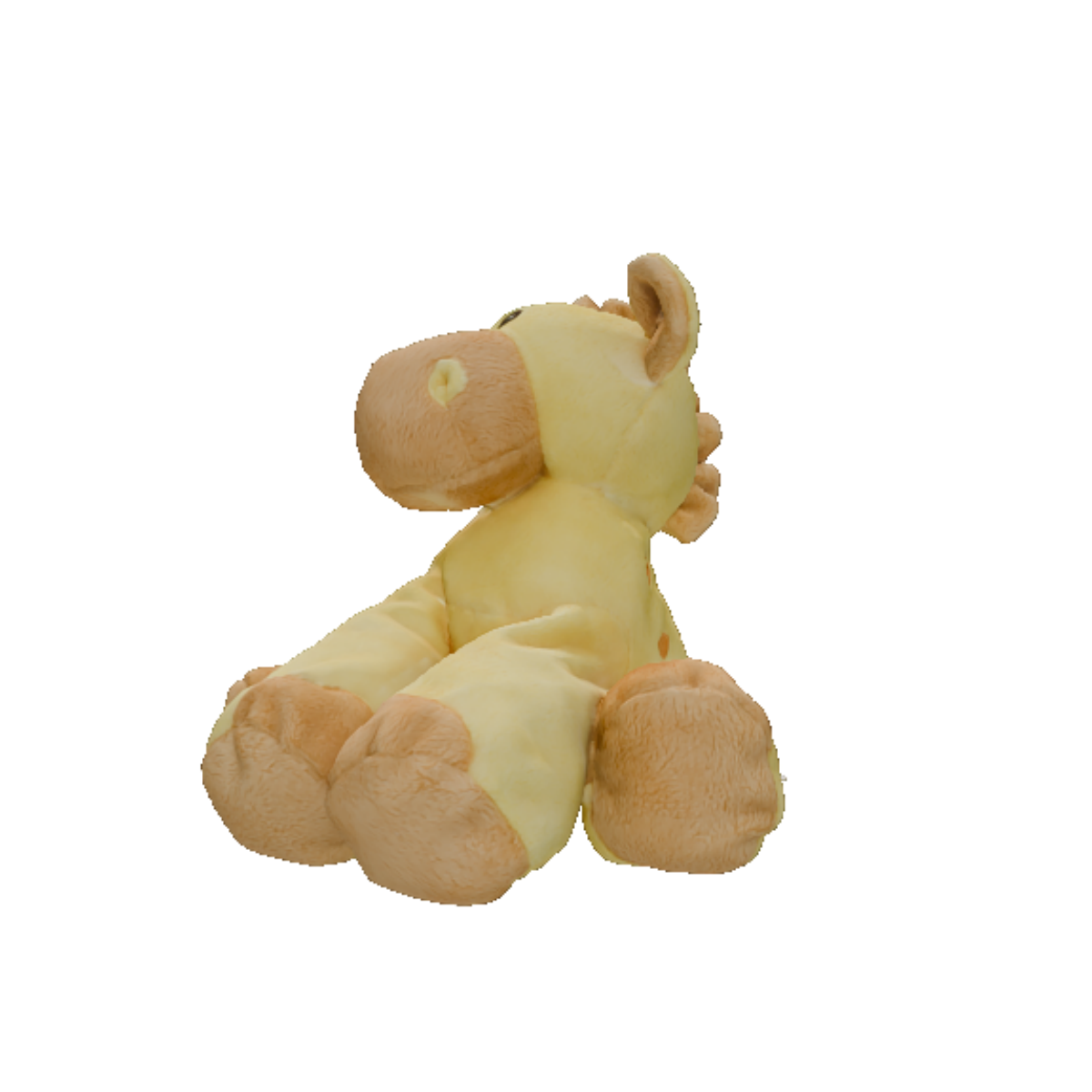}
    		\end{subfigure}&
    		\begin{subfigure}[t]{.133\linewidth}
    			\centering
    			\includegraphics[width=\linewidth]{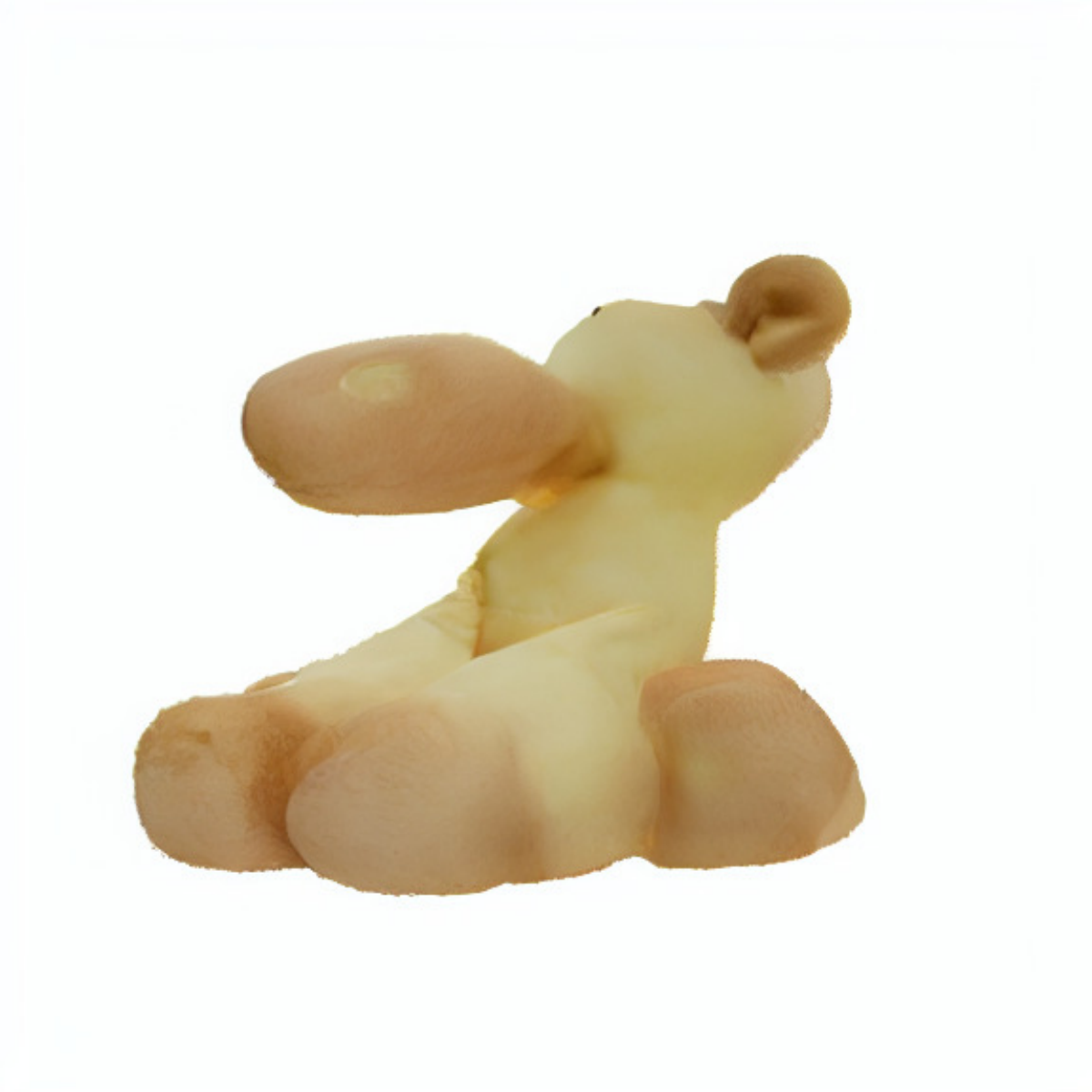}
    		\end{subfigure}&
    		\begin{subfigure}[t]{.133\linewidth}
    			\centering
    			\includegraphics[width=\linewidth]{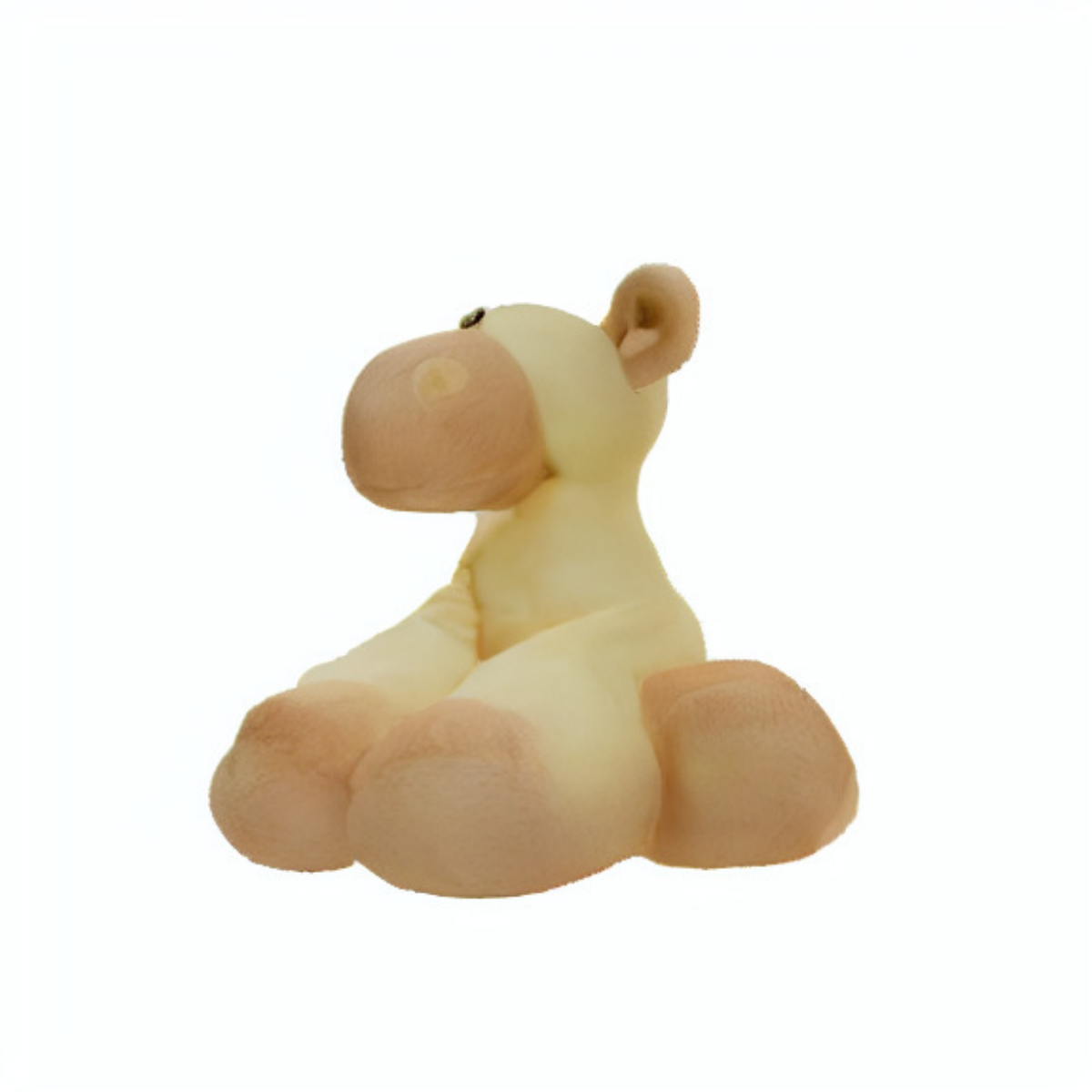}
    		\end{subfigure}&
    		\begin{subfigure}[t]{.133\linewidth}
    			\centering
    			\includegraphics[width=\linewidth]{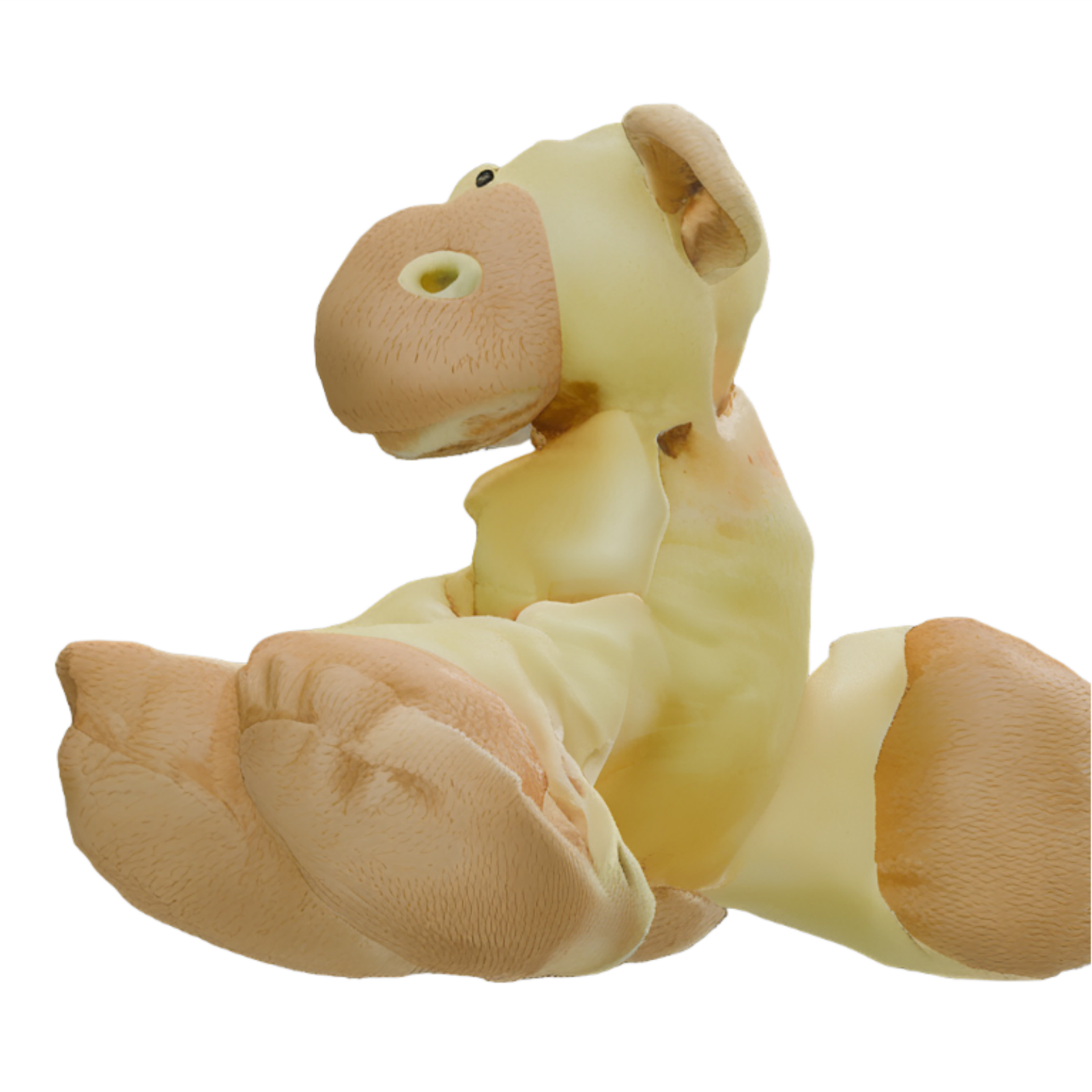}
    		\end{subfigure}&
    		\begin{subfigure}[t]{.133\linewidth}
    			\centering
    			\includegraphics[width=\linewidth]{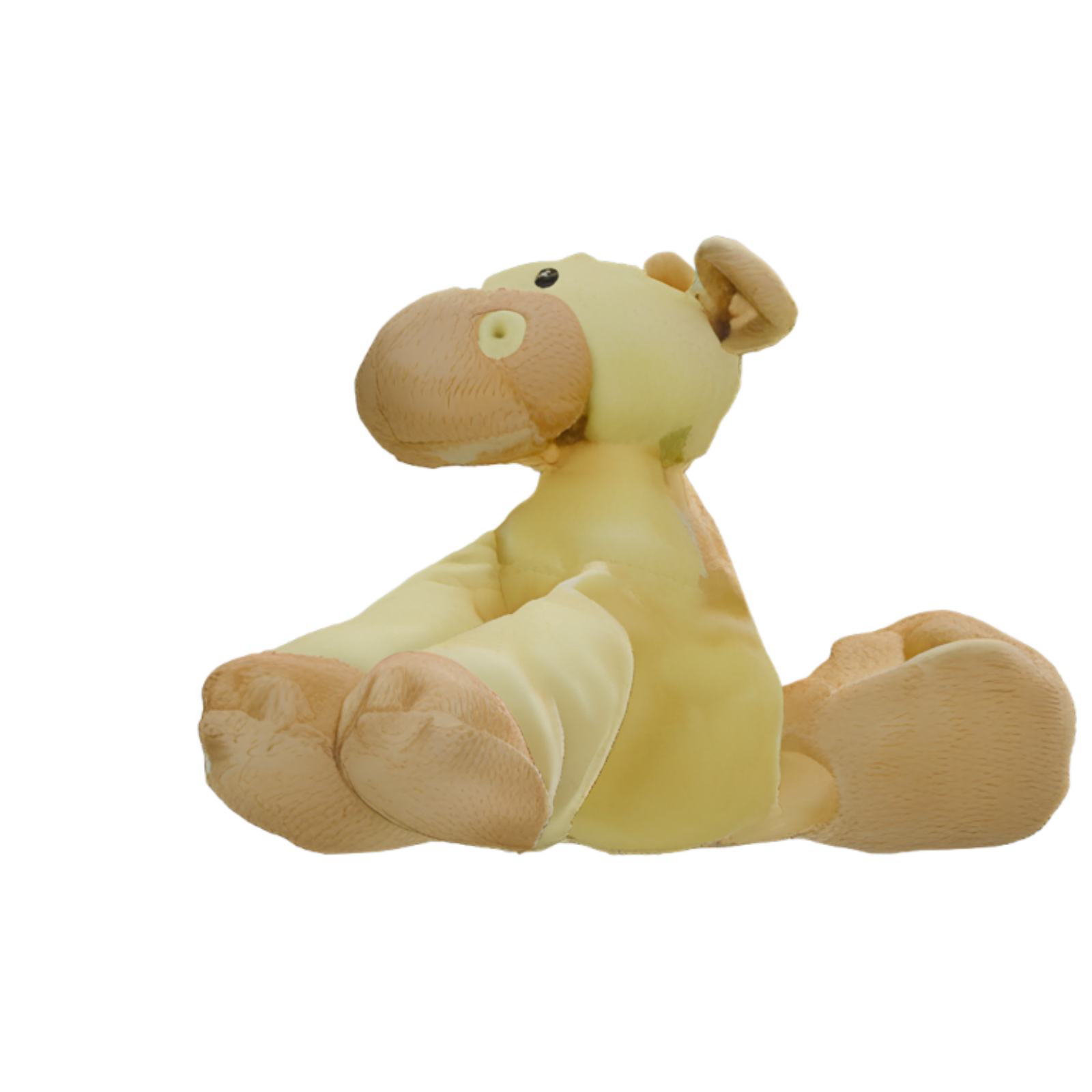}
    		\end{subfigure}\\
    		\begin{subfigure}[t]{.133\linewidth}
    			\centering
    			\includegraphics[width=\linewidth]{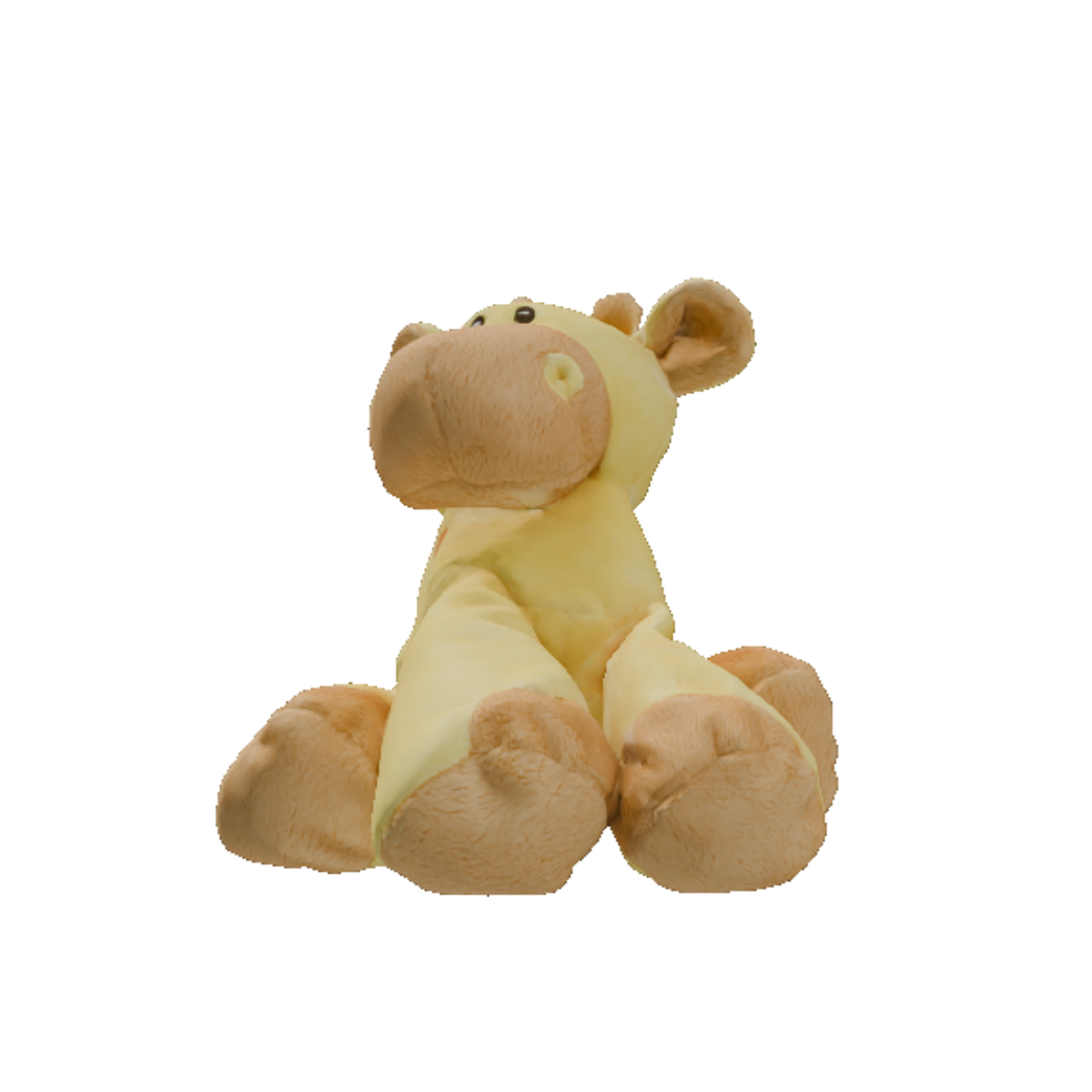}
    		\end{subfigure}&
    		\begin{subfigure}[t]{.133\linewidth}
    			\centering
    			\includegraphics[width=\linewidth]{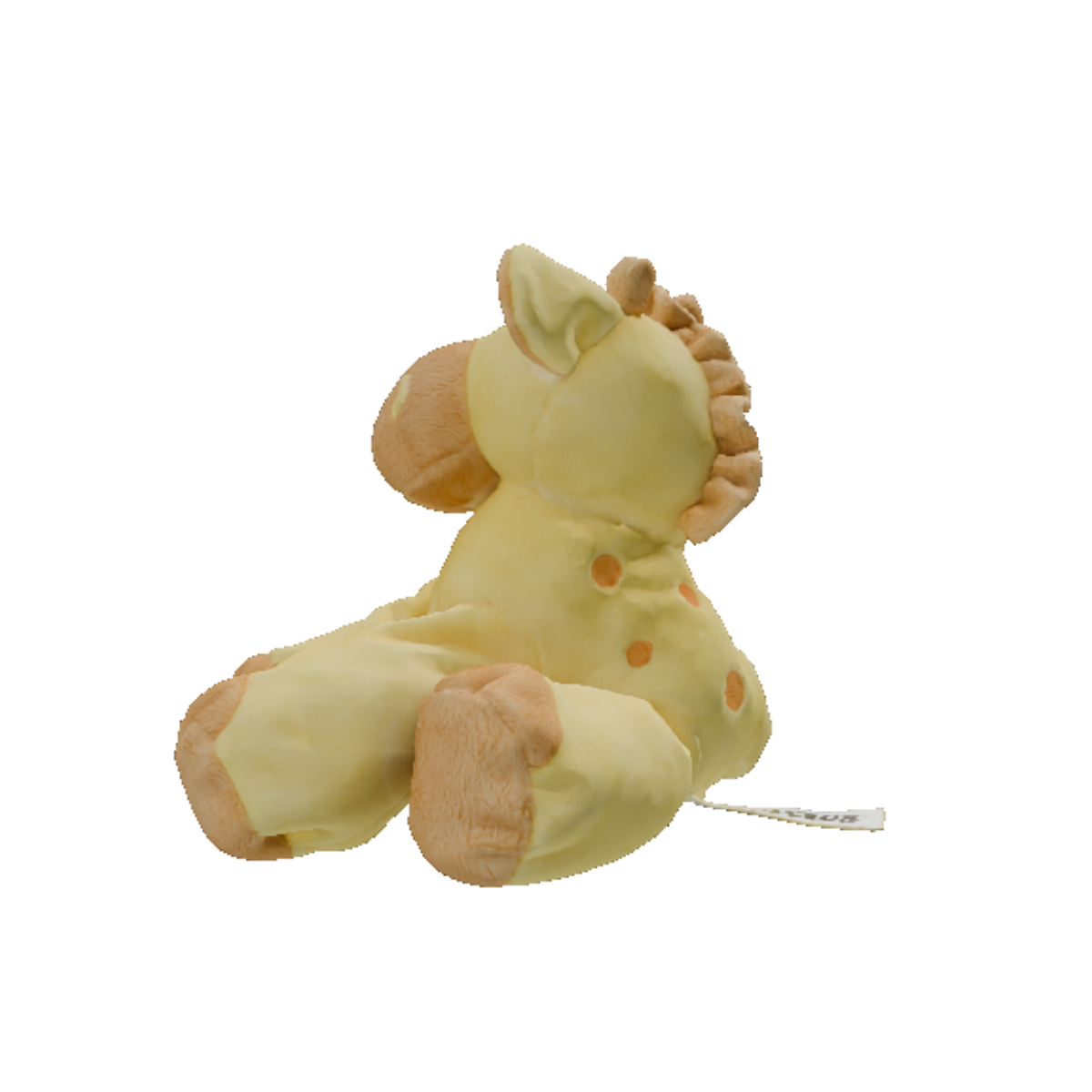}
    		\end{subfigure}&
    		\begin{subfigure}[t]{.133\linewidth}
    			\centering
    			\includegraphics[width=\linewidth]{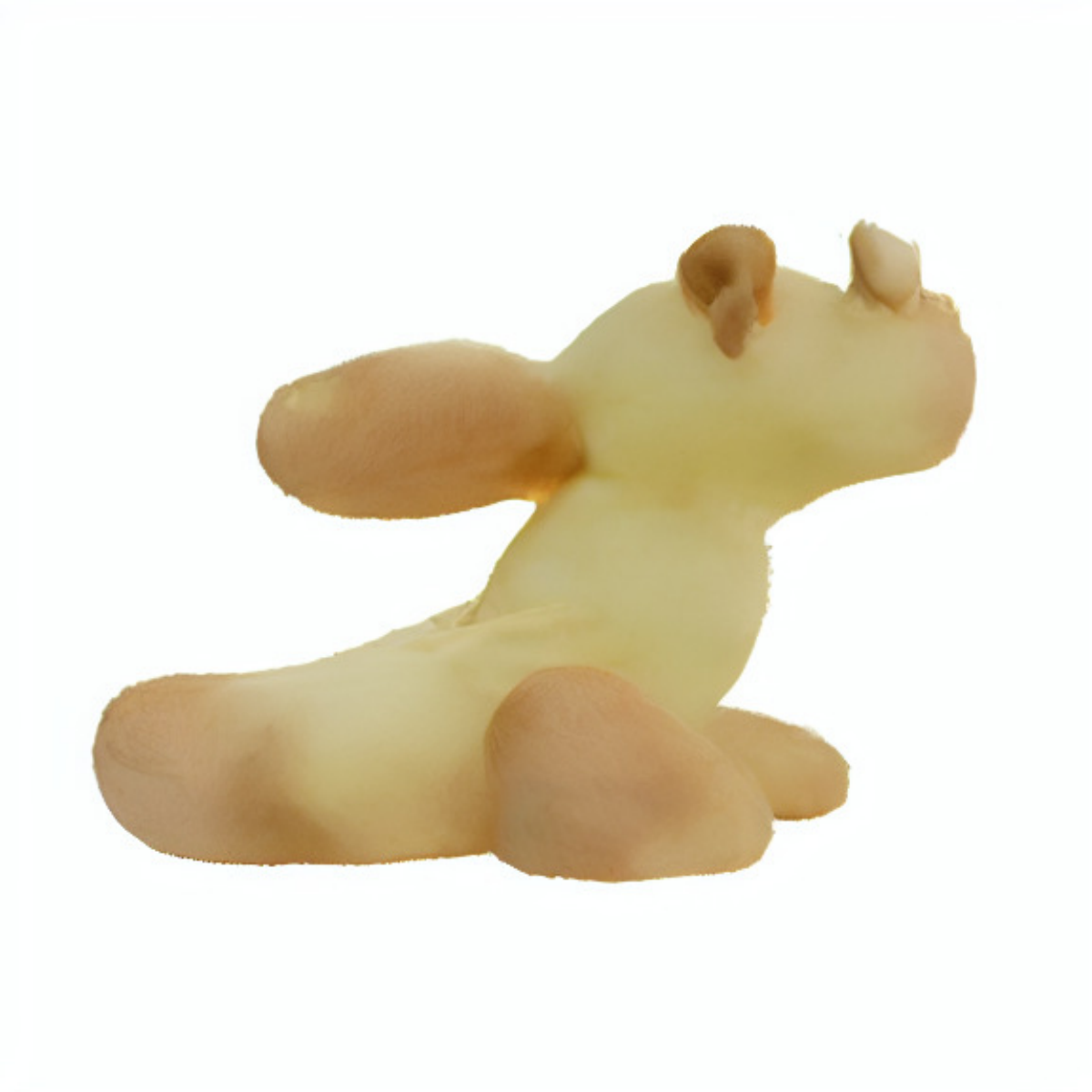}
    		\end{subfigure}&
    		\begin{subfigure}[t]{.133\linewidth}
    			\centering
    			\includegraphics[width=\linewidth]{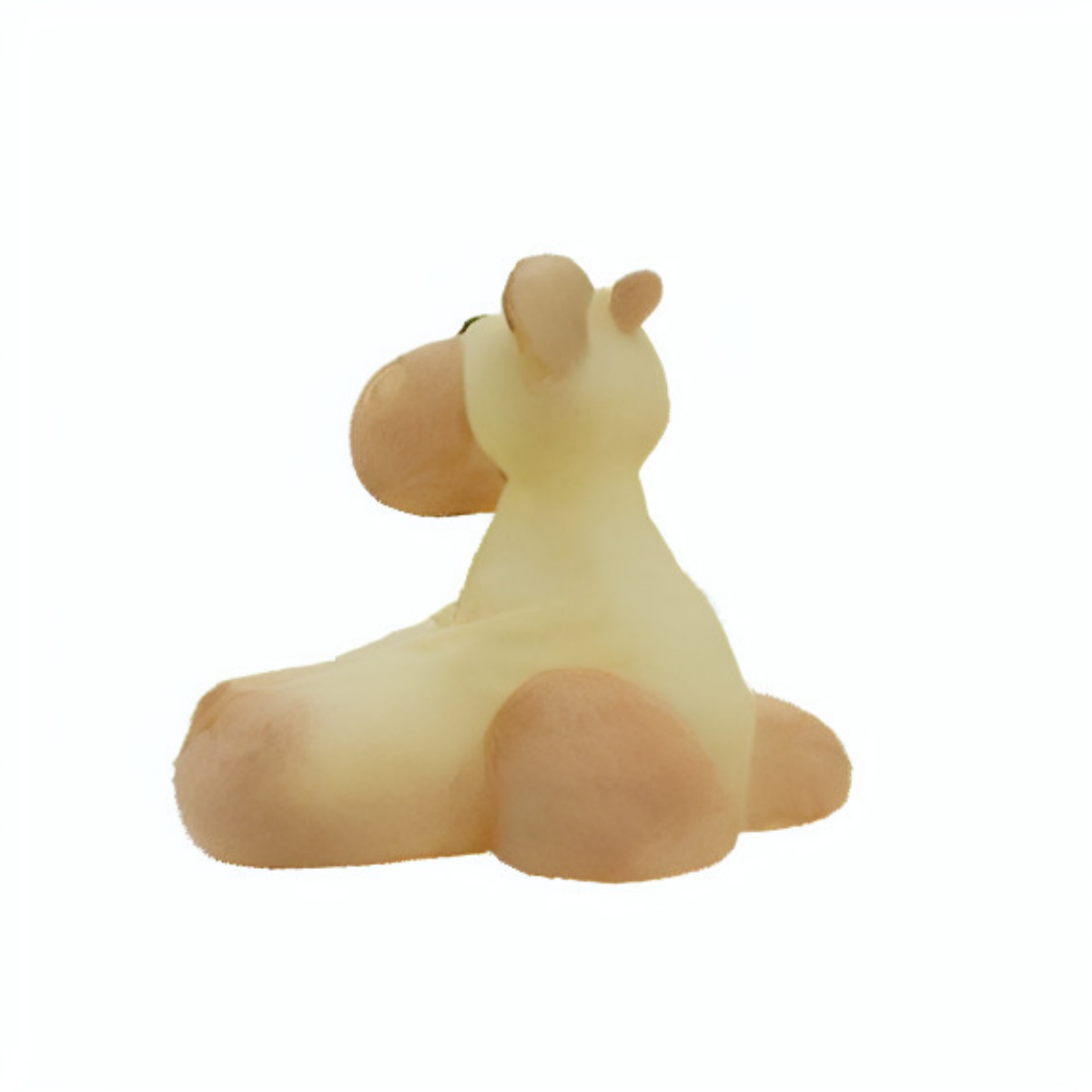}
    		\end{subfigure}&
    		\begin{subfigure}[t]{.133\linewidth}
    			\centering
    			\includegraphics[width=\linewidth]{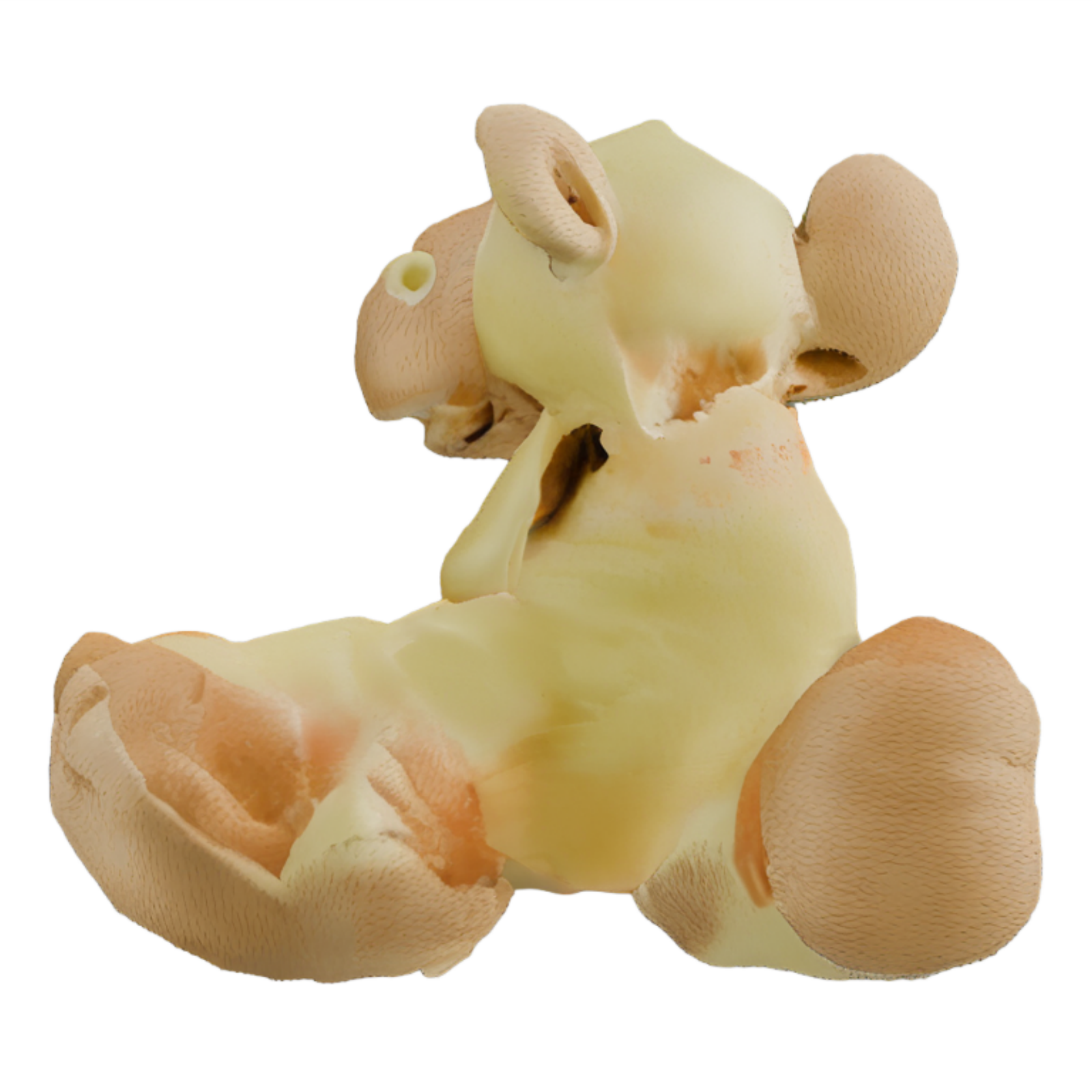}
    		\end{subfigure}&
    		\begin{subfigure}[t]{.133\linewidth}
    			\centering
    			\includegraphics[width=\linewidth]{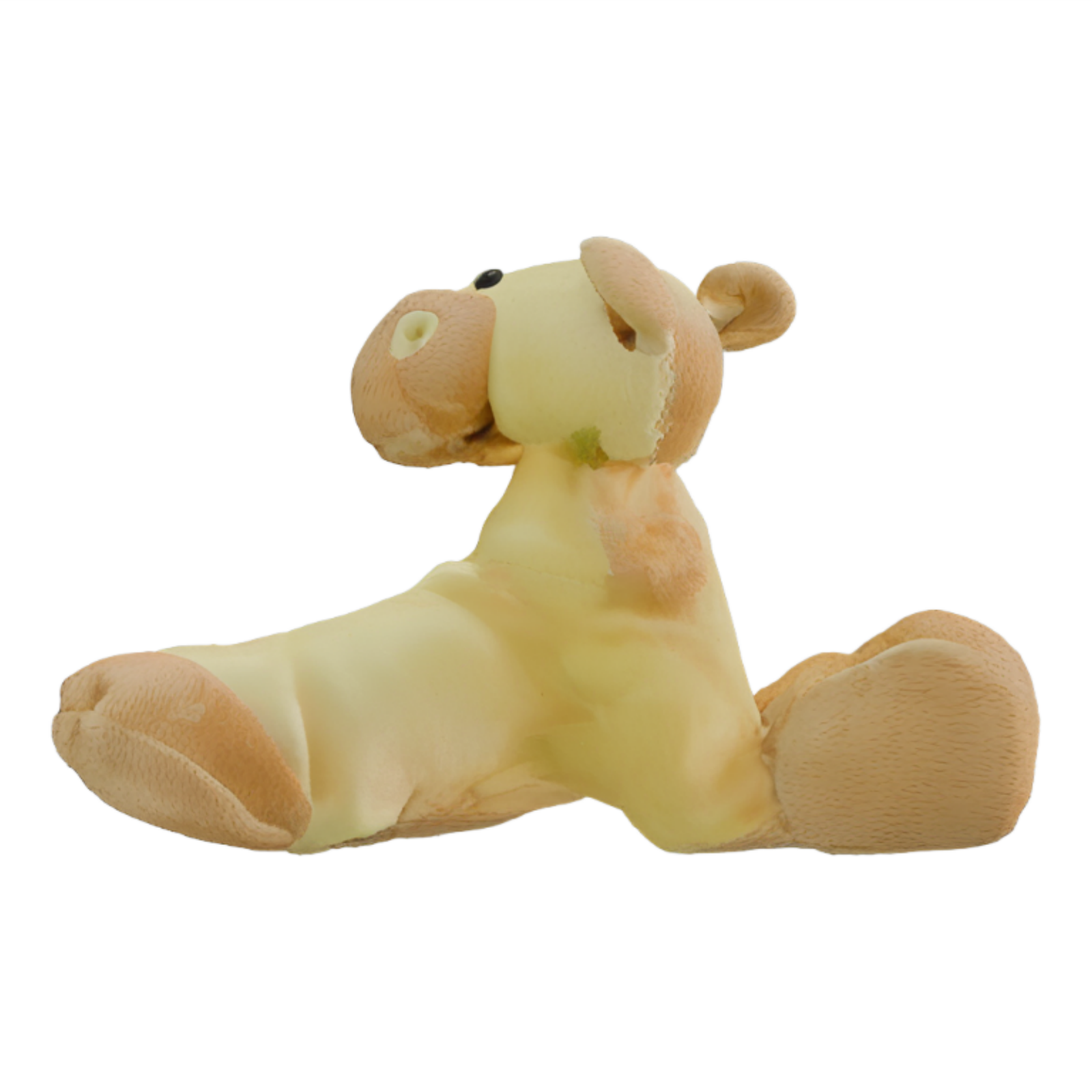}
    		\end{subfigure}\\
    		
    		&
    		\begin{subfigure}[t]{.133\linewidth}
    			\centering
    			\includegraphics[width=\linewidth]{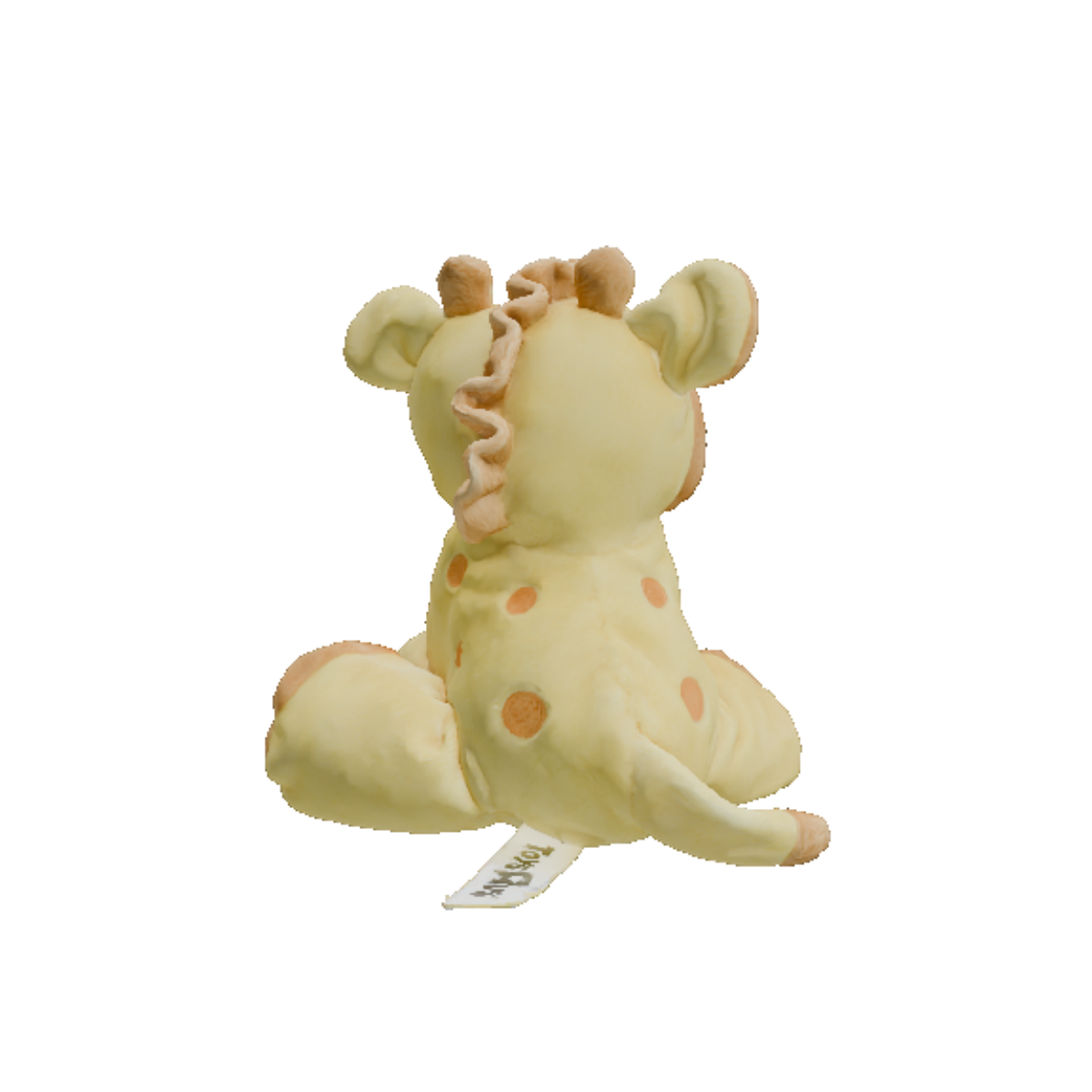}
    		\end{subfigure}&
    		\begin{subfigure}[t]{.133\linewidth}
    			\centering
    			\includegraphics[width=\linewidth]{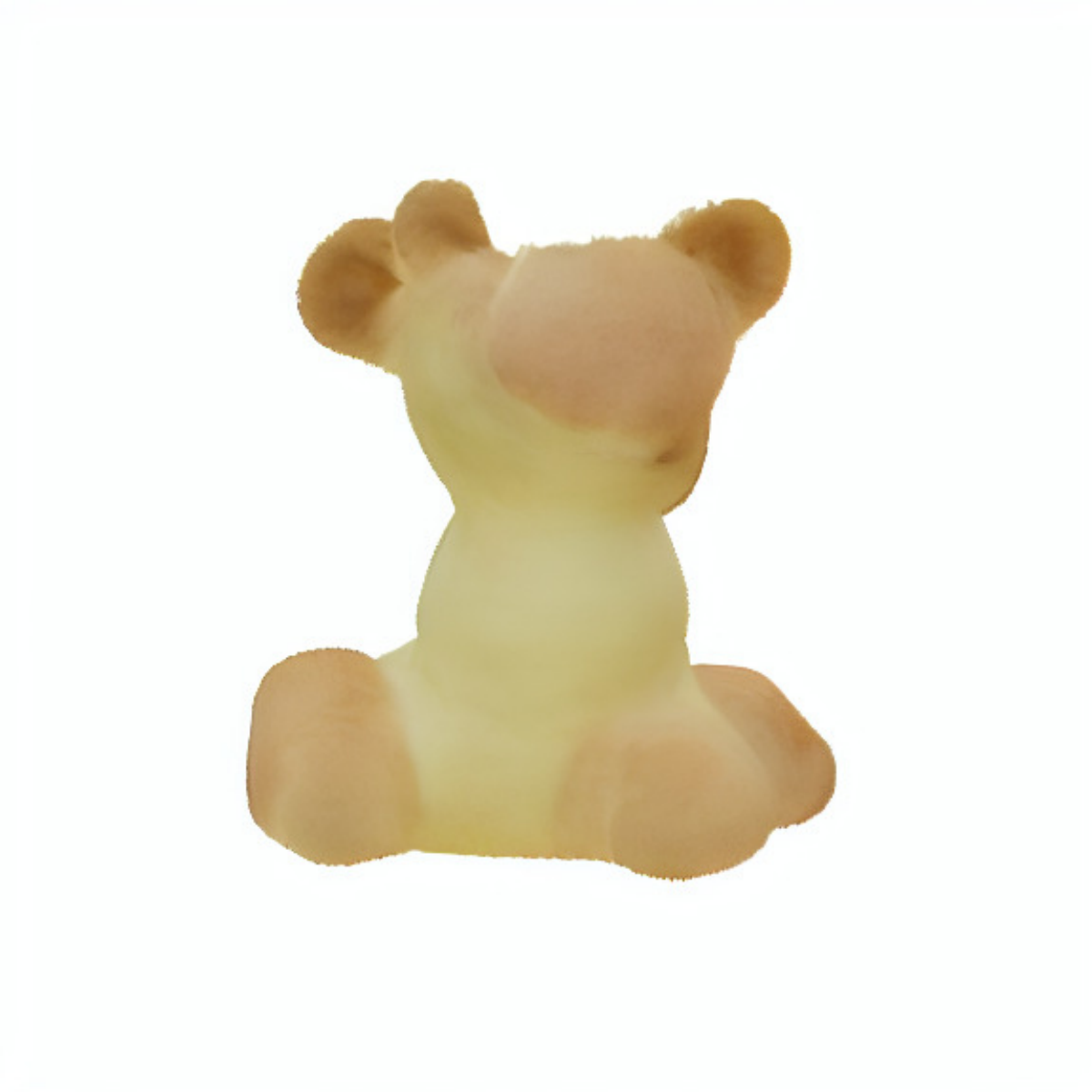}
    		\end{subfigure}&
    		\begin{subfigure}[t]{.133\linewidth}
    			\centering
    			\includegraphics[width=\linewidth]{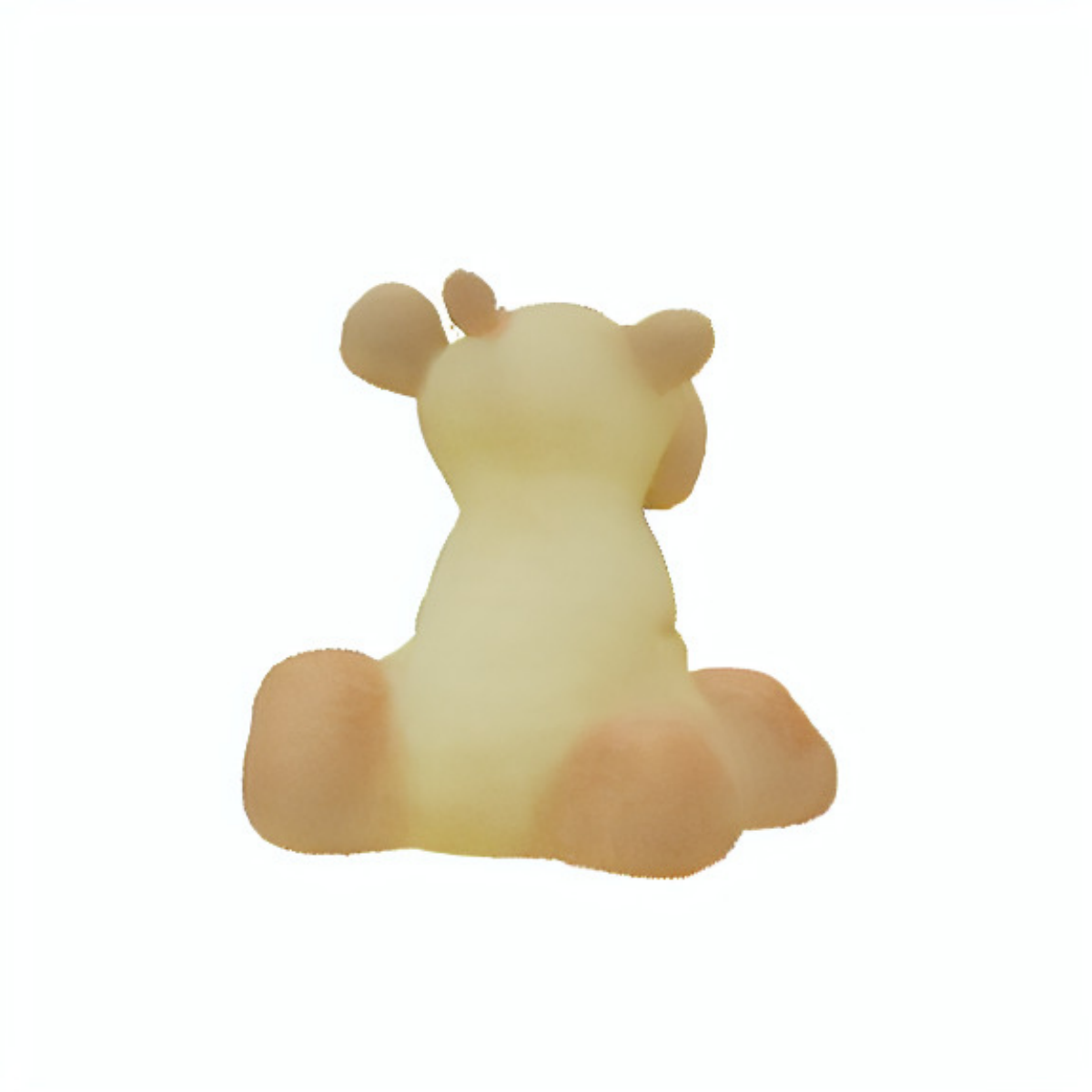}
    		\end{subfigure}&
    		\begin{subfigure}[t]{.133\linewidth}
    			\centering
    			\includegraphics[width=\linewidth]{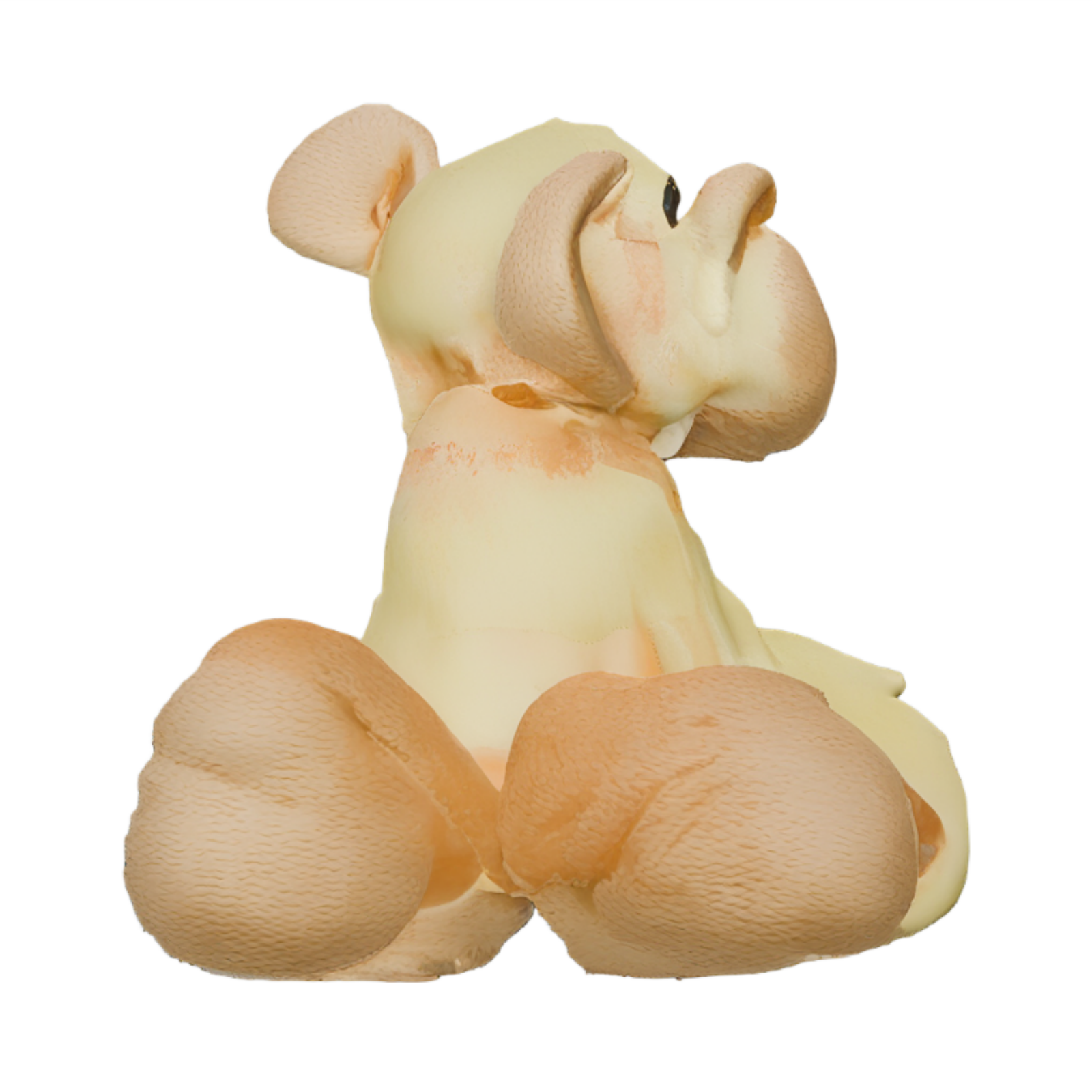}
    		\end{subfigure}&
    		\begin{subfigure}[t]{.133\linewidth}
    			\centering
    			\includegraphics[width=\linewidth]{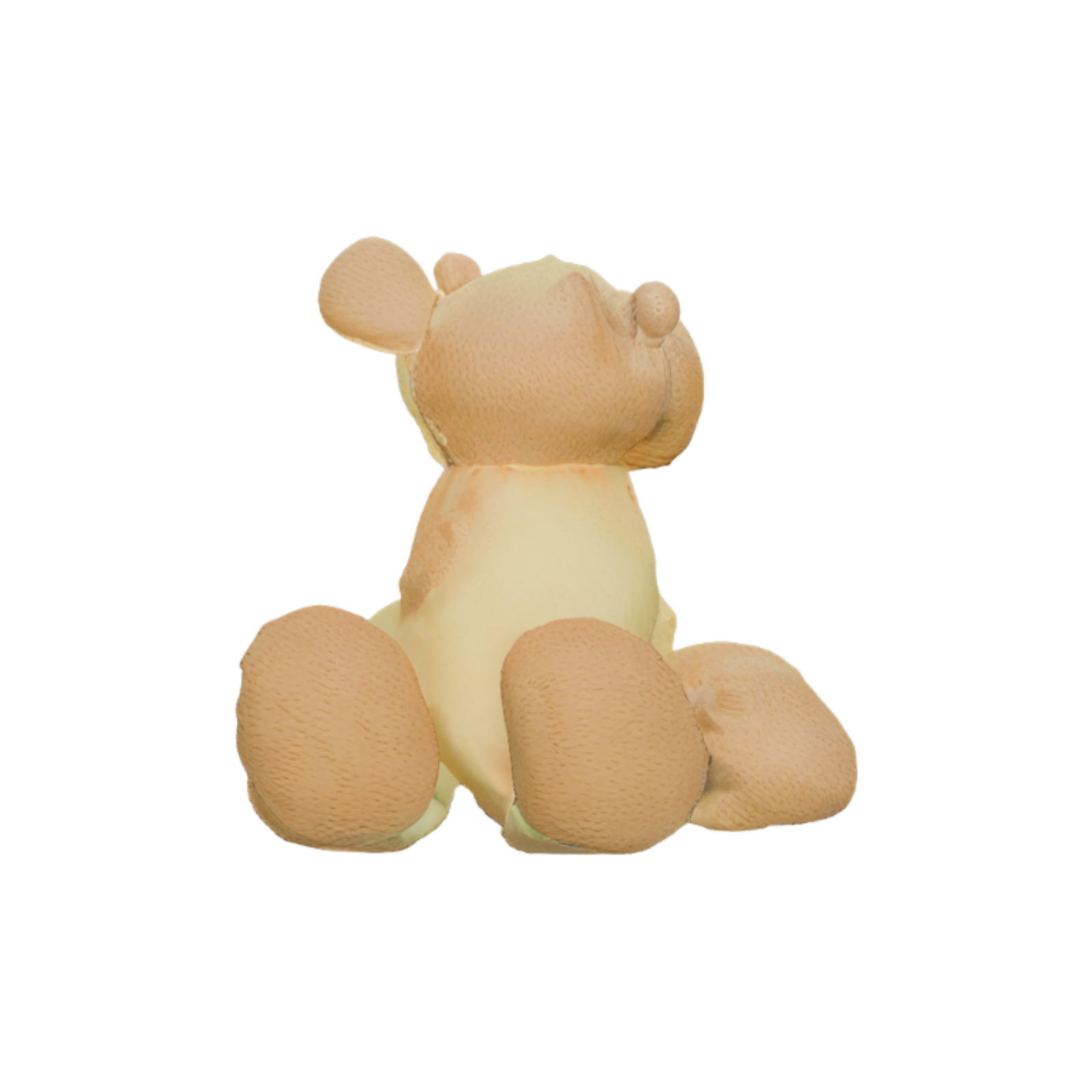}
    		\end{subfigure}\\
    		Input&GT&SV3D&SV3D \textbf{with EDN}&Mv-Adapter&Mv-Adapter \textbf{with EDN}
    	\end{tabular}
    	\caption{Results of two NVS models generated from random Gaussian noise and our EDN-optimized noise, respectively. Images are generated using the same random seed and camera poses. Images synthesized with EDN exhibit better consistency with the ground truth in both appearance contours and local details.}
    	\label{EDN_outputs_camparison}
    \end{figure*}
    \begin{abstract}
    Single-view novel view synthesis (NVS) models based on diffusion models have recently attracted increasing attention, as they can generate a series of novel view images from a single image prompt and camera pose information as conditions. It has been observed that in diffusion models, certain high-quality initial noise patterns lead to better generation results than others. However, there remains a lack of dedicated learning frameworks that enable NVS models to learn such high-quality noise. To obtain high-quality initial noise from random Gaussian noise, we make the following contributions. First, we design a discretized Euler inversion method to inject image semantic information into random noise, thereby constructing paired datasets of random and high-quality noise. Second, we propose a learning framework based on an encoder-decoder network (EDN) that directly transforms random noise into high-quality noise. Experiments demonstrate that the proposed EDN can be seamlessly plugged into various NVS models, such as SV3D and MV-Adapter, achieving significant performance improvements across multiple datasets. Code is available at: https://github.com/zhihao0512/EDN.
    \end{abstract}
    \section{Introduction}
    \IEEEPARstart{S}{ingle}-view novel view synthesis (NVS), involving generating 3D object images from other viewpoints using only one reference image, is a long-standing and valuable research challenge in computer vision. This task has numerous applications in areas such as game design \cite{xu2024sketch2scene} and virtual/augmented reality \cite{asish2025synthesizing}.
	
    Recent advancements in diffusion models \cite{rombach2022high,song2020denoising} have spurred several related studies. Zero-1-to-3 \cite{liu2023zero} is the first to apply diffusion models for zero-shot novel view synthesis. However, it generates only a single image for each specified viewpoint. To address this limitation, several models, including Syncdreamer \cite{liu2023syncdreamer}, Zero123++ \cite{shi2023zero123++}, and MvDiffusion++ \cite{tang2024mvdiffusion++}, have leveraged attention mechanisms to generate multiple images from different viewpoints simultaneously using text-to-image diffusion models. Despite these advancements, the generated views are still constrained by fixed viewpoints. Recent studies \cite{gao2024cat3d, zheng2024free3d} have resolved the issue of controlling camera poses. In pursuit of better generation effects, some methods \cite{kwak2024vivid,voleti2024sv3d} have been improved based on video diffusion models to achieve better multi-view consistency. Currently, generating high-resolution images has become a central research focus, with models like Mv-Adapter \cite{huang2025mv} and Pippo \cite{kant2025pippo} showing notable progress.
	
	Previous works in this area typically require fine-tuning the diffusion model architecture to perform NVS, which is computationally expensive. However, image synthesis is driven not only by the image prompt but also by the noise input. Recent studies \cite{lichenzigzag, guo2024initno, qi2024not} have shown that in text-to-image models, certain selected or optimized noise patterns can generate images that are more consistent with the given prompt. Similar observations have been found in text-to-video models \cite{kim2025model,oshima2025inference}. We hypothesize that this phenomenon, where certain initial noise leads to better generation results, also holds for NVS models built upon text-to-image and text-to-video diffusion models.
	
    In \cite{zhou2025golden}, a learning framework was proposed to convert random Gaussian noise into optimized noise by adding a small desirable perturbation derived from the text prompt, thereby enhancing the generation results of text-to-image diffusion models. This perturbation can be regarded as the semantic information corresponding to the given text prompt. The improved random noise can be considered as noise prompt. Inspired by \cite{zhou2025golden}, in this paper, we focus on improving the quality of NVS results by optimizing the initial noise through a learning framework. Our main contributions are as follows:
	
	First, we introduce the concept of ``high-quality noise'' for NVS model and propose a noise learning framework based on an encoder-decoder network (EDN). Compared with random Gaussian noise, high-quality noise incorporates image semantic information from the reference image, yielding NVS results that are more consistent with the reference. The trained EDN is a plug-and-play module that integrates seamlessly into the inference process of NVS models without modifying the original U-Net architecture. EDN requires minimal computational resources and does not significantly slow down the inference speed. 
	
	Second, we design a high-quality noise collection process based on NVS model, using an ``inference-inversion'' method to inject image semantic information into the initial noise. For this purpose, we design a discretized Euler inversion method based on the principle of Denoising Diffusion Implict Model (DDIM) inversion \cite{mokady2023null}. We also implement a data filtering mechanism to remove high-quality noise with poor performance, improving the quality of the training dataset.
	
	Third, the high-quality noise collection process and the EDN can be applied to multiple NVS models based on diffusion, discretized Euler inference method, and classifier-free guidance (CFG), such as SV3D \cite{voleti2024sv3d} and Mv-Adapter. As shown in Fig. \ref{EDN_outputs_camparison}, the results generated by the EDN-optimized noise significantly improve both local details and contour consistency compared to the original models.
	
	The remainder of this paper is organized as follows: Section \ref{Related_Work} reviews related work, Section \ref{Preliminaries} introduces the preliminary concepts of discretized Euler inference and inversion methods and CFG, Section \ref{High-Quality_Noise_Learning} details the high-quality noise collection process and EDN design, and Section \ref{Experimental_Results} presents experimental results, including quantitative, qualitative, ablation studies, and a series of supplementary experiments to validate the generalization and effectiveness of the proposed method. 
	
	\section{Related Work}
	\label{Related_Work}
	\subsection{Novel View Synthesis from a Single Image}
	Diffusion model has become a key approach for addressing NVS tasks with a single image input. Zero-1-to-3 \cite{liu2023zero} pioneers open-world single-image-to-3D conversion through zero-shot novel view synthesis. It utilizes the image prior knowledge from the Stable Diffusion \cite{rombach2022high} model and proposes a framework that, given a single image, generates images of the target object from different perspectives by adjusting the camera pose conditions. Zero-1-to-3 XL improves generation quality by using a larger dataset \cite{deitke2023objaversexl}. However, both methods can only generate a single image from one perspective at a time; thus, when generating multiple views of the same object, they suffer from a lack of consistency among the generated images.
	
   Several studies have focused on generating multiple images from different perspectives simultaneously, aiming to improve multi-view consistency. Syncdreamer \cite{liu2023syncdreamer} employs 3D volumes and depth-wise attention to ensure consistency across views. Zero123++ \cite{shi2023zero123++} uses 3D self-attention to condition multi-view image generation on a single image. Mvdiffusion++ \cite{tang2024mvdiffusion++} extends the multi-branch U-Net architecture from MV-Diffusion \cite{deng2023mv}, using global self-attention to achieve 3D consistency. While these models generate multiple views simultaneously, the viewpoints are fixed, limiting controllability over the camera pose conditions.
	
	Some studies are dedicated to making the camera pose conditions controllable. Free3D \cite{zheng2024free3d} introduces lightweight multi-view attention layers and a ray conditioning normalization (RCN) layer. Cat3d \cite{gao2024cat3d} uses 3D self-attention and a camera ray representation (``ray map'') to represent camera pose conditions. These methods allow users to specify the pose conditions for multi-view image generation. 
	
	To further enhance performance, some methods leverage the powerful temporal prior knowledge within video diffusion models for better multi-view consistency and generalization. SV3D \cite{voleti2024sv3d} adapts an image-to-video diffusion model \cite{blattmann2023stable} for novel multi-view synthesis, incorporating explicit camera pose control. However, the model sometimes suffers from blurred local details in the generated images. Vivid-1-to-3 \cite{kwak2024vivid} combines predicted noises from the view-conditioned diffusion model Zero-1-to-3 with the video diffusion model Zeroscope\footnote{Available at: https://huggingface.co/cerspense/zeroscope\_v2\_XL} for denoising. Despite these improvements, it still struggles with multi-view consistency for some objects.
	
	High-resolution image generation has become a key focus in recent research. Mv-Adapter \cite{huang2025mv} proposes an adapter-based solution for multi-view image generation. It updates fewer parameters, making training more efficient with lower computational resource consumption, while being capable of generating high-resolution ($768\times 768$) multi-view images. However, the model still faces challenges in generating realistic image details. Pippo \cite{kant2025pippo} is designed with a diffusion transformer architecture aimed at improving multi-view generation performance and viewpoint control. It can generate multiple 1K-resolution images with consistent multi-view alignment during inference. However, it is limited to generating human images.
	\subsection{Noise Search for Diffusion Models}
	Fine-tuning diffusion models requires substantial computational resources. As a result, many studies focus on improving the generation quality of diffusion models without fine-tuning. Among these efforts, noise search in diffusion models is an emerging research direction. 
	
	Some researchers \cite{lichenzigzag,guo2024initno,qi2024not} have pointed out that certain initial noise patterns can lead to better generation results in text-to-image diffusion models. Bai \textit{et al.} \cite{lichenzigzag} improves image generation by repeatedly adding ``semantic information'' during inference to optimize the noise. Guo \textit{et al.} \cite{guo2024initno} adopts a two-step noise search algorithm: it first optimizes a random noise sample, and then, after obtaining multiple optimized noises, selects a globally optimal noise. Qi \textit{et al.} \cite{qi2024not} proposes two types of search algorithms: noise selection, which picks the optimal noise from a set of random samples, and noise optimization, which refines an existing noise using gradient descent based on evaluation metrics. Building on previous research, Ma \textit{et al.} \cite{ma2025inference} explains noise search from the perspective of inference-time scaling and proposes a systematic noise search framework, exploring the effects of different evaluation functions and search algorithms. 
	
	Researchers have extended noise search methods to video diffusion models. Kim \textit{et al.} \cite{kim2025model} proposes an active noise selection framework based on a principled Bayesian formulation of attention-based uncertainty. Oshima \textit{et al.} \cite{oshima2025inference} presents a diffusion latent space beam search method with a forward-looking estimator, which maximizes the given alignment reward by selecting a better diffusion latent space during inference. 
	
    To reduce inference time, Zhou \textit{et al.} \cite{zhou2025golden} proposes a machine learning framework that improves the initial random noise in text-to-image diffusion models, enhancing the alignment between text and generated images.

	This paper builds upon advanced NVS models and incorporates noise search techniques from diffusion models to optimize the initial random noise in NVS pipelines, thereby improving generation quality. The resulting optimized noise is referred to as ``high-quality noise''. This optimization method avoids the substantial computational cost of fine-tuning NVS models. Additionally, it extends noise search research from text-to-image and text-to-video settings to the domain of NVS.
	\section{Preliminaries}
	\label{Preliminaries}
	The core idea of this paper is to collect paired samples of random noise and high-quality noise based on the difference in CFG scales during the discretized Euler inference and inversion processes within the NVS model, and then use these pairs to train an encoder-decoder network. This section introduces discretized Euler inference method and CFG, and derives the discretized Euler inversion formulation inspired by the DDIM inversion principle.
	\subsection{Discretized Euler Inference Method}
	For NVS with diffusion model, different sampling strategies can be applied during the reverse process, among which the discretized Euler inference method \cite{karras2022elucidating} is a widely used option. This method uses Euler's scheme to solve the deterministic ordinary differential equation (ODE), introducing a time-varying scaling term that constrains the noisy data distribution within a unit-standard-deviation tunnel as time progresses. We define the total number of denoising steps \(T\), the image prompt \(\mathbf{c}\), which is an encoding of the input image and the pose prompt \(\mathbf{p}\), which is an encoding of the desired camera poses. For models employing discretized Euler inference, the initial random noise \(\mathbf{z}_{T}\) is produced by scaling a standard Gaussian noise sample with an initial scaling factor \(q\), and is subsequently rescaled at each timestep. Let \(\mathbf{z}_{t}\) denote the latent for desired views of images at timestep \(t\), \(\mathbf{z}_{t}'\) denote the rescaled version of \(\mathbf{z}_{t}\), i.e., \(\mathbf{z}_{t}' = \mathrm{DEScaler}(\mathbf{z}_{t}) =\mathbf{z}_{t}/ \sqrt{\sigma_{t}^{2}+1  } \). Here, \(\sigma _{t}\) is a pre-defined parameter for scheduling the scales of adding noises, the factor \(q= \max_{t=0}^T\sqrt{\sigma_{t}^{2}+1}\). The reverse diffusion process can be written as \(\mathbf{z}_{t-1} =\Phi  ( \mathbf{z}_{t} ,\varepsilon _{\theta } ( \mathbf{z}_{t}',t,\mathbf{c},\mathbf{p}  )   ) \), where \(\Phi  ( \cdot   ) \) is the discretized Euler sampling rule, the term \(\varepsilon _{\theta }(\mathbf{z}_{t}',t,\mathbf{c} ,\mathbf{p})\) denotes the predicted noise by the denoising network \(\theta\) at timestep \(t\). 
	
	The discretized Euler inference method supports multiple prediction types, including ``sample'', ``epsilon'', and ``v-prediction'' \cite{von-platen-etal-2022-diffusers}. In this work, we focus on the ``v-prediction'' and ``epsilon'' cases, as SV3D and Mv-Adapter employ these two prediction techniques respectively. The reverse process can thus be represented as follows:
	
	\noindent for ``v-prediction'':
	\begin{equation}
		 \mathbf{z}_{t-1} =\frac{1+\sigma _{t} \sigma  _{t-1}}{\sigma _{t}^{2} +1} \mathbf{z}_{t} +\frac{\sigma  _{t-1}-\sigma _{t}}{\sqrt{\sigma _{t}^{2} +1} }\varepsilon _{\theta }(\mathbf{z}_{t}',t,\mathbf{c},\mathbf{p} ),\label{1}
	\end{equation}
    and for ``epsilon'':
    \begin{equation}
     \mathbf{z}_{t-1} =\mathbf{z}_{t} + ( \sigma  _{t-1}-\sigma _{t}  ) \varepsilon _{\theta }(\mathbf{z}_{t}',t,\mathbf{c},\mathbf{p}).\label{13}
    \end{equation}
	\indent The process of generating samples from random noise using the discretized Euler inference method is deterministic, as no additional stochastic noise is injected during the intermediate steps. Due to this property, its inversion process becomes feasible. 
	
	The essence of DDIM inversion is to reverse the forward update from \(\mathbf{z}_{t}\) to \(\mathbf{z}_{t-1}\) into an update that recovers \(\mathbf{z}_{t}\) from \(\mathbf{z}_{t-1}\). Following this idea, we derive the discretized Euler inversion process by adding predicted noise:
	\(\tilde{\mathbf{z}}_{t}   =\Psi   ( \tilde{\mathbf{z}}_{t-1} ,\varepsilon _{\theta } (\tilde{\mathbf{z}}_{t-1}',t,\mathbf{c},\mathbf{p}  )   )\).
	
	\noindent For ``v-prediction'': the inversion is derived from Eq. \eqref{1}, given by
	\begin{equation}
		\tilde{\mathbf{z}}_{t} =\frac{\sigma _{t}^{2} +1}{1+\sigma _{t} \sigma  _{t-1}} \tilde{\mathbf{z}}_{t-1} -\frac{\sqrt{\sigma _{t}^{2} +1} }{1+\sigma _{t} \sigma  _{t-1}}(\sigma  _{t-1}-\sigma _{t} )\varepsilon _{\theta } (\tilde{\mathbf{z}}_{t-1}',t,\mathbf{c},\mathbf{p}  ), \label{3}
	\end{equation}
	and for ``epsilon'': the inversion is derived from Eq. \eqref{13}, given by
	\begin{equation}
		\tilde{\mathbf{z}}_{t} =\tilde{\mathbf{z}}_{t-1} - ( \sigma  _{t-1}-\sigma _{t}  ) \varepsilon _{\theta } (\tilde{\mathbf{z}}_{t-1}',t,\mathbf{c},\mathbf{p}  ).  \label{14}
	\end{equation}
	
	Here, along the inversion, we approximate the predicted noise at timestep \(t\) using the estimate at timestep \(t-1\), i.e., \(\varepsilon _{\theta } (\tilde{\mathbf{z}}_{t-1}',t,\mathbf{c},\mathbf{p}  ) \approx \varepsilon _{\theta } (\tilde{\mathbf{z}}_{t}',t,\mathbf{c},\mathbf{p}  ) \).
	\subsection{Classifier-Free Guidance (CFG)}
	Classifier-free guidance \cite{ho2022classifier} allows us to control the generation process, balancing the quality and diversity of the generated samples. By interpolating between image-prompted and non-image-prompted predictions, \(\varepsilon _{\theta } ( \mathbf{z}_{t}',t,\mathbf{c},\mathbf{p}  )\) can be expressed as:
	\begin{align}
		\nonumber&\varepsilon _{\theta } ( \mathbf{z}_{t}',t,\mathbf{c},\mathbf{p}  )\\=&\mu _{\theta }(\mathbf{z}_{t}',t,\mathbf{\emptyset},\mathbf{p} )+\gamma  [\mu _{\theta }(\mathbf{z}_{t}',t,\mathbf{c},\mathbf{p})-\mu _{\theta }(\mathbf{z}_{t}',t,\mathbf{\emptyset},\mathbf{p} ) ].\label{2}
	\end{align}
	\indent Here, \(\mu _{\theta }\) is the noise predictor, \(\gamma\) denotes the CFG scale, \(\mathbf{\emptyset}\) denotes an empty image prompt, which refers to an entity that has the same shape as \(\mathbf{c}\) but all its values are zero. 
	\section{High-Quality Noise Learning}
	\label{High-Quality_Noise_Learning}
	In this section, we present the methodology of high-quality noise learning, including data collection, data filter, and network training. 
	
	Recent studies \cite{lichenzigzag,guo2024initno,qi2024not,kim2025model,oshima2025inference} have shown that, in diffusion models, certain high-quality initial noise patterns yield better generation results than others, including in both text-to-image and text-to-video settings. We hypothesize that NVS models exhibit a similar phenomenon, where the initial noise can also be regarded as a prompt. Motivated by this intuition, we aim to transform random Gaussian noise into high-quality noise by adding a desirable perturbation derived from the image prompt. In this way, the resulting high-quality noise carries semantic information from the reference image, thereby enhancing its guiding effect in the diffusion process and enabling the generation of outputs that are both higher in quality and more consistent with the reference image. Inspired by \cite{zhou2025golden}, we propose an initial noise optimization learning framework that turns random noise into the high-quality noise. Given the training dataset \(D= \{ \mathbf{z}_{Ti}, \tilde{\mathbf{z}}_{Ti},\mathbf{I}_{i}   \} _{i=1}^{ |  D | }\) (including source noise \(\mathbf{z}_{T}\), target noise \(\tilde{\mathbf{z}}_{T}\), and the VAE embedding \(\mathbf{I}\) of reference image), a loss function \(\zeta \), and a neural network \(\phi \), the general formula for the noise prompt learning task is as follows: 
	\begin{equation}
		\phi ^{*} = \mathrm{argmin}_{\phi } \mathbb{E} _{(\mathbf{z}_{Ti}, \tilde{\mathbf{z}}_{Ti},\mathbf{I}_{i})\sim D }  [  \zeta  ( \phi  (  \mathbf{z}_{Ti},\mathbf{I}_{i} ) , \tilde{\mathbf{z}}_{Ti} )  ].\label{11}
	\end{equation}
	\indent Our goal is to train an optimal neural network model \(\phi ^{*}\) using the training dataset \(D\). Fig. \ref{collect_dataset_process} illustrates a workflow with three phases: data collection, training, and inference.
	
	\subsection{High-Quality Noise Collection}
	\subsubsection{Collection Process Based on Diffusion Model}
	\begin{figure*}[htbp] 
		\centering 
		\includegraphics[width=0.9\textwidth]{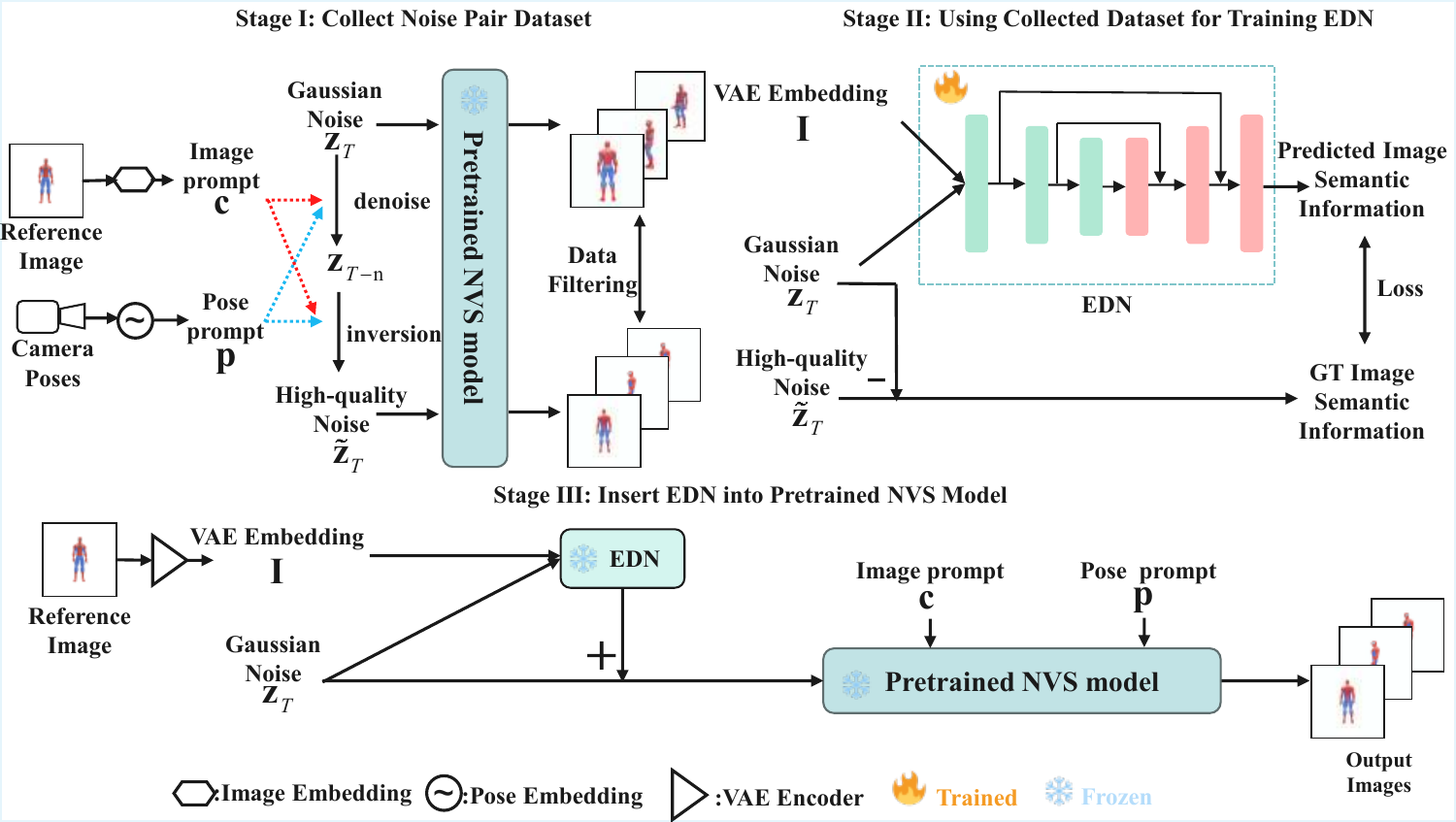} 
		\caption{The workflow of our high-quality noise learning framework with three stages. Stage I: We first denoise the initial Gaussian noise \(\mathbf{z} _{T}\)​ to obtain \(\mathbf{z} _{T-n}\). Then, using the discretized Euler inversion method, we derive the inverted noise \(\tilde{\mathbf{z}}_{T}\), which is infused with semantic information from the reference image. The resulting samples are further filtered to ensure that constructed training dataset is both diverse and representative. Stage II: The initial noise \(\mathbf{z} _{T}\) and the VAE embedding \(\mathbf{I}\) of the reference image are concatenated and fed into the EDN. The EDN decoder then predicts a semantic information map, which is used to compute the loss based on its differences from both \(\mathbf{z} _{T}\) and the inverted noise \(\tilde{\mathbf{z}}_{T}\). Stage III: During inference, the EDN injects the predicted image semantic information into the initial random noise before it enters the diffusion reverse process. This produces high-quality noise that enhances the generation performance of the pretrained NVS model.} 
		\label{collect_dataset_process} 
	\end{figure*}
	How can we inject image semantic information into the initial Gaussian noise to obtain high-quality noise? Meng \textit{et al.} \cite{meng2023distillation} suggests that adding random noise at each timestep during sampling, followed by re-denoising, can significantly enhance the semantic fidelity of synthesized images. In \cite{zhou2025golden}, a straightforward method called ``re-denoising sampling'' is introduced, which combines DDIM inversion with CFG to embed semantic information into the initial noise. Inspired by these approaches, we use the ``inference-inversion'' method to inject image semantics into the initial random noise. Specifically, we first perform \(n\) steps of discretized Euler inference (\(\Phi(\cdot )\)) with a CFG scale of \(\gamma _{1}\), and then apply \(n\) steps of discretized Euler inversion (\(\Psi(\cdot )\)) with a CFG scale of \(\gamma _{2}\). The CFG formulations used in the inference and inversion processes are expressed as:
	\begin{align}
		\nonumber&\varepsilon _{\theta, 1 } ( \mathbf{z}_{t}',t,\mathbf{c},\mathbf{p} )\\=&\mu _{\theta }(\mathbf{z}_{t}',t,\mathbf{\emptyset},\mathbf{p} )+\gamma_{1}  [\mu _{\theta }(\mathbf{z}_{t}',t,\mathbf{c},\mathbf{p})-\mu _{\theta }(\mathbf{z}_{t}',t,\mathbf{\emptyset},\mathbf{p} ) ],\label{16}
	\end{align}
	and
	\begin{align}
		\nonumber&\varepsilon  _{\theta, 2 }(\tilde{\mathbf{z}}_{t-1}',t,\mathbf{c},\mathbf{p} )\\=&\mu _{\theta }(\tilde{\mathbf{z}}_{t-1}',t,\mathbf{\emptyset},\mathbf{p} )+\gamma_{2}  [\mu _{\theta }(\tilde{\mathbf{z}}_{t-1}',t,\mathbf{c},\mathbf{p})-\mu _{\theta }(\tilde{\mathbf{z}}_{t-1}',t,\mathbf{\emptyset},\mathbf{p} ) ],\label{17}
	\end{align}
	respectively.
	
	This process transforms the initial random noise \(\mathbf{z} _{T}\) into \(\tilde{\mathbf{z}}_{T}=\Psi_{t=T-n+1}^{T}(\Phi_{t=T}^{T-n+1}  (\mathbf{z} _{T}))\). Throughout the process, the CFG scale \(\gamma _{1}\) in \(\Phi(\cdot )\) is always greater than \(\gamma _{2}\) in \(\Psi(\cdot )\), which gradually injects image semantic information into the noise. The detailed theoretical derivation is provided in Appendix \ref{Theory}.
	
	It is important to note that, in the discretized Euler scheduler, the initial noise is obtained by scaling standard Gaussian noise, and an additional \(\mathrm{DEScaler}(\cdot)\) is applied at each inference step. Therefore, it is necessary to record the mean and standard deviation of the intermediate variables during the inference process. These statistics are then reused to scale the corresponding intermediate variables during the inversion process. We define a function that aligns the distribution of \(\mathbf{z}\) to that of \(\mathbf{x}\) (i.e., matching the mean and standard deviation), given as:
	\begin{equation}
		\mathrm{Align}(\mathbf{z}|\mathbf{x})=\sigma _{\mathbf{x}} \cdot \frac{\mathbf{z}-\mu _{\mathbf{z}} }{\sigma _{\mathbf{z}}} +\mu _{\mathbf{x}}.\label{10}
	\end{equation}
	\indent Here, \(\mu _{\mathbf{x}}\) and \(\sigma _{\mathbf{x}}\) represent the mean and standard deviation of \(\mathbf{x}\), respectively, while \(\mu _{\mathbf{z}}\) and \(\sigma _{\mathbf{z}}\) represent those of \(\mathbf{z}\).
	
	As shown in the Algorithm \ref{alg1} and Stage I of Fig. \ref{collect_dataset_process}, the process for collecting paired samples of random noise and high-quality noise is illustrated.
	\begin{algorithm}[!h]
		\caption{Noise pair collection based on NVS model}
		\label{alg1}
		\renewcommand{\algorithmicrequire}{\textbf{Input:}}
		\renewcommand{\algorithmicensure}{\textbf{Output:}}
		
		\begin{algorithmic}[1]
			\REQUIRE Random Gaussian noise \(\mathbf{z} _{	T}\), inference steps \(n\), image prompt \(\mathbf{c}\), camera pose prompt \(\mathbf{p}\), noise predictor \(\mu _{\theta }(\cdot)\), CFG scales \(\gamma_{1}\) and \(\gamma_{2}\).
			\ENSURE Source noise \(\mathbf{z} _{T}\), target noise \(\tilde{\mathbf{z}}_{T}\).  
			
			\textbf{Inference:}
			\FOR{each \(t\in[T,\cdots,T-n+1]\)}
			\STATE \(\mathbf{z}_{t}'\gets \mathrm{DEScaler}(\mathbf{z}_{t})\)
			\STATE  Calculate \(\varepsilon _{\theta,1 } ( \mathbf{z}_{t}',t,\mathbf{c},\mathbf{p} )\) by Eq. \eqref{16};
			\STATE  Calculate \( \mathbf{z}_{t-1}\) by Eq. \eqref{1} or Eq. \eqref{13};
			\ENDFOR  
			\STATE  \(\tilde{\mathbf{z}}_{T-n}\gets \mathbf{z}_{T-n}\)
			
			\textbf{Inversion:}
			\FOR{each \(t\in[T-n+1,\cdots,T]\)} 
			\STATE  \(\tilde{\mathbf{z}}_{t-1}'\gets \mathrm{DEScaler}(\tilde{\mathbf{z}}_{t-1})\)
			\STATE  \(\tilde{\mathbf{z}}_{t-1}'\gets \mathrm{Align}(\tilde{\mathbf{z}}_{t-1}'|\mathbf{z}_{t}')\)
			\STATE  \(\mu _{\theta }(\tilde{\mathbf{z}}_{t-1}',t,\mathbf{c},\mathbf{p})\gets \mathrm{Align}(\mu _{\theta }(\tilde{\mathbf{z}}_{t-1}',t,\mathbf{c},\mathbf{p})|\mu _{\theta }(\mathbf{z}_{t}',t,\mathbf{c},\mathbf{p})) \)
			\STATE  \(\mu _{\theta }(\tilde{\mathbf{z}}_{t-1}',t,\mathbf{\emptyset},\mathbf{p})			
			\gets \mathrm{Align}(\mu _{\theta }(\tilde{\mathbf{z}}_{t-1}',t,\mathbf{\emptyset},\mathbf{p})|\mu _{\theta }(\mathbf{z}_{t}',t,\mathbf{\emptyset},\mathbf{p})) \)
			\STATE  Calculate \(\varepsilon _{\theta,2 } ( \tilde{\mathbf{z}}_{t-1}',t,\mathbf{c},\mathbf{p} )\) by Eq. \eqref{17};
			\STATE  Calculate \( \tilde{\mathbf{z}}_{t}\) by Eq. \eqref{3} or Eq. \eqref{14};
			\STATE  \(\tilde{\mathbf{z}}_{t}\gets \mathrm{Align}(\tilde{\mathbf{z}}_{t}|\mathbf{z}_{t})\)
			\ENDFOR		
			\RETURN \(\mathbf{z} _{T}\), \(\tilde{\mathbf{z}}_{T}\)
		\end{algorithmic}
	\end{algorithm}
   	\subsubsection{Data Filtering}
   	\indent Although we have successfully collected pairs of random noise and high-quality noise using the collection process based on diffusion model, not all high-quality noise produces satisfactory results. Therefore, it is necessary to filter the collected data. The filtering criterion is defined as \(s_{rd} > s_{hq} + m\), where \(m\) is the filtering threshold, and \(s_{rd}\) and \(s_{hq}\) are the Learned Perceptual Image Patch Similarity (LPIPS) \cite{zhang2018unreasonable} scores between the generated image \(\mathbf{P}_{pred}\) (from the random noise and the high-quality noise, respectively) and the corresponding ground-truth image \(\mathbf{P}_{gt}\), i.e., \(s=\frac{1}{N} \sum_{i=1}^{N} \mathrm{LPIPS}(\mathbf{P}_{pred}^{i} , \mathbf{P}_{gt}^{i}) \), where \(N\) denotes the number of images generated by NVS model. Noise pairs that satisfy this criterion are considered valuable and retained for further use.
   	\subsection{Encoder-Decoder Network (EDN)}
   	\subsubsection{Architecture}
   	As shown in Fig. \ref{collect_dataset_process} Stage II, the input of the EDN consists of the VAE embedding \(\mathbf{I}\) of the reference image and the initial random noise \(\mathbf{z} _{T}\). The output is the corresponding image semantic information to be added to \(\mathbf{z} _{T}\). The EDN adopts an encoder-decoder architecture \cite{kumar2020fisheyedistancenet} based on the U-Net architecture \cite{ronneberger2015u} with skip connections. We use ResNet18 \cite{he2016deep} as the encoder to balance lightweight deployment with the preservation of detailed features. To maintain feature integrity after decoding, convolutional feature upsampling is performed using Pixel Shuffle \cite{shi2016real}. Additional details of the EDN are provided in the Supplementary Material.
   	
   	\subsubsection{Training Loss}
   	We use the Smooth L1 loss \cite{girshick2015fast} as the loss function for the network. During training, the loss is calculated using two quantities: the predicted image semantic information \(\mathbf{S}_{pred}\) from the EDN and the ground-truth (GT) semantic information \(\mathbf{S}_{gt}=\tilde{\mathbf{z}}_{T}-\mathbf{z} _{T}\), which represents the difference between the GT high-quality noise and the random noise. 
   	\subsubsection{Insert EDN into Pretrained NVS model}
   	As shown in Stage III of Fig. \ref{collect_dataset_process}, the input to the EDN consists of the VAE embedding \(\mathbf{I}\) of the reference image and the initial random noise \(\mathbf{z} _{T}\). Adding the EDN’s output to the initial noise produces the final high-quality noise \(\tilde{\mathbf{z}}_{T-pred}=\mathbf{z} _{T}+\mathbf{S}_{pred}\). This high-quality noise can directly replace the initial noise and be fed into the diffusion inference process.
   	\section{Experimental Results}
   	\label{Experimental_Results}
   	In this section, we evaluate the effectiveness, generalization and efficiency of our EDN. We conduct experiments across various datasets on two models, SV3D and Mv-Adapter. Specific details of both models are provided in the Supplementary Material. 
   	\subsection{Datasets and Evaluation Metrics}
   	\noindent \textbf{Metrics.}  We evaluate the differences between the generated images and the ground-truth images using three metrics: Peak Signal-to-Noise Ratio (PSNR) \cite{fardo2016formal}, Structural Similarity Index (SSIM) \cite{wang2004image}, and Learned Perceptual Image Patch Similarity (LPIPS) \cite{zhang2018unreasonable}.
   	
   	\noindent \textbf{Datasets.}  We use Objaverse \cite{deitke2023objaverse} as the training dataset to collect pairs of random noise and high-quality noise. All selected objects are rendered using Blender\footnote{Available at: http://www.blender.org} under consistent lighting. For SV3D, we randomly sample 1,765 3D objects and render 21 views per object. The camera azimuth angles are uniformly distributed from \(0^{\circ} \) to \(360^{\circ} \), while the elevation angle for each object remains fixed. The evaluation starts at \(-24^{\circ} \) and increases by \(3^{\circ} \) for every 100 objects. For Mv-Adapter, we randomly sample 1,752 3D objects and render 6 views per object. All objects use uniformly sampled azimuth angles and a fixed elevation angle of \(0^{\circ} \). Each object is assigned a unique random seed (1--1765 for SV3D and 1--1752 for Mv-Adapter). For both models, the reference image used during noise-pair generation is the view rendered at an azimuth and elevation of \(0^{\circ} \). After data filtering, 359 noise pairs are retained for SV3D and 638 for Mv-Adapter.
   	
   	We use the Google Scanned Objects (GSO) dataset \cite{downs2022google}, Objaverse, and OmniObject3D \cite{wu2023omniobject3d} as test datasets. From each dataset, we randomly select and render 100 objects. We evaluate the NVS model under two camera-pose orbits: static and dynamic. In the static orbit, each object is rendered with azimuth angles uniformly distributed from \(0^{\circ} \) to \(360^{\circ} \), while the elevation angle is fixed at \(0^{\circ} \). In the dynamic orbit, the azimuth angles follow the same uniform distribution, but the elevation angles vary in a sine-like pattern, from \(20^{\circ} \) to \(-20^{\circ} \). Unless otherwise specified, we use a random seed of 23 for static-orbit experiments and 801-900 for dynamic-orbit experiments.
   	\subsection{Implementation Details}
   	All experiments are conducted on an NVIDIA RTX 3090 GPU (24GB). For SV3D, the inference CFG scale \(\gamma_{1}\) varies from 6.0 to 2.5 (decreasing linearly from the front view to the back view and then increasing back to 6.0) to balance the triangular wave–shaped CFG scaling strategy \cite{voleti2024sv3d}, while the inversion CFG scale \(\gamma_{2}\) is set to 0.0. The number of inference-inversion steps \(n\) is set to 16. For Mv-Adapter, \(\gamma_{1}=13.0\), \(\gamma_{2}=0.0\), and \(n=25\). During data filtering, both models use a filtering threshold of 0.0. For EDN training, the Adam optimizer \cite{adam2014method} is used with an initial learning rate of 0.0003, a batch size of 8, and a learning rate decay factor of 0.8 every 200 epochs. The total training duration is 600 epochs. 	
   	\subsection{Comparison} 
\begin{figure*}[htbp]
    	\centering
    	\fontsize{9pt}{12pt}\selectfont 
    	\setlength{\tabcolsep}{0pt}    
    	\begin{tabular}{c c c c c c c c c c}
    		\renewcommand{\arraystretch}{0}
    		\makecell{Input\\image} &
    		GSO &
    		\begin{subfigure}[c]{0.104\textwidth}  
    			\centering
    			\includegraphics[width=\linewidth]{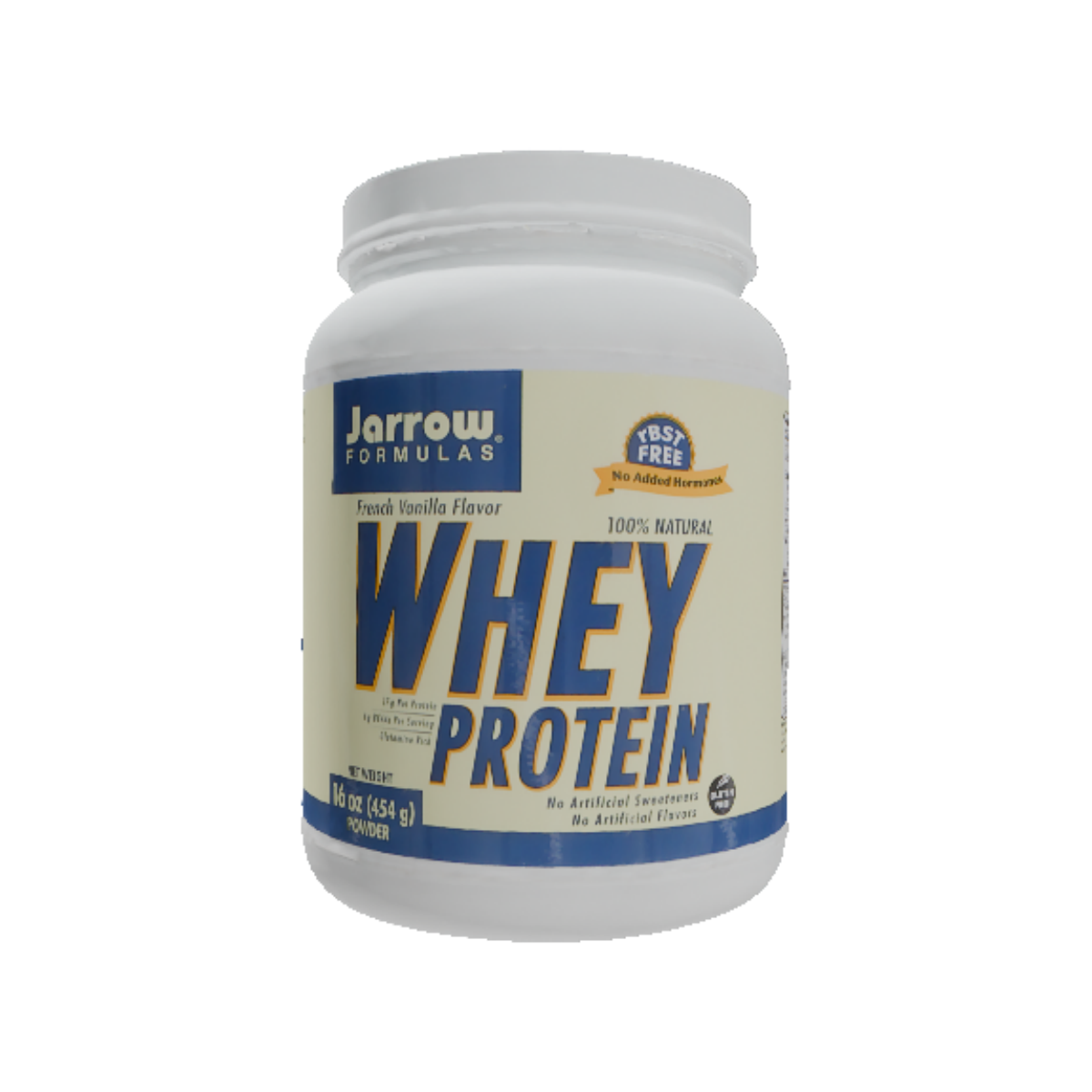}
    		\end{subfigure} &
    		&
    		Objaverse &
    		\begin{subfigure}[c]{0.104\textwidth}  
    			\centering
    			\includegraphics[width=\linewidth]{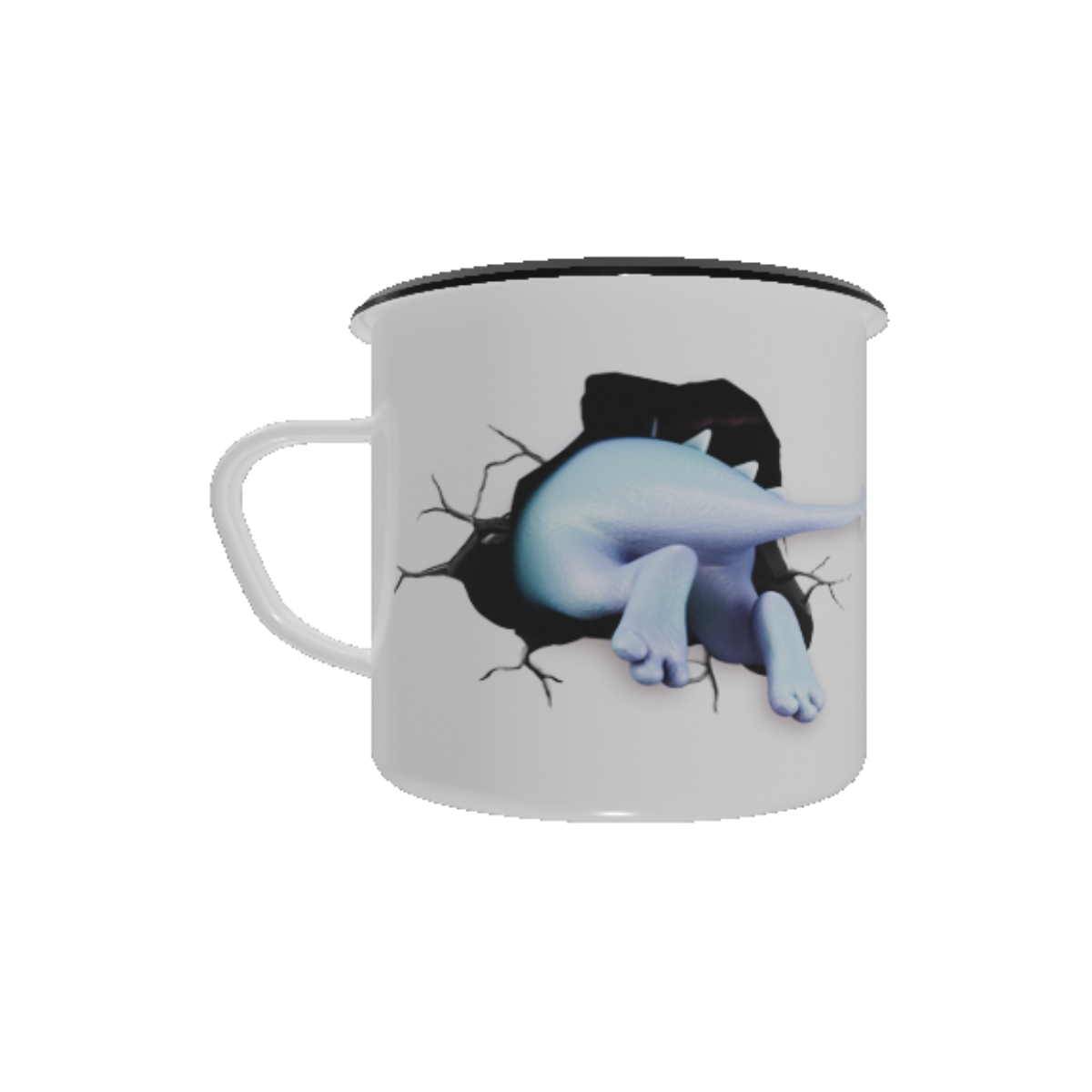}
    		\end{subfigure} &
    		&
    		\makecell{Omni\\Object3D} &
    		\begin{subfigure}[c]{0.104\textwidth}  
    			\centering
    			\includegraphics[width=\linewidth]{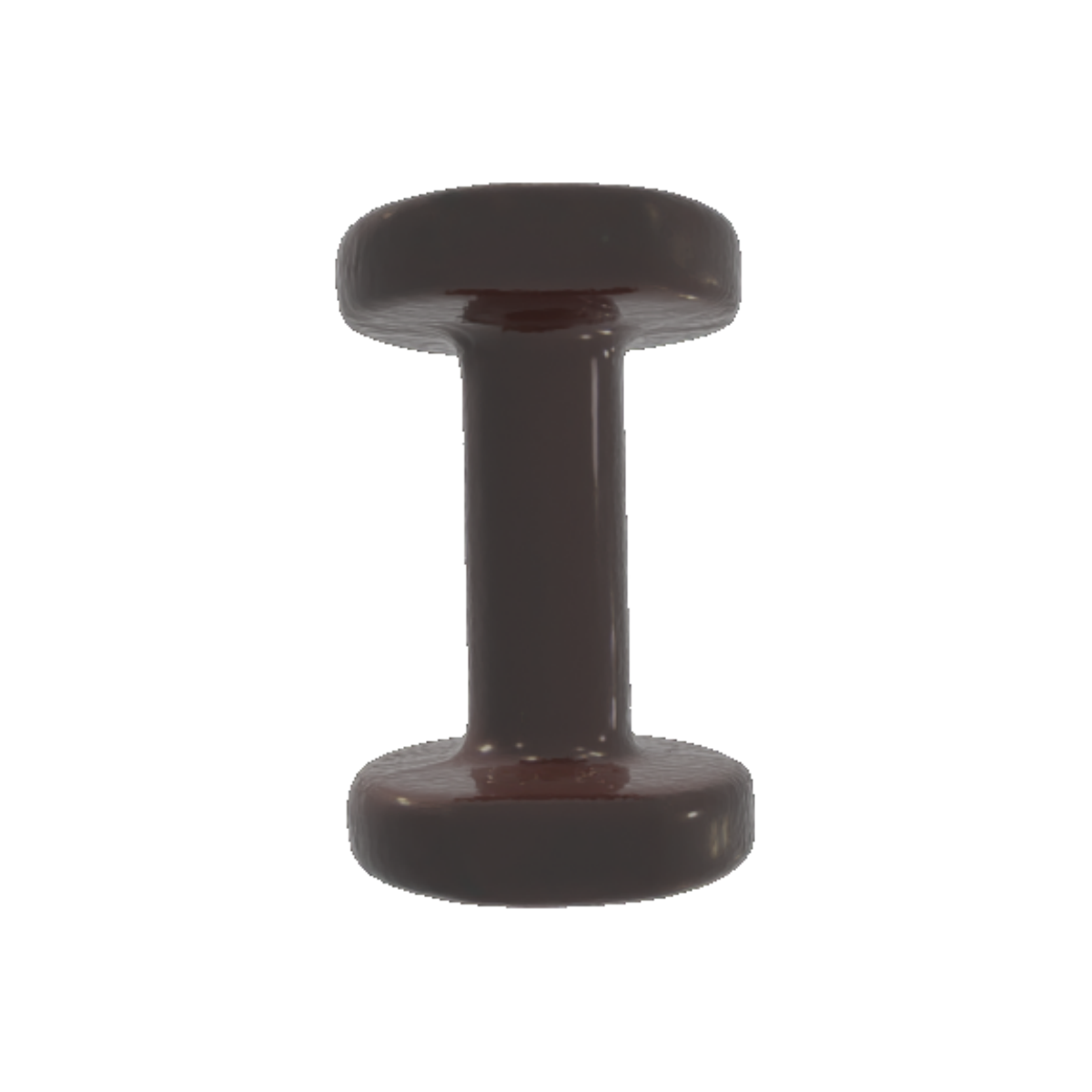}
    		\end{subfigure} &\\
    		GT &
    		\begin{subfigure}[c]{0.104\textwidth}  
    			\centering
    			\includegraphics[width=\linewidth]{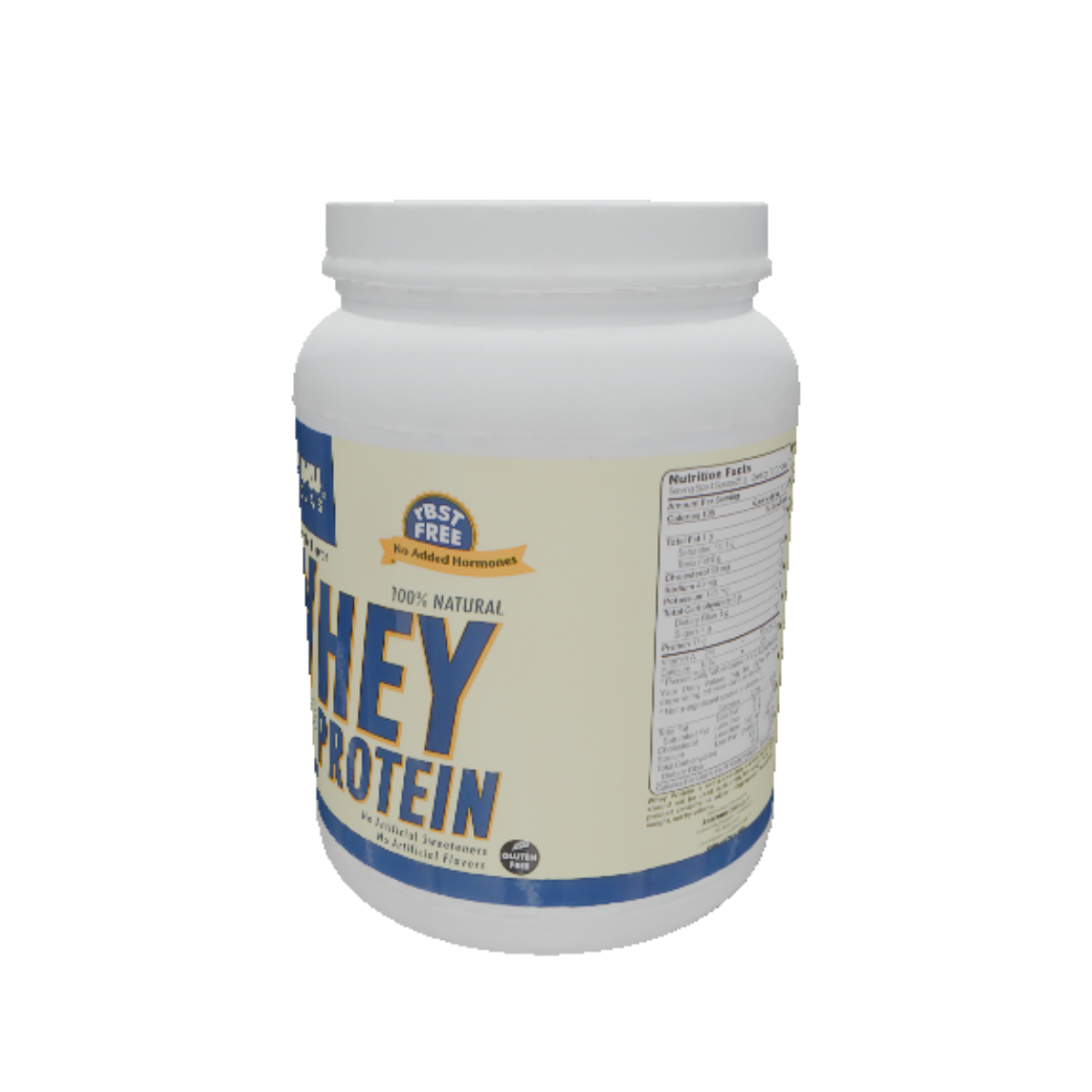}
    		\end{subfigure} &
    		\begin{subfigure}[c]{0.104\textwidth}  
    			\centering
    			\includegraphics[width=\linewidth]{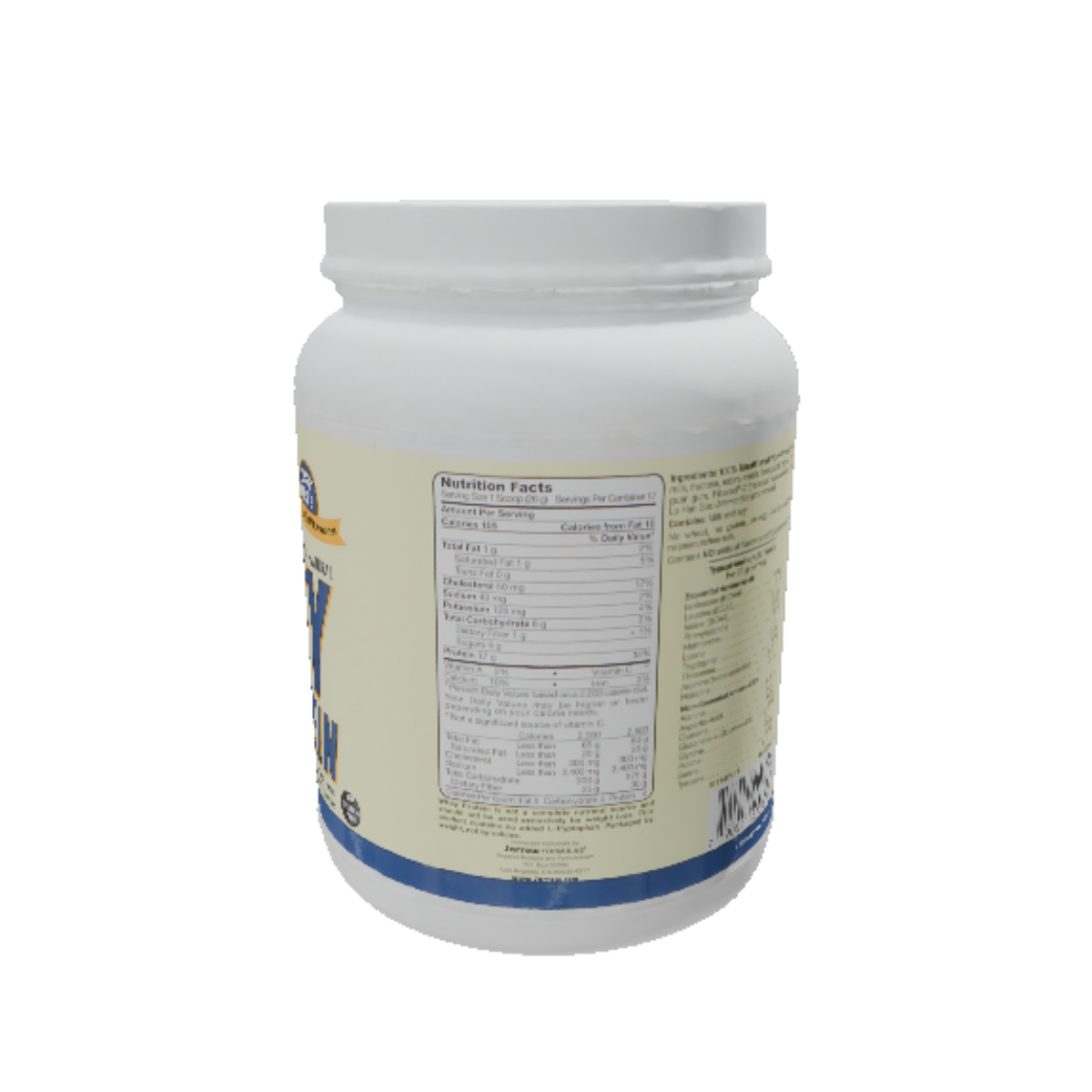}
    		\end{subfigure} &
    		\begin{subfigure}[c]{0.104\textwidth}  
    			\centering
    			\includegraphics[width=\linewidth]{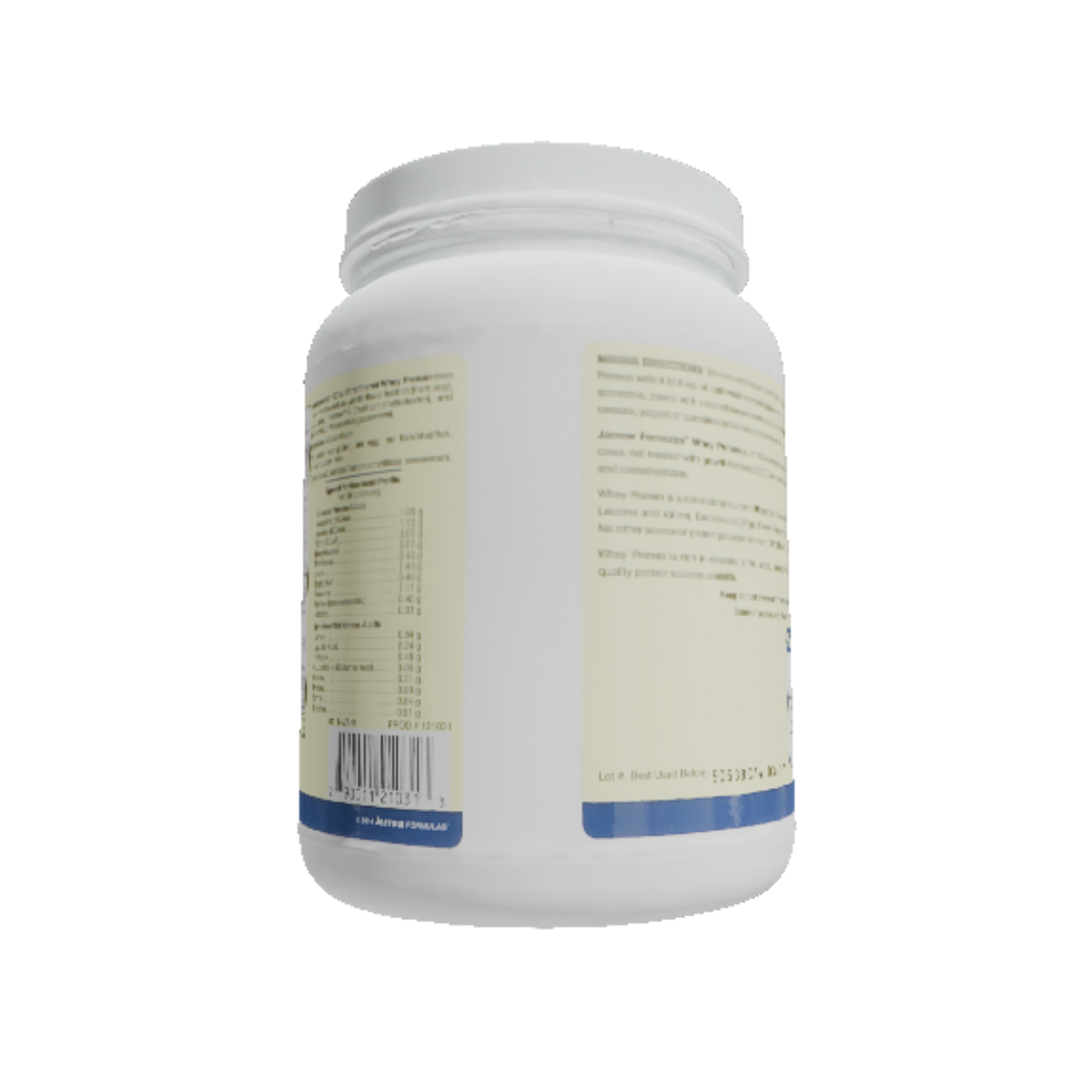}
    		\end{subfigure} &
    		\begin{subfigure}[c]{0.104\textwidth}  
    			\centering
    			\includegraphics[width=\linewidth]{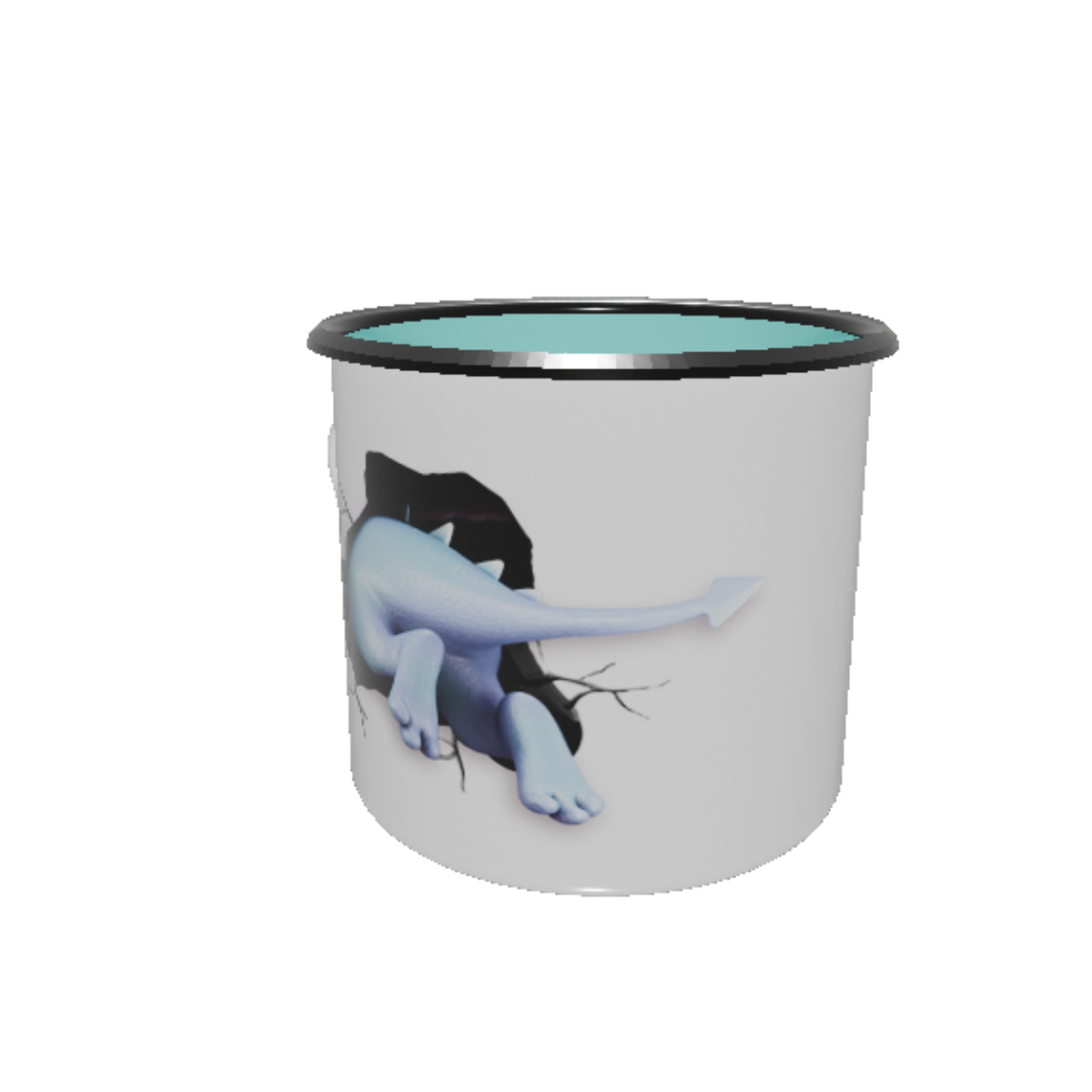}
    		\end{subfigure} &
    		\begin{subfigure}[c]{0.104\textwidth}  
    			\centering
    			\includegraphics[width=\linewidth]{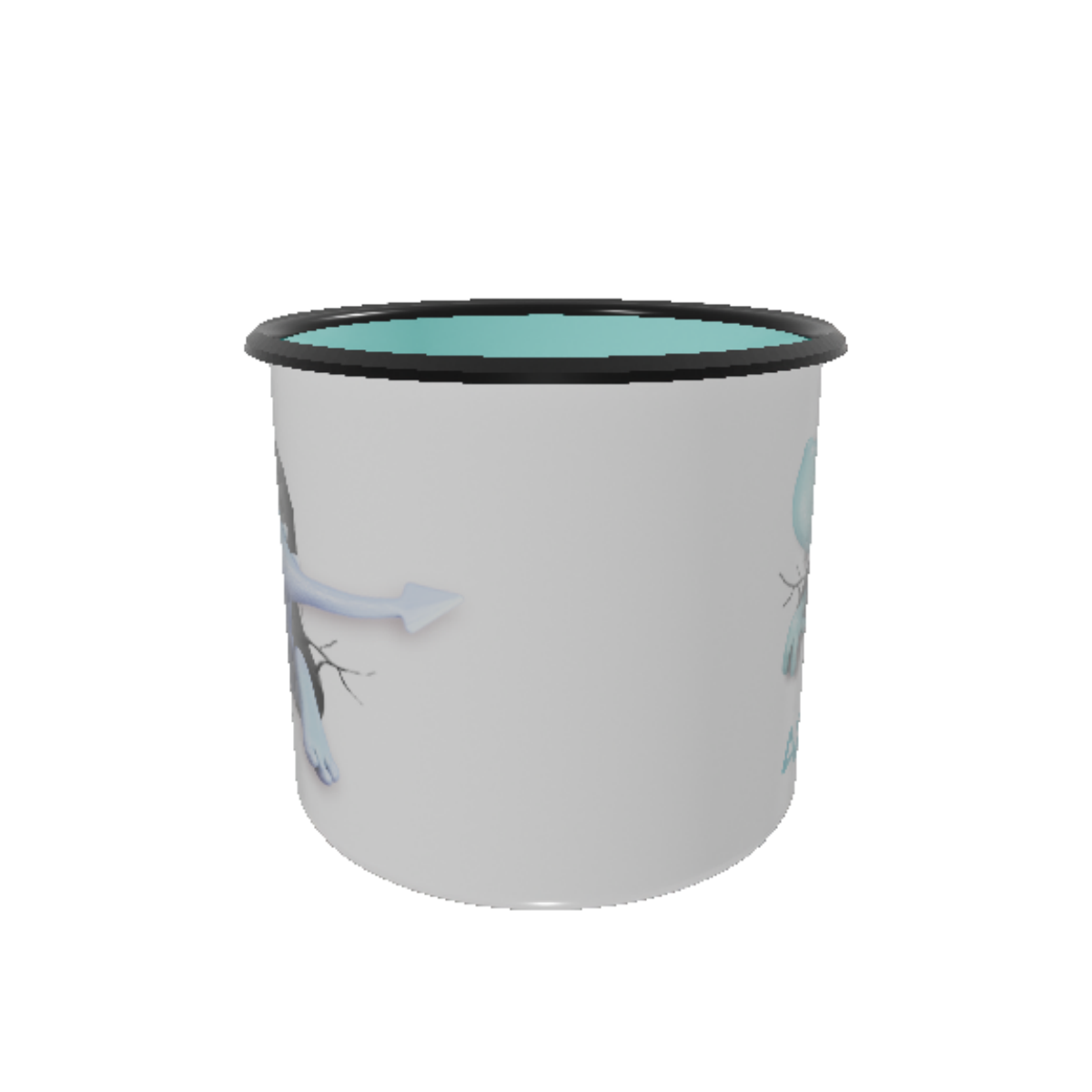}
    		\end{subfigure} &
    		\begin{subfigure}[c]{0.104\textwidth} 
    			\centering
    			\includegraphics[width=\linewidth]{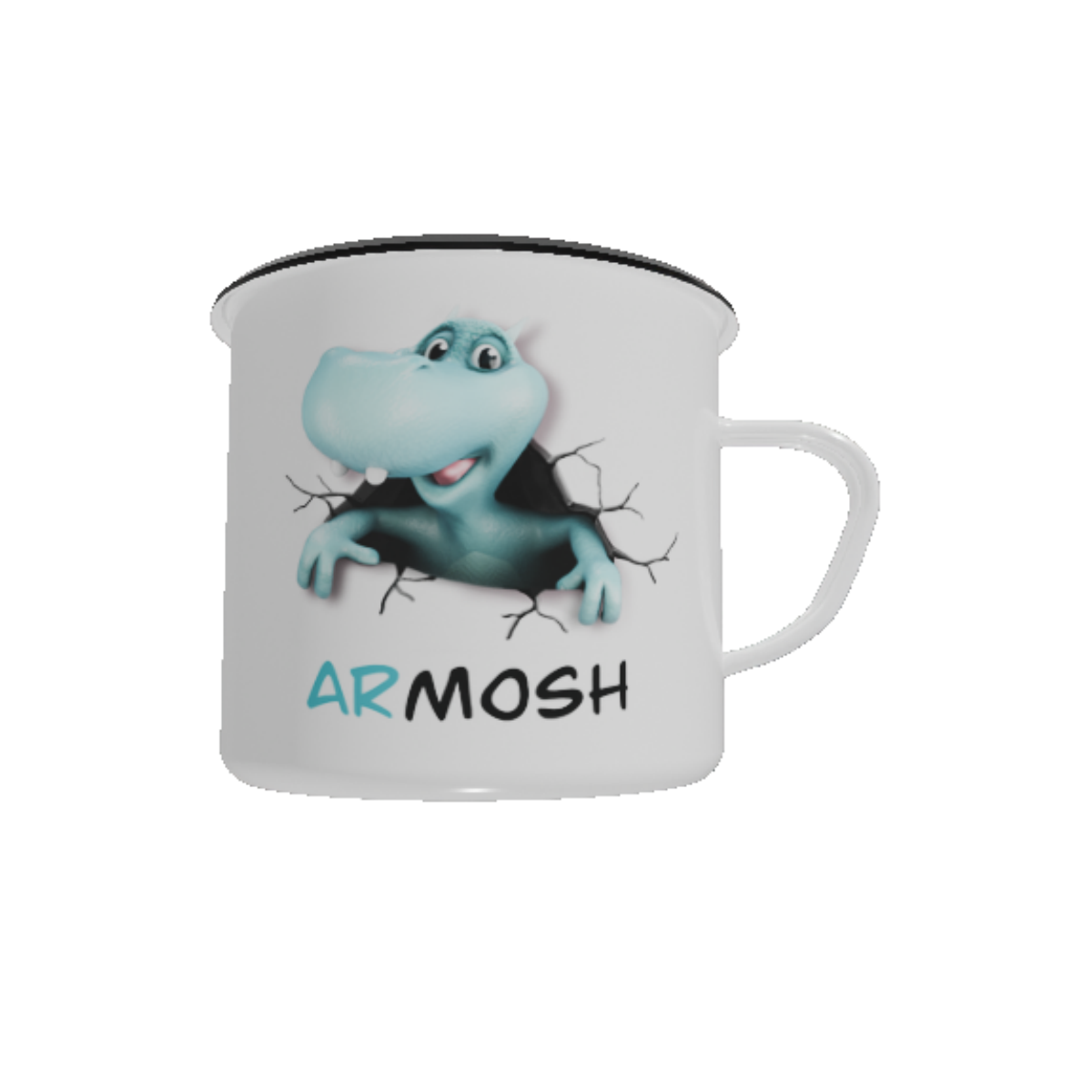}
    		\end{subfigure} &
    		\begin{subfigure}[c]{0.104\textwidth}  
    			\centering
    			\includegraphics[width=\linewidth]{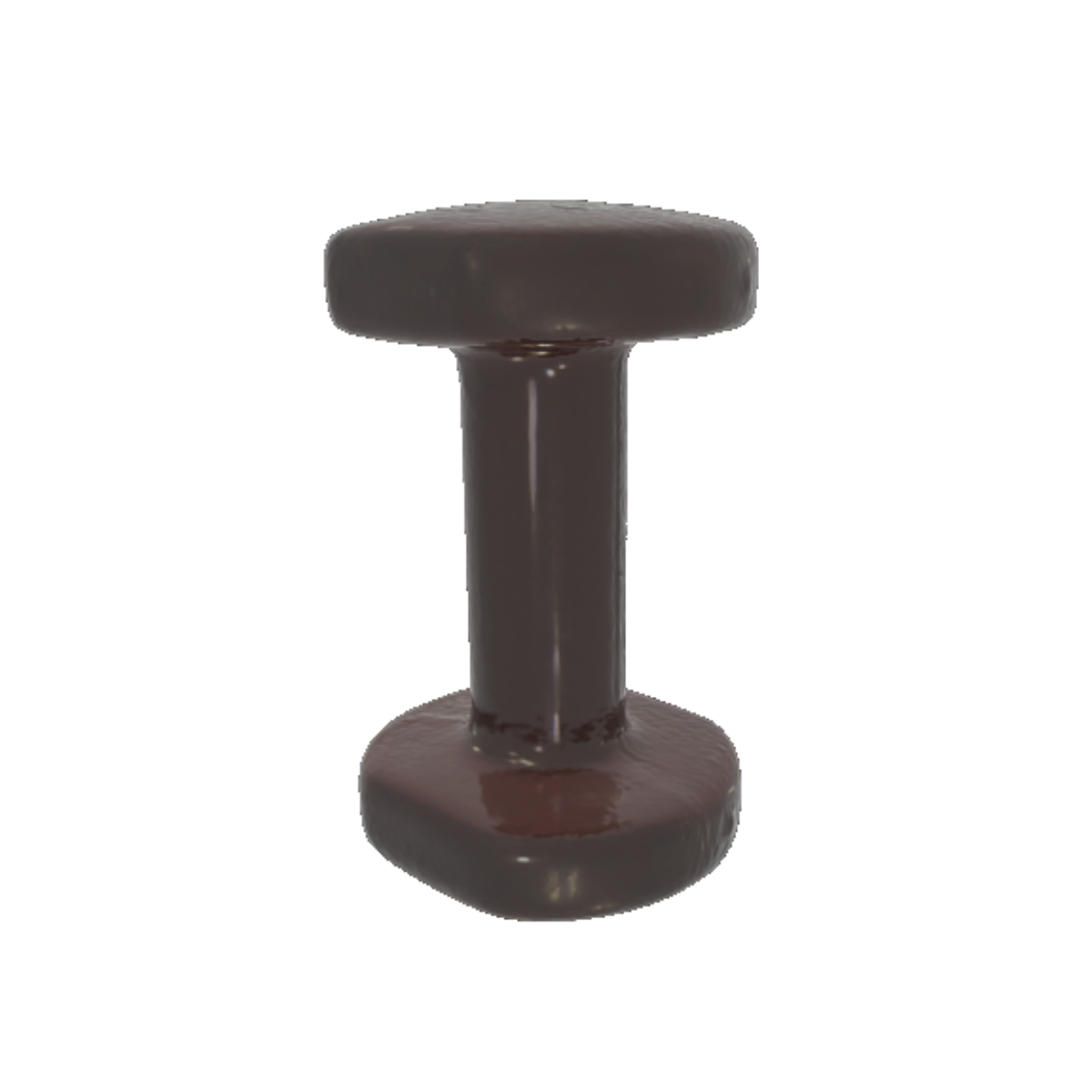}
    		\end{subfigure} &
    		\begin{subfigure}[c]{0.104\textwidth}  
    			\centering
    			\includegraphics[width=\linewidth]{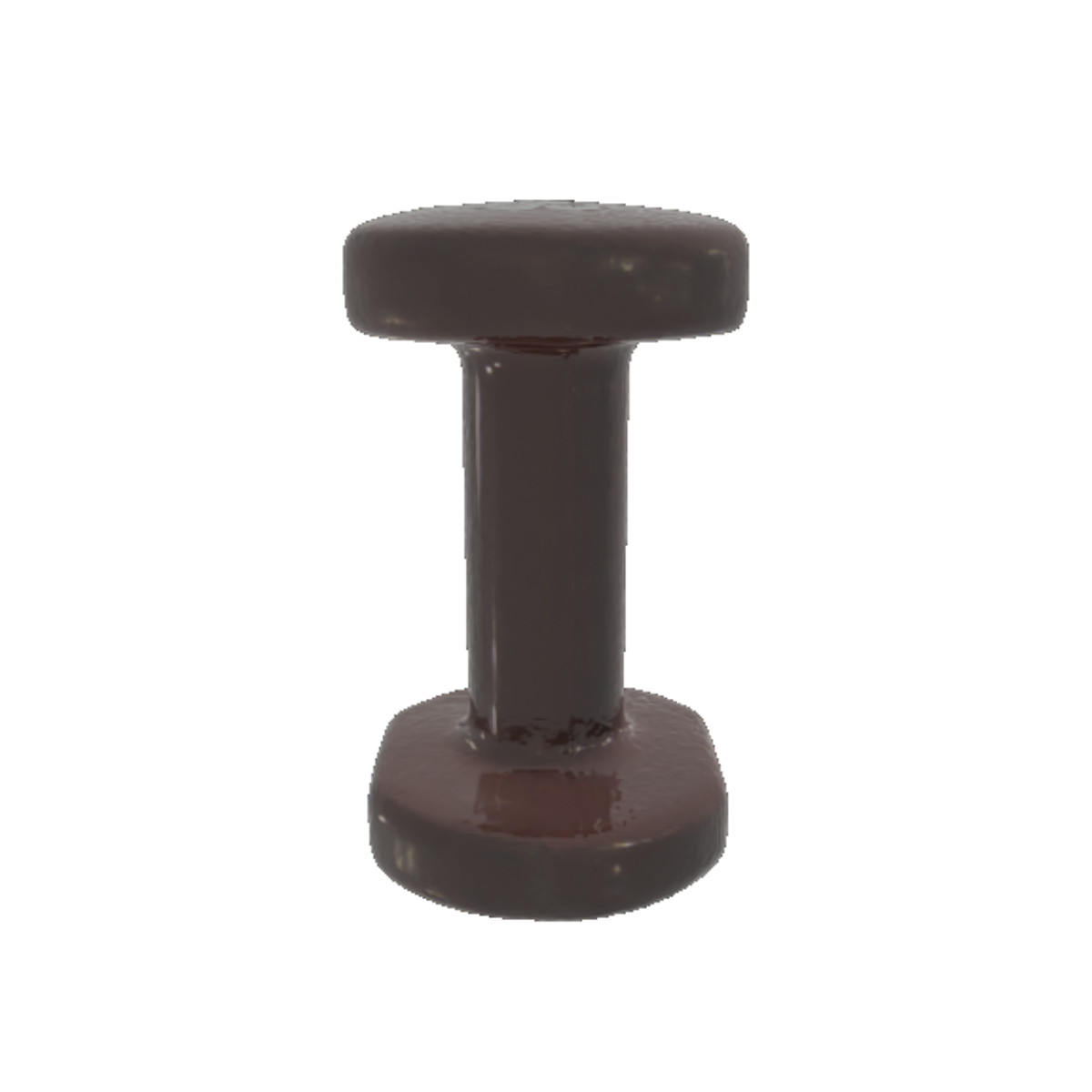}
    		\end{subfigure}&
    		\begin{subfigure}[c]{0.104\textwidth}  
    			\centering
    			\includegraphics[width=\linewidth]{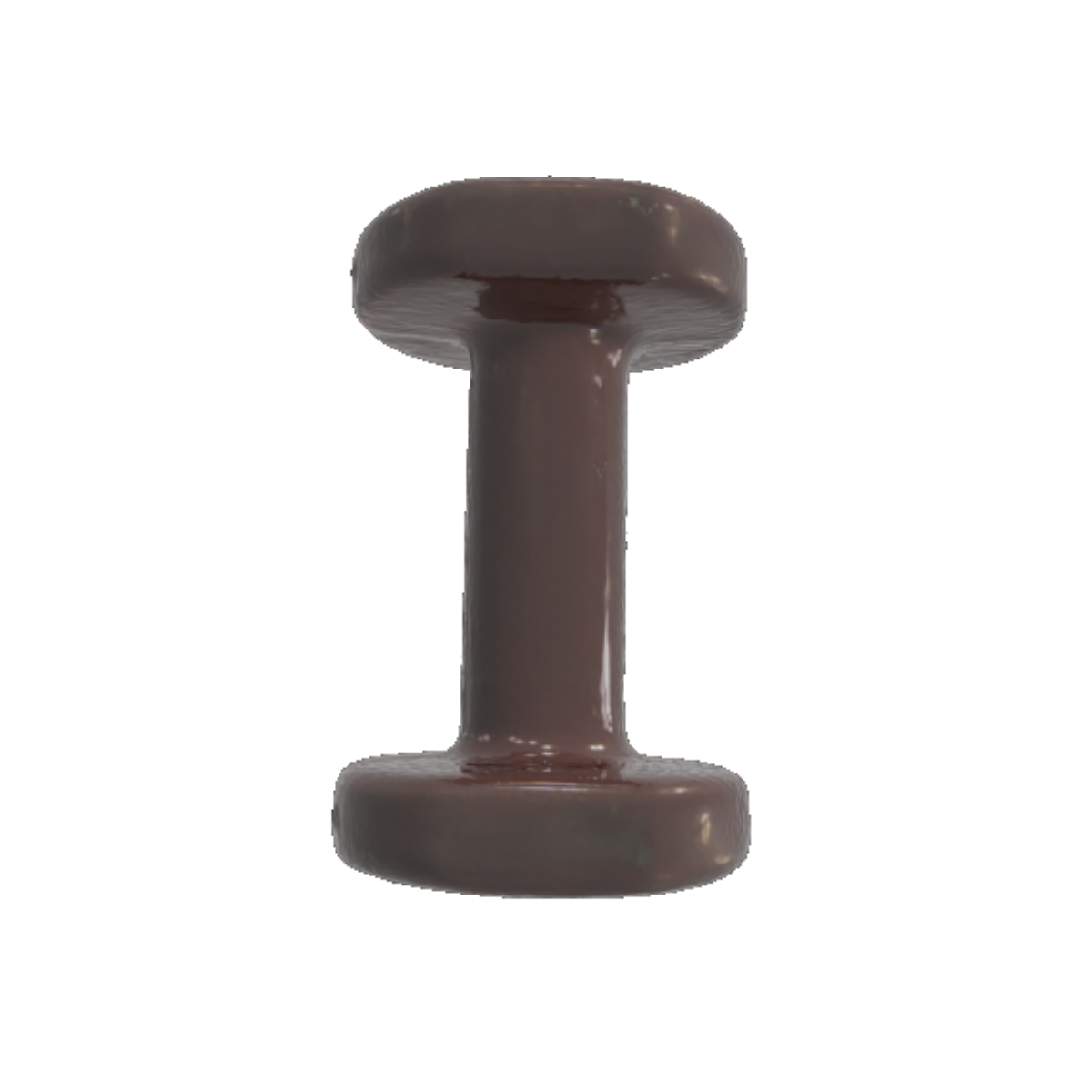}
    		\end{subfigure} \\
    		SV3D &
    		\begin{subfigure}[c]{0.104\textwidth}  
    			\centering
    			\includegraphics[width=\linewidth]{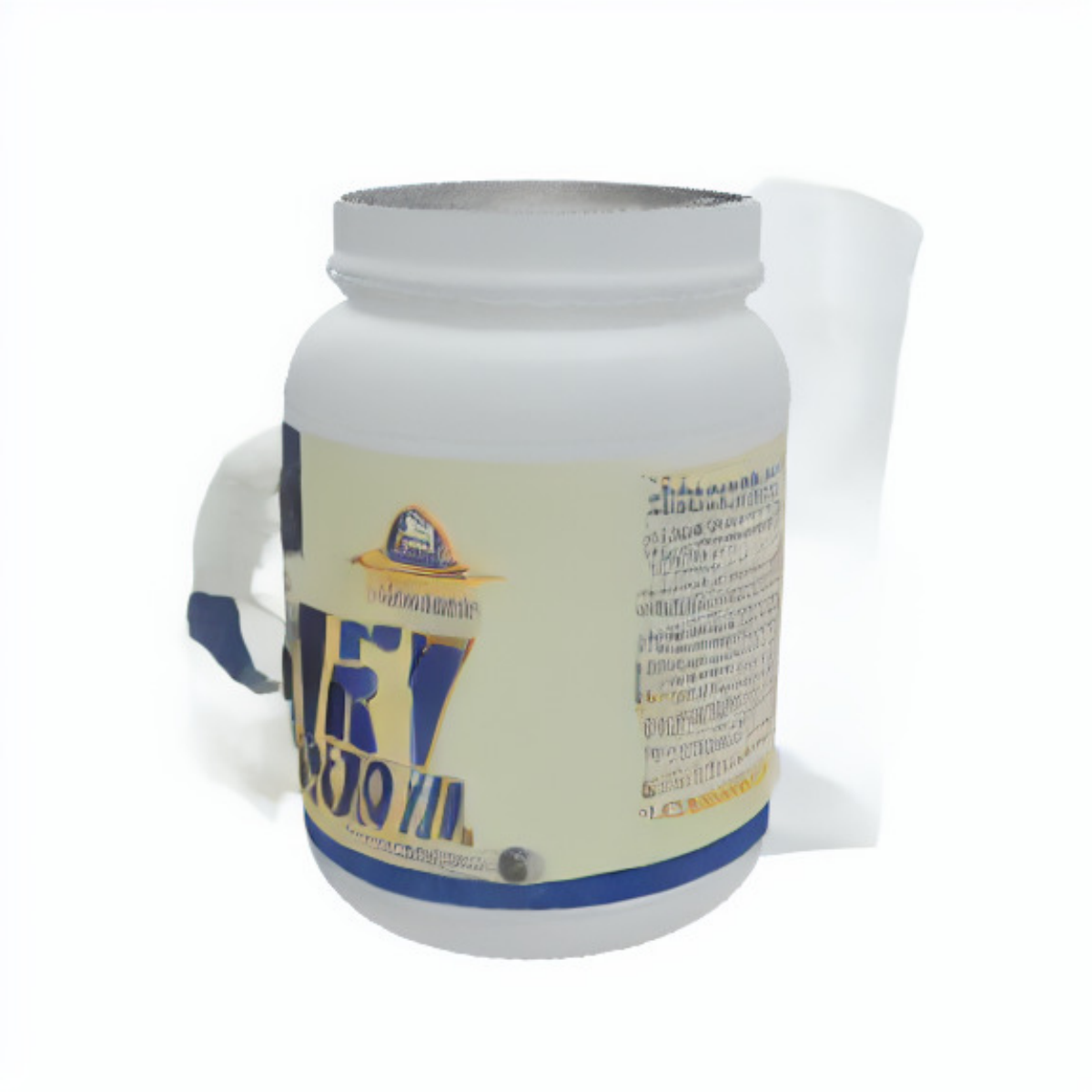}
    		\end{subfigure} &
    		\begin{subfigure}[c]{0.104\textwidth}  
    			\centering
    			\includegraphics[width=\linewidth]{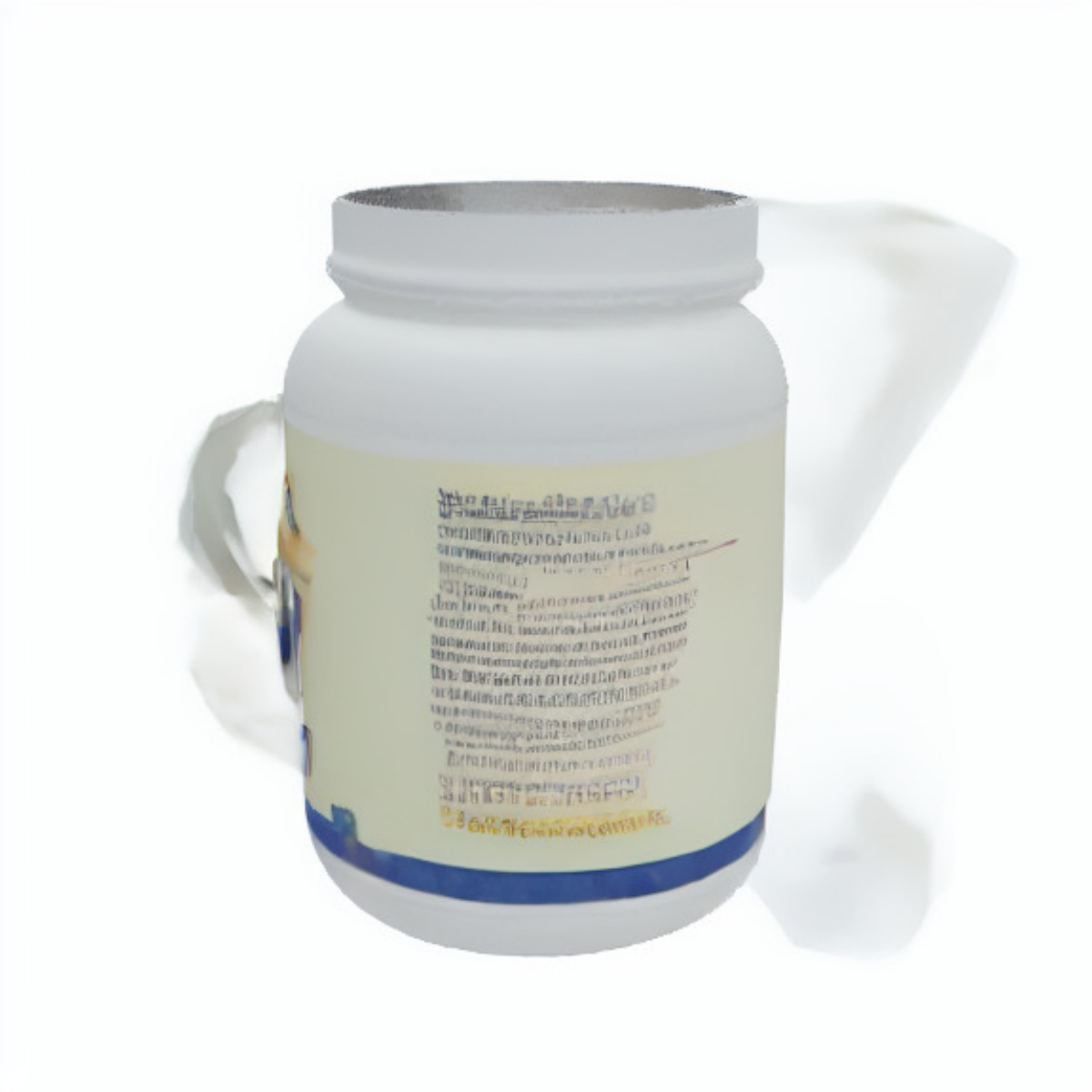}
    		\end{subfigure} &
    		\begin{subfigure}[c]{0.104\textwidth}  
    			\centering
    			\includegraphics[width=\linewidth]{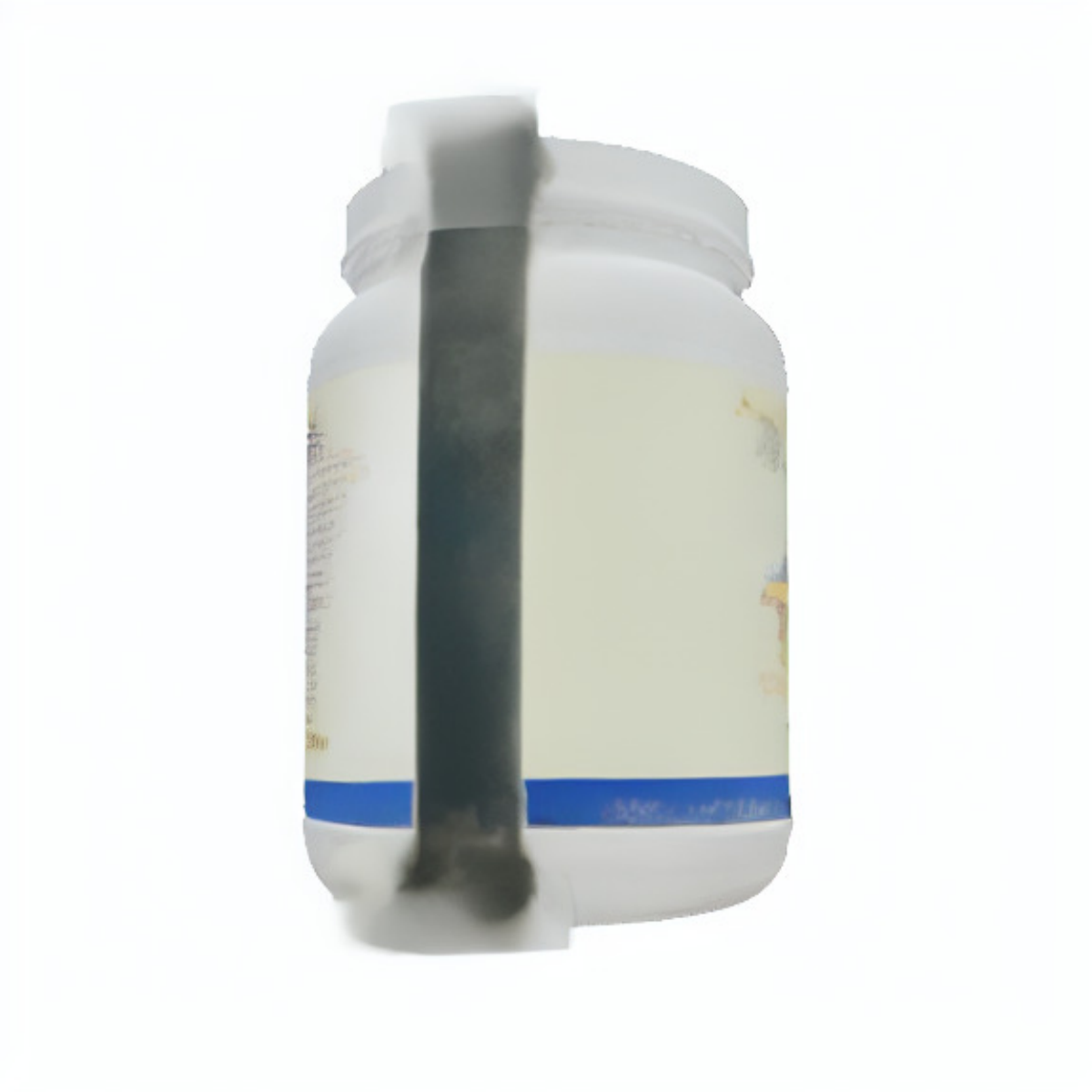}
    		\end{subfigure} &
    		\begin{subfigure}[c]{0.104\textwidth}  
    			\centering
    			\includegraphics[width=\linewidth]{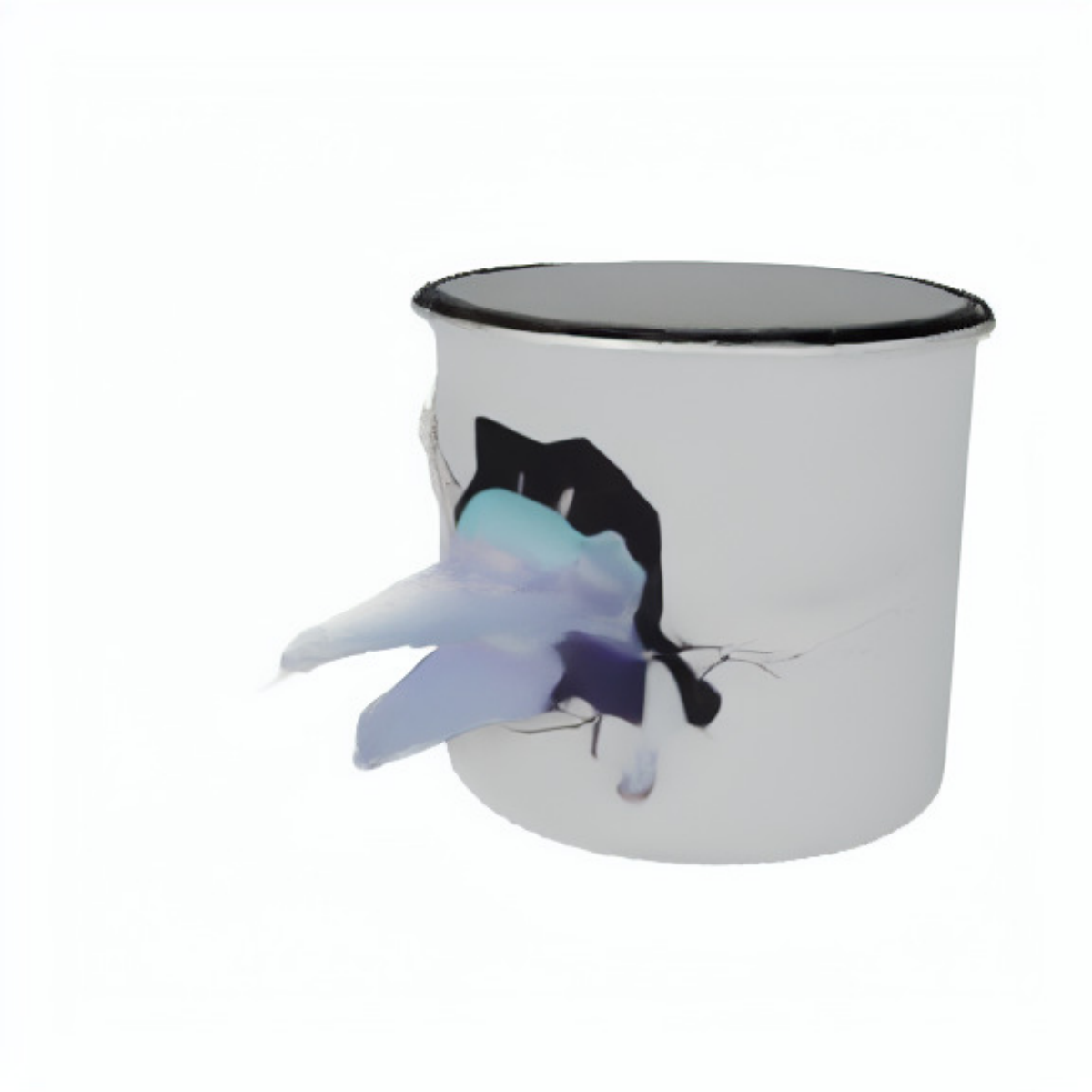}
    		\end{subfigure} &
    		\begin{subfigure}[c]{0.104\textwidth}  
    			\centering
    			\includegraphics[width=\linewidth]{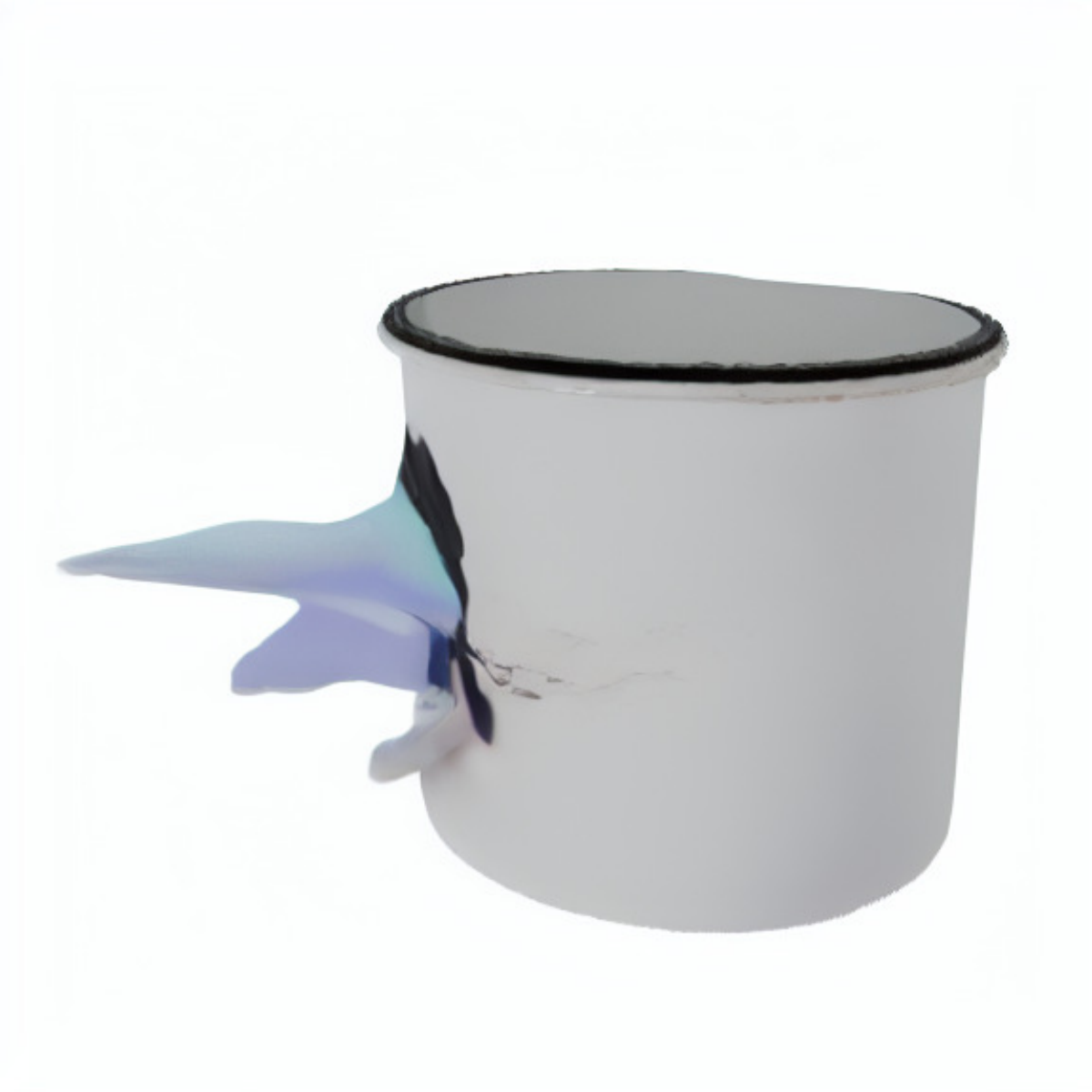}
    		\end{subfigure} &
    		\begin{subfigure}[c]{0.104\textwidth} 
    			\centering
    			\includegraphics[width=\linewidth]{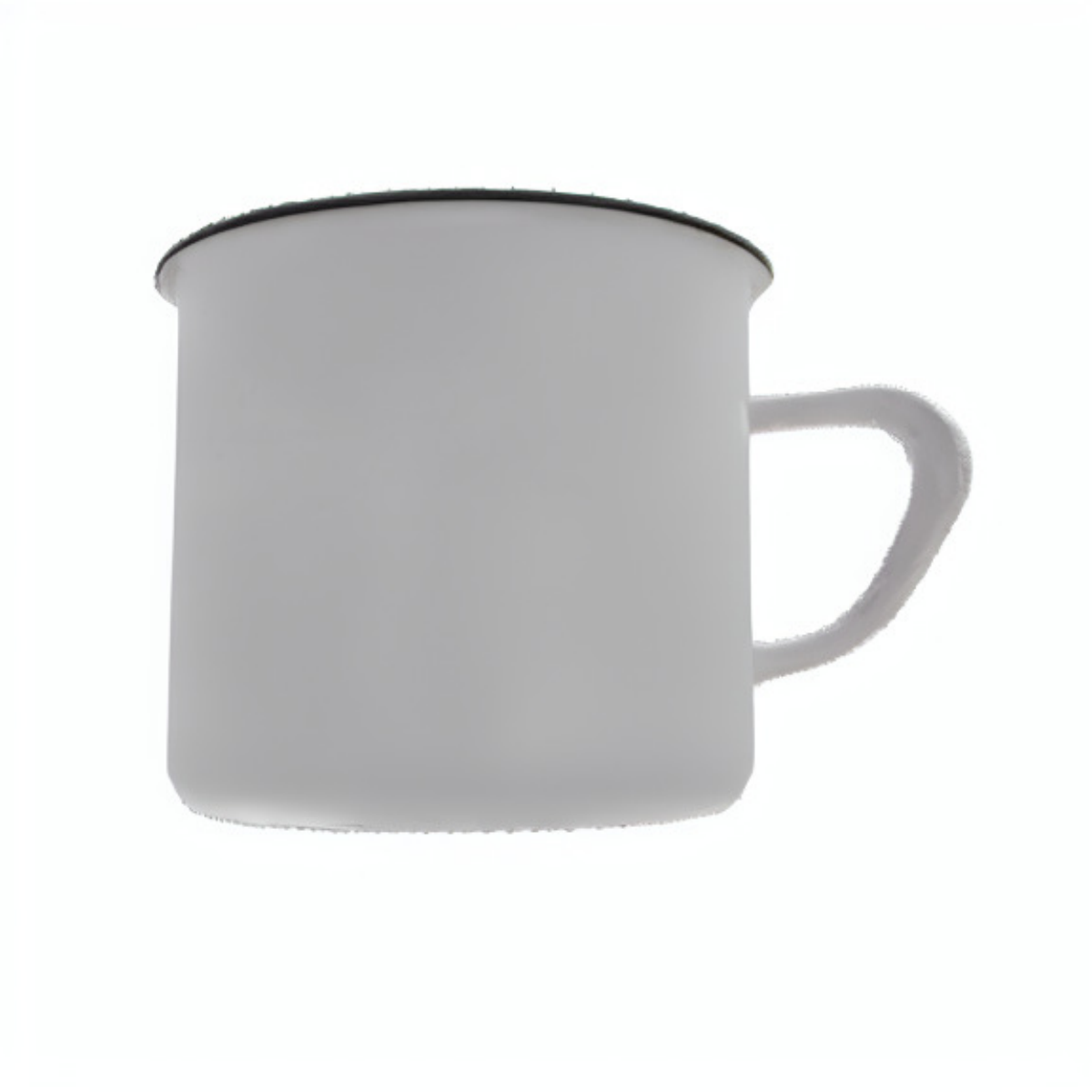}
    		\end{subfigure} &
    		\begin{subfigure}[c]{0.104\textwidth}  
    			\centering
    			\includegraphics[width=\linewidth]{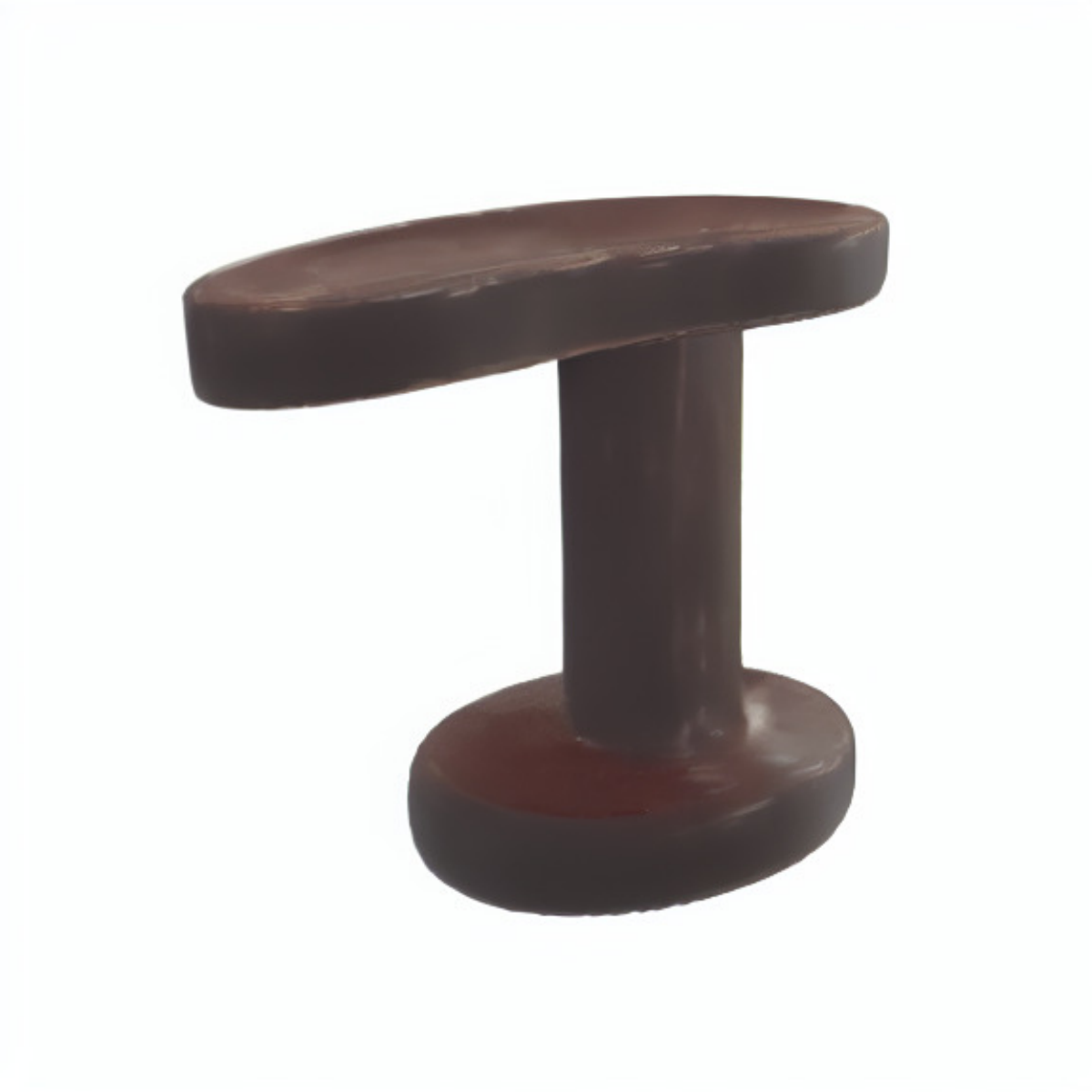}
    		\end{subfigure} &
    		\begin{subfigure}[c]{0.104\textwidth}  
    			\centering
    			\includegraphics[width=\linewidth]{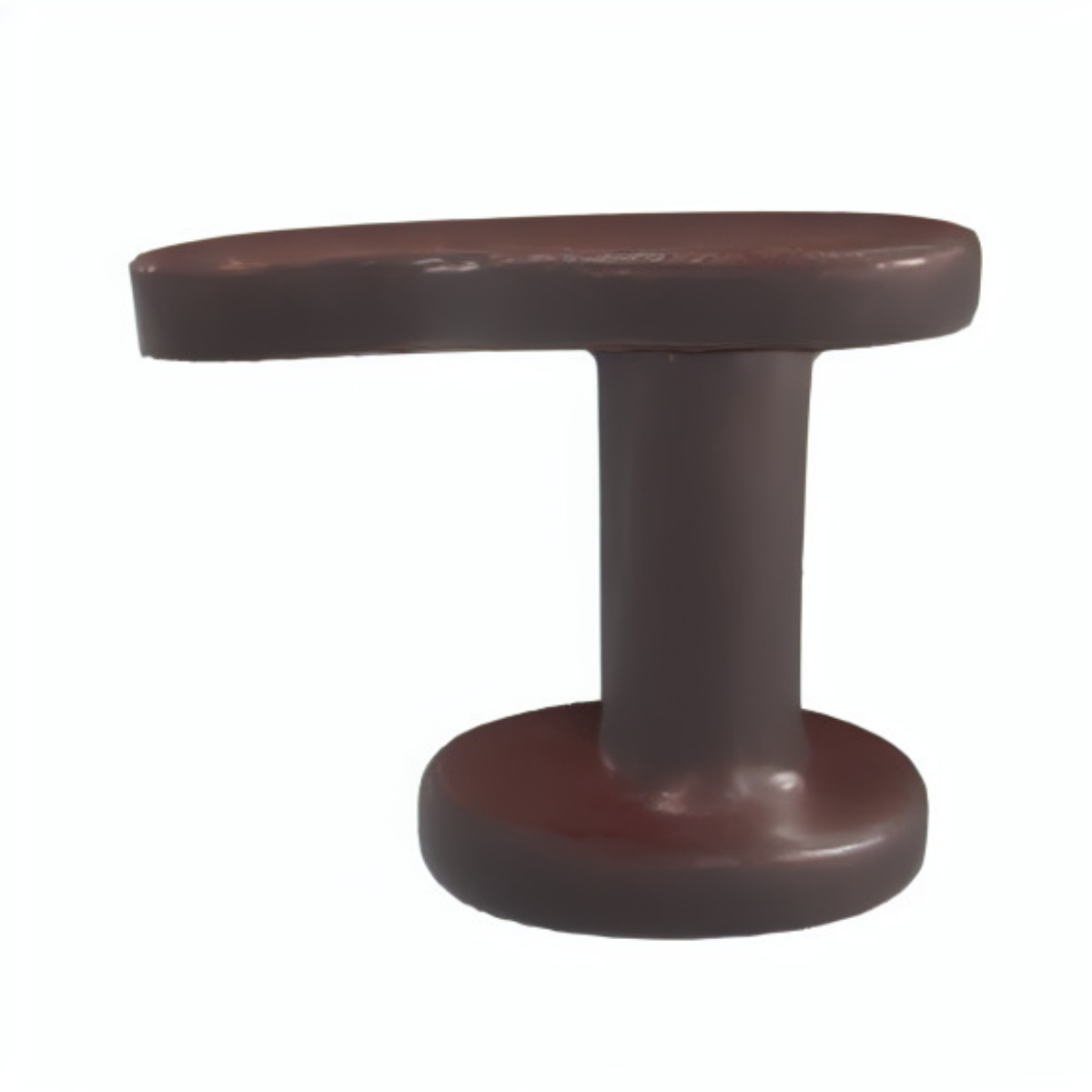}
    		\end{subfigure}&
    		\begin{subfigure}[c]{0.104\textwidth}  
    			\centering
    			\includegraphics[width=\linewidth]{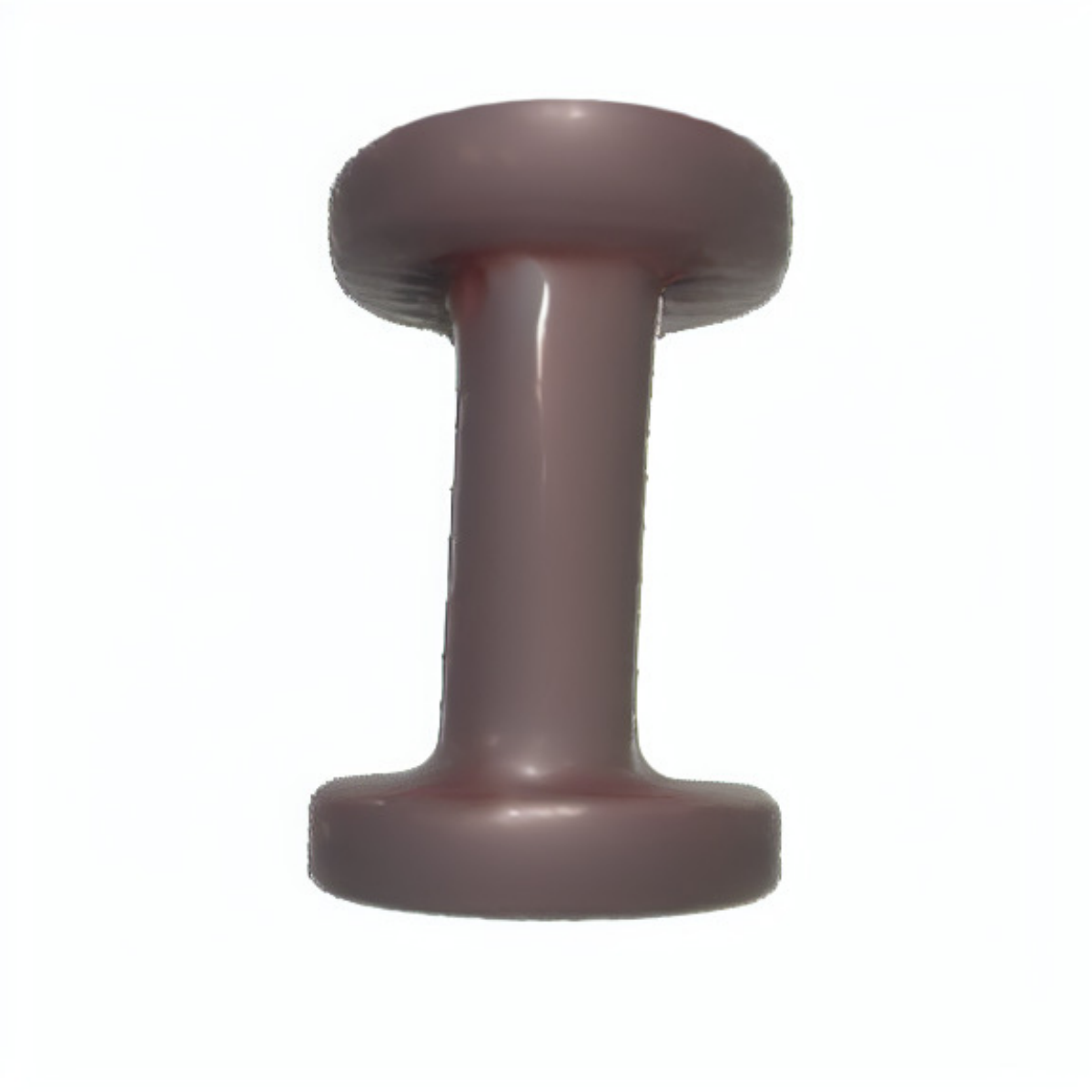}
    		\end{subfigure} \\
    		\makecell{SV3D\\\textbf{with}\\\textbf{EDN}} &
    		\begin{subfigure}[c]{0.104\textwidth}  
    			\centering
    			\includegraphics[width=\linewidth]{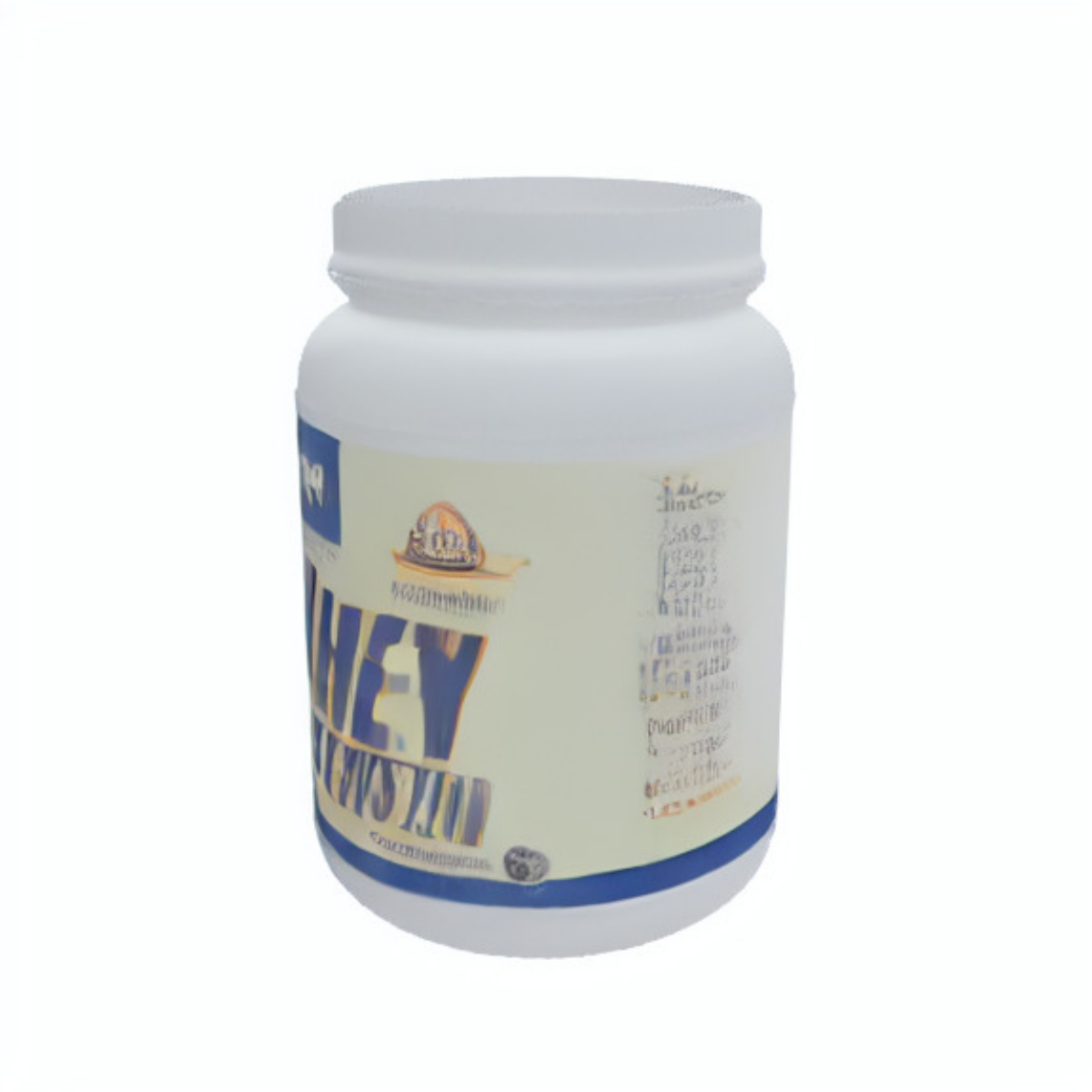}
    		\end{subfigure} &
    		\begin{subfigure}[c]{0.104\textwidth}  
    			\centering
    			\includegraphics[width=\linewidth]{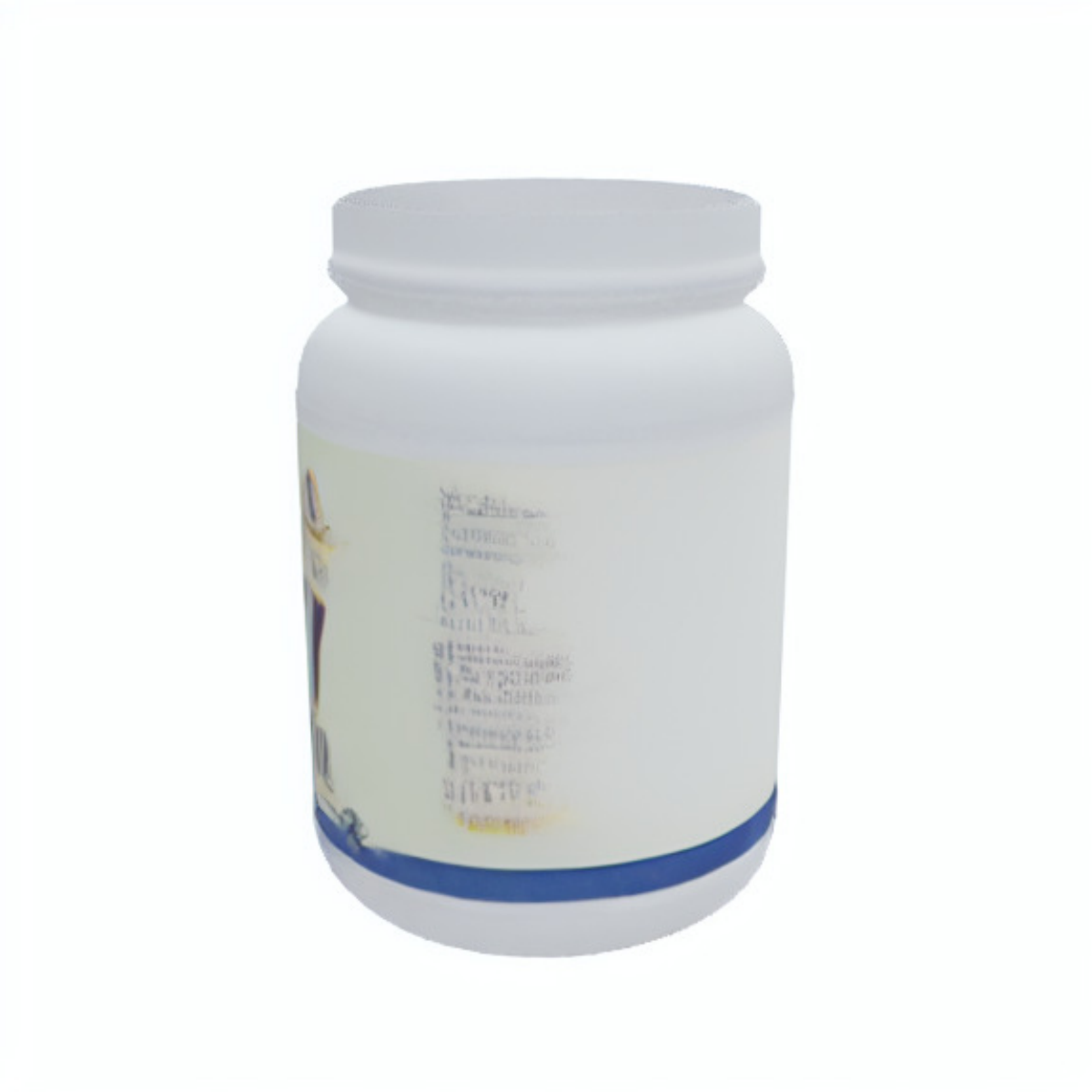}
    		\end{subfigure} &
    		\begin{subfigure}[c]{0.104\textwidth}  
    			\centering
    			\includegraphics[width=\linewidth]{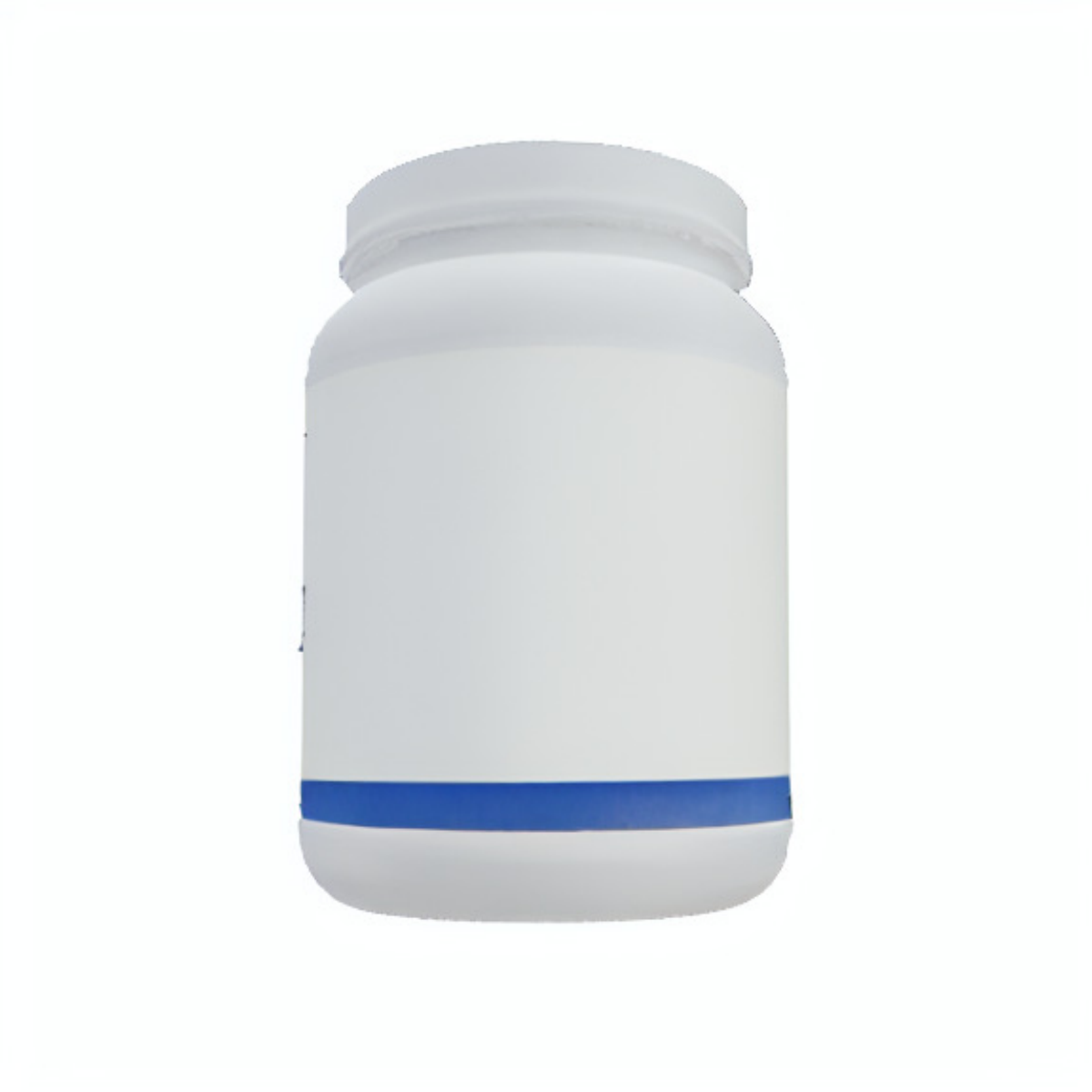}
    		\end{subfigure} &
    		\begin{subfigure}[c]{0.104\textwidth}  
    			\centering
    			\includegraphics[width=\linewidth]{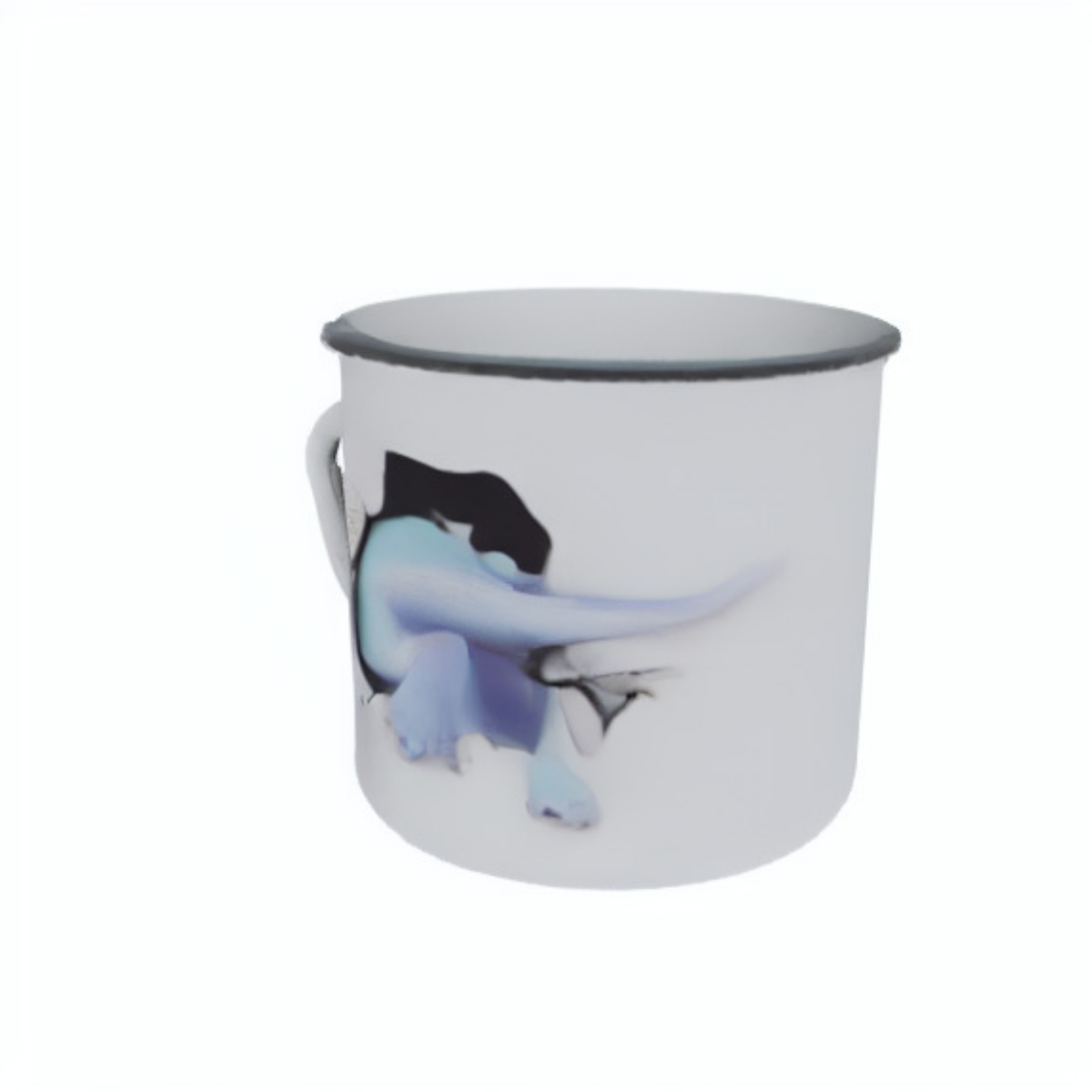}
    		\end{subfigure} &
    		\begin{subfigure}[c]{0.104\textwidth}  
    			\centering
    			\includegraphics[width=\linewidth]{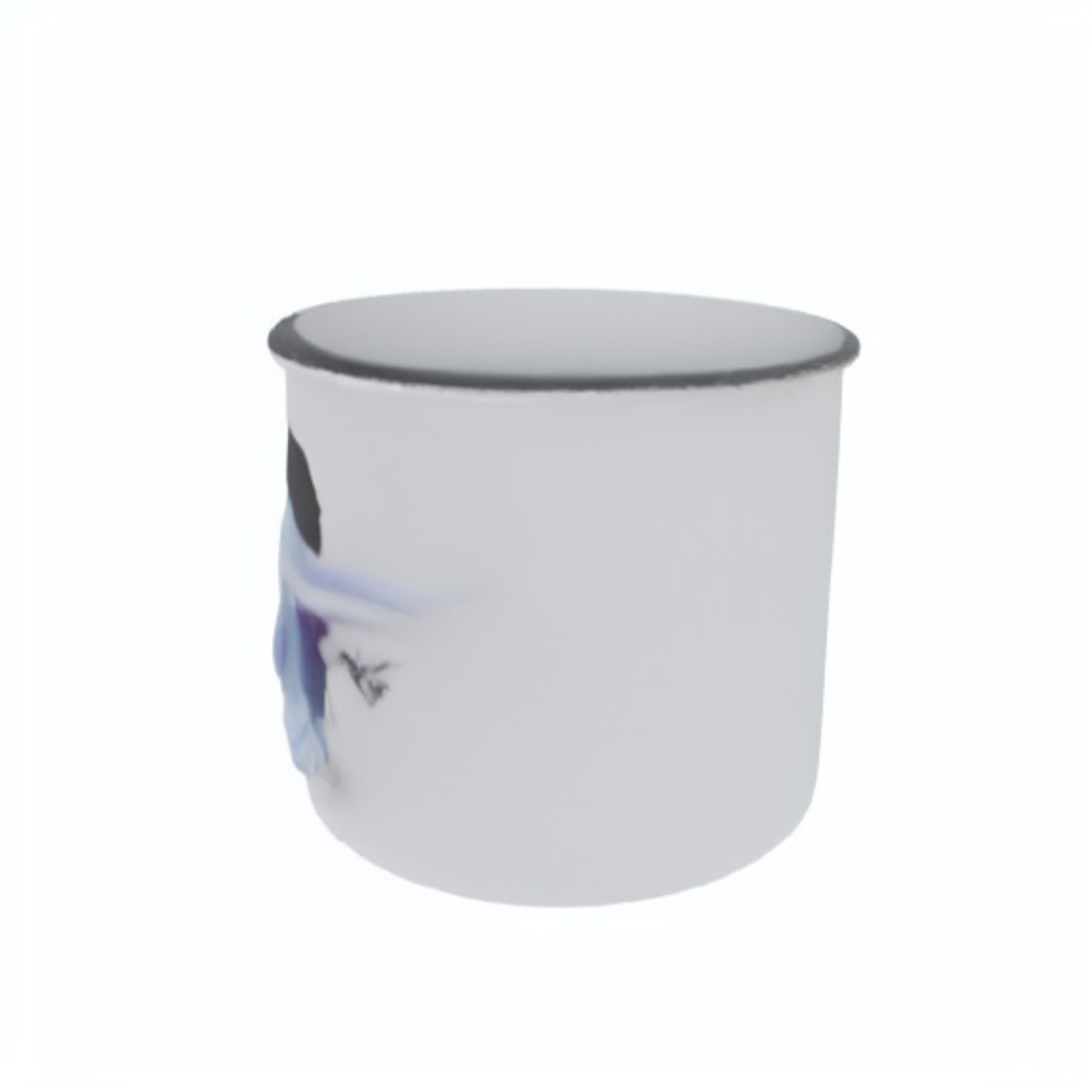}
    		\end{subfigure} &
    		\begin{subfigure}[c]{0.104\textwidth} 
    			\centering
    			\includegraphics[width=\linewidth]{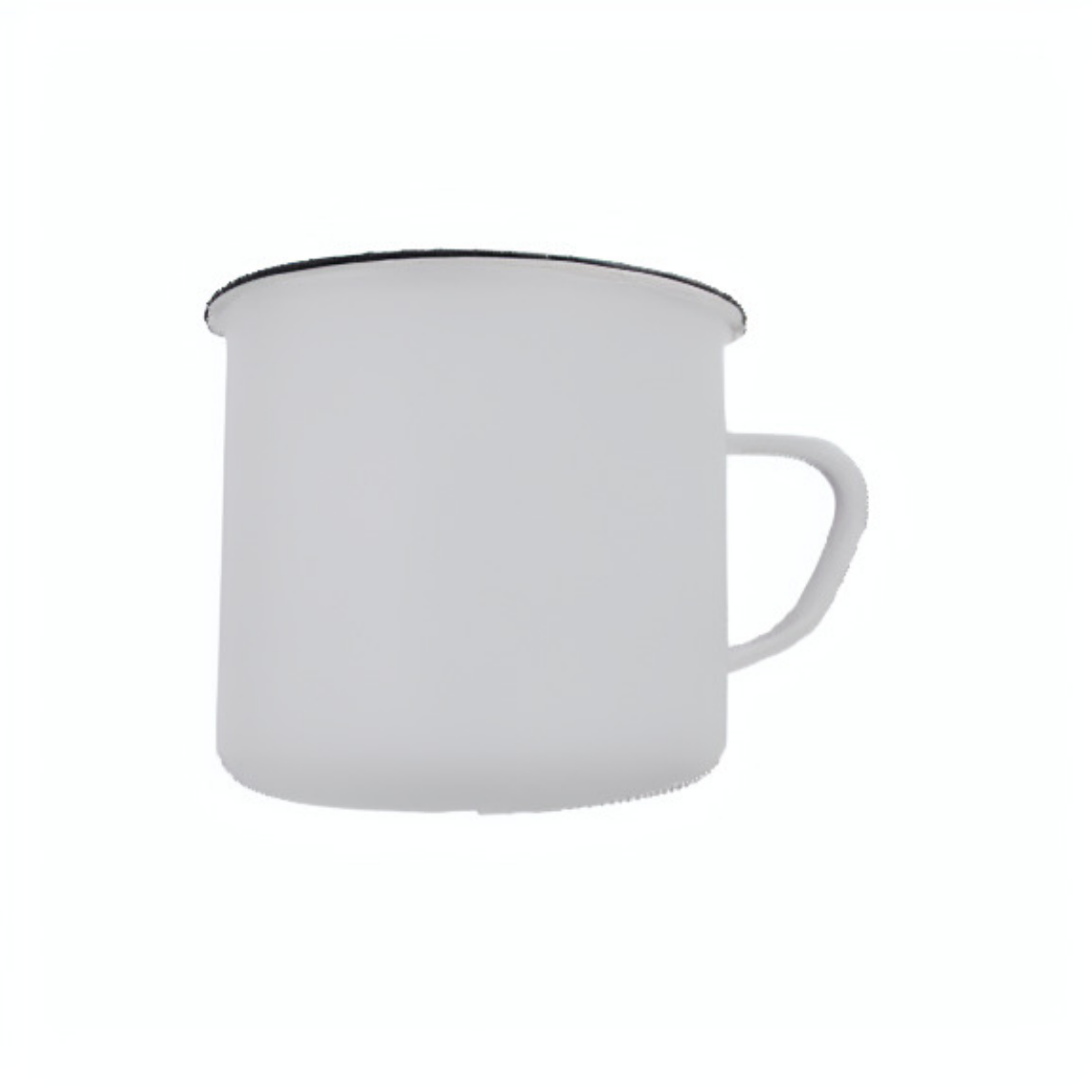}
    		\end{subfigure} &
    		\begin{subfigure}[c]{0.104\textwidth}  
    			\centering
    			\includegraphics[width=\linewidth]{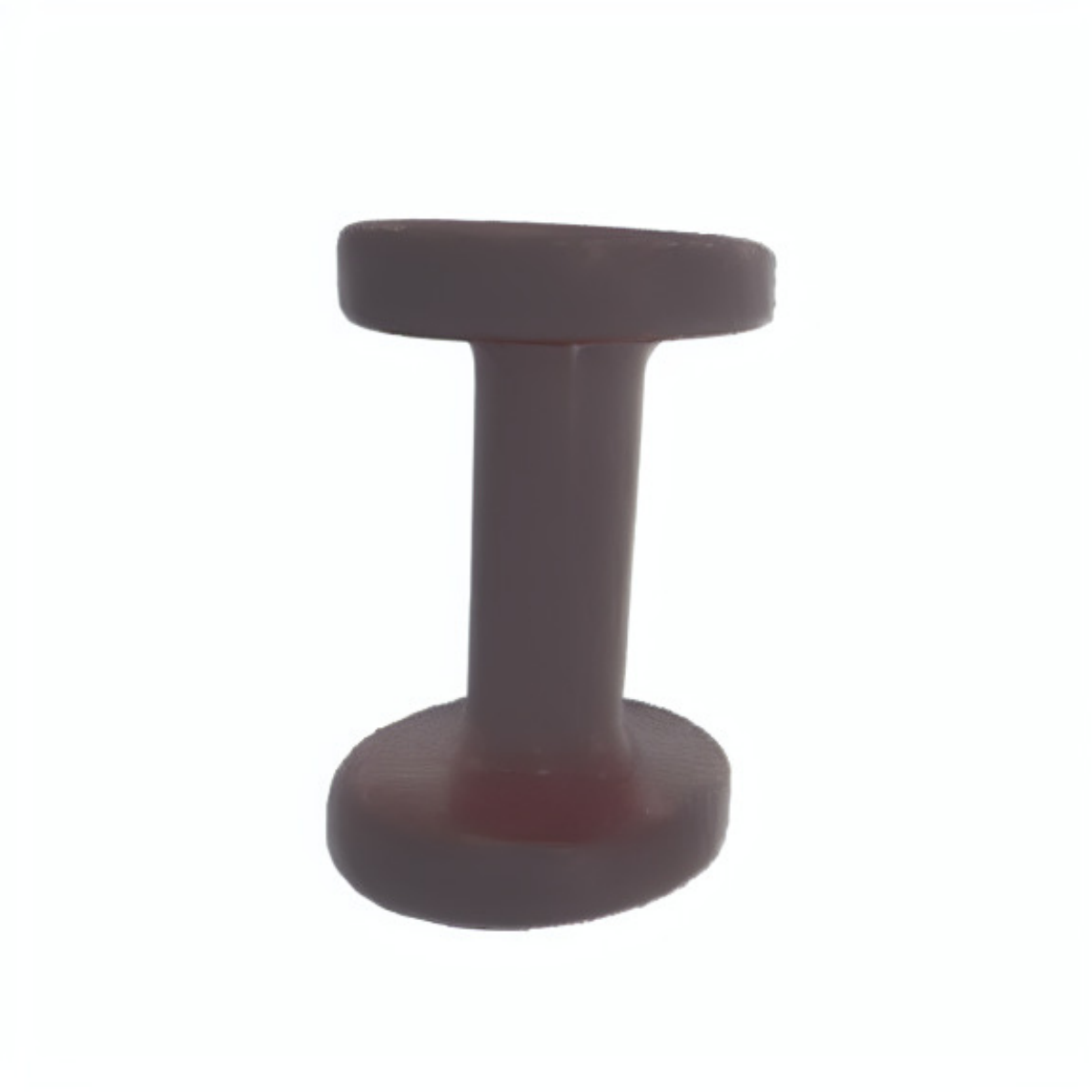}
    		\end{subfigure} &
    		\begin{subfigure}[c]{0.104\textwidth}  
    			\centering
    			\includegraphics[width=\linewidth]{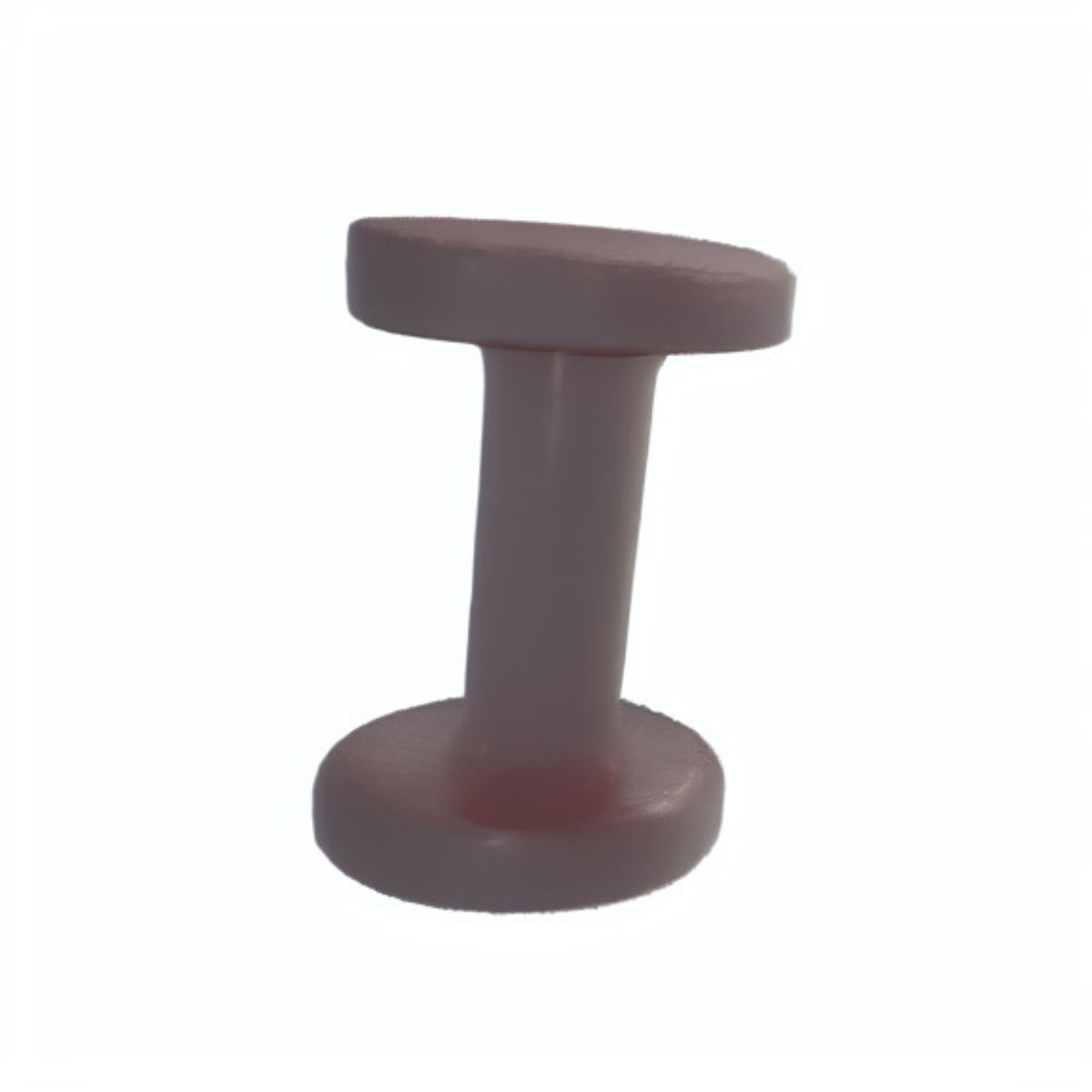}
    		\end{subfigure}&
    		\begin{subfigure}[c]{0.104\textwidth}  
    			\centering
    			\includegraphics[width=\linewidth]{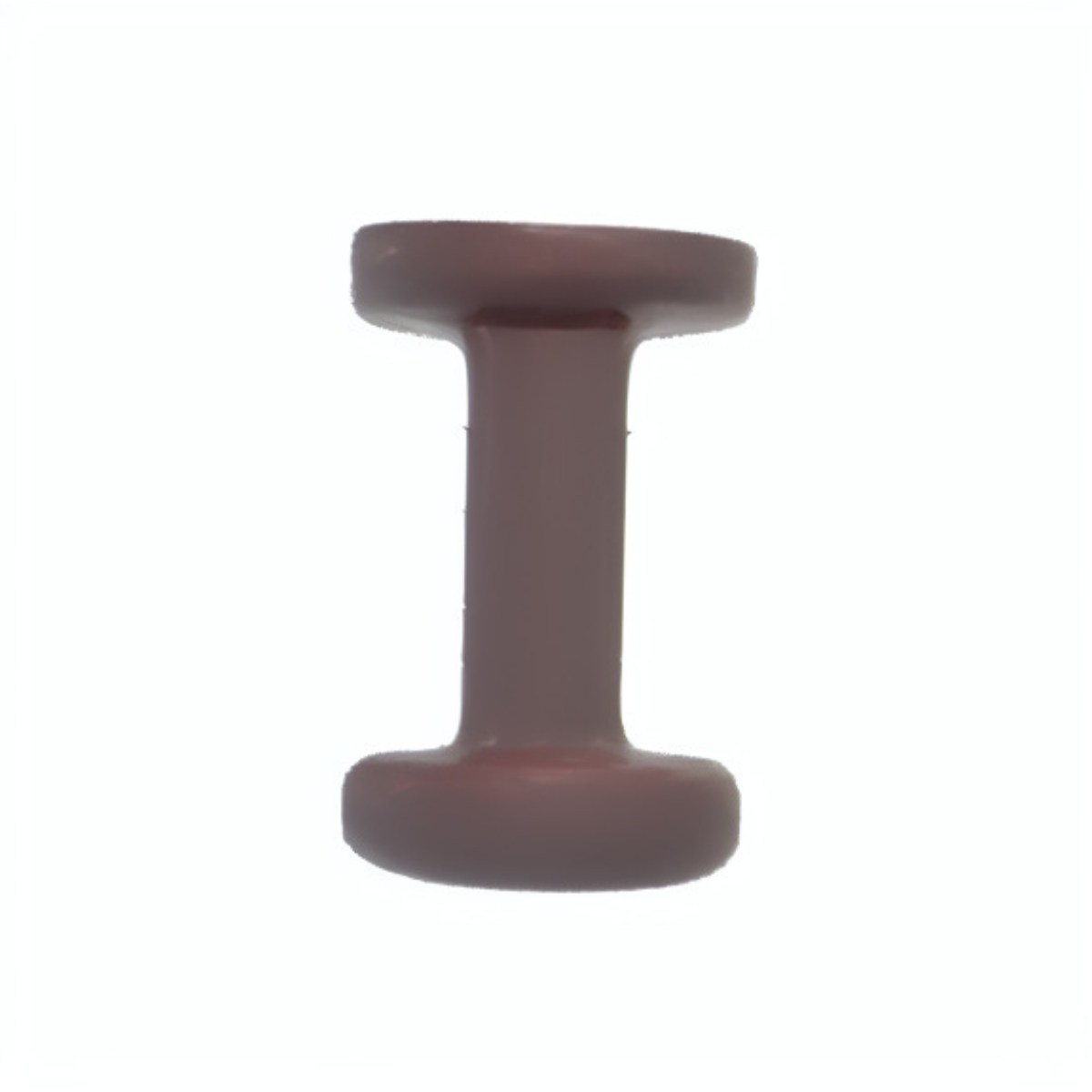}
    		\end{subfigure} \\
    		\makecell{Vivid\\1-to-3} &
    		\begin{subfigure}[c]{0.104\textwidth}  
    			\centering
    			\includegraphics[width=\linewidth]{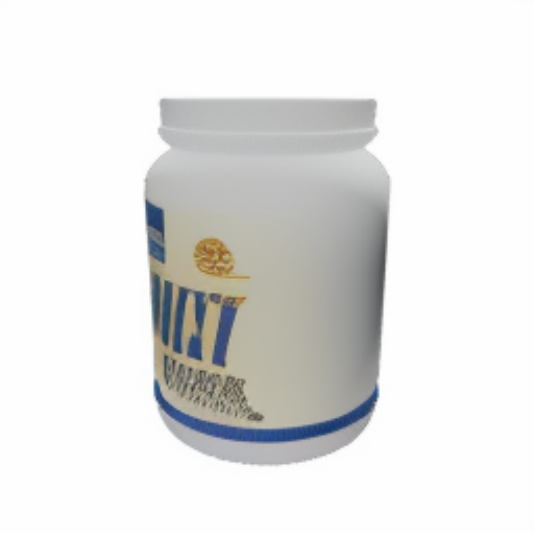}
    		\end{subfigure} &
    		\begin{subfigure}[c]{0.104\textwidth}  
    			\centering
    			\includegraphics[width=\linewidth]{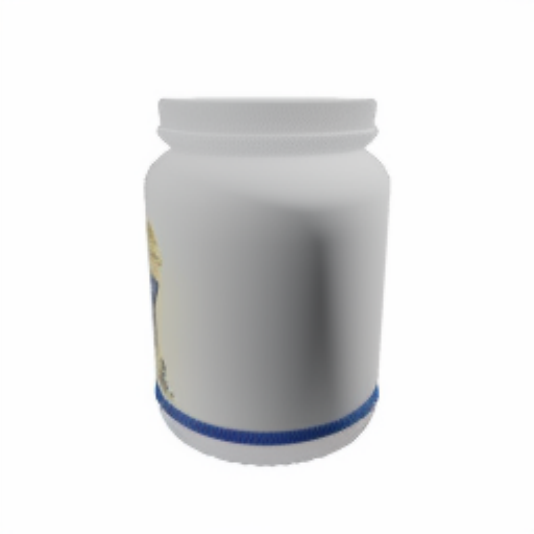}
    		\end{subfigure} &
    		\begin{subfigure}[c]{0.104\textwidth}  
    			\centering
    			\includegraphics[width=\linewidth]{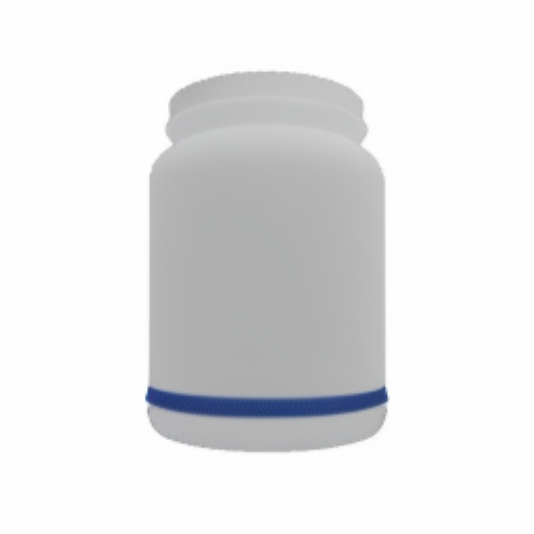}
    		\end{subfigure} &
    		\begin{subfigure}[c]{0.104\textwidth}  
    			\centering
    			\includegraphics[width=\linewidth]{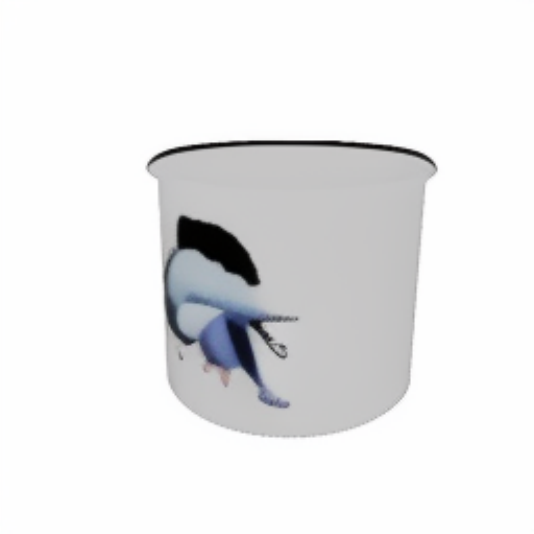}
    		\end{subfigure} &
    		\begin{subfigure}[c]{0.104\textwidth}  
    			\centering
    			\includegraphics[width=\linewidth]{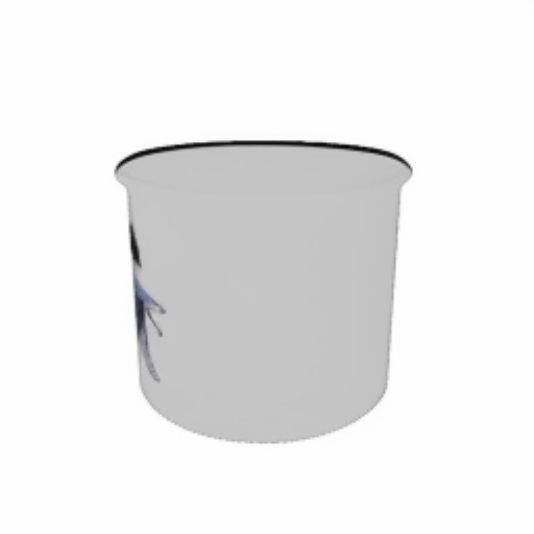}
    		\end{subfigure} &
    		\begin{subfigure}[c]{0.104\textwidth} 
    			\centering
    			\includegraphics[width=\linewidth]{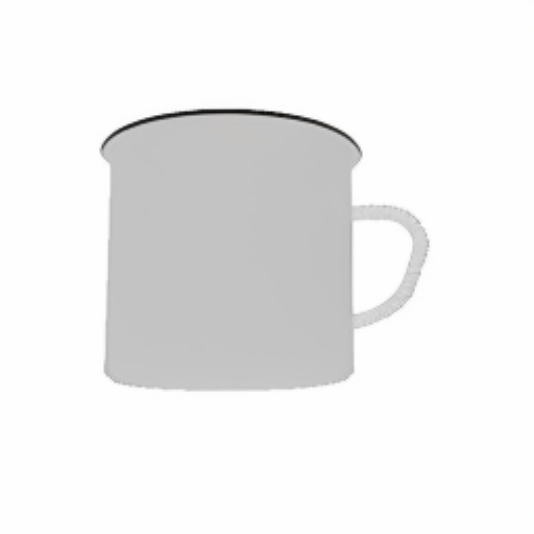}
    		\end{subfigure} &
    		\begin{subfigure}[c]{0.104\textwidth}  
    			\centering
    			\includegraphics[width=\linewidth]{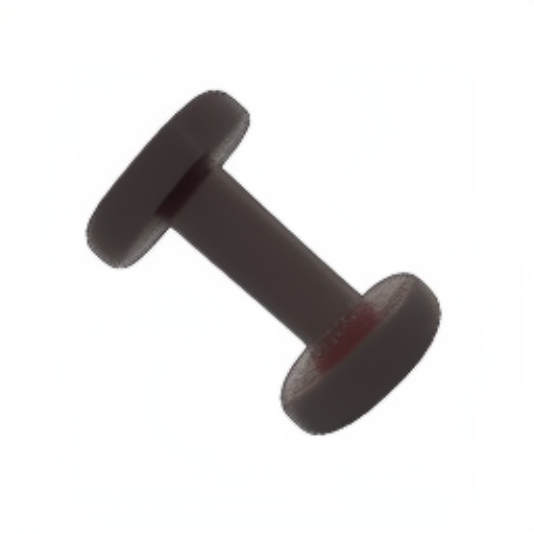}
    		\end{subfigure} &
    		\begin{subfigure}[c]{0.104\textwidth}  
    			\centering
    			\includegraphics[width=\linewidth]{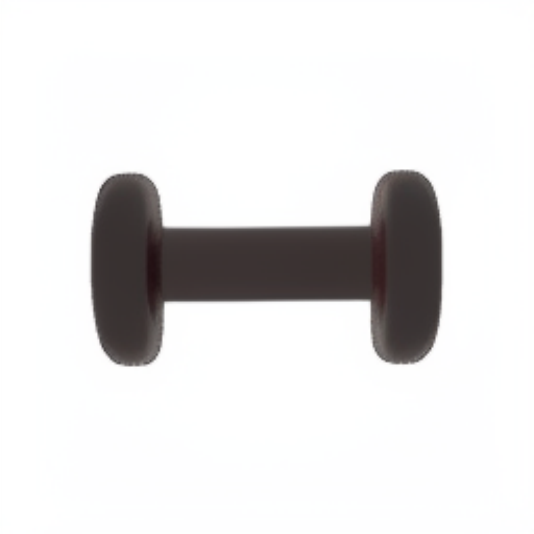}
    		\end{subfigure}&
    		\begin{subfigure}[c]{0.104\textwidth}  
    			\centering
    			\includegraphics[width=\linewidth]{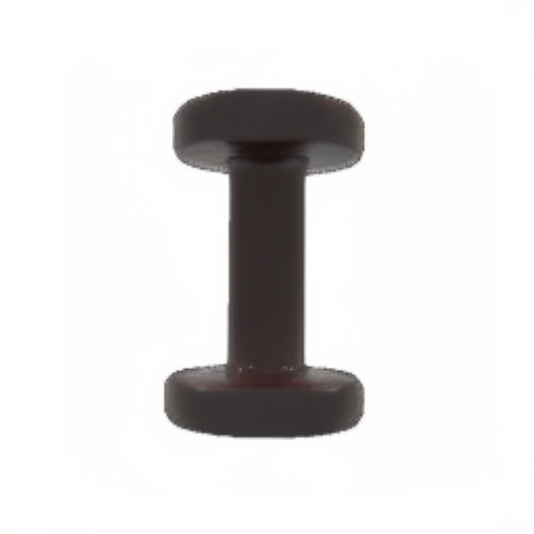}
    		\end{subfigure} \\
    		\makecell{Zero\\1-to-3\\XL} &
    		\begin{subfigure}[c]{0.104\textwidth}  
    			\centering
    			\includegraphics[width=\linewidth]{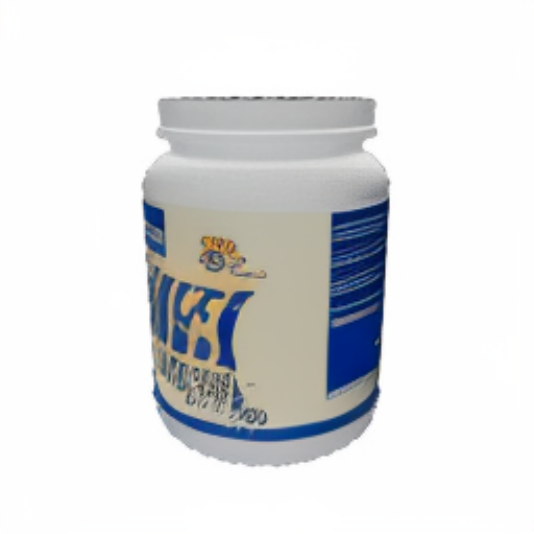}
    		\end{subfigure} &
    		\begin{subfigure}[c]{0.104\textwidth}  
    			\centering
    			\includegraphics[width=\linewidth]{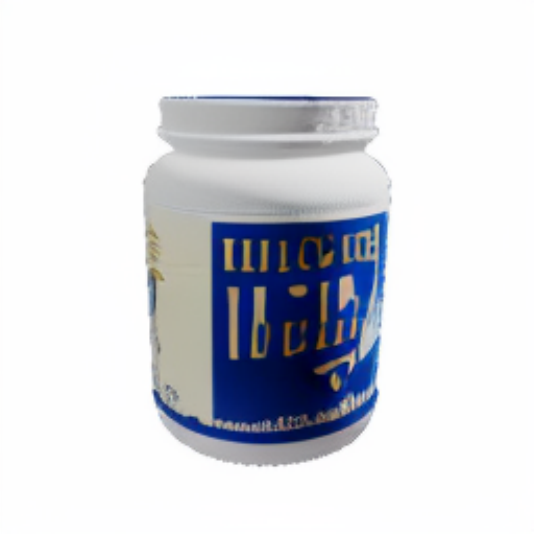}
    		\end{subfigure} &
    		\begin{subfigure}[c]{0.104\textwidth}  
    			\centering
    			\includegraphics[width=\linewidth]{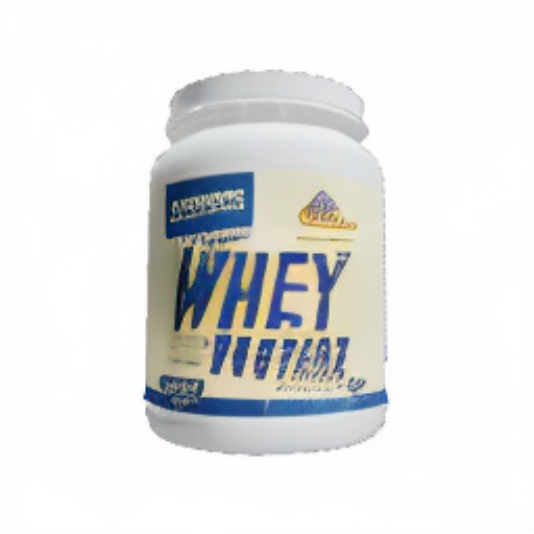}
    		\end{subfigure} &
    		\begin{subfigure}[c]{0.104\textwidth}  
    			\centering
    			\includegraphics[width=\linewidth]{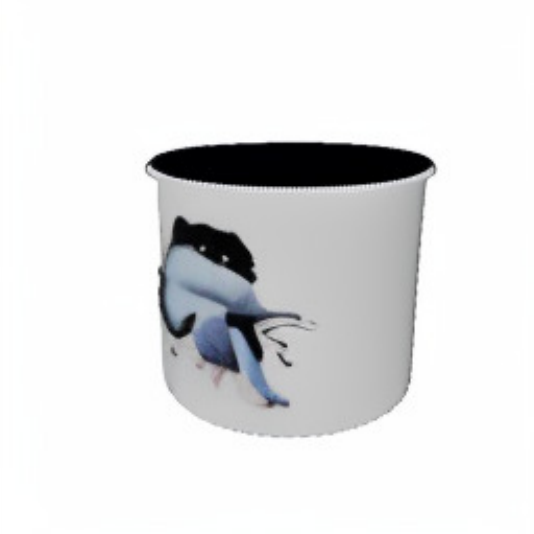}
    		\end{subfigure} &
    		\begin{subfigure}[c]{0.104\textwidth}  
    			\centering
    			\includegraphics[width=\linewidth]{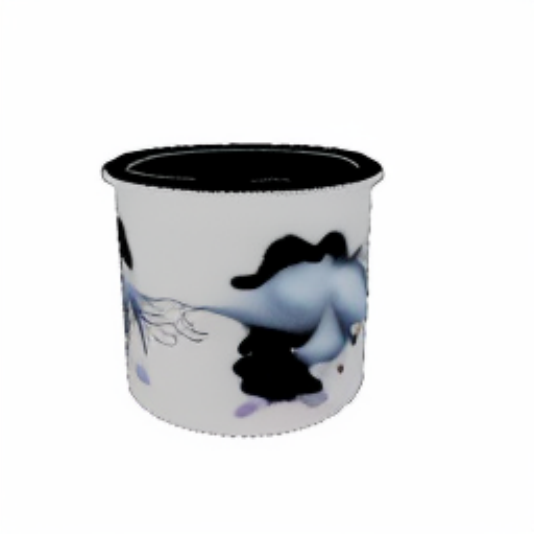}
    		\end{subfigure} &
    		\begin{subfigure}[c]{0.104\textwidth} 
    			\centering
    			\includegraphics[width=\linewidth]{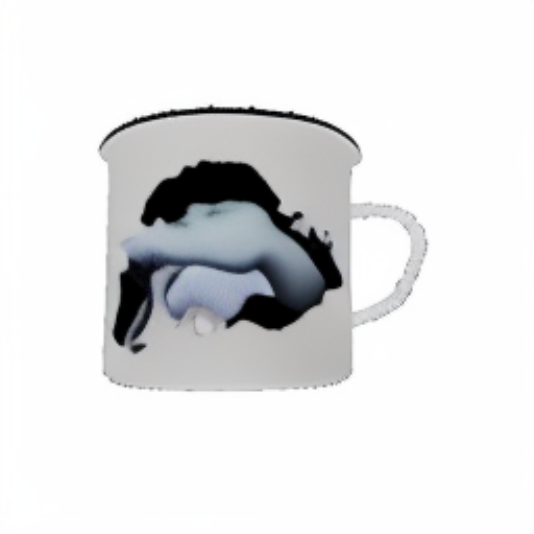}
    		\end{subfigure} &
    		\begin{subfigure}[c]{0.104\textwidth}  
    			\centering
    			\includegraphics[width=\linewidth]{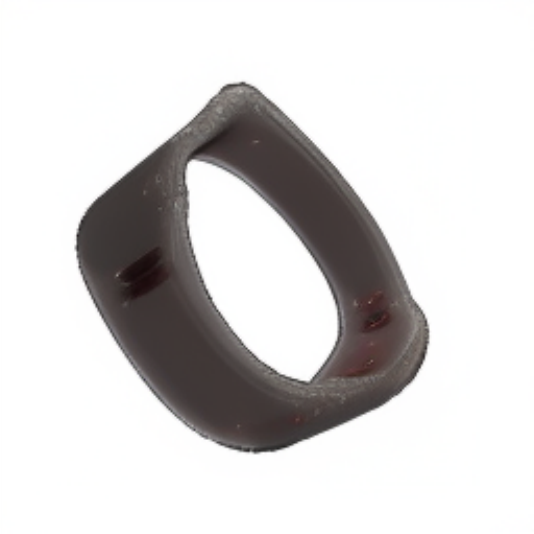}
    		\end{subfigure} &
    		\begin{subfigure}[c]{0.104\textwidth}  
    			\centering
    			\includegraphics[width=\linewidth]{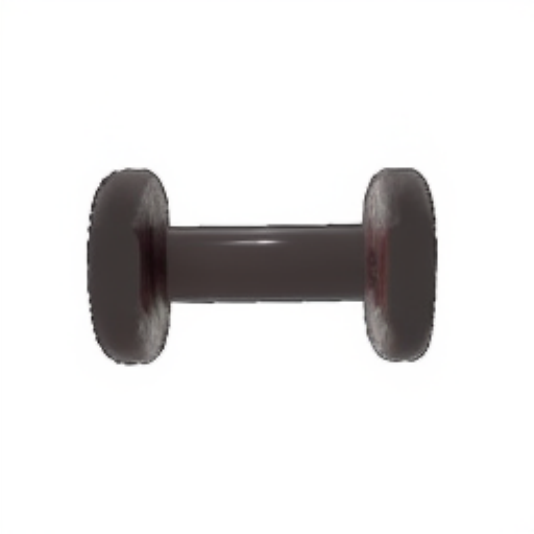}
    		\end{subfigure}&
    		\begin{subfigure}[c]{0.104\textwidth}  
    			\centering
    			\includegraphics[width=\linewidth]{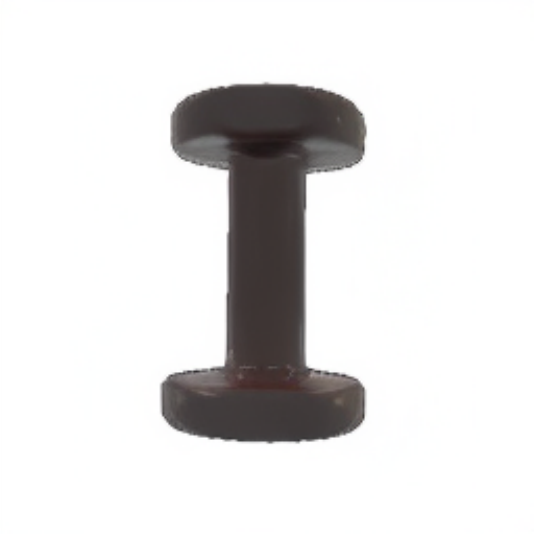}
    		\end{subfigure} 
    	\end{tabular}
    	\caption{Visual results of different novel view synthesis models on dynamic orbits.}
    	\label{qualitative_results_under_dynamic_orbit}
    \end{figure*}

   	We campare our method (with EDN) against the original SV3D and Mv-Adapter models, as well as several NVS baselines, including Zero-1-to-3 \cite{liu2023zero}, Zero-1-to-3 XL \cite{deitke2023objaversexl}, and Vivid-1-to-3 \cite{kwak2024vivid}, across the three test datasets. Among the multiple versions provided by the SV3D authors, we use \(\mathrm{SV3D}^{p}\), which performs the best. Since the authors of Mv-Adapter only provide model weights for fixed viewpoints, Mv-Adapter is evaluated only on the static orbit. Quantitative results for static and dynamic orbits are reported in Tables \ref{compare_with_baselines} and \ref{compare_with_baselines_dynamic_orbit}, respectively. Visual results are provided in the Supplementary Material for the static orbit and in Fig. \ref{qualitative_results_under_dynamic_orbit} for the dynamic orbit.  Both SV3D with EDN and Mv-Adapter with EDN outperform their original counterparts and other NVS methods on both static and dynamic orbits. Fig. \ref{qualitative_results_under_dynamic_orbit} illustrates that Zero-1-to-3 XL exhibits inconsistent multi-view results because each view is generated independently. Vivid-1-to-3, which is also based on Zero-1-to-3 XL, occasionally suffers from the same issue. SV3D with EDN notably improves over the original SV3D in terms of local detail, appearance size, and multi-view consistency.
    \begin{table}[htbp]
    	\caption{Evaluation of seven models on static orbits of three different datasets.}
    	\label{compare_with_baselines}
    	\centering
    	\begin{tabular}{ccccc}
    		\toprule
    		Dataset&Method&PSNR\(\uparrow \)&SSIM\(\uparrow \)&LPIPS\(\downarrow  \)\\
    		\midrule
    		&Zero-1-to-3&18.801&0.8362&0.1471\\
    		\cmidrule{2-5}
    		&Zero-1-to-3 XL&19.058&0.8375&0.1418\\
    		\cmidrule{2-5}
    		&Vivid-1-to-3&20.066&0.8673&0.1366\\
    		\cmidrule{2-5}
    		GSO&SV3D&20.349&0.8909&0.1364\\
    		\cmidrule{2-5}
    		& SV3D with EDN (ours)&21.659&\textbf{0.9070}&\textbf{0.1208}\\
    		\cmidrule{2-5}
    		&Mv-Adapter&20.456&0.8537&0.2105 \\
    		\cmidrule{2-5}
    		&\makecell{Mv-Adapter\\with EDN (ours)}&\textbf{22.316}&0.8856&0.1619 \\
    		\cmidrule{1-5}
    		&Zero-1-to-3&19.495&0.8441&0.1454\\
    		\cmidrule{2-5}
    		&Zero-1-to-3 XL&19.881&0.8365&0.1541\\
    		\cmidrule{2-5}
    		&Vivid-1-to-3&21.303&0.9024&0.1330\\
    		\cmidrule{2-5}
    		Objaverse&SV3D&23.292&0.9208&0.1061\\
    		\cmidrule{2-5}
    		& SV3D with EDN (ours)&\textbf{24.641}&\textbf{0.9373}&\textbf{0.0911}\\
    		\cmidrule{2-5}
    		&Mv-Adapter&21.838&0.8973&0.1570\\
    		\cmidrule{2-5}
    		&\makecell{Mv-Adapter\\with EDN (ours)}&23.547&0.9194&0.1237 \\
    		\cmidrule{1-5}
    		&Zero-1-to-3&17.501&0.8253&0.1809\\
    		\cmidrule{2-5}
    		&Zero-1-to-3 XL&16.163&0.7846&0.2057\\
    		\cmidrule{2-5}
    		&Vivid-1-to-3&17.293&0.8535&0.1820\\
    		\cmidrule{2-5}
    		\multirow{8}{*}{\raisebox{1.7\totalheight}[0pt][0pt]{\makecell{Omni\\Object3D}}}&SV3D&18.640&0.8759&0.l746\\
    		\cmidrule{2-5}
    		& SV3D with EDN (ours)&19.847&\textbf{0.9059}&\textbf{0.1382}\\
    		\cmidrule{2-5}
    		&Mv-Adapter&20.368&0.8453&0.2321 \\
    		\cmidrule{2-5}
    		&\makecell{Mv-Adapter\\with EDN (ours)}&\textbf{22.754}&0.8944&0.1674 \\
    		\bottomrule
    	\end{tabular}
    \end{table}
    \begin{table}[htbp]
    	\caption{Evaluation of five models on dynamic orbits of three different datasets.}
    	\label{compare_with_baselines_dynamic_orbit}
    	\centering
    	\begin{tabular}{ccccc}
    		\toprule
    		Dataset&Method&PSNR\(\uparrow \)&SSIM\(\uparrow \)&LPIPS\(\downarrow  \)\\
    		\midrule
    		&Zero-1-to-3&18.682&0.8358&0.1480\\
    		\cmidrule{2-5}
    		&Zero-1-to-3 XL&18.779&0.8343&0.1445\\
    		\cmidrule{2-5}
    		GSO&Vivid-1-to-3&19.657&0.8633&0.1416\\
    		\cmidrule{2-5}
    		&SV3D&19.823&0.8868&0.1442\\
    		\cmidrule{2-5}
    		&SV3D with EDN (ours)&\textbf{21.231}&\textbf{0.9045}&\textbf{0.1277}\\
    		\cmidrule{1-5}
    		&Zero-1-to-3&20.081&0.8553&0.1376\\
    		\cmidrule{2-5}
    		&Zero-1-to-3 XL&19.522&0.8357&0.1547 \\
    		\cmidrule{2-5}
    		Objaverse&Vivid-1-to-3&20.649&0.8942&0.1404\\
    		\cmidrule{2-5}
    		&SV3D&21.879&0.9126&0.1184\\
    		\cmidrule{2-5}
    		&SV3D with EDN (ours)&\textbf{23.314}&\textbf{0.9325}&\textbf{0.0996}\\
    		\cmidrule{2-5}
    		\cmidrule{1-5}
    		&Zero-1-to-3&17.482&0.8287&0.1777\\
    		\cmidrule{2-5}
    		&Zero-1-to-3 XL&16.751&0.7950&0.2003\\
    		\cmidrule{2-5}
    		\multirow{5}{*}{\raisebox{1.0\totalheight}[0pt][0pt]{\makecell{Omni\\Object3D}}}&Vivid-1-to-3&17.363&0.8542&0.1790\\
    		\cmidrule{2-5}
    		&SV3D&18.383&0.8780&0.1747\\
    		\cmidrule{2-5}
    		&SV3D with EDN (ours)&\textbf{19.794}&\textbf{0.9059}&\textbf{0.1398} \\
    		\bottomrule
    	\end{tabular}
    \end{table}
    \subsection{Ablation Studies}
    \label{Ablation Studies}
    In this subsection, we conduct ablation studies to evaluate the effectiveness of the EDN architecture and the training dataset. The ablation experiments are performed on the SV3D model, with results on the dynamic orbits of GSO reported in Table \ref{Ablation_Studies_dynamic_orbit}, and those on the static orbits provided in the Supplementary Material.
    \begin{table}[htbp]
    	\caption{Ablation experiments on dynamic orbits of GSO dataset on SV3D.}
    	\label{Ablation_Studies_dynamic_orbit}
    	\centering
    	\begin{tabular}{cccc}
    		\toprule
    		Method&PSNR\(\uparrow \)&SSIM\(\uparrow \)&LPIPS\(\downarrow  \)\\
    		\midrule 
    		standard&19.823&0.8868&0.1442\\
    		\cmidrule{1-4}
    		EDN w/o elevation w/o filter&20.910&0.9008&0.1309\\
    		\cmidrule{1-4}
    		EDN w elevation w/o filter&21.069&0.9035&0.1291\\
    		\cmidrule{1-4}
    		EDN w/o elevation w filter&21.176&0.9030&0.1294\\
    		\cmidrule{1-4}
    		EDN w  sine-based pose embedding&20.403&0.8968&0.1326\\
    		\cmidrule{1-4}
    		EDN w ray map&20.390&0.8976&0.1321\\
    		\cmidrule{1-4}
    		EDN w transposed conv&20.724&0.8973&0.1349\\
    		\cmidrule{1-4}
    		EDN&21.231&0.9045&0.1277\\
    		\bottomrule
    	\end{tabular}
    \end{table}    
    
    First, during the training data collection process, we evaluate the impact of removing elevation variation (i.e. fixing the elevation angle at \(0^{\circ} \) for all objects) and omitting data filtering, as shown in Rows 3--5 of Table \ref{Ablation_Studies_dynamic_orbit}. Compared with the full method in Row 9, incorporating elevation variation and applying data filtering lead to a slight but consistent performance improvement.   
    \begin{figure*}[h]
    	\centering
    	\begin{subfigure}[b]{0.51\textwidth}
    		\includegraphics[width=\textwidth]{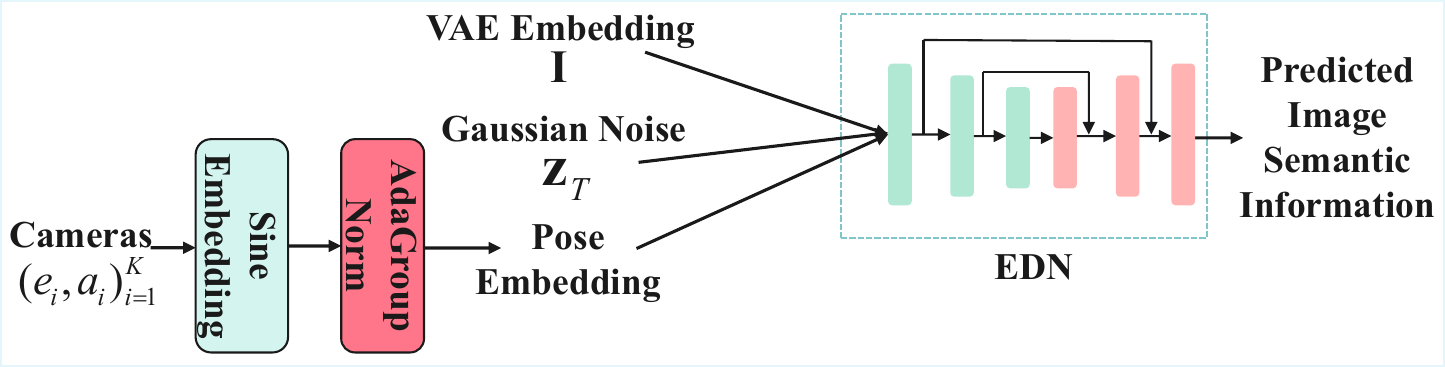}
    		\caption{}
    		\label{EDN_with_pose_1}
    	\end{subfigure}
    	\hfill
    	\begin{subfigure}[b]{0.48\textwidth}
    		\includegraphics[width=\textwidth]{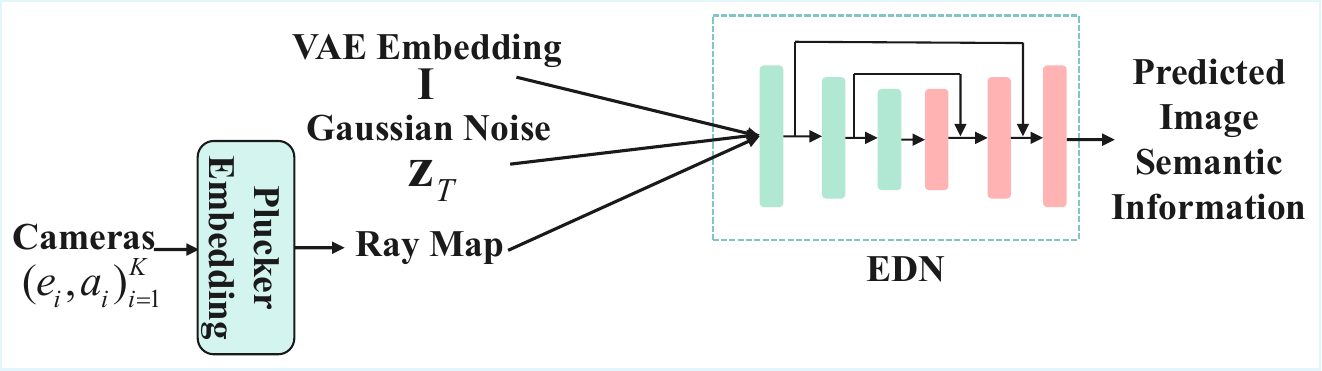}
    		\caption{}
    		\label{EDN_with_pose_2}
    	\end{subfigure}
    	\caption{EDN with pose prompts. (a) EDN with sine-based pose embedding. The camera's azimuth and elevation angles are encoded using sine embedding, and then, transformed through AdaGroupNorm into a tensor matching the shape of the Gaussian noise. This tensor is concatenated with the VAE embedding of the reference image and the Gaussian noise, before being fed into the EDN. (b) EDN with ray map embedding. The camera pose is encoded using Plücker embedding, and converted into a ray map aligned with the Gaussian noise. The ray map is concatenated with the VAE embedding and the Gaussian noise, and the combined input is passed into the EDN.}
    	\label{EDN_with_pose}
    \end{figure*}
    
    Second, we investigate incorporating an additional pose prompt into the EDN architecture, aiming to inject view-dependent pose semantics into the initial noise and thereby improve multi-view image generation quality. We experiment with two pose embedding strategies --- sine embedding and ray map --- as illustrated in Figs. \ref{EDN_with_pose_1} and \ref{EDN_with_pose_2}. We train EDN using datasets of identical size and evaluate the results. However, as shown in Rows 6--7 of Table \ref{Ablation_Studies_dynamic_orbit}, EDN with pose prompts performs worse than EDN without pose prompts. Therefore, we do not include pose information in the EDN input in our final method.
    
    Third, we replace the Pixel Shuffle upsampling layer in the EDN decoder with transposed convolution for comparison. As shown in Row 8 of Table \ref{Ablation_Studies_dynamic_orbit}, Pixel Shuffle yields better performance than transposed convolution. Therefore, we adopt Pixel Shuffle in the decoder.  
    \subsection{Generalization and Effectiveness}
    \label{Supplementary Experiments}
    In this section, we perform additional experiments to validate the generalization and effectiveness of the proposed model.
    \begin{table}[htbp]
    	\caption{Effect of initial noise on SV3D and Mv-Adapter, over the static orbits of three datasets.}
    	\label{compare_with_different_models}
    	\centering
    	\begin{tabular}{cccccc}
    		\toprule
    		Model&Dataset&Method&PSNR\(\uparrow \)&SSIM\(\uparrow \)&LPIPS\(\downarrow  \)\\
    		\midrule 
    		&&standard&20.349&0.8909&0.1364 \\
    		&GSO&inversion&18.706&0.8794&0.1386\\
    		&&with EDN&21.659&0.9070&0.1208\\
    		\cmidrule{2-6}
    		&&standard&23.292&0.9208&0.1061\\
    		SV3D&Objaverse&inversion&20.657&0.9103&0.1117\\
    		&&with EDN&24.641&0.9373&0.0911\\
    		\cmidrule{2-6}
    		&&standard&18.640&0.8759&0.1746\\
    		&\multirow{3}{*}{\raisebox{0.5\totalheight}[0pt][0pt]{\makecell{Omni\\Object3D}}}&inversion&17.337&0.8624&0.1790\\
    		&&with EDN&19.847&0.9059&0.1382\\
    		\cmidrule{1-6}
    		&&standard&20.456&0.8537&0.2105\\
    		&GSO&inversion&19.417&0.8457&0.2138\\
    		&&with EDN&22.316&0.8856&0.1619\\
    		\cmidrule{2-6}
    		&&standard&21.838&0.8973&0.1570\\
    		Mv-Adapter&Objaverse&inversion&19.761&0.8797&0.1691\\
    		&&with EDN&23.547&0.9194&0.1237\\
    		\cmidrule{2-6}
    		&&standard&20.368&0.8453&0.2321\\
    		&\multirow{3}{*}{\raisebox{0.5\totalheight}[0pt][0pt]{\makecell{Omni\\Object3D}}}&inversion&18.356&0.8116&0.2512\\
    		&&with EDN&22.754&0.8944&0.1674\\
    		\bottomrule
    	\end{tabular}
    \end{table} 
       
    We evaluate the impact of different types of initial noise, including standard Gaussian noise (standard), the high-quality noise obtained directly via the discretized Euler inference and inversion (inversion), and the high-quality noise generated by EDN. Table \ref{compare_with_different_models} reports the results on SV3D and Mv-Adapter across three datasets. The results show that EDN consistently improves generation quality across different models and datasets. In contrast, the high-quality noise obtained directly through Euler inference and inversion degrades overall performance. This is because most collected noise samples are actually harmful for generation, as indicated by the high filtering rate. In practice, only a small portion of noise pairs pass the data filtering stage (359 out of 1765 for SV3D and 638 out of 1752 for Mv-Adapter). With data filtering, the network learns exclusively from truly beneficial noise samples, leading to better generation quality. Table \ref{Ablation_Studies_dynamic_orbit} further shows that even without data filtering, the network can still extract useful patterns from the noise dataset, although with reduced stability. Similar trends are also observed under the dynamic orbit, and detailed results are provided in the Supplementary Material.
    \begin{table}[htbp]
    	\caption{Effect of training dataset size on SV3D under dynamic orbits on GSO dataset.}
    	\label{compare_with_different_select_dynamic_orbit}
    	\centering
    	\begin{tabular}{ccccc}
    		\toprule
    		Method&Size&PSNR\(\uparrow \)&SSIM\(\uparrow \)&LPIPS\(\downarrow  \)\\
    		\midrule
    		standard&&19.823&0.8868&0.1442\\
    		EDN&90&20.815&0.8987&0.1322\\
    		EDN&180&21.105&0.9020&0.1298\\
    		EDN&359&21.231&0.9045&0.1277\\
    		\bottomrule
    	\end{tabular}
    \end{table} 
    
    We evaluate the impact of training dataset size on EDN performance, as shown in Table \ref{compare_with_different_select_dynamic_orbit}. As the training size increases, all three evaluation metrics for SV3D with EDN improve consistently. 
    \begin{table}[htbp]
    	\caption{Effect of initial noise across different random seeds on SV3D under dynamic orbits on GSO dataset.}
    	\label{compare_with_different_seeds_dynamic_orbit}
    	\centering
    	\begin{tabular}{ccccc}
    		\toprule
    		Seed&Method&PSNR\(\uparrow \)&SSIM\(\uparrow \)&LPIPS\(\downarrow  \)\\
    		\midrule
    		\multirow{2}{*}{\raisebox{0.005\totalheight}[0pt][0pt]{\makecell{801--900\\(Original)}}}&standard&19.823&0.8868&0.1442 \\
    		&with EDN&21.231&0.9045&0.1277\\
    		\cmidrule{1-5}
    		&standard&19.822&0.8864&0.1435\\
    		\multirow{2}{*}{\raisebox{1.5\totalheight}[0pt][0pt]{1851--1950}}
    		&with EDN&21.277&0.9046&0.1256\\
    		\cmidrule{1-5}
    		&standard&19.869&0.8882&0.1422\\
    		\multirow{2}{*}{\raisebox{1.5\totalheight}[0pt][0pt]{5851--5950}}
    		&with EDN&21.158&0.9040&0.1267\\
    		\bottomrule
    	\end{tabular}
    \end{table} 
    
    We also assess whether EDN maintains stable performance across different random seeds, as summarized in Table \ref{compare_with_different_seeds_dynamic_orbit}. For the trained EDN, seeds both inside and outside the training range consistently deliver performance improvements under dynamic orbits. 
    \begin{table}[htbp]
    	\caption{Effect of data filtering threshold on SV3D under dynamic orbits on GSO dataset.}
    	\label{compare_with_different_m_dynamic_orbit}
    	\centering
    	\begin{tabular}{ccccc}
    		\toprule
    		Threshold&Filtering rate (\%)&PSNR\(\uparrow \)&SSIM\(\uparrow \)&LPIPS\(\downarrow  \)\\
    		\midrule
    		0.000&20.34&21.231&0.9045&0.1277\\
    		0.005&8.16&21.100&0.9033&0.1292\\
    		0.010&4.31&21.093&0.9039&0.1283\\
    		\bottomrule
    	\end{tabular}
    \end{table} 
    
    Different filtering thresholds \(m\) result in different training dataset sizes. Therefore, we conduct experiments to determine an appropriate threshold. As shown in Table \ref{compare_with_different_m_dynamic_orbit}, although a higher threshold yields better quality samples, it also reduces the dataset size and ultimately degrades training performance. Thus, we set the filtering threshold to 0. 
    
   The results of the above three experiments for static orbits on GSO dataset are consistent with those discussed earlier; additional details can be found in the Supplementary Material.
    
    \begin{table}[htbp]
    	\caption{Inference time of EDN on SV3D and Mv-Adapter.}
    	\label{inference_time}
    	\centering
    	\begin{tabular}{ccc}
    		\toprule
    		Model&Inference time (s)&EDN time (ms)\\
    		\midrule 
    		SV3D&42&2.08\\
    		Mv-Adapter&63&2.10\\
    		\bottomrule
    	\end{tabular}
    \end{table} 
    In theory, EDN does not affect the inference efficiency of NVS models. We validate this through experiments. As shown in Table \ref{inference_time}, EDN requires minimal computational resources, and its inference time is negligible.
    
    When collecting high-quality noise, the number of inference-inversion steps \(n\), the inference CFG scale \(\gamma _{1}\), and the inversion CFG scale \(\gamma _{2}\) jointly determine the strength of the injected image semantic information. When \(\gamma _{1}\) and \(\gamma _{2}\) are fixed, increasing \(n\) produces stronger semantic injection. Conversely, when \(n\) is fixed, a larger gap between \(\gamma _{1}\) and \(\gamma _{2}\) also leads to stronger semantic injection. We validate the choices of \(n\) for SV3D and Mv-Adapter through visualization experiments, with results provided in the Supplementary Material. For SV3D, we set \(n=16\) and the initial scaling factor \(q=700.0007\). With fixed CFG scales, too few steps inject insufficient image semantic information, while too many steps can introduce excessive information, causing noisy generation results. For Mv-Adapter, we set \(n=25\) and \(q=36.4351\) to balance collection speed and noise quality. Under a fixed CFG scale, even a small \(n\) improves generation quality without artifacts, but excessively large \(n\) slows the high-quality noise collection process. 
   	\section{Conclusion}
   	This paper proposes a high-quality noise learning framework for NVS models. Our method injects image semantic information into the initial random noise by leveraging the difference in CFG scales during the discretized Euler inference and inversion processes of a diffusion model. A data filter mechanism is then applied to construct a high-quality noise dataset. Using the self-constructed dataset, we train a plug-and-play encoder-decoder network that transforms the initial random noise into high-quality noise. Experimental results demonstrate that the high-quality noise optimized by the proposed EDN enables the NVS model to generate images that are more consistent and closer to the ground truth than those produced by state-of-the-art methods. Moreover, the EDN is lightweight, requires minimal computational resources, and has a negligible impact on inference speed.
   	
   	This paper still has certain limitations. The discretized Euler inversion method approximates the predicted noise at timestep \(t\) using the estimate from timestep \(t–1\), which means that the reconstructed high-quality noise is theoretically only an approximation. Developing more accurate noise reconstruction methods will be an important direction for future work.
	\bibliography{reference}

@inproceedings{ho2022classifier,
	title={Classifier-Free Diffusion Guidance},
	author={Ho, Jonathan and Salimans, Tim},
	booktitle={NeurIPS 2021 Workshop Deep Gener. Models Downstr. Appl.},
	year={2021}
}

@inproceedings{song2020denoising,
	title={Denoising Diffusion Implicit Models},
	author={Song, Jiaming and Meng, Chenlin and Ermon, Stefano},
	booktitle={Int. Conf. Learn. Represent.},
	year={2021}
}

@inproceedings{rombach2022high,
	title={High-resolution image synthesis with latent diffusion models},
	author={Rombach, Robin and Blattmann, Andreas and Lorenz, Dominik and others},
	booktitle={Proc. IEEE/CVF Conf. Comput. Vis. Pattern Recognit.},
	pages={10684--10695},
	year={2022}
}

@inproceedings{liu2023zero,
	title={Zero-1-to-3: Zero-shot one image to 3d object},
	author={Liu, Ruoshi and Wu, Rundi and Van Hoorick, Basile and others},
	booktitle={Proc. IEEE/CVF Int. Conf. Comput. Vis.},
	pages={9298--9309},
	year={2023}
}

@inproceedings{liu2023syncdreamer,
	title={SyncDreamer: Generating Multiview-consistent Images from a Single-view Image},
	author={Liu, Yuan and Lin, Cheng and Zeng, Zijiao and others},
	booktitle={Int. Conf. Learn. Represent.},
	year={2024}
}

@article{shi2023zero123++,
	title={Zero123++: a single image to consistent multi-view diffusion base model},
	author={Shi, Ruoxi and Chen, Hansheng and Zhang, Zhuoyang and others},
	journal={arXiv preprint arXiv:2310.15110},
	year={2023}
}

@inproceedings{zheng2024free3d,
	title={Free3d: Consistent novel view synthesis without 3d representation},
	author={Zheng, Chuanxia and Vedaldi, Andrea},
	booktitle={Proc. IEEE/CVF Conf. Comput. Vis. Pattern Recognit.},
	pages={9720--9731},
	year={2024}
}

@inproceedings{tang2024mvdiffusion++,
	title={Mvdiffusion++: A dense high-resolution multi-view diffusion model for single or sparse-view 3d object reconstruction},
	author={Tang, Shitao and Chen, Jiacheng and Wang, Dilin and others},
	booktitle={Eur. Conf. Comput. Vis.},
	pages={175--191},
	year={2024},
}

@inproceedings{deng2023mv,
	title={MV-Diffusion: Motion-aware video diffusion model},
	author={Deng, Zijun and He, Xiangteng and Peng, Yuxin and others},
	booktitle={Proc. 31st ACM Int. Conf. Multimedia},
	pages={7255--7263},
	year={2023}
}

@inproceedings{deitke2023objaverse,
	title={Objaverse: A universe of annotated 3d objects},
	author={Deitke, Matt and Schwenk, Dustin and Salvador, Jordi and others},
	booktitle={Proc. IEEE/CVF Conf. Comput. Vis. Pattern Recognit.},
	pages={13142--13153},
	year={2023}
}

@inproceedings{voleti2024sv3d,
	title={Sv3d: Novel multi-view synthesis and 3d generation from a single image using latent video diffusion},
	author={Voleti, Vikram and Yao, Chun-Han and Boss, Mark and others},
	booktitle={Eur. Conf. Comput. Vis.},
	pages={439--457},
	year={2024},
}

@article{blattmann2023stable,
	title={Stable video diffusion: Scaling latent video diffusion models to large datasets},
	author={Blattmann, Andreas and Dockhorn, Tim and Kulal, Sumith and others},
	journal={arXiv preprint arXiv:2311.15127},
	year={2023}
}

@inproceedings{huang2025mv,
	title={Mv-adapter: Multi-view consistent image generation made easy},
	author={Huang, Zehuan and Guo, Yuan-Chen and Wang, Haoran and others},
	booktitle={Proc. IEEE/CVF Int. Conf. Comput. Vis.},
	pages={16377--16387},
	year={2025}
}

@inproceedings{kwak2024vivid,
	title={Vivid-1-to-3: Novel view synthesis with video diffusion models},
	author={Kwak, Jeong-gi and Dong, Erqun and Jin, Yuhe and others},
	booktitle={Proc. IEEE/CVF Conf. Comput. Vis. Pattern Recognit.},
	pages={6775--6785},
	year={2024}
}

@article{ma2025inference,
	title={Inference-time scaling for diffusion models beyond scaling denoising steps},
	author={Ma, Nanye and Tong, Shangyuan and Jia, Haolin and others},
	journal={arXiv preprint arXiv:2501.09732},
	year={2025}
}

@article{kim2025model,
	title={Model Already Knows the Best Noise: Bayesian Active Noise Selection via Attention in Video Diffusion Model},
	author={Kim, Kwanyoung and Kim, Sanghyun},
	journal={arXiv preprint arXiv:2505.17561},
	year={2025}
}

@inproceedings{zhou2025golden,
	title={Golden noise for diffusion models: A learning framework},
	author={Zhou, Zikai and Shao, Shitong and Bai, Lichen and others},
	booktitle={Proc. IEEE/CVF Int. Conf. Comput. Vis.},
	pages={17688--17697},
	year={2025}
}

@inproceedings{lichenzigzag,
	title={Zigzag Diffusion Sampling: Diffusion Models Can Self-Improve via Self-Reflection},
	author={Bai, Lichen and Shao, Shitong and Zhou, Zikai and others},
	booktitle={13th Int. Conf. Learn. Represent.},
	year={2025}
}

@inproceedings{mokady2023null,
	title={Null-text inversion for editing real images using guided diffusion models},
	author={Mokady, Ron and Hertz, Amir and Aberman, Kfir and others},
	booktitle={Proc. IEEE/CVF Conf. Comput. Vis. Pattern Recognit.},
	pages={6038--6047},
	year={2023}
}

@article{karras2022elucidating,
	title={Elucidating the design space of diffusion-based generative models},
	author={Karras, Tero and Aittala, Miika and Aila, Timo and others},
	journal={Adv. Neural Inf. Process. Syst.},
	volume={35},
	pages={26565--26577},
	year={2022}
}

@inproceedings{zhang2018unreasonable,
	title={The unreasonable effectiveness of deep features as a perceptual metric},
	author={Zhang, Richard and Isola, Phillip and Efros, Alexei A and others},
	booktitle={Proc. IEEE Conf. Comput. Vis. Pattern Recognit.},
	pages={586--595},
	year={2018}
}

@inproceedings{downs2022google,
	title={Google scanned objects: A high-quality dataset of 3d scanned household items},
	author={Downs, Laura and Francis, Anthony and Koenig, Nate and others},
	booktitle={2022 IEEE Int. Conf. Robot. Autom. (ICRA)},
	pages={2553--2560},
	year={2022},
}

@article{deitke2023objaversexl,
	title={Objaverse-xl: A universe of 10m+ 3d objects},
	author={Deitke, Matt and Liu, Ruoshi and Wallingford, Matthew and others},
	journal={Adv. Neural Inf. Process. Syst.},
	volume={36},
	pages={35799--35813},
	year={2023}
}

@inproceedings{gao2024cat3d,
	title={CAT3D: create anything in 3D with multi-view diffusion models},
	author={Gao, Ruiqi and Holynski, Aleksander and Henzler, Philipp and others},
	booktitle={Proc. 38th Int. Conf. Neural Inf. Process. Syst.},
	pages={75468--75494},
	year={2024}
}

@inproceedings{guo2024initno,
	title={Initno: Boosting text-to-image diffusion models via initial noise optimization},
	author={Guo, Xiefan and Liu, Jinlin and Cui, Miaomiao and others},
	booktitle={Proc. IEEE/CVF Conf. Comput. Vis. Pattern Recognit.},
	pages={9380--9389},
	year={2024}
}

@article{qi2024not,
	title={Not all noises are created equally: Diffusion noise selection and optimization},
	author={Qi, Zipeng and Bai, Lichen and Xiong, Haoyi and others},
	journal={arXiv preprint arXiv:2407.14041},
	year={2024}
}

@article{oshima2025inference,
	title={Inference-time text-to-video alignment with diffusion latent beam search},
	author={Oshima, Yuta and Suzuki, Masahiro and Matsuo, Yutaka and others},
	journal={arXiv preprint arXiv:2501.19252},
	year={2025}
}

@inproceedings{girshick2015fast,
	title={Fast r-cnn},
	author={Girshick, Ross},
	booktitle={Proc. IEEE Int. Conf. Comput. Vis.},
	pages={1440--1448},
	year={2015}
}

@article{fardo2016formal,
	title={A formal evaluation of PSNR as quality measurement parameter for image segmentation algorithms},
	author={Fardo, Fernando A and Conforto, Victor H and De Oliveira, Francisco C and others},
	journal={arXiv preprint arXiv:1605.07116},
	year={2016}
}

@article{wang2004image,
	title={Image quality assessment: from error visibility to structural similarity},
	author={Wang, Zhou and Bovik, Alan C and Sheikh, Hamid R and others},
	journal={IEEE Trans. Image Process.},
	volume={13},
	number={4},
	pages={600--612},
	year={2004},
	publisher={IEEE}
}

@inproceedings{kant2025pippo,
	title={Pippo: High-resolution multi-view humans from a single image},
	author={Kant, Yash and Weber, Ethan and Kim, Jin Kyu and others},
	booktitle={Proc. Comput. Vis. Pattern Recognit. Conf.},
	pages={16418--16429},
	year={2025}
}

@article{adam2014method,
	title={A method for stochastic optimization},
	author={Adam, Kingma DP Ba J and others},
	journal={arXiv preprint arXiv:1412.6980},
	year={2014}
}

@inproceedings{wu2023omniobject3d,
	title={Omniobject3d: Large-vocabulary 3d object dataset for realistic perception, reconstruction and generation},
	author={Wu, Tong and Zhang, Jiarui and Fu, Xiao and others},
	booktitle={Proc. IEEE/CVF Conf. Comput. Vis. Pattern Recognit.},
	pages={803--814},
	year={2023}
}

@inproceedings{meng2023distillation,
	title={On distillation of guided diffusion models},
	author={Meng, Chenlin and Rombach, Robin and Gao, Ruiqi and others},
	booktitle={Proc. IEEE/CVF Conf. Comput. Vis. Pattern Recognit.},
	pages={14297--14306},
	year={2023}
}

@inproceedings{shi2016real,
	title={Real-time single image and video super-resolution using an efficient sub-pixel convolutional neural network},
	author={Shi, Wenzhe and Caballero, Jose and Husz{\'a}r, Ferenc and others},
	booktitle={Proc. IEEE Conf. Comput. Vis. Pattern Recognit.},
	pages={1874--1883},
	year={2016}
}

@inproceedings{he2016deep,
	title={Deep residual learning for image recognition},
	author={He, Kaiming and Zhang, Xiangyu and Ren, Shaoqing and others},
	booktitle={Proc. IEEE Conf. Comput. Vis. Pattern Recognit.},
	pages={770--778},
	year={2016}
}

@inproceedings{kumar2020fisheyedistancenet,
	title={Fisheyedistancenet: Self-supervised scale-aware distance estimation using monocular fisheye camera for autonomous driving},
	author={Kumar, Varun Ravi and Hiremath, Sandesh Athni and Bach, Markus and others},
	booktitle={2020 IEEE Int. Conf. Robot. Autom. (ICRA)},
	pages={574--581},
	year={2020},
}

@inproceedings{ronneberger2015u,
	title={U-net: Convolutional networks for biomedical image segmentation},
	author={Ronneberger, Olaf and Fischer, Philipp and Brox, Thomas},
	booktitle={Int. Conf. Med. Image Comput. Comput.-Assisted Interv.},
	pages={234--241},
	year={2015},
}

@misc{von-platen-etal-2022-diffusers,
	author = {Patrick von Platen and Suraj Patil and Anton Lozhkov and others},
	title = {Diffusers: State-of-the-art diffusion models},
	year = {2022},
	publisher = {GitHub},
	journal = {GitHub repository},
	howpublished = {\url{https://github.com/huggingface/diffusers}}
}

@article{xu2024sketch2scene,
	title={Sketch2Scene: Automatic Generation of Interactive 3D Game Scenes from User's Casual Sketches},
	author={Xu, Yongzhi and Ng, Yonhon and Wang, Yifu and others},
	journal={arXiv preprint arXiv:2408.04567},
	year={2024}
}

@inproceedings{asish2025synthesizing,
	title={Synthesizing Six Years of AR/VR Research: A Systematic Review of Machine and Deep Learning Applications},
	author={Asish, Sarker M and Karki, Bhoj B and Kolahchi, Niloofar and others},
	booktitle={2025 IEEE Conf. Virtual Reality 3D User Interfaces (VR)},
	pages={175--185},
	year={2025},
}

@inproceedings{podellsdxl,
	title={SDXL: Improving Latent Diffusion Models for High-Resolution Image Synthesis},
	author={Podell, Dustin and English, Zion and Lacey, Kyle and others},
	booktitle={12th Int. Conf. Learn. Represent.},
	year={2024}
}
	
	{\appendices
	\section{Theory}
	\label{Theory}
	In this section, we theoretically demonstrate that the difference between the CFG scale \(\gamma_{1}\) and 
	\(\gamma_{2}\) during the discretized Euler inference and inversion process injects image semantic information into random noise. For clarity, all instances of \(\varepsilon _{\theta }(\mathbf{z}_{t}',t,\mathbf{c},\mathbf{p} )\) in the appendix are approximated as \(\varepsilon _{\theta }(\mathbf{z}_{t}',t )\).
	
    Suppose we start with an initial noise \(\mathbf{z}_{t}\). After one inference step at timestep \(t\), we obtain the noise \(\mathbf{z}_{t-1}\), as expressed in Eq. \eqref{1} and Eq. \eqref{13}. We then set \(\mathbf{z}_{t-1}=\tilde{\mathbf{z}}_{t-1} \). Next, applying one inversion step to \(\tilde{\mathbf{z}}_{t-1}\) yields \(\tilde{\mathbf{z}}_{t} \), which corresponds to the same state as the initial noise \(\mathbf{z}_{t}\). The results of this inversion process are shown in Eq. \eqref{3} and Eq. \eqref{14}. By substituting the \(\mathbf{z}_{t-1}\) from Eq. \eqref{1} and Eq. \eqref{13} into Eq. \eqref{3} and Eq. \eqref{14}, we obtain:
	
	\noindent for ``v-prediction'':
	\begin{equation}
		\tilde{\mathbf{z}}_{t}  =\mathbf{z}_{t}+\frac{\sqrt{\sigma _{t}^{2} +1}(\sigma  _{t-1}-\sigma _{t}) }{1+\sigma _{t} \sigma  _{t-1}} [\varepsilon _{\theta, 1 }(\mathbf{z}_{t}',t )-\varepsilon _{\theta, 2 }(\tilde{\mathbf{z}}_{t-1}',t)],\label{4}
	\end{equation}
	and for ``epsilon'':
	\begin{equation}
		\tilde{\mathbf{z}}_{t}  =\mathbf{z}_{t}+ (  \sigma  _{t-1}-\sigma _{t}  ) [\varepsilon _{\theta, 1 }(\mathbf{z}_{t}',t )-\varepsilon _{\theta, 2 }(\tilde{\mathbf{z}}_{t-1}',t )].\label{15}
	\end{equation}
	\indent Subtracting \(\mathbf{z}_{t}\) from \(\tilde{\mathbf{z}}_{t} \) and noting that \(\sigma  _{t}\) is a pre-defined parameter, both \(\frac{\sqrt{\sigma _{t}^{2} +1}(\sigma  _{t-1}-\sigma _{t}) }{1+\sigma _{t} \sigma  _{t-1}}\) and \( (  \sigma  _{t-1}-\sigma _{t}  )\) are fixed constants. Therefore, the key term of interest is \([\varepsilon _{\theta, 1 }(\mathbf{z}_{t}',t )-\varepsilon _{\theta, 2 }(\tilde{\mathbf{z}}_{t-1}',t )]\). We further approximate the predicted noise at timestep \(t-1\) using the prediction at timestep \(t\), i.e., \(\varepsilon _{\theta, 2 }(\mathbf{z}_{t}',t  ) \approx \varepsilon _{\theta, 2 } (\tilde{\mathbf{z}}_{t-1}',t  ) \). \(\varepsilon _{\theta, 1 }(\mathbf{z}_{t}',t )-\varepsilon _{\theta, 2 }(\tilde{\mathbf{z}}_{t-1}',t)\) can be approximated by \(\varepsilon _{\theta, 1 }(\mathbf{z}_{t}',t  )-\varepsilon _{\theta, 2 } (\mathbf{z}_{t}',t  )\). 
    
    According to the CFG formulation in Eq. \eqref{17}, \(\varepsilon _{\theta, 2 } (\mathbf{z}_{t}',t  )\) can be expressed as
	\begin{equation}
		\varepsilon _{\theta,2 } (\mathbf{z}_{t}',t  )=\mu _{\theta }(\mathbf{z}_{t}',t,\mathbf{\emptyset},\mathbf{p} )+\gamma_{2}  [\mu _{\theta }(\mathbf{z}_{t}',t,\mathbf{c},\mathbf{p})-\mu _{\theta }(\mathbf{z}_{t}',t,\mathbf{\emptyset},\mathbf{p} ) ].\label{7}
	\end{equation}	 
	\indent Substituting Eq. \eqref{16} and Eq. \eqref{7} into \(\varepsilon _{\theta, 1 }(\mathbf{z}_{t}',t )-\varepsilon _{\theta, 2 }(\mathbf{z}_{t}',t  )\), we derive 
	\begin{align}
		\nonumber&\varepsilon _{\theta, 1 }(\mathbf{z}_{t}',t )-\varepsilon _{\theta, 2 }(\mathbf{z}_{t}',t  )\\=&(\gamma _{1} -\gamma _{2} )[\mu _{\theta }(\mathbf{z}_{t}',t,\mathbf{c},\mathbf{p})-\mu _{\theta }(\mathbf{z}_{t}',t,\mathbf{\emptyset},\mathbf{p} ) ].\label{9}
	\end{align}
	\indent Here, \(\gamma _{1}\) and \(\gamma _{2}\) represent the CFG scale used during inference and inversion, respectively. Since \(\gamma _{1}\) is always greater than \(\gamma _{2}\), we can conclude that the updated noise \(\tilde{\mathbf{z}}_{t} \) contains relative to \(\mathbf{z}_{t}\), an additional term reflecting the discrepancy between the predicted noise conditioned on the reference image and the unconditioned prediction. This discrepancy can be interpreted as the semantic information of the reference image injected into the noise.}
\end{document}


\maketitle 
	\section{Introduction to NVS Models}
	\subsection{SV3D}
\begin{figure*}[htbp] 
		\centering 
		\includegraphics[width=0.8\textwidth]{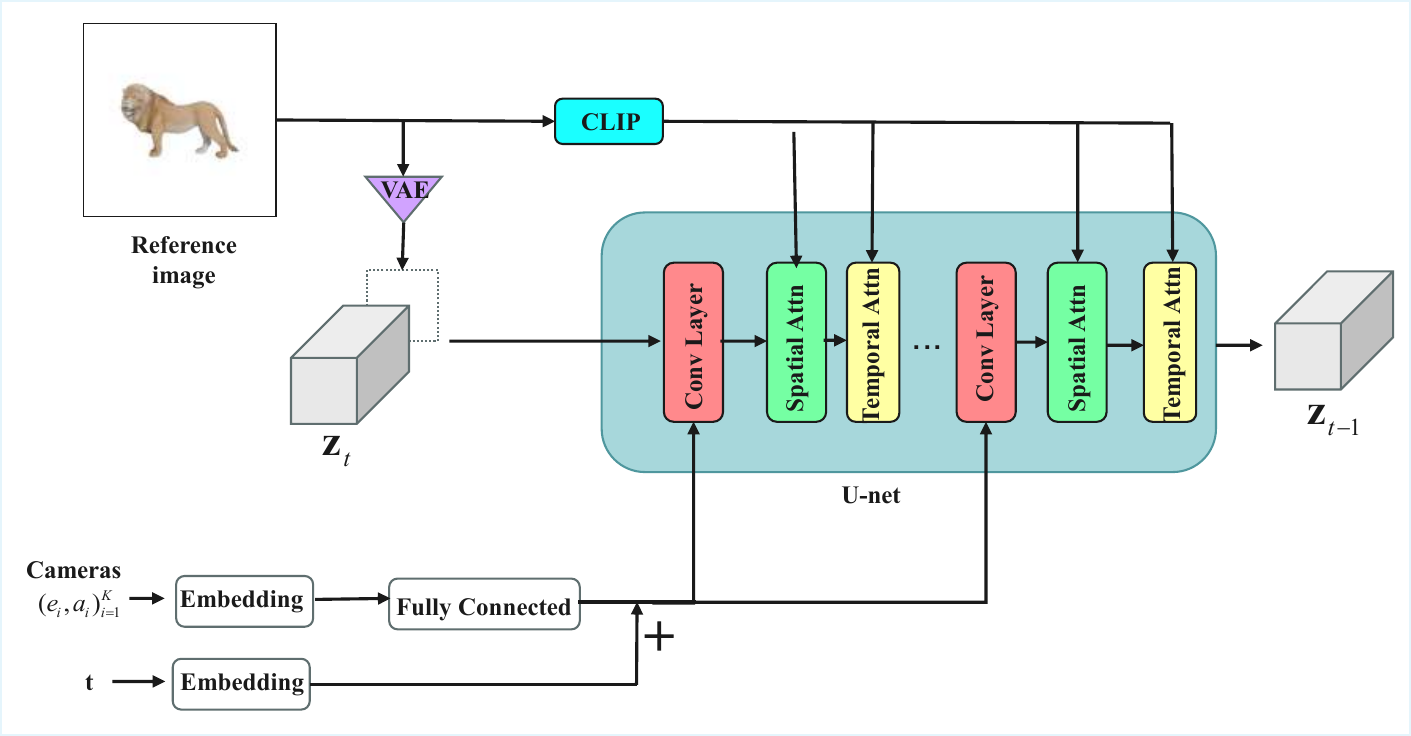} 
		\caption{SV3D architecture. The sinusoidal embeddings of the camera pose elevation and azimuth angles $(e, a)$, along with the timestep $t$, are summed and fed into the convolutional layers of the U-Net. The VAE embedding of the reference image is concatenated with the initial noise $\mathbf{z} _{t}$ and then passed into the U-net. Meanwhile, the CLIP embedding of the reference image is supplied to the spatial and temporal attention blocks within the U-Net.} 
		\label{sv3d_architecture} 
	\end{figure*}

	As shown in Fig. \ref{sv3d_architecture}, the architecture of SV3D \cite{voleti2024sv3d} is an enhanced version of SVD \cite{blattmann2023stable}. The SVD model adopts a multi-layer U-Net \cite{ronneberger2015u} structure, where each layer comprises a convolutional block, a spatial attention block, and a temporal attention block.
	
	During inference, SV3D employs a triangular wave–shaped CFG \cite{ho2022classifier} scaling strategy: the CFG scale starts from a small value at the front view, increases linearly to a larger value at the back view, and then decreases linearly back to the initial value when returning to the front view.
	
	For each inference, SV3D generates 21 images at a resolution of \(576\times 576\) from different viewpoints using a single reference image. Moreover, the elevation and azimuth angles of the camera poses are fully controlled.
	\subsection{Mv-Adapter}
\begin{figure*}[htbp] 
		\centering
		\includegraphics[width=0.8\textwidth]{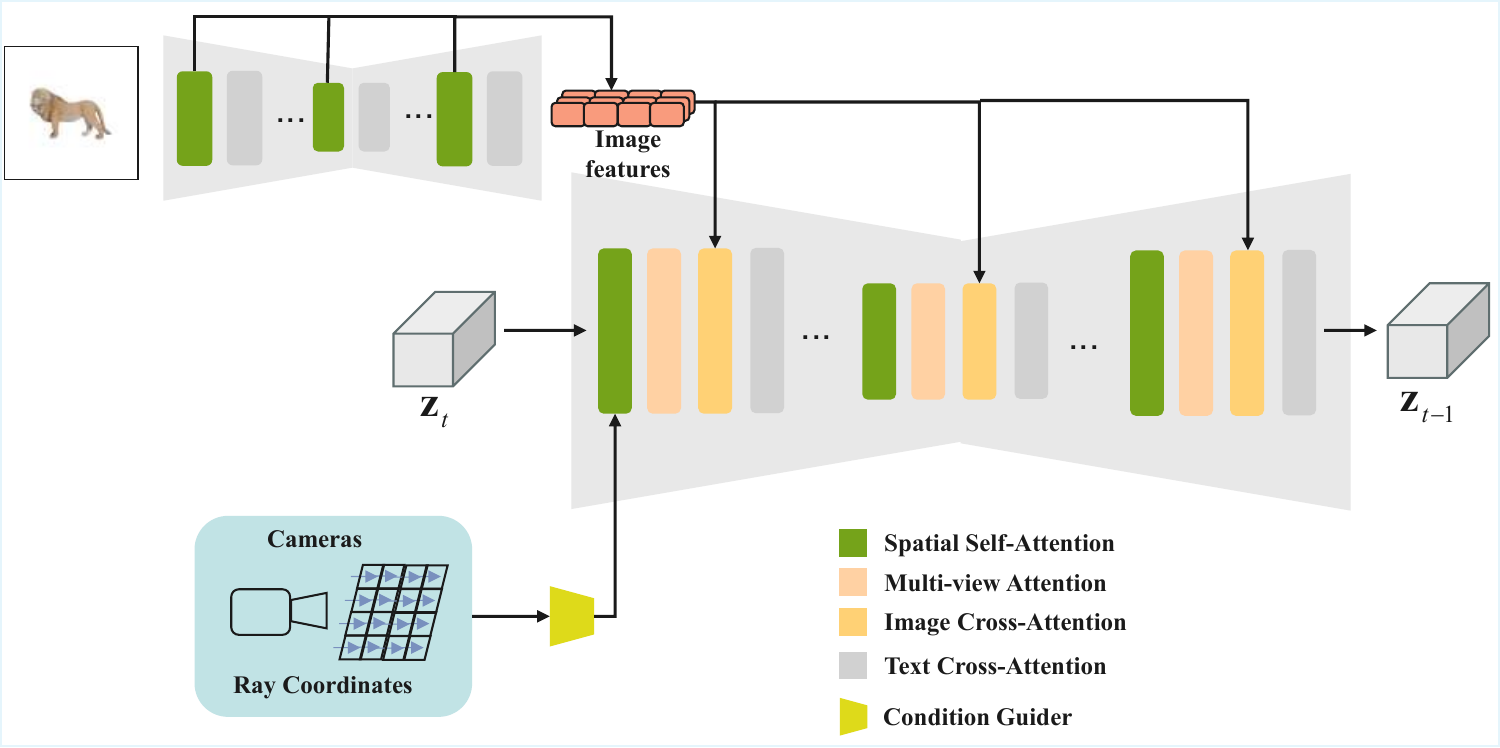} 
		\caption{Mv-Adapter architecture. After the VAE embedding of the reference image is processed by the original Stable Diffusion XL U-Net, the resulting multi-level image features are sequentially fed into the image cross-attention layers of the Mv-Adapter U-Net. Mv-Adapter controls the viewing angle of the synthesized novel view using a camera ray representation (``ray map''), which is processed by the condition guider before being injected into the Mv-Adapter U-Net. } 
		\label{mvadapter_architecture} 
	\end{figure*}

	As shown in Fig. \ref{mvadapter_architecture}, the structure of Mv-Adapter \cite{huang2025mv} is built upon the Stable Diffusion XL \cite{podellsdxl} with additional multi-view attention layers and image cross-attention layers inserted after the original spatial self-attention layers and text cross-attention layers.
	
	During inference, Mv-Adapter can take text conditions as input. For consistency with other novel view synthesis models, we set the text condition to empty. 
	
	For each inference, Mv-Adapter generates 6 images of size \(768\times 768\) from different viewpoints using a single reference image.
	\section{Encoder-Decoder Network Details}
	\label{Encoder-Decoder Network Details}
\begin{figure*}[htb] 
		\centering 
		\includegraphics[width=0.8\textwidth]{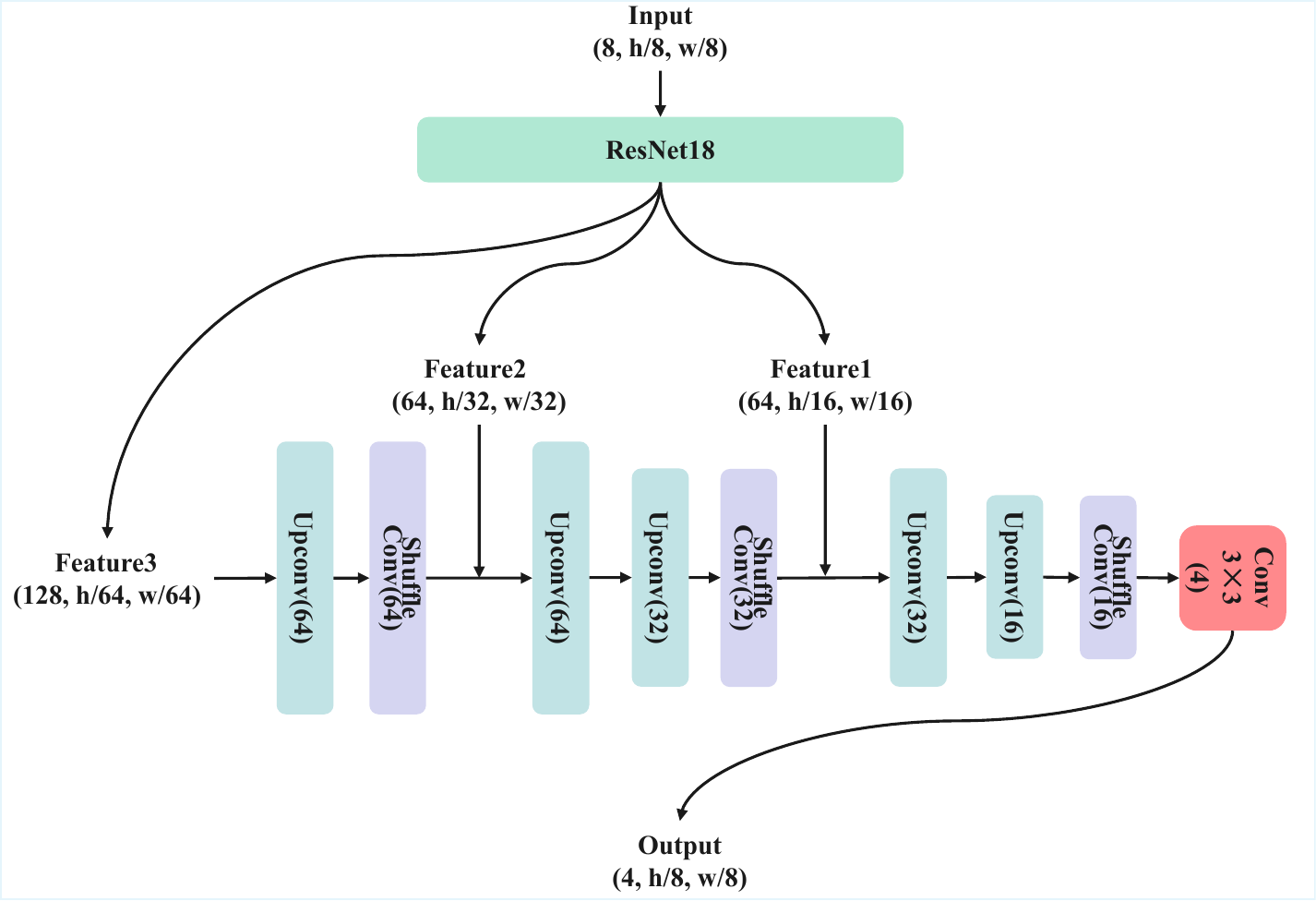} 
		\caption{EDN architecture receives the concatenation of the VAE embedding of the reference image and the initial random Gaussian noise as input. The combined tensor is processed by the encoder to obtain three feature maps, which are then decoded to produce the final output.} 
		\label{EDN_architecture}
	\end{figure*}

	The EDN takes as input the VAE embedding of the reference image \((4,h/8,w/8)\) and the initial random Gaussian noise of the same shape \((4,h/8,w/8)\), concatenated along the channel dimension, to form a tensor of shape \((8,h/8,w/8)\). The encoder processes this tensor to produce three feature maps with shapes \((64,h/16,w/16)\), \((64,h/32,w/32)\), and \((128,h/64,w/64)\). These feature maps are then fed into the decoder, which outputs a tensor of shape \((4,h/8,w/8)\) representing the predicted image semantic information. The detailed architecture is illustrated in Fig. \ref{EDN_architecture}. The following subsections describe the encoder and decoder structures in detail.
	\subsection{Encoder Structure}
	We adopt ResNet18 \cite{he2016deep} as the encoder, which primarily consists of Conv2d, BatchNorm2d, MaxPool, and ReLU layers. For generating the second and third feature maps, a parallel branch structure is used, with the outputs of the two branches summed together.
	\subsection{Decoder Structure}
	The decoder primarily consists of Conv2D, ELU, and PixelShuffle \cite{shi2016real} layers. The third feature map is used as the initial input, and the second and first feature maps are progressively concatenated with intermediate results along the channel dimension.
	\section{Datasets}
	\textbf{Objaverse} \cite{deitke2023objaverse} is a large-scale open dataset of approximately 800,000 3D objects, each accompanied by text descriptions. 
	
	\textbf{GSO} \cite{downs2022google} is a high-quality dataset of 1,030 commonly used household items across 17 categories, captured via 3D scanning.
	
	\textbf{OmniObject3D} \cite{wu2023omniobject3d} is a high-quality, large-scale dataset, which contains approximately 6000 real-scanned 3D objects covering 190 daily categories. 
	\section{Additional Experiment Results}
	\label{Additional Experiment Results}
	In this section, we will supplement the results of the relevant experiments.
	\begin{table}[htbp]
		\caption{Ablation experiments on static orbits of GSO dataset on SV3D.}
		\label{Ablation_Studies_datasets}
		\centering
		\begin{tabular}{cccc}
			\toprule
			Method&PSNR\(\uparrow \)&SSIM\(\uparrow \)&LPIPS\(\downarrow  \)\\
			\midrule 
			standard&20.349&0.8909&0.1364\\
			\cmidrule{1-4}
			EDN w/o elevation w/o filter&21.228&0.9033&0.1233\\
			\cmidrule{1-4}
			EDN w elevation w/o filter&21.563&0.9066&0.1215\\
			\cmidrule{1-4}
			EDN w/o elevation w filter&21.596&0.9059&0.1208\\
			\cmidrule{1-4}
			EDN w sine-based pose embedding&20.836&0.9010&0.1248\\
			\cmidrule{1-4}
			EDN w ray map&20.655&0.9001&0.1273\\
			\cmidrule{1-4}
			EDN w transposed conv&21.220&0.9013&0.1271\\
			\cmidrule{1-4}
			EDN&21.659&0.9070&0.1208\\
			\bottomrule
		\end{tabular}
	\end{table} 
	\begin{table}[htbp]
		\caption{Effect of initial noise across different random seeds on SV3D under static orbits on GSO dataset.}
		\label{compare_with_different_seeds}
		\centering
		\begin{tabular}{ccccc}
			\toprule
			Seed&Method&PSNR\(\uparrow \)&SSIM\(\uparrow \)&LPIPS\(\downarrow  \)\\
			\midrule
			\multirow{2}{*}{\raisebox{0.005\totalheight}[0pt][0pt]{\makecell{801--900\\(Original)}}}&standard&20.441&0.8912&0.1374 \\
			&with EDN&21.607&0.9075&0.1226\\
			\cmidrule{1-5}
			&standard&20.461&0.8917&0.1354\\
			\multirow{2}{*}{\raisebox{1.5\totalheight}[0pt][0pt]{1851--1950}}
			&with EDN&21.704&0.9074&0.1212\\
			\cmidrule{1-5}
			&standard&20.525&0.8937&0.1340\\
			\multirow{2}{*}{\raisebox{1.5\totalheight}[0pt][0pt]{5851--5950}}&with EDN&21.597&0.9073&0.1206\\
			\bottomrule
		\end{tabular}
	\end{table} 
	\begin{table}[htbp]
		\caption{Effect of initial noise on SV3D over the dynamic orbits of three datasets.}
		\label{compare_with_dynamic_orbit}
		\centering
		\begin{tabular}{ccccc}
			\toprule
			Dataset&Method&PSNR\(\uparrow \)&SSIM\(\uparrow \)&LPIPS\(\downarrow  \)\\
			\midrule
			&standard&19.823&0.8868&0.1442 \\
			GSO&inversion&18.212&0.8737&0.1471\\
			&with EDN&21.231&0.9045&0.1277\\
			\cmidrule{1-5}
			&standard&21.879&0.9126&0.1184\\
			Objaverse&inversion&19.622&0.9010&0.1235\\
			&with EDN&23.314&0.9325&0.0996\\
			\cmidrule{1-5}
			&standard&18.383&0.8780&0.1747\\
			OmniObject3D&inversion&17.240&0.8637&0.1812\\
			&with EDN&19.794&0.9059&0.1398\\
			\bottomrule
		\end{tabular}
	\end{table}
	\begin{table}[htbp]
		\caption{Effect of data filtering threshold on SV3D under static orbits on GSO dataset.} 
		\label{compare_with_different_m}
		\centering
		\begin{tabular}{ccccc}
			\toprule
			Threshold&Filtering rate(\%)&PSNR\(\uparrow \)&SSIM\(\uparrow \)&LPIPS\(\downarrow  \)\\
			\midrule
			0.000&20.34&21.659&0.9070&0.1208\\
			0.005&8.16&21.552&0.9059&0.1225\\
			0.010&4.31&21.537&0.9066&0.1218\\
			\bottomrule
		\end{tabular}
	\end{table} 
	\begin{table}[htbp]
		\caption{Effect of training dataset size on SV3D under static orbits on GSO dataset.}
		\label{compare_with_different_select}
		\centering
		\begin{tabular}{ccccc}
			\toprule
			Method&Size&PSNR\(\uparrow \)&SSIM\(\uparrow \)&LPIPS\(\downarrow  \)\\
			\midrule 
			standard&&20.349&0.8909&0.1364\\
			EDN&90&21.167&0.9017&0.1266\\
			EDN&180&21.537&0.9051&0.1226\\
			EDN&359&21.659&0.9070&0.1208\\
			\bottomrule
		\end{tabular}
	\end{table} 
	\bibliography{reference_supply}
	\begin{figure*}[htbp]
		\centering
		\fontsize{9pt}{12pt}\selectfont  
		\begin{tabular}{c c c c c c}
			&SV3D& & &Mv-Adapter&\\
			standard&
			\begin{subfigure}[c]{0.15\textwidth}  
				\centering
				\includegraphics[width=\linewidth]{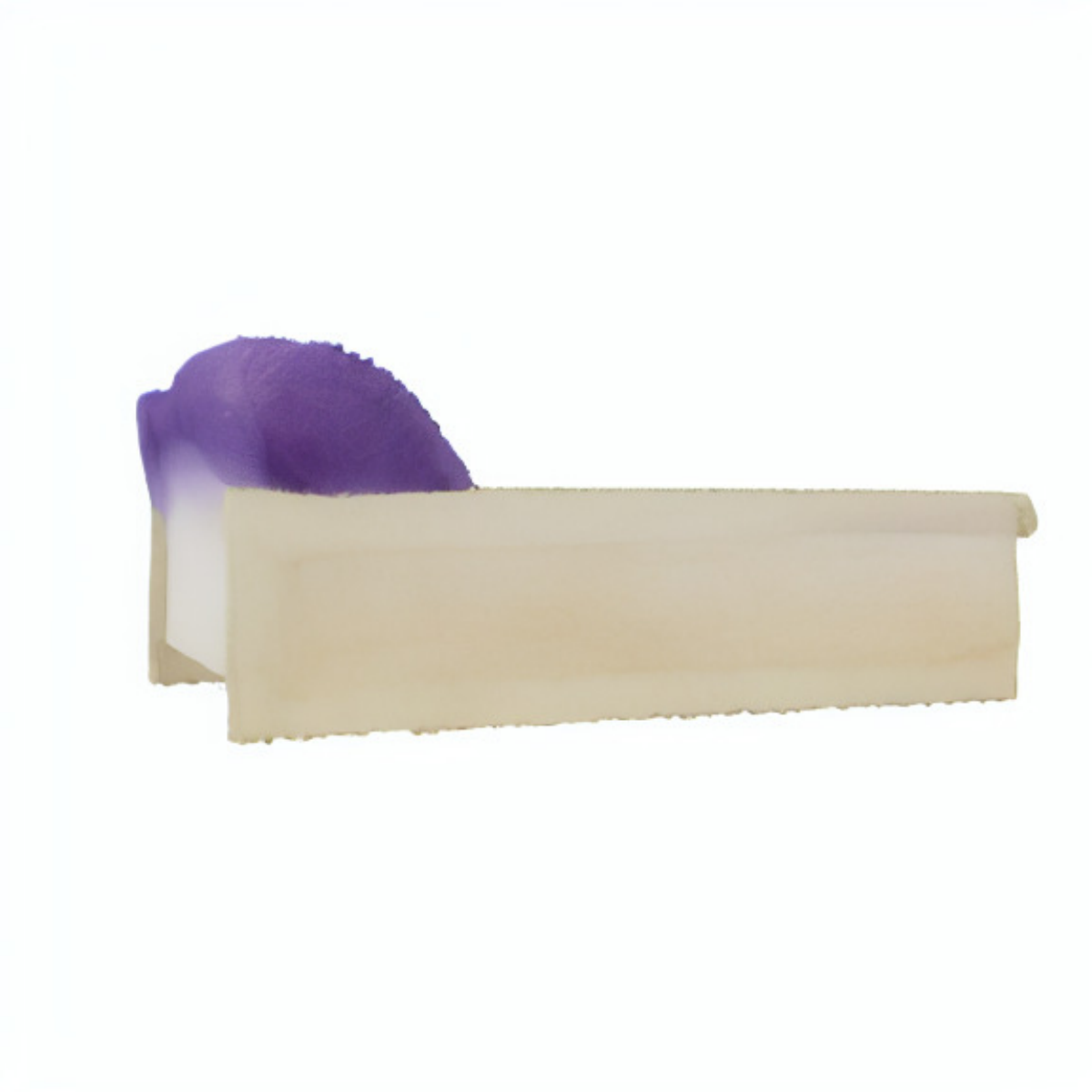}
			\end{subfigure} &
			\begin{subfigure}[c]{0.15\textwidth}  
				\centering
				\includegraphics[width=\linewidth]{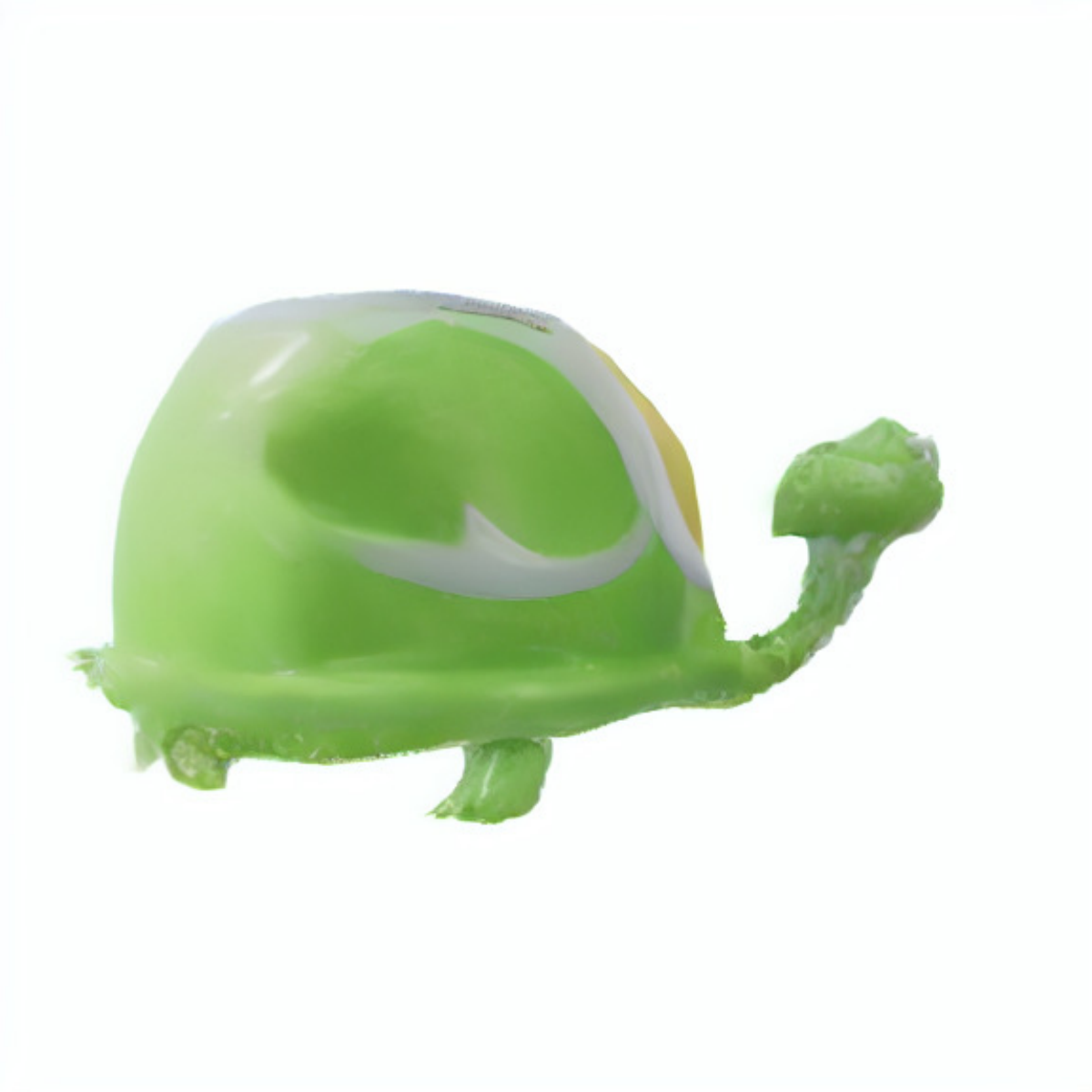}
			\end{subfigure} &
			standard&
			\begin{subfigure}[c]{0.15\textwidth}  
				\centering
				\includegraphics[width=\linewidth]{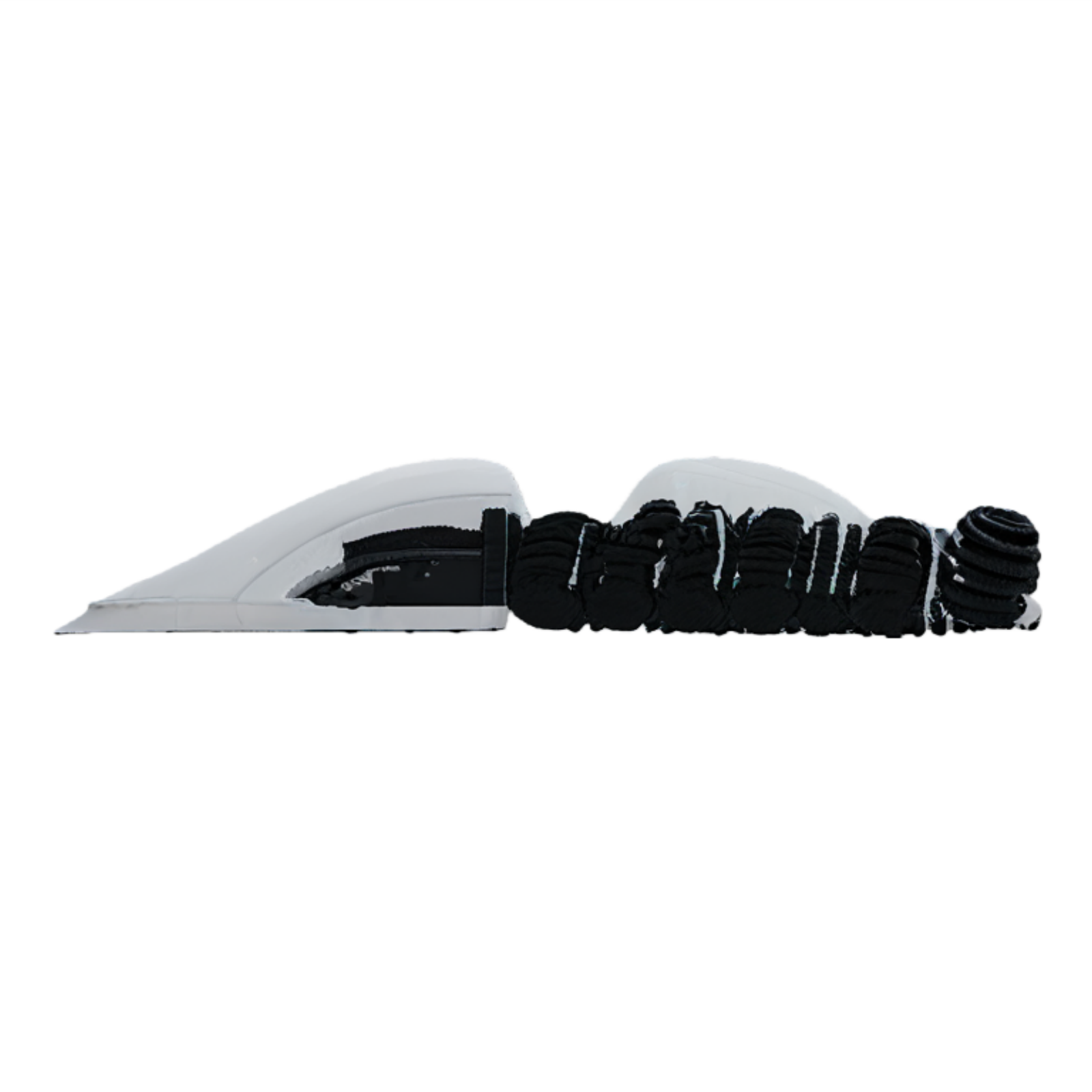}
			\end{subfigure} &
			\begin{subfigure}[c]{0.15\textwidth}  
				\centering
				\includegraphics[width=\linewidth]{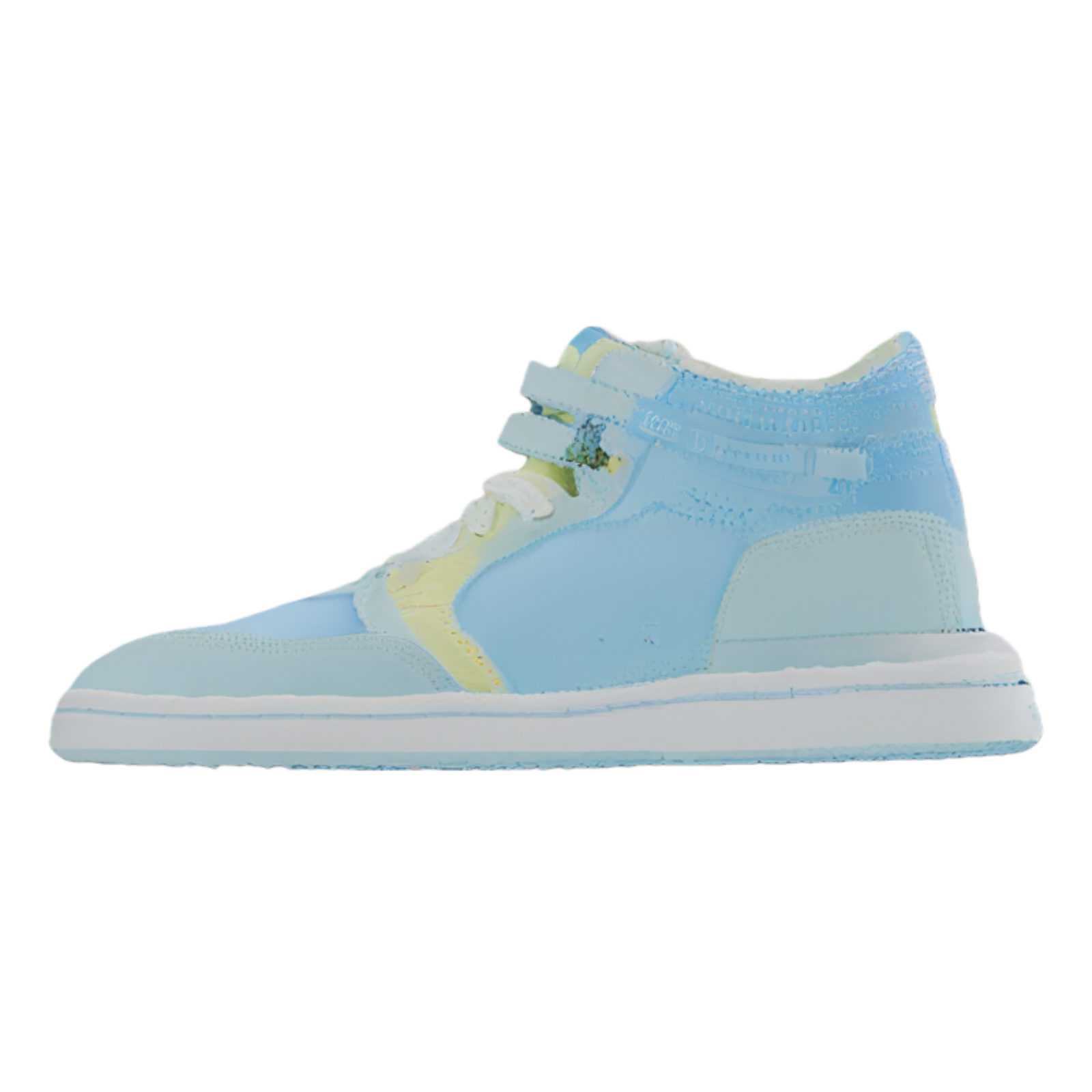}
			\end{subfigure}  \\
			n = 2&
			\begin{subfigure}[c]{0.15\textwidth}  
				\centering
				\includegraphics[width=\linewidth]{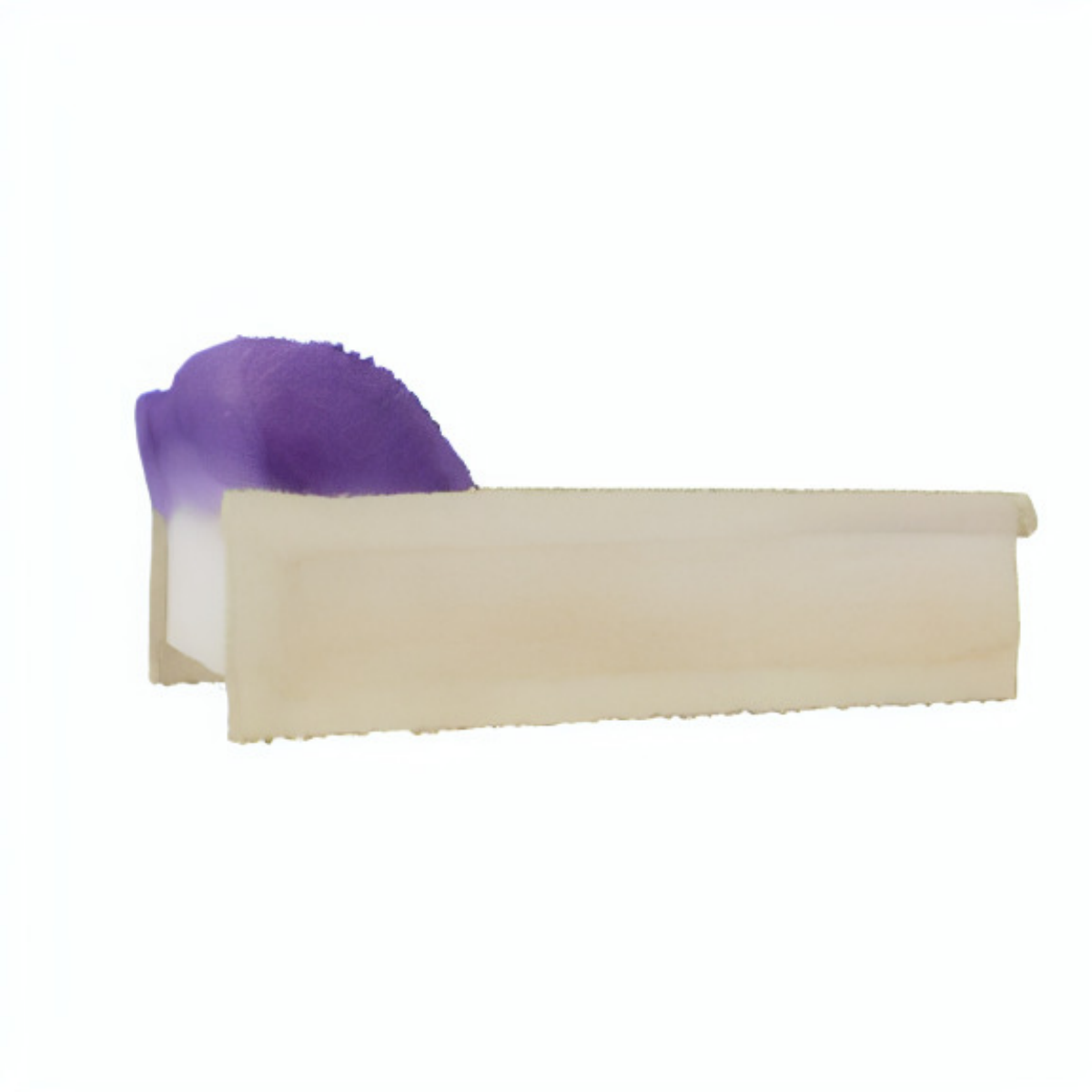}
			\end{subfigure} &
			\begin{subfigure}[c]{0.15\textwidth}  
				\centering
				\includegraphics[width=\linewidth]{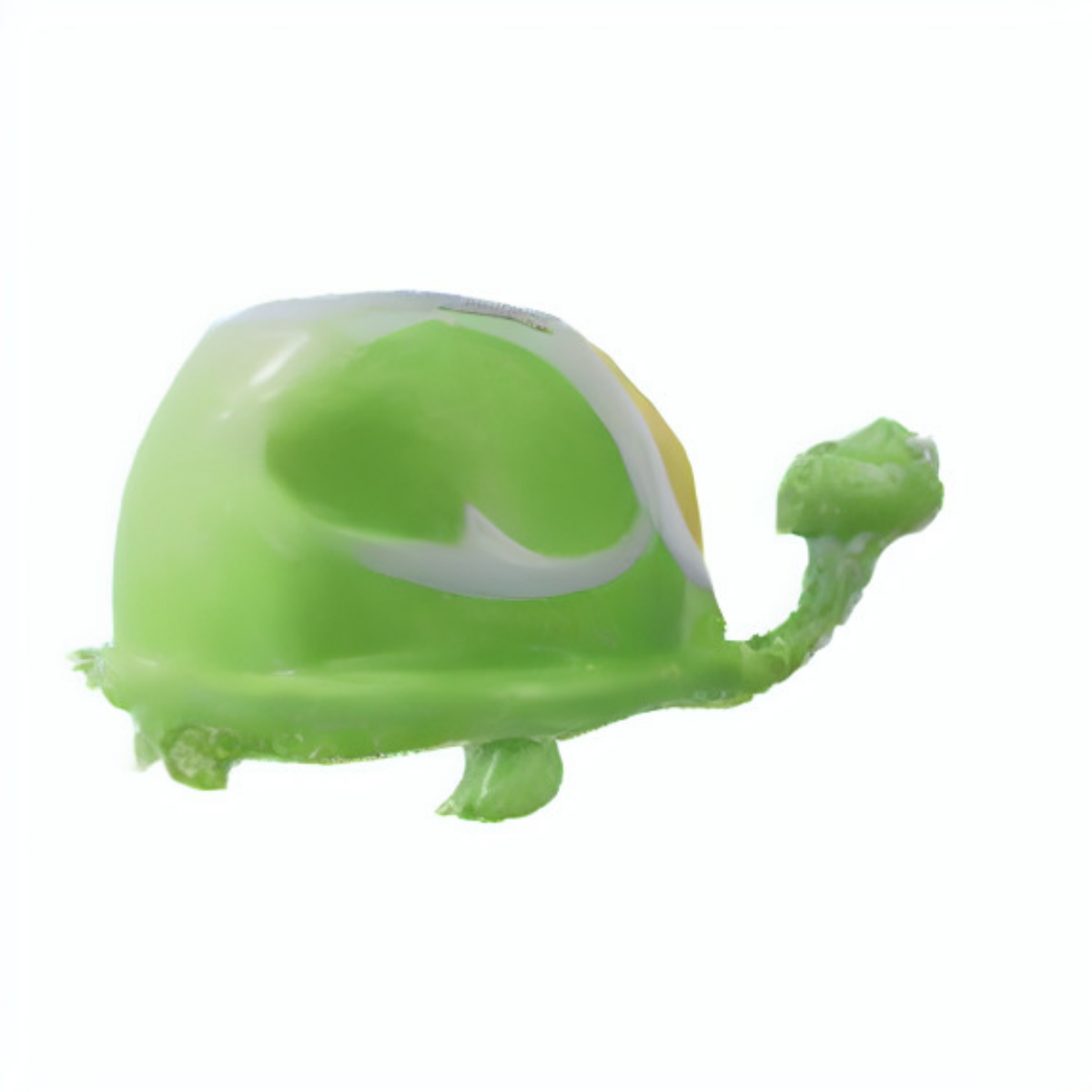}
			\end{subfigure} &
			n =5&
			\begin{subfigure}[c]{0.15\textwidth}  
				\centering
				\includegraphics[width=\linewidth]{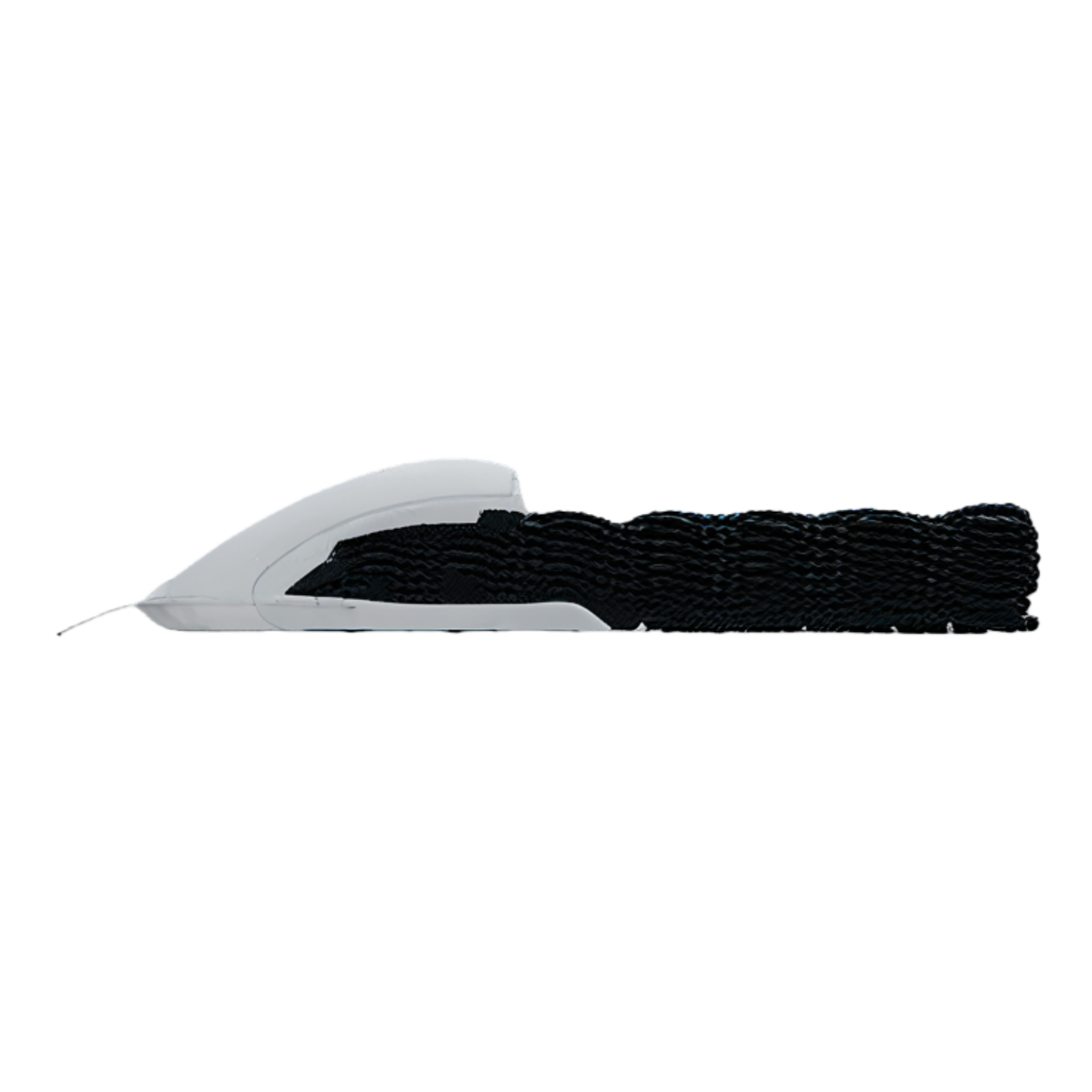}
			\end{subfigure} &
			\begin{subfigure}[c]{0.15\textwidth}  
				\centering
				\includegraphics[width=\linewidth]{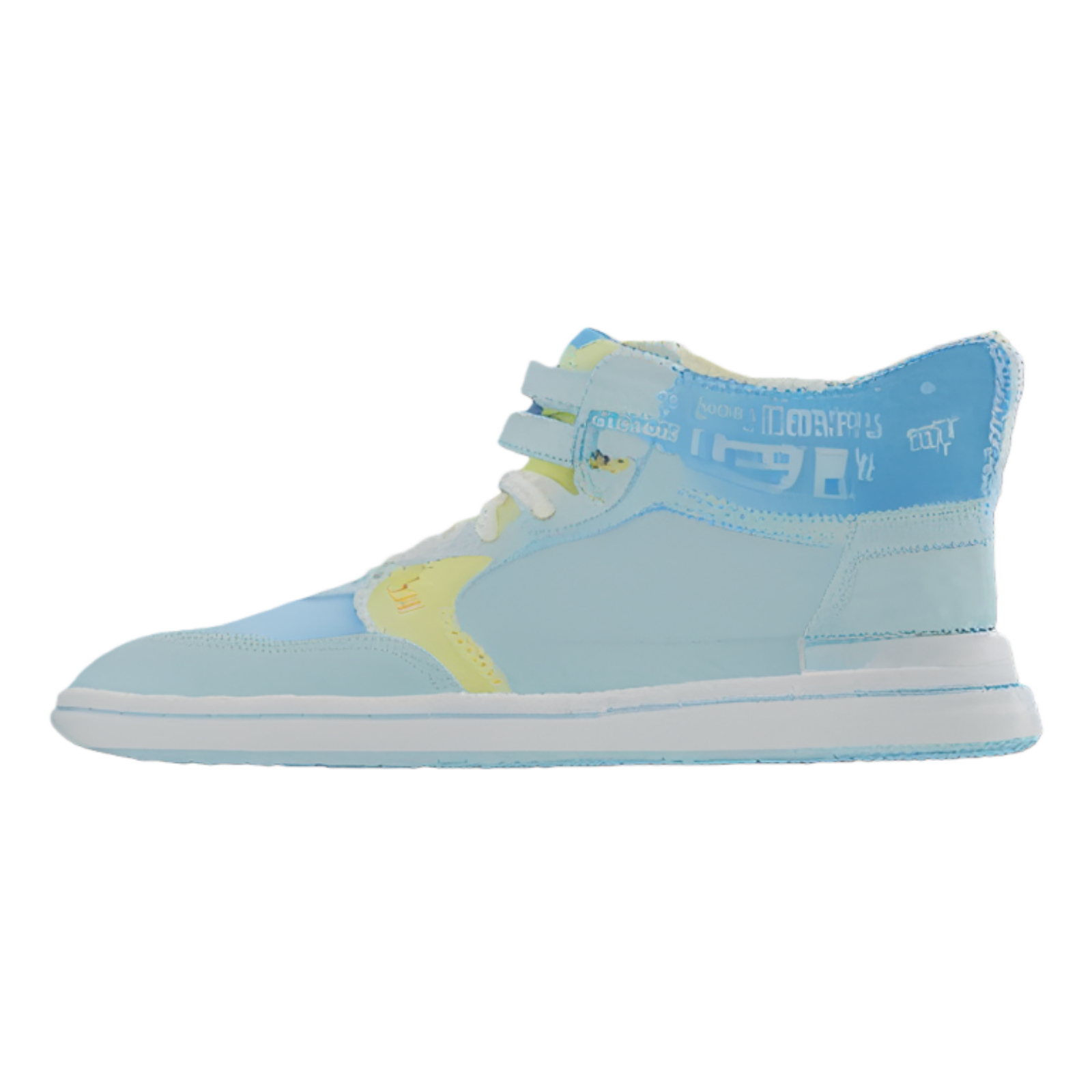}
			\end{subfigure}  \\
			n = 8&
			\begin{subfigure}[c]{0.15\textwidth}  
				\centering
				\includegraphics[width=\linewidth]{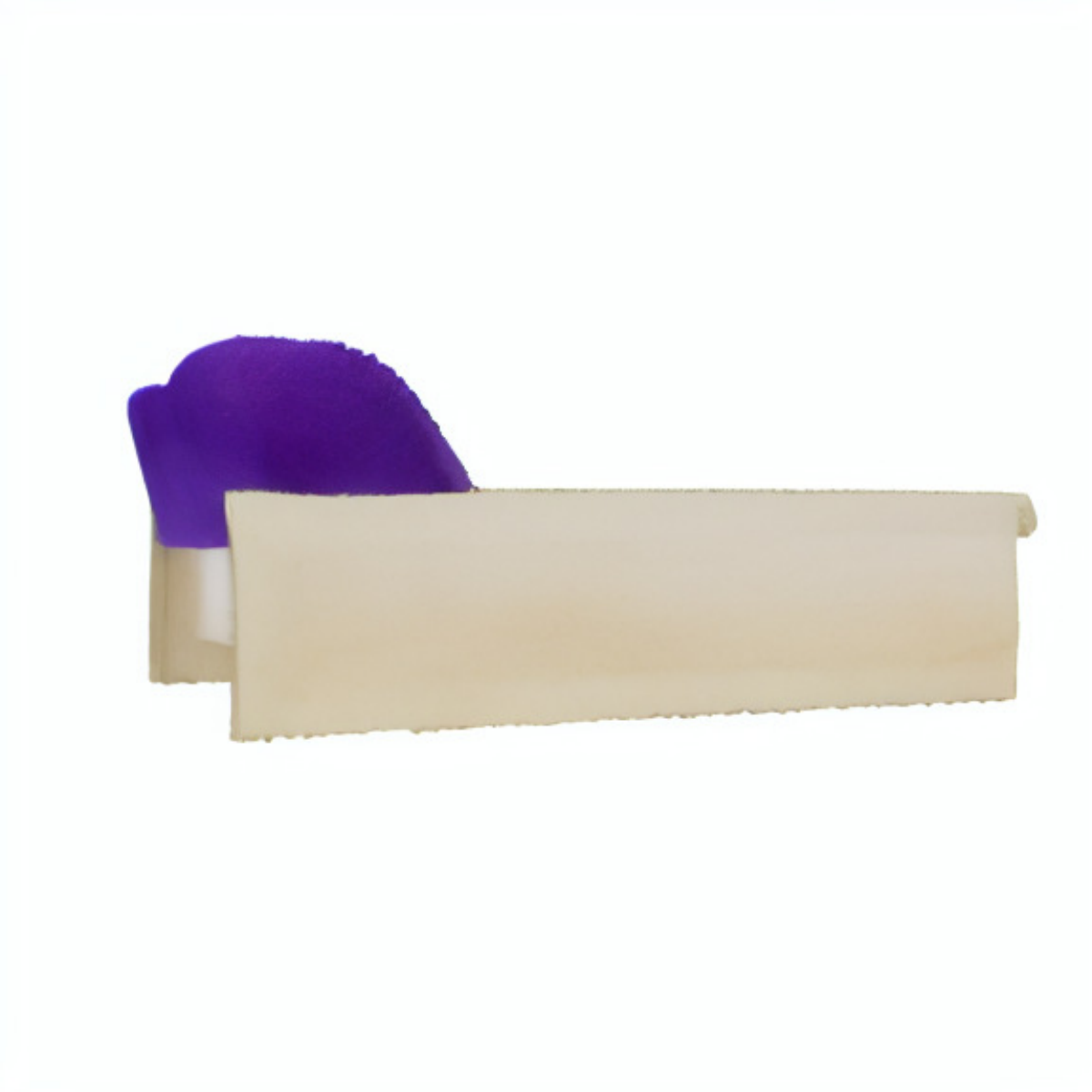}
			\end{subfigure} &
			\begin{subfigure}[c]{0.15\textwidth}  
				\centering
				\includegraphics[width=\linewidth]{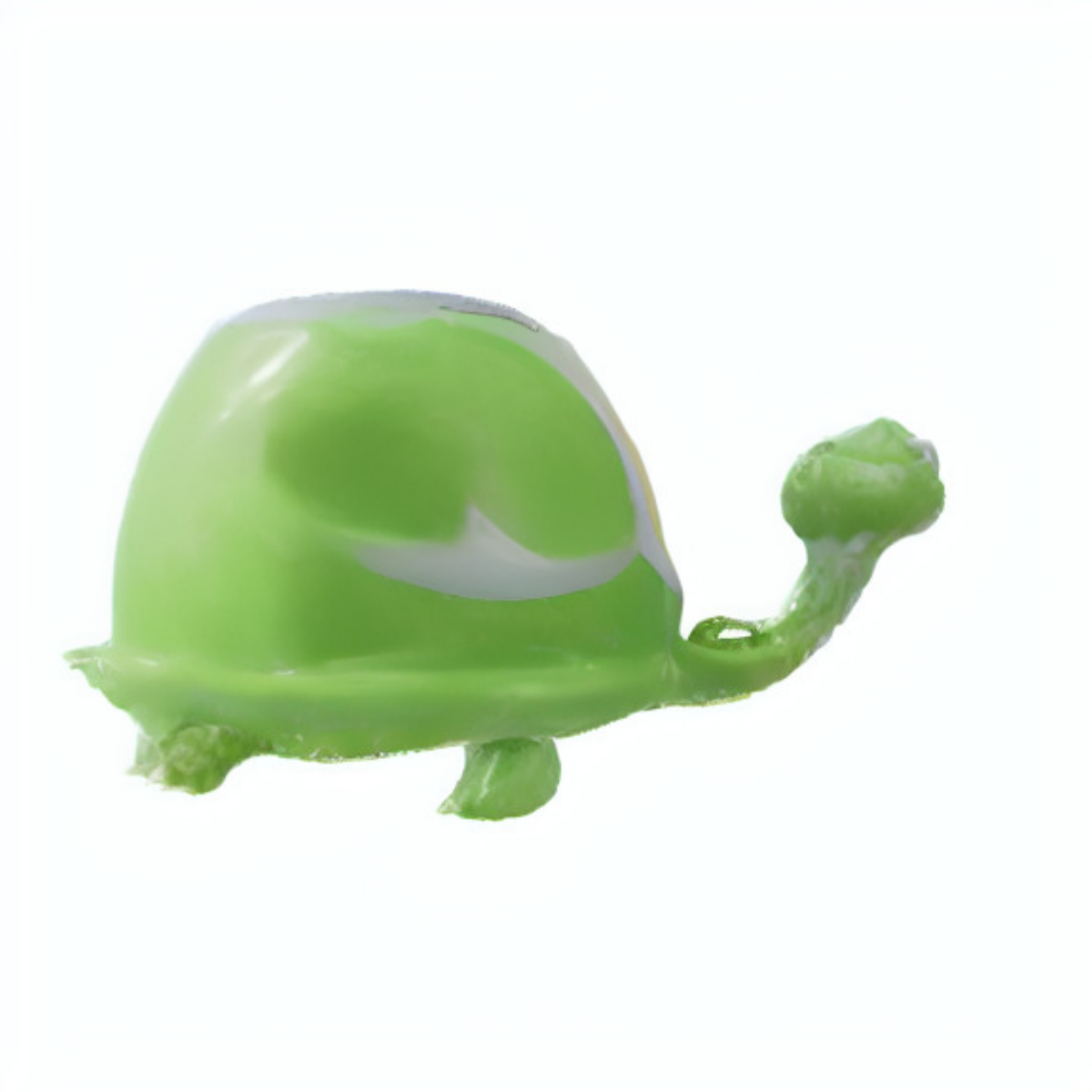}
			\end{subfigure} &
			n =10&
			\begin{subfigure}[c]{0.15\textwidth}  
				\centering
				\includegraphics[width=\linewidth]{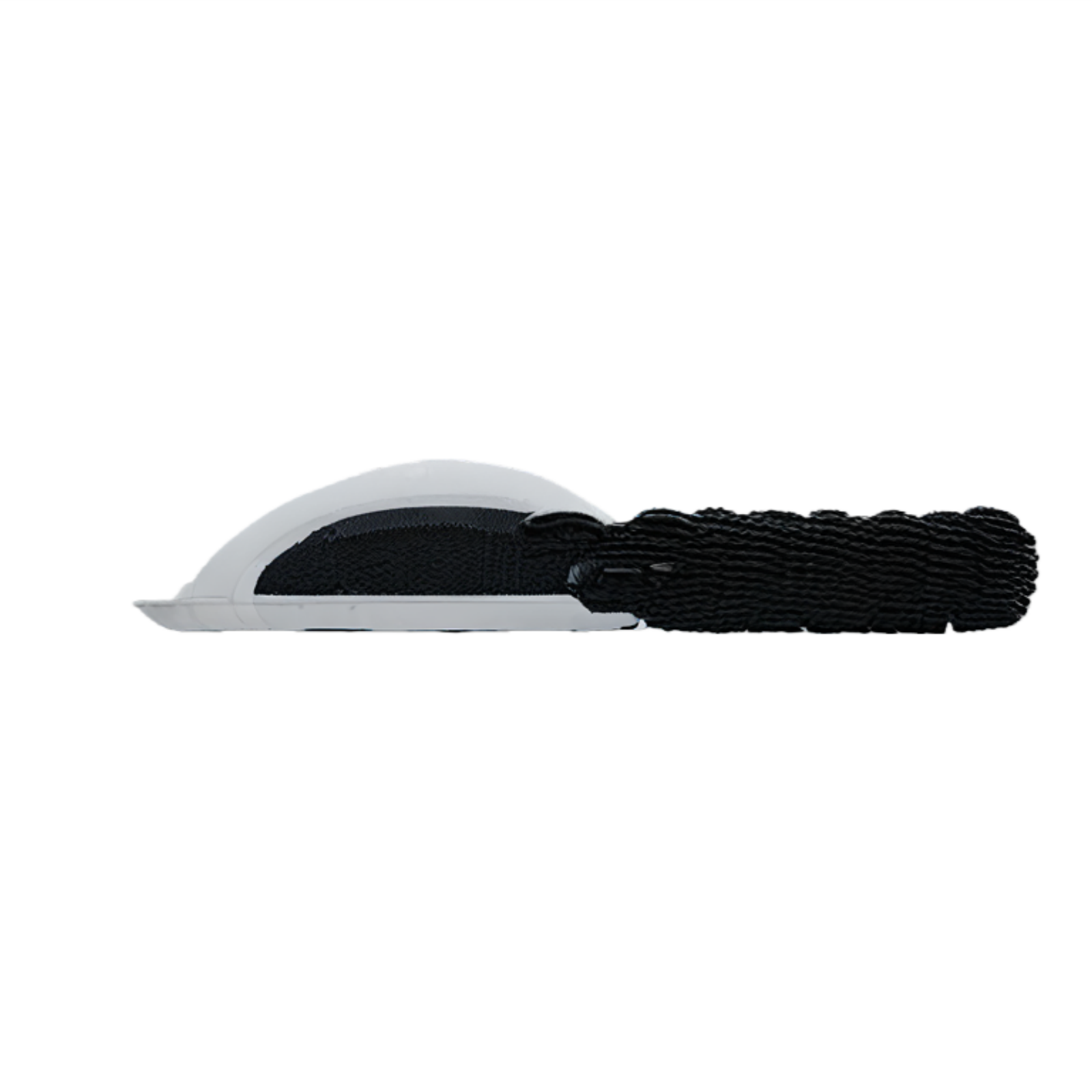}
			\end{subfigure} &
			\begin{subfigure}[c]{0.15\textwidth}  
				\centering
				\includegraphics[width=\linewidth]{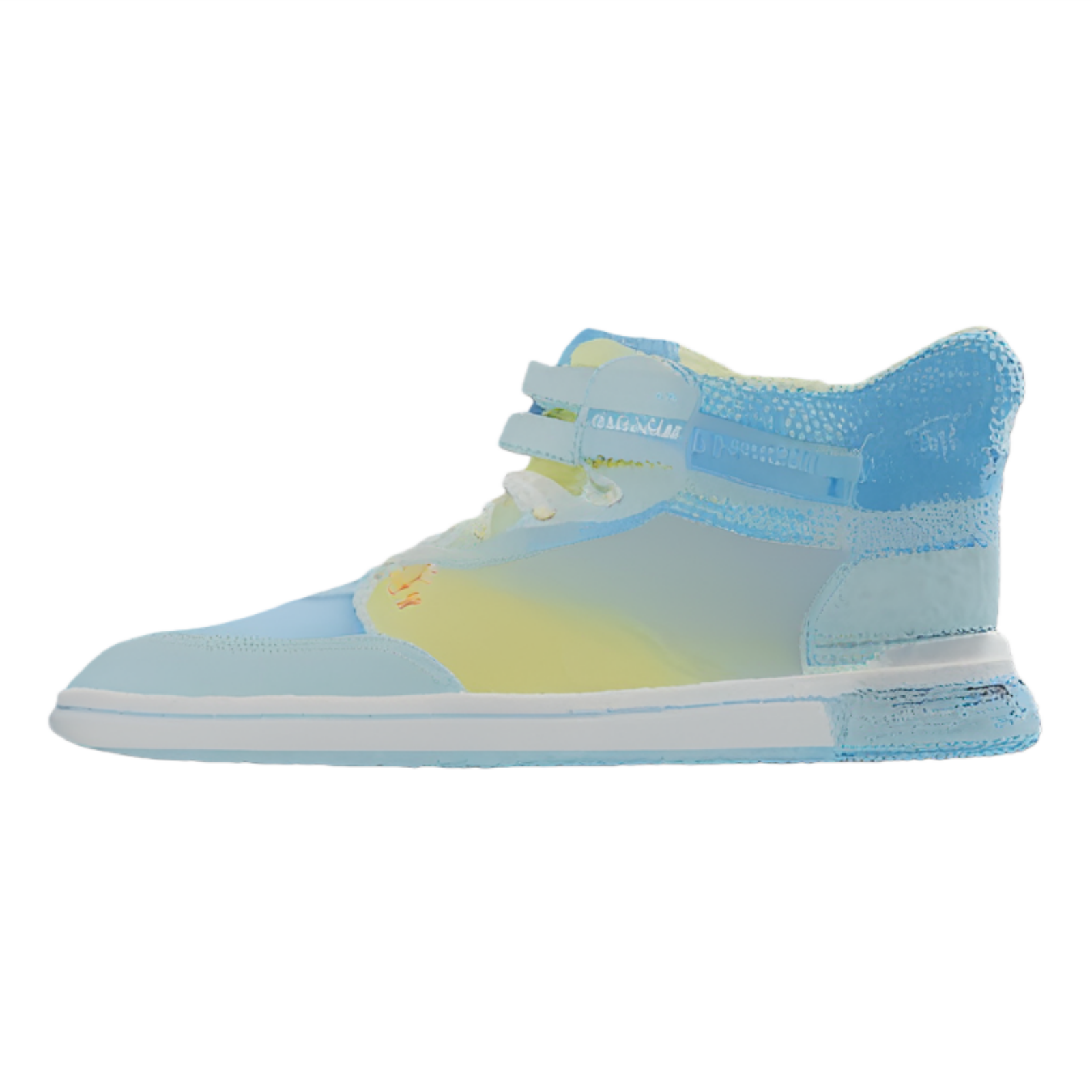}
			\end{subfigure}  \\
			n = 16&
			\begin{subfigure}[c]{0.15\textwidth}  
				\centering
				\includegraphics[width=\linewidth]{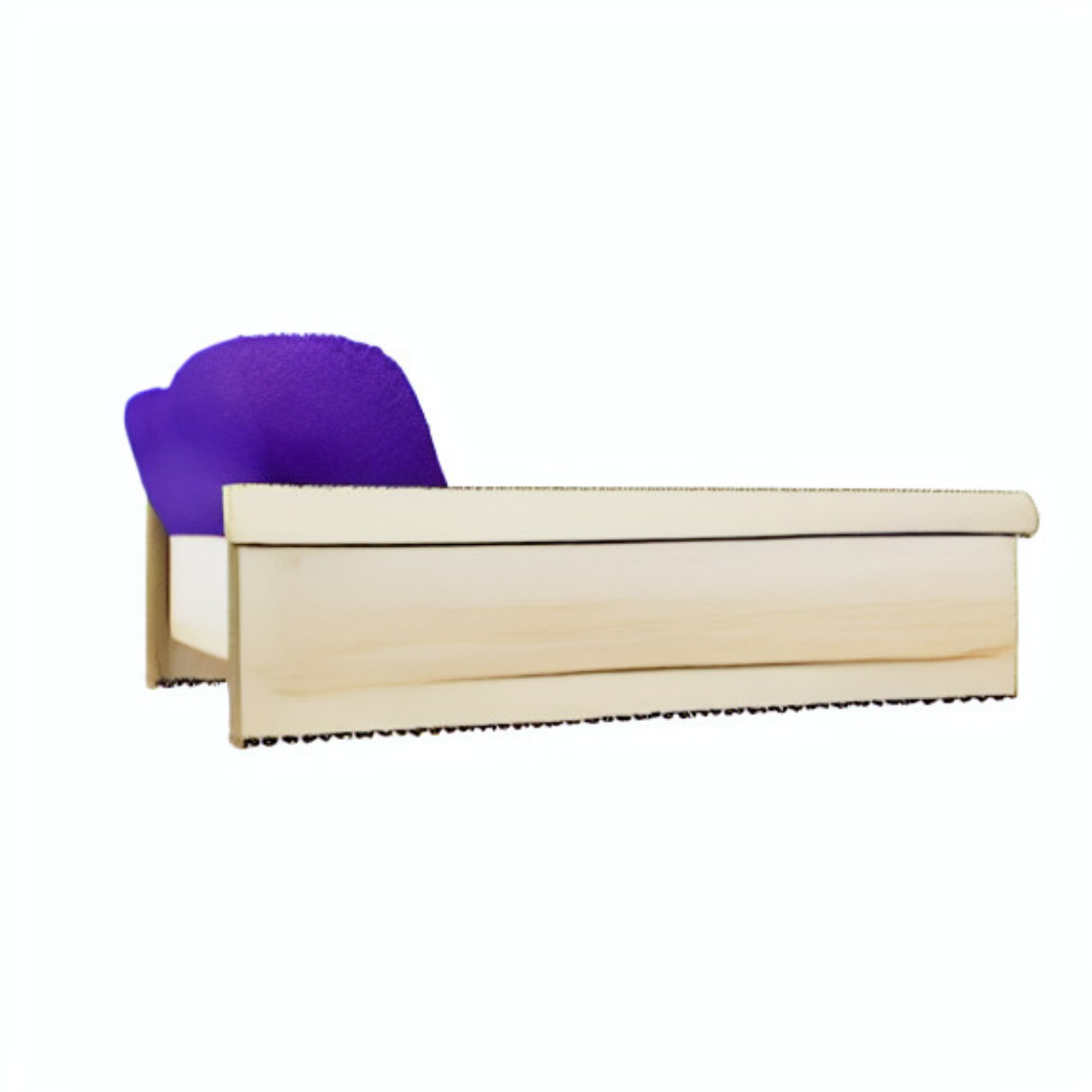}
			\end{subfigure} &
			\begin{subfigure}[c]{0.15\textwidth}  
				\centering
				\includegraphics[width=\linewidth]{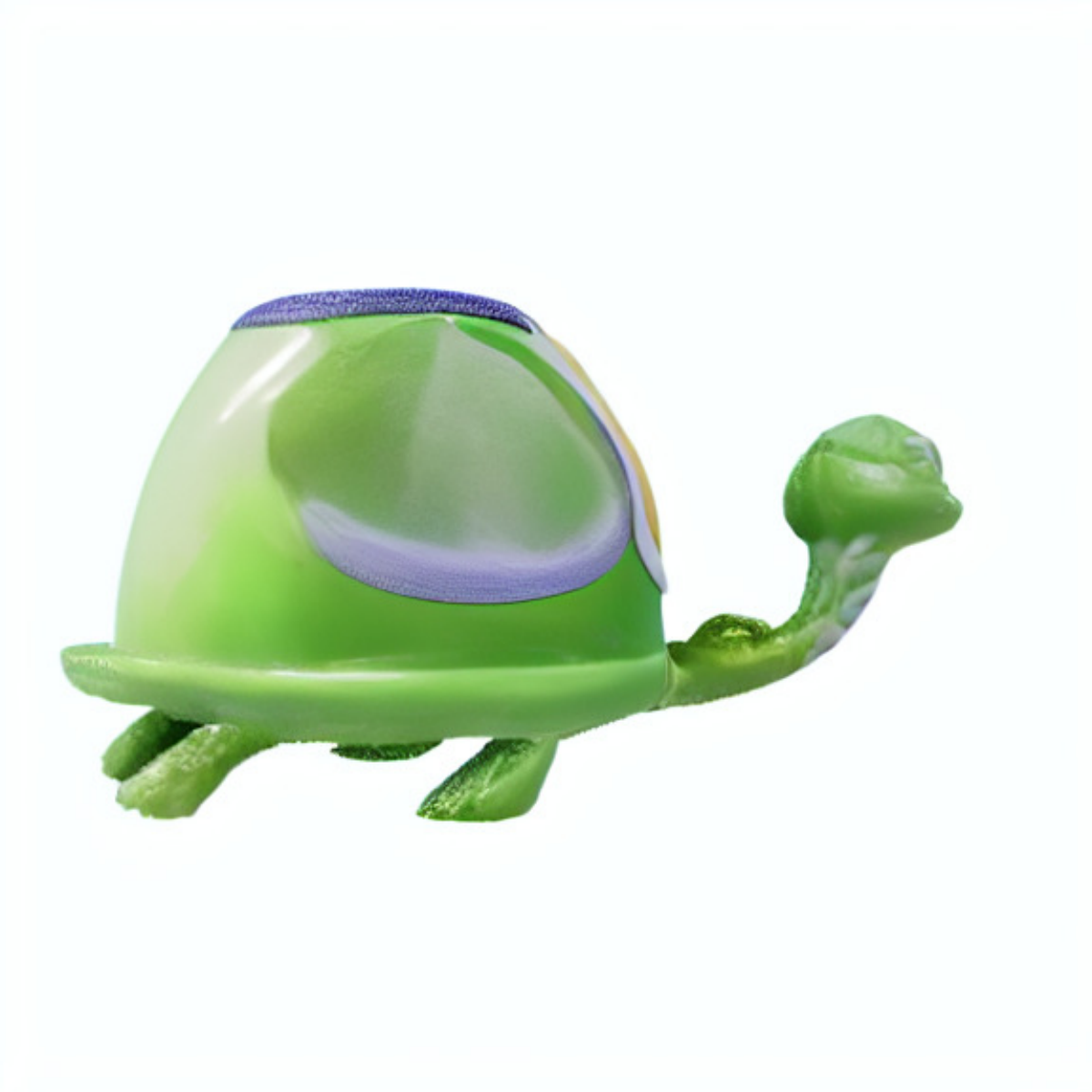}
			\end{subfigure} &
			n =25&
			\begin{subfigure}[c]{0.15\textwidth}  
				\centering
				\includegraphics[width=\linewidth]{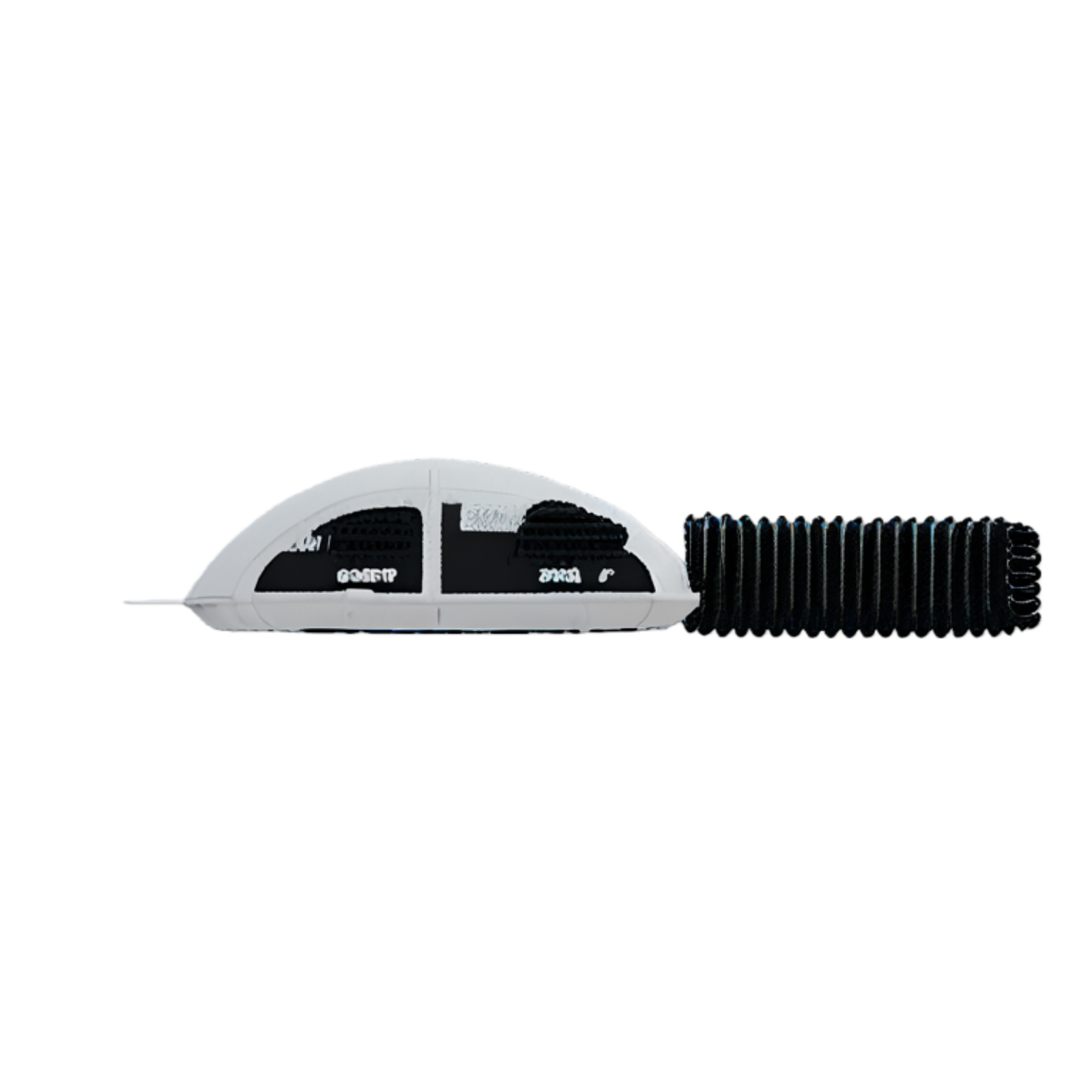}
			\end{subfigure} &
			\begin{subfigure}[c]{0.15\textwidth}  
				\centering
				\includegraphics[width=\linewidth]{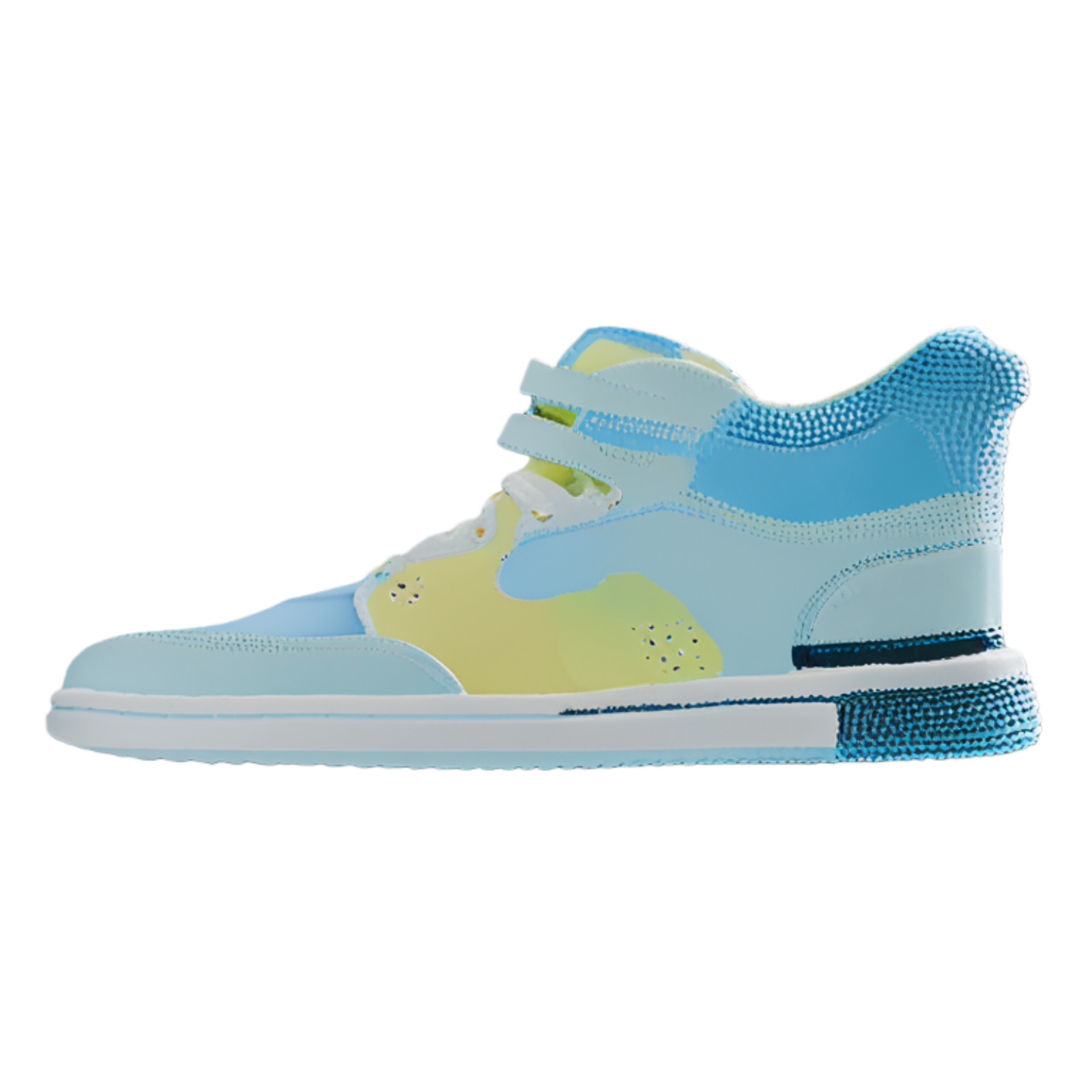}
			\end{subfigure}  \\
			n = 24&
			\begin{subfigure}[c]{0.15\textwidth}  
				\centering
				\includegraphics[width=\linewidth]{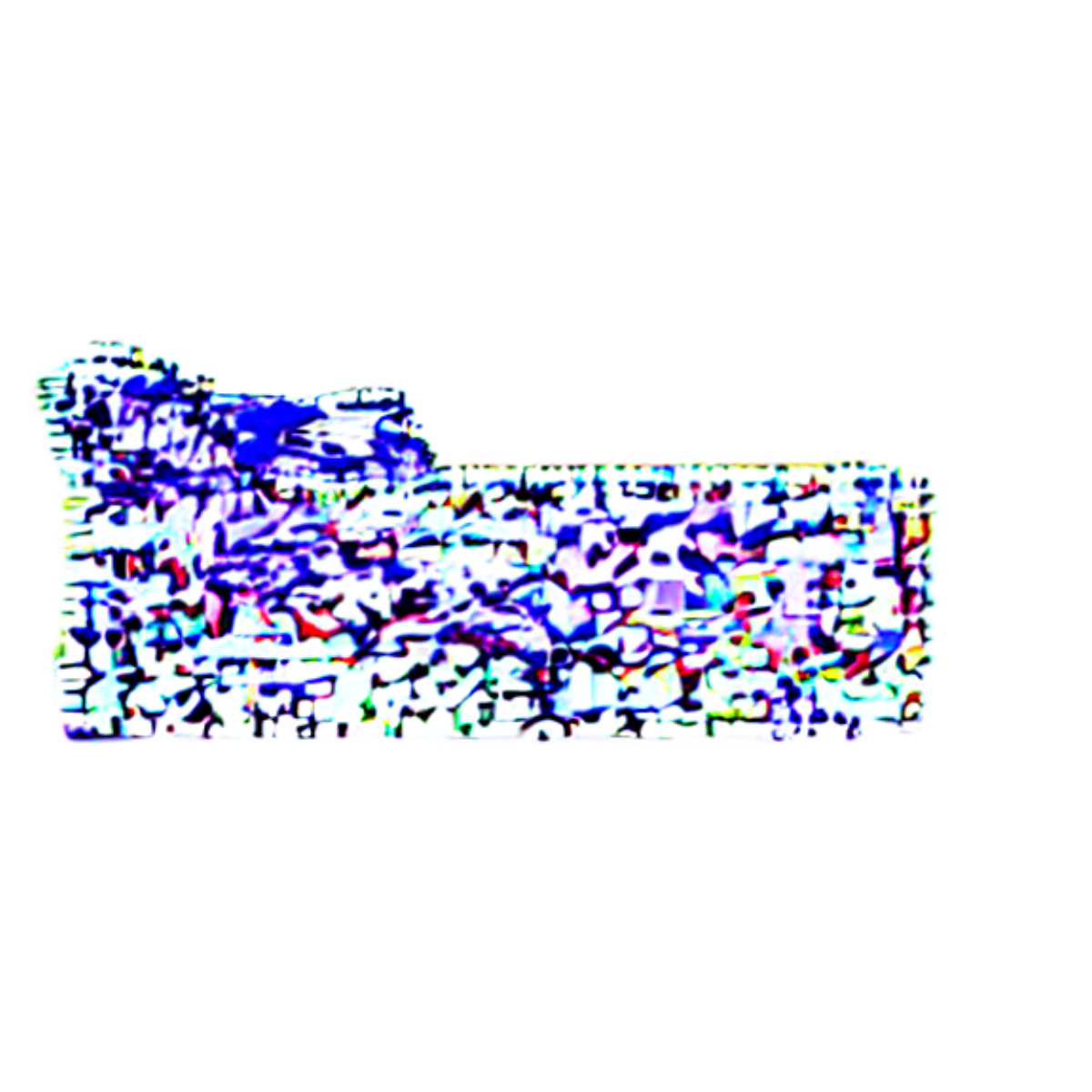}
			\end{subfigure} &
			\begin{subfigure}[c]{0.15\textwidth}  
				\centering
				\includegraphics[width=\linewidth]{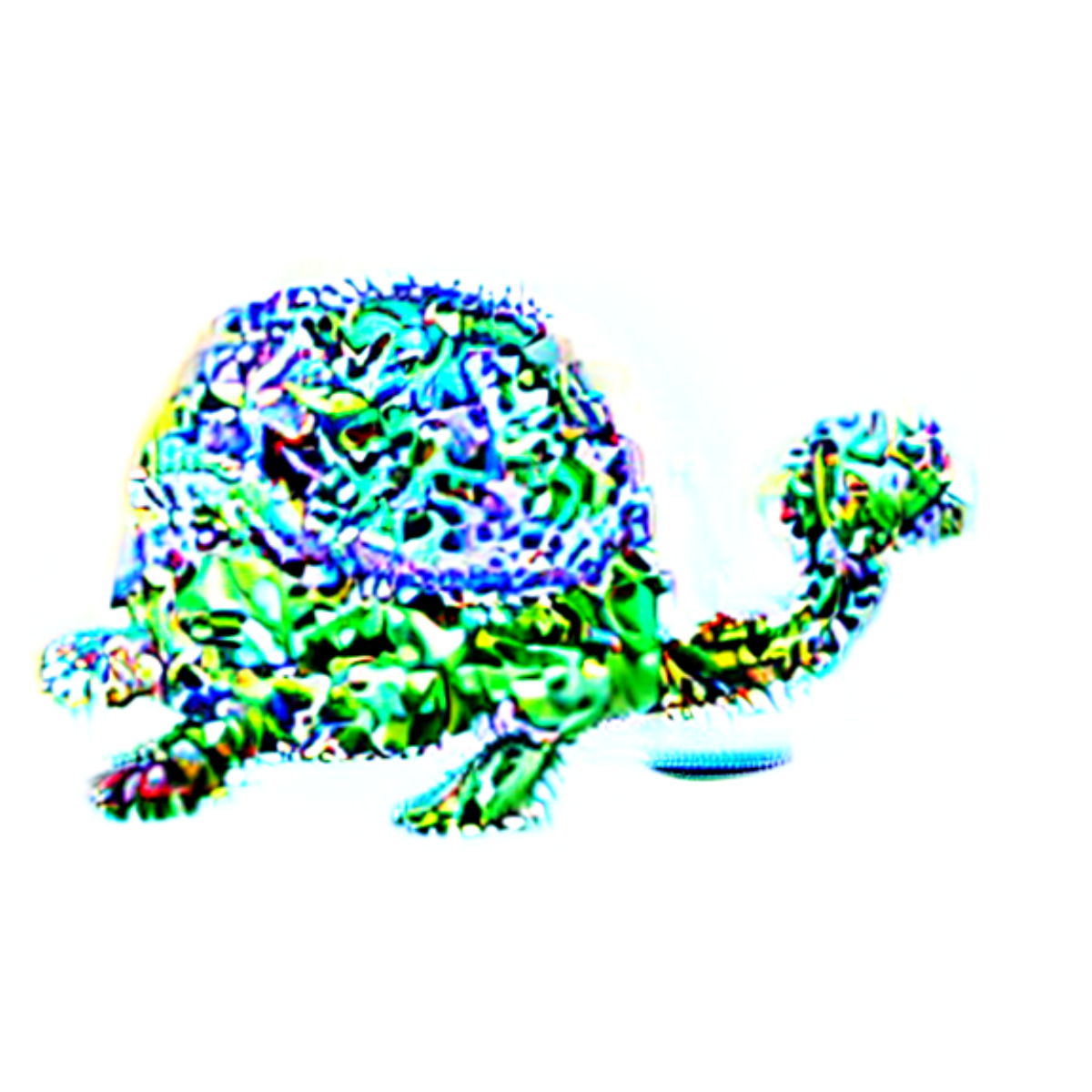}
			\end{subfigure} &
			n =45&
			\begin{subfigure}[c]{0.15\textwidth}  
				\centering
				\includegraphics[width=\linewidth]{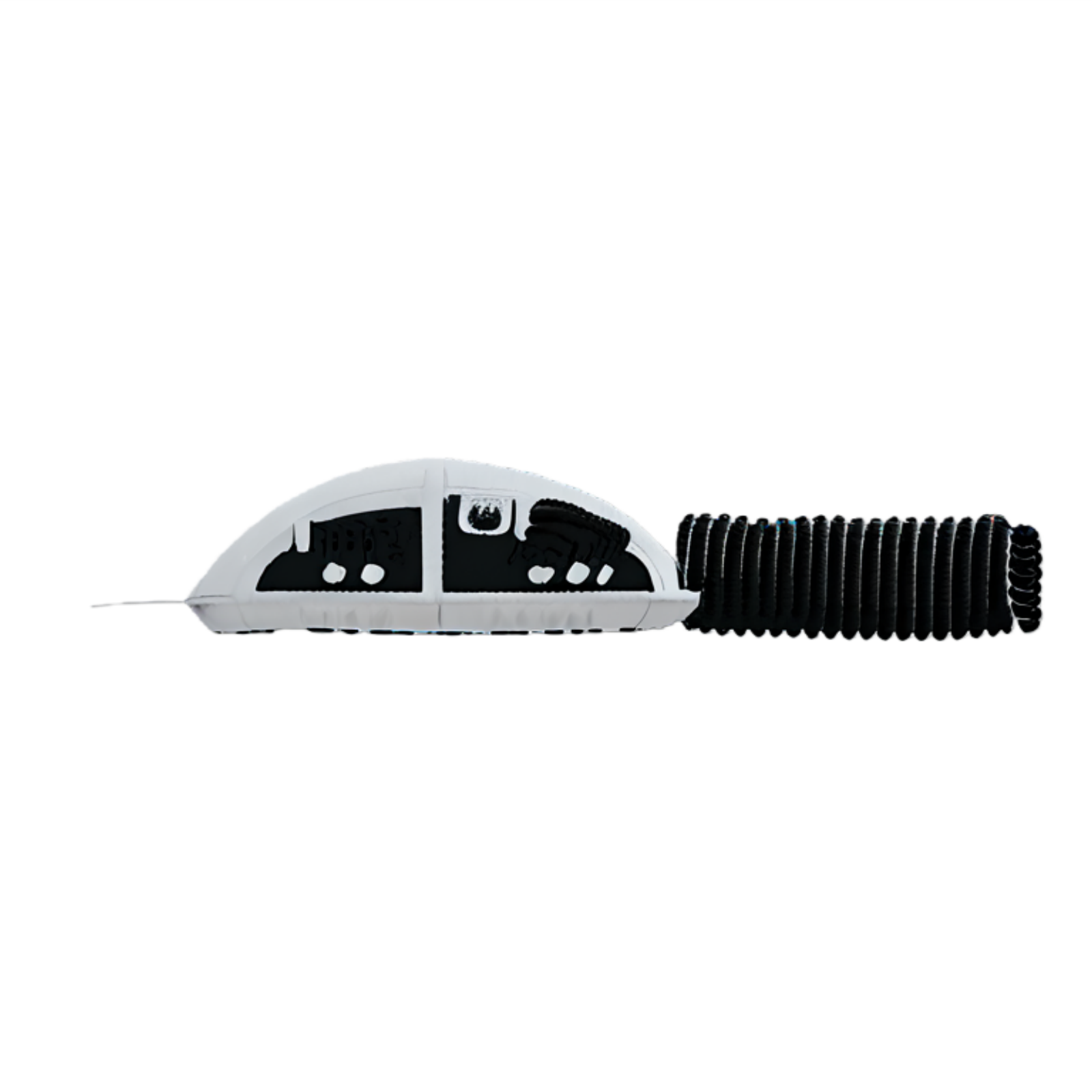}
			\end{subfigure} &
			\begin{subfigure}[c]{0.15\textwidth}  
				\centering
				\includegraphics[width=\linewidth]{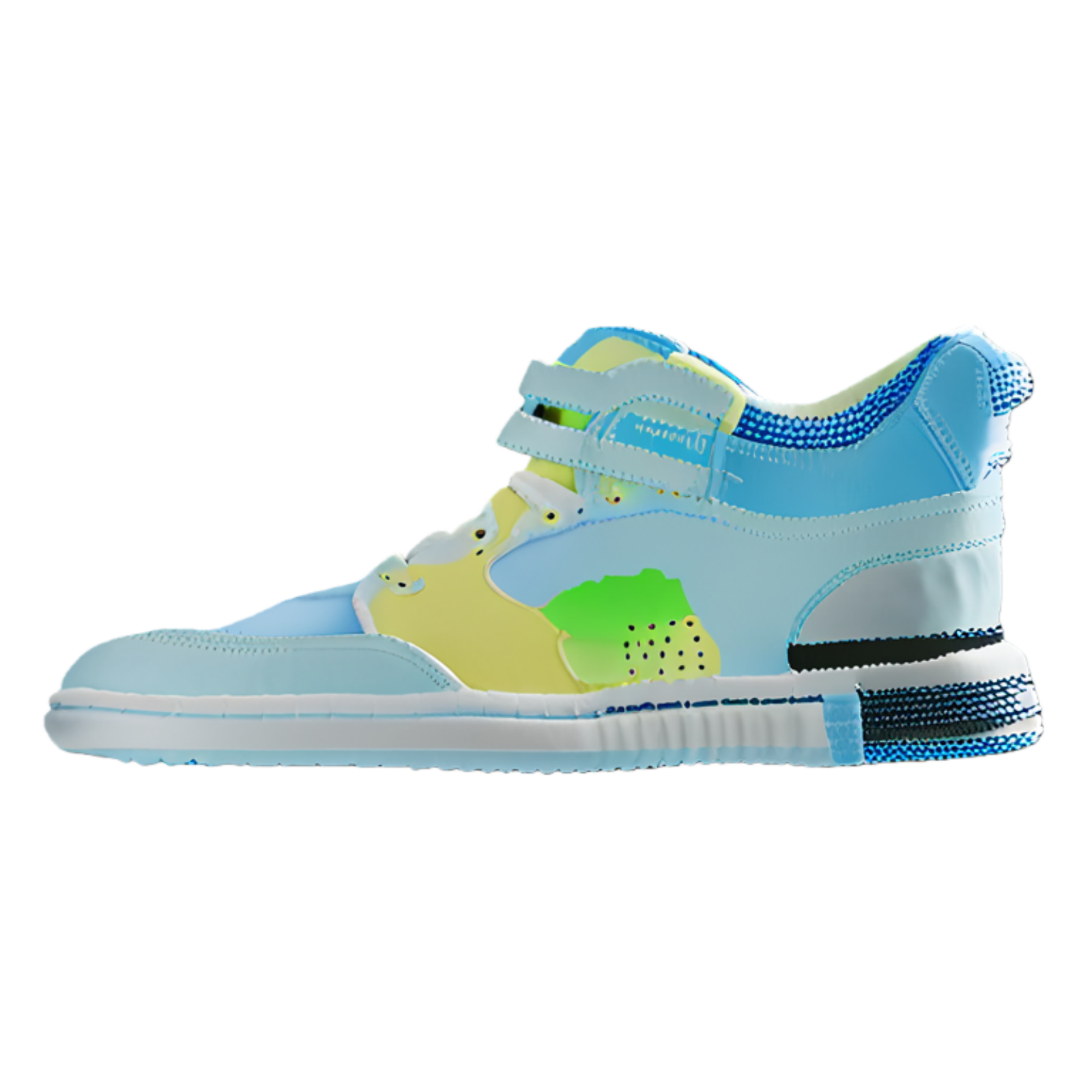}
			\end{subfigure}  \\
		\end{tabular}
		\caption{Comparison of generated results obtained with different numbers of inference-inversion steps in high-quality noise collection process.}
		\label{compare_with_different_n}
	\end{figure*} 
	\begin{figure*}[htbp]
		\centering
		\fontsize{9pt}{12pt}\selectfont  
		\setlength{\tabcolsep}{0pt}  
		\begin{tabular}{c c c c c c c c c c}
			\renewcommand{\arraystretch}{0}
			\makecell{Input\\image} &
			GSO &
			\begin{subfigure}[c]{0.104\textwidth}  
				\centering
				\includegraphics[width=\linewidth]{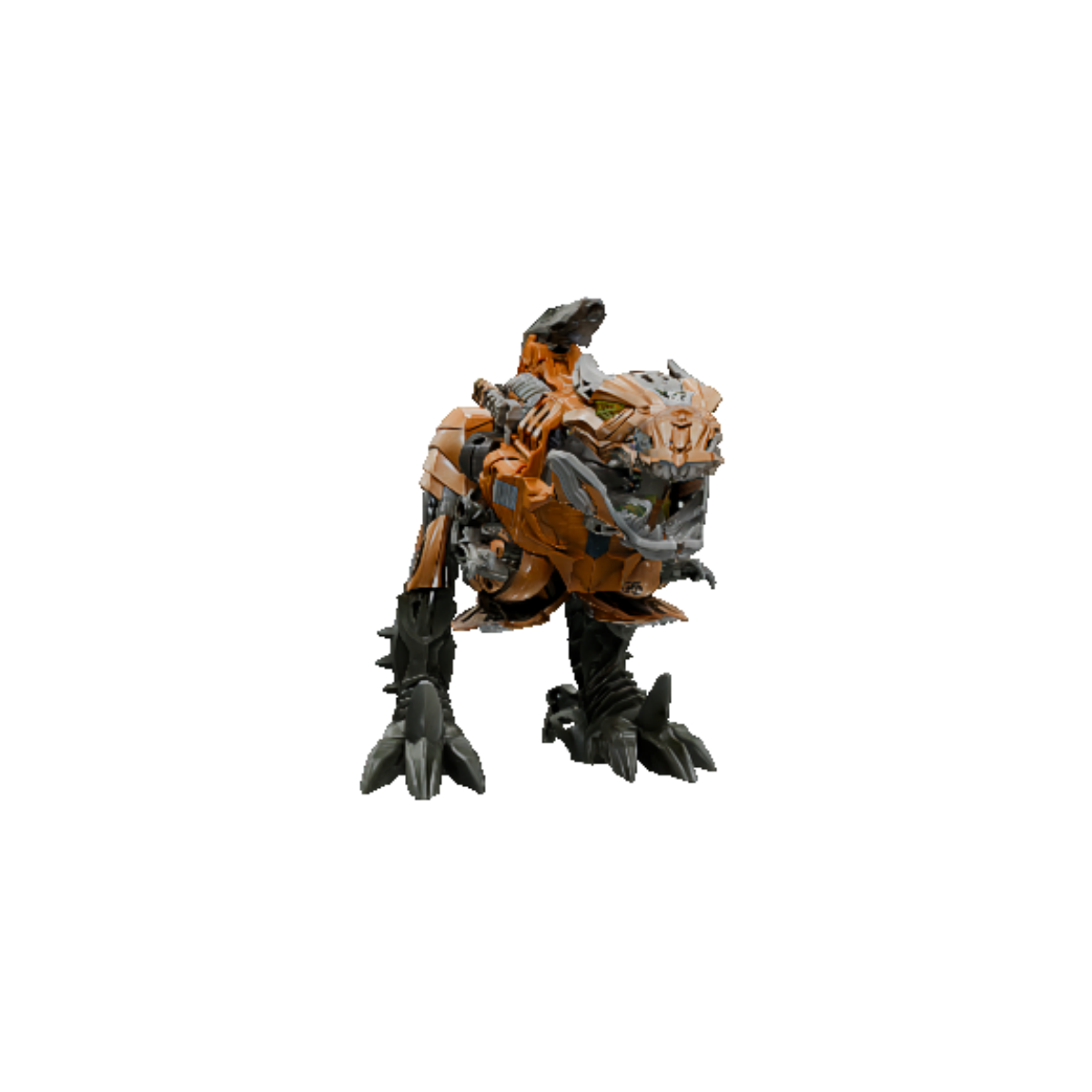}
			\end{subfigure} &
			&
			Objaverse &
			\begin{subfigure}[c]{0.104\textwidth}  
				\centering
				\includegraphics[width=\linewidth]{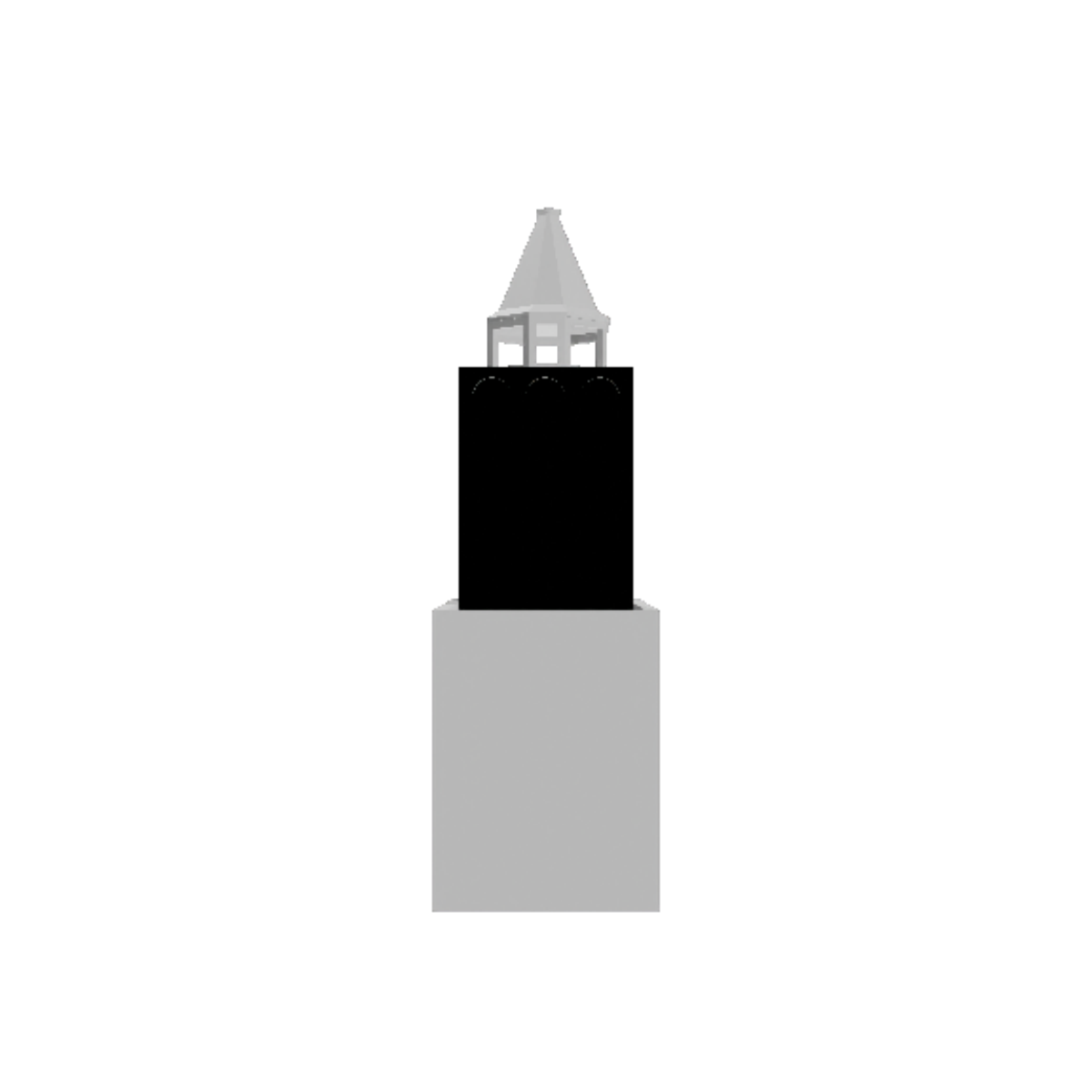}
			\end{subfigure} &
			&
			\makecell{Omni\\Object3D} &
			\begin{subfigure}[c]{0.104\textwidth}  
				\centering
				\includegraphics[width=\linewidth]{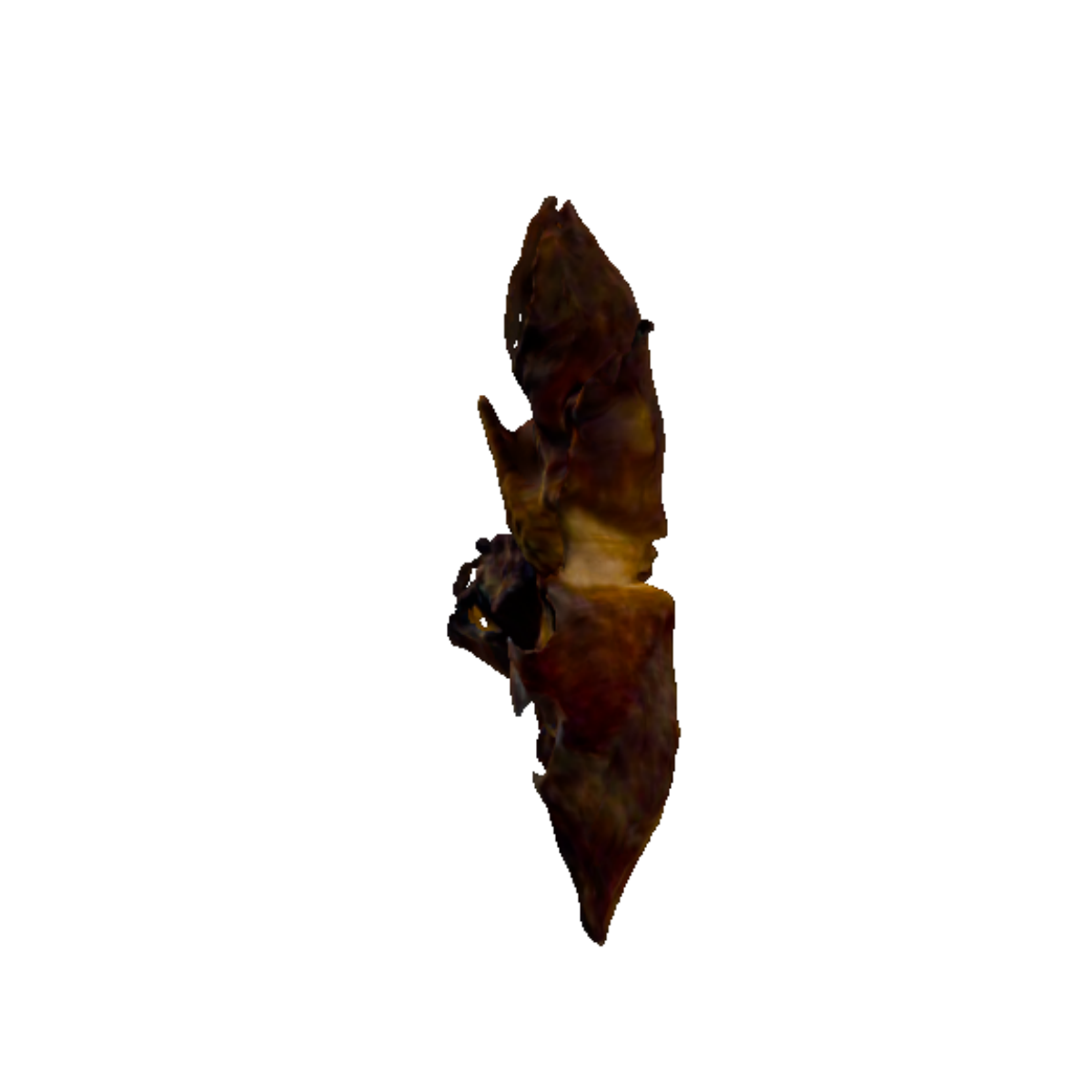}
			\end{subfigure} &\\
			GT &
			\begin{subfigure}[c]{0.104\textwidth}  
				\centering
				\includegraphics[width=\linewidth]{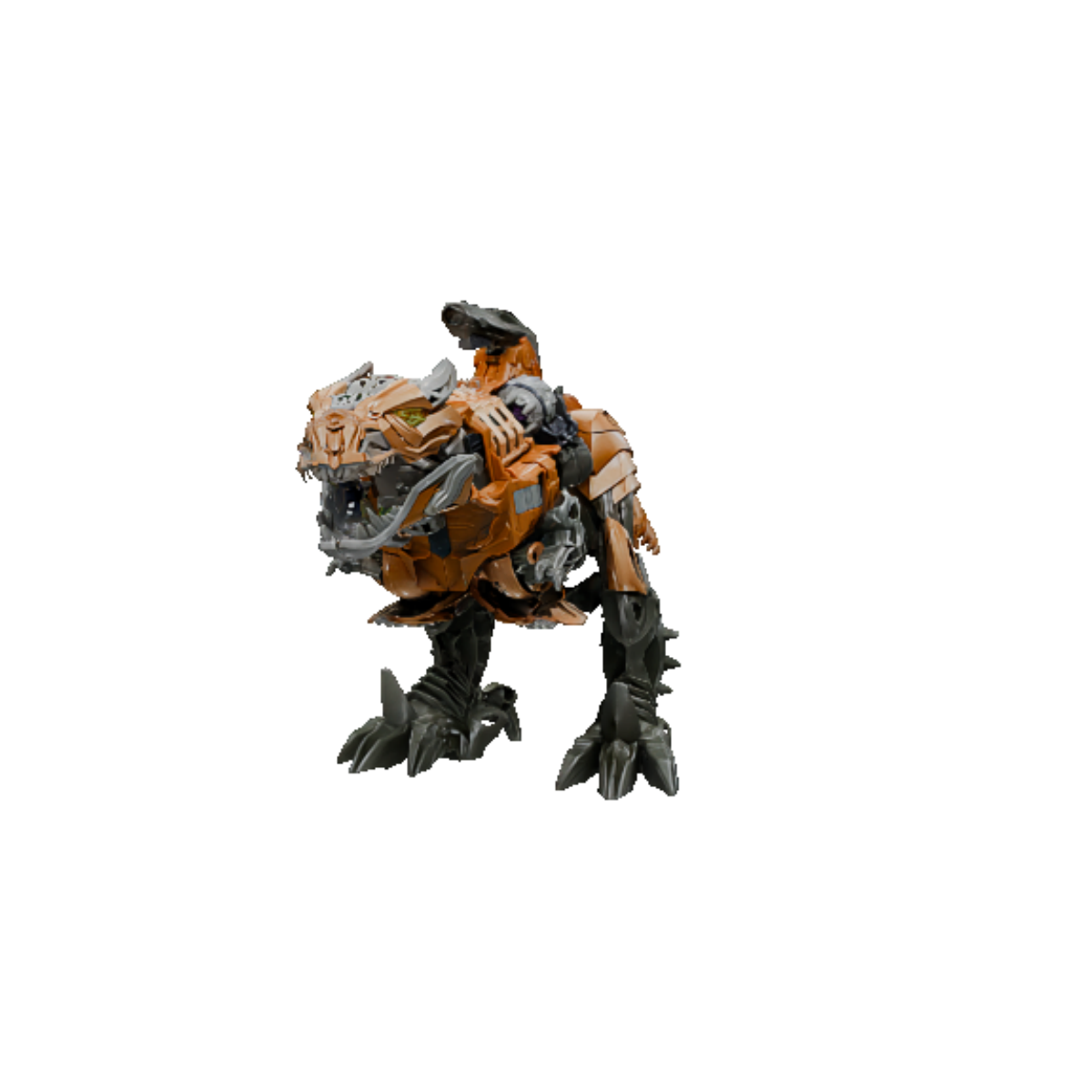}
			\end{subfigure} &
			\begin{subfigure}[c]{0.104\textwidth}  
				\centering
				\includegraphics[width=\linewidth]{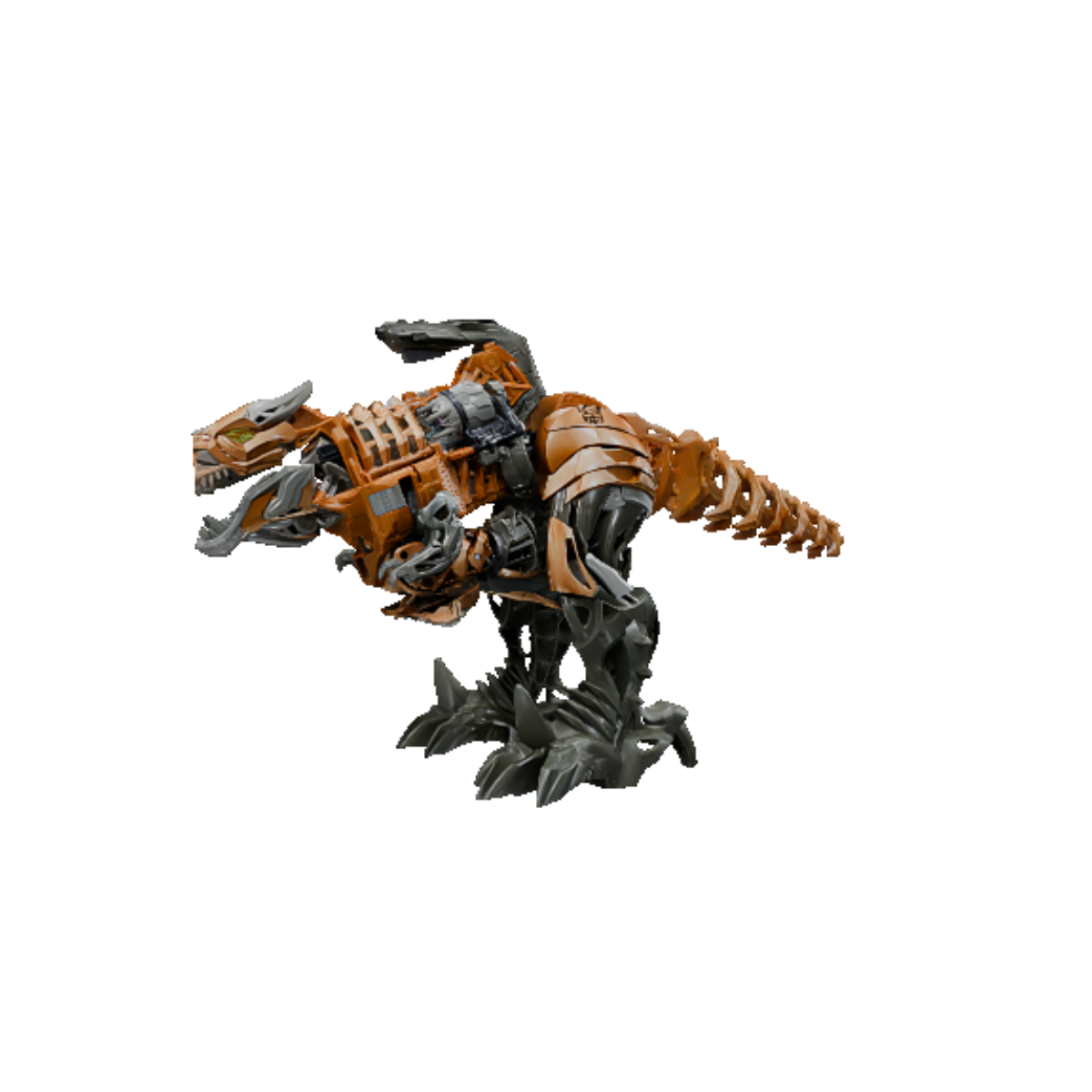}
			\end{subfigure} &
			\begin{subfigure}[c]{0.104\textwidth}  
				\centering
				\includegraphics[width=\linewidth]{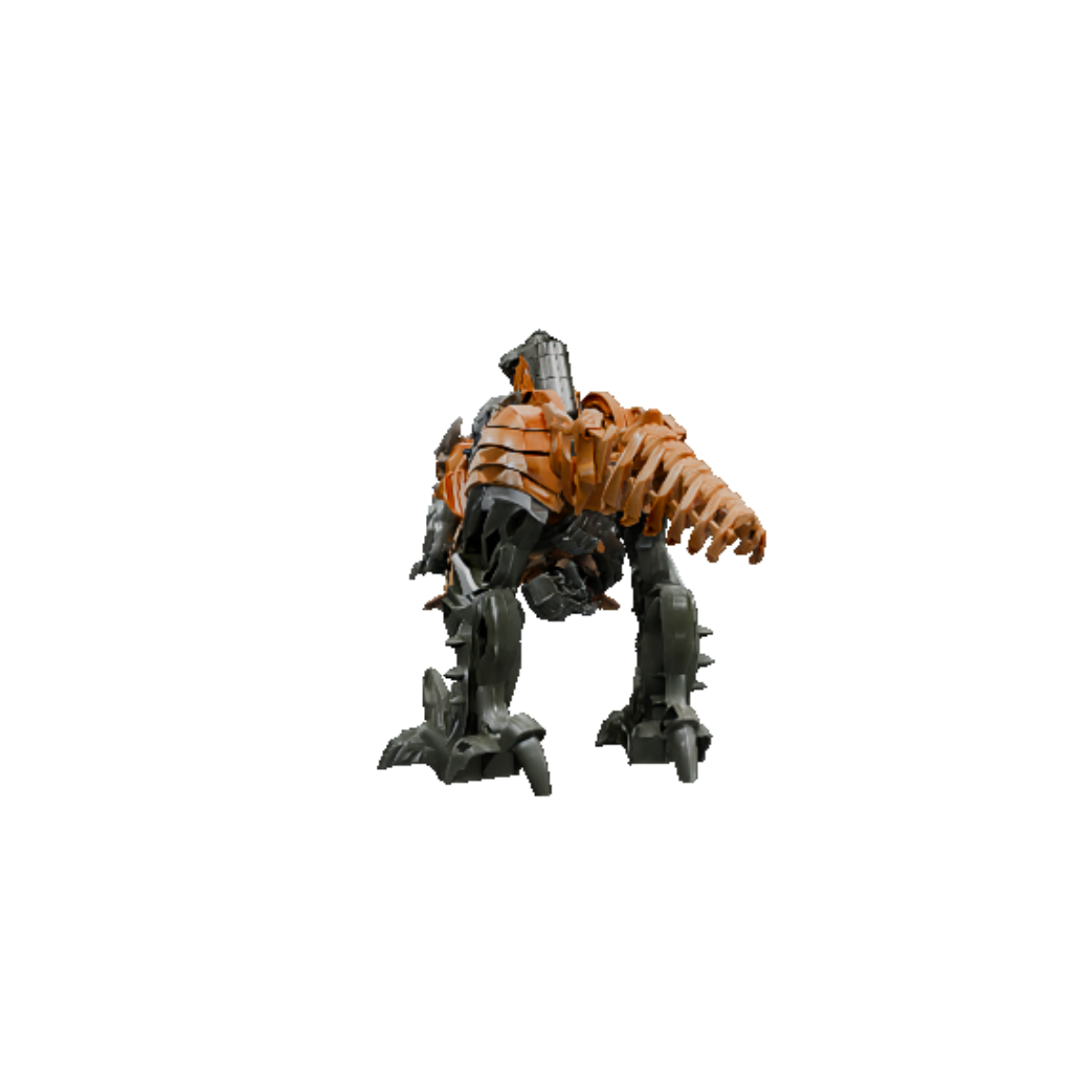}
			\end{subfigure} &
			\begin{subfigure}[c]{0.104\textwidth}  
				\centering
				\includegraphics[width=\linewidth]{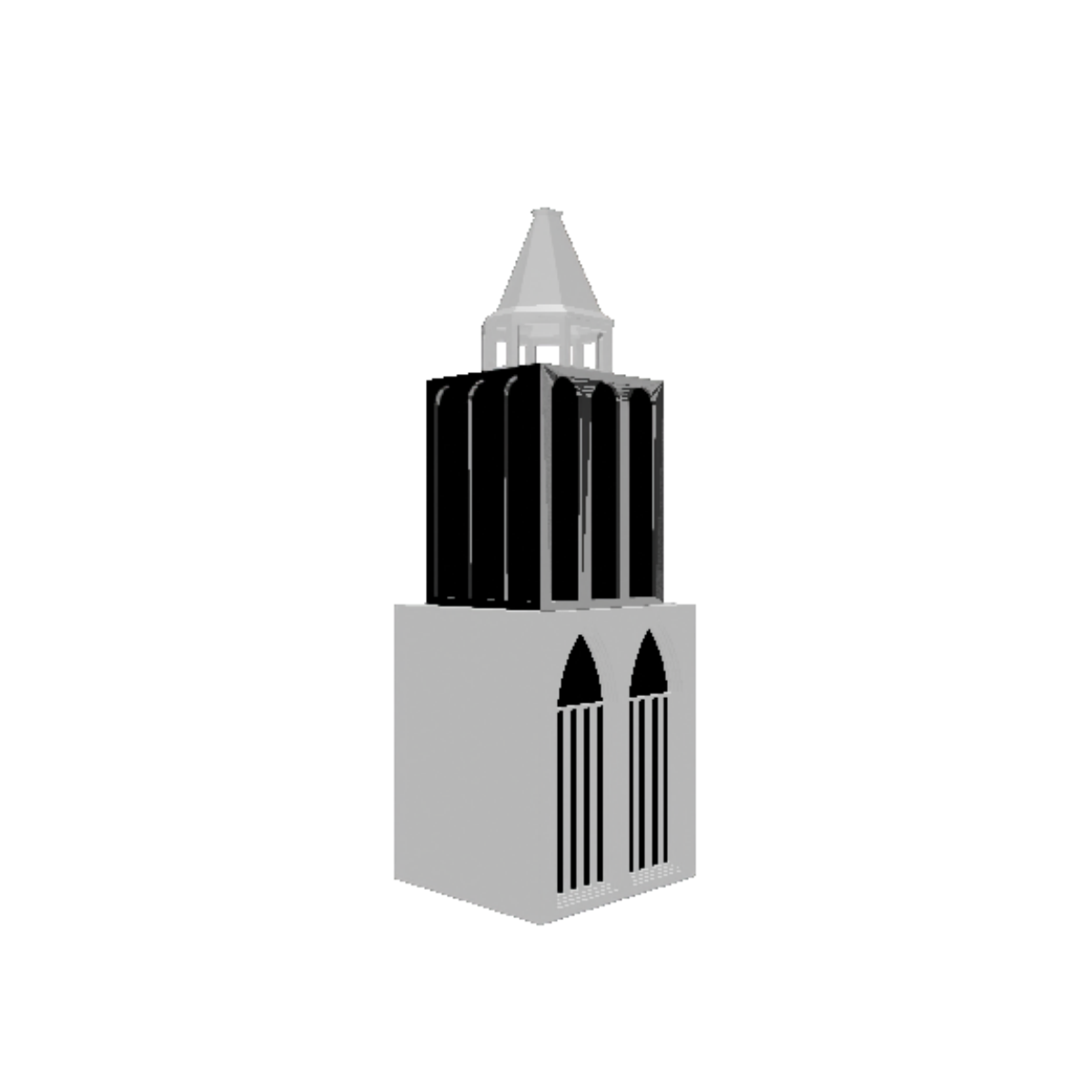}
			\end{subfigure} &
			\begin{subfigure}[c]{0.104\textwidth}  
				\centering
				\includegraphics[width=\linewidth]{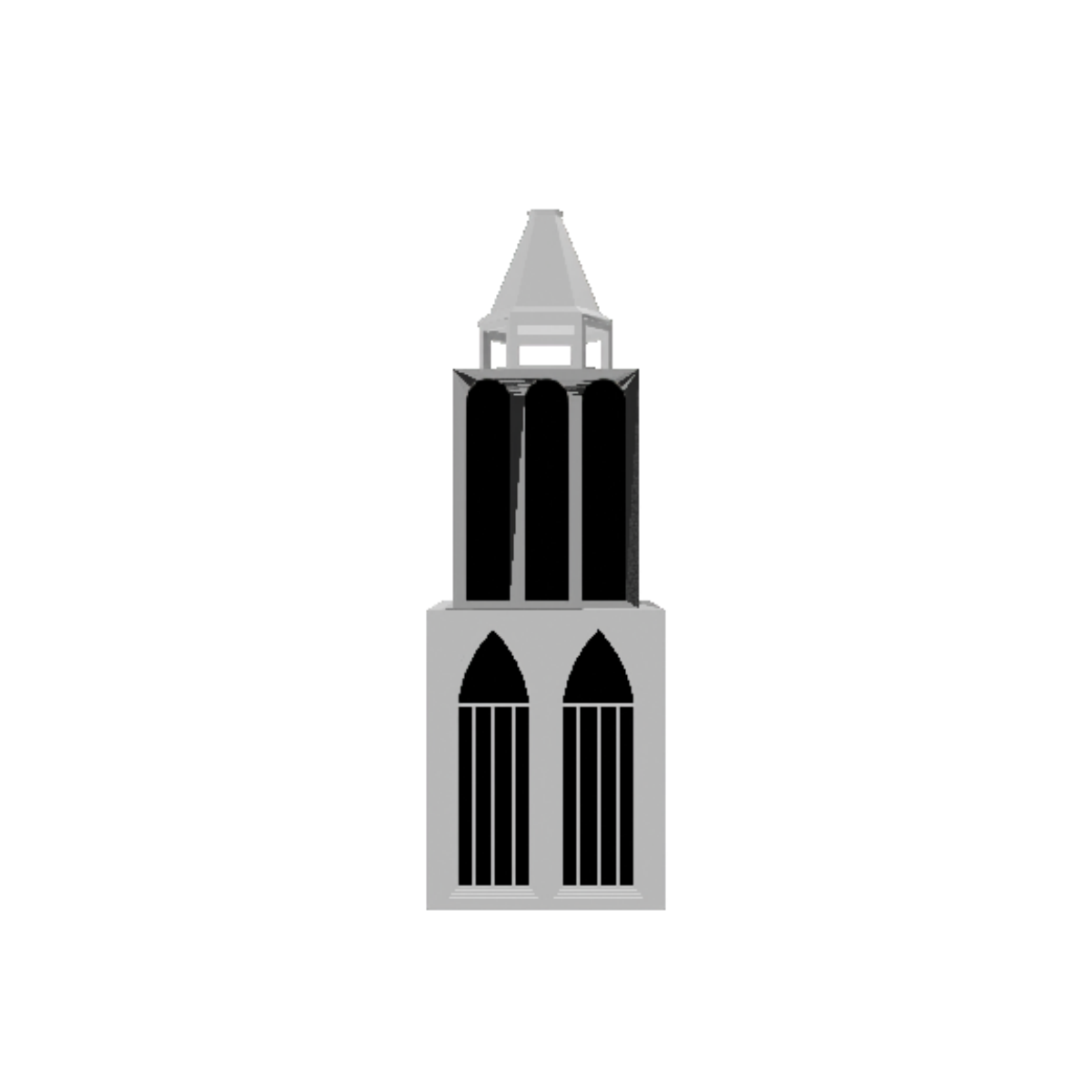}
			\end{subfigure} &
			\begin{subfigure}[c]{0.104\textwidth} 
				\centering
				\includegraphics[width=\linewidth]{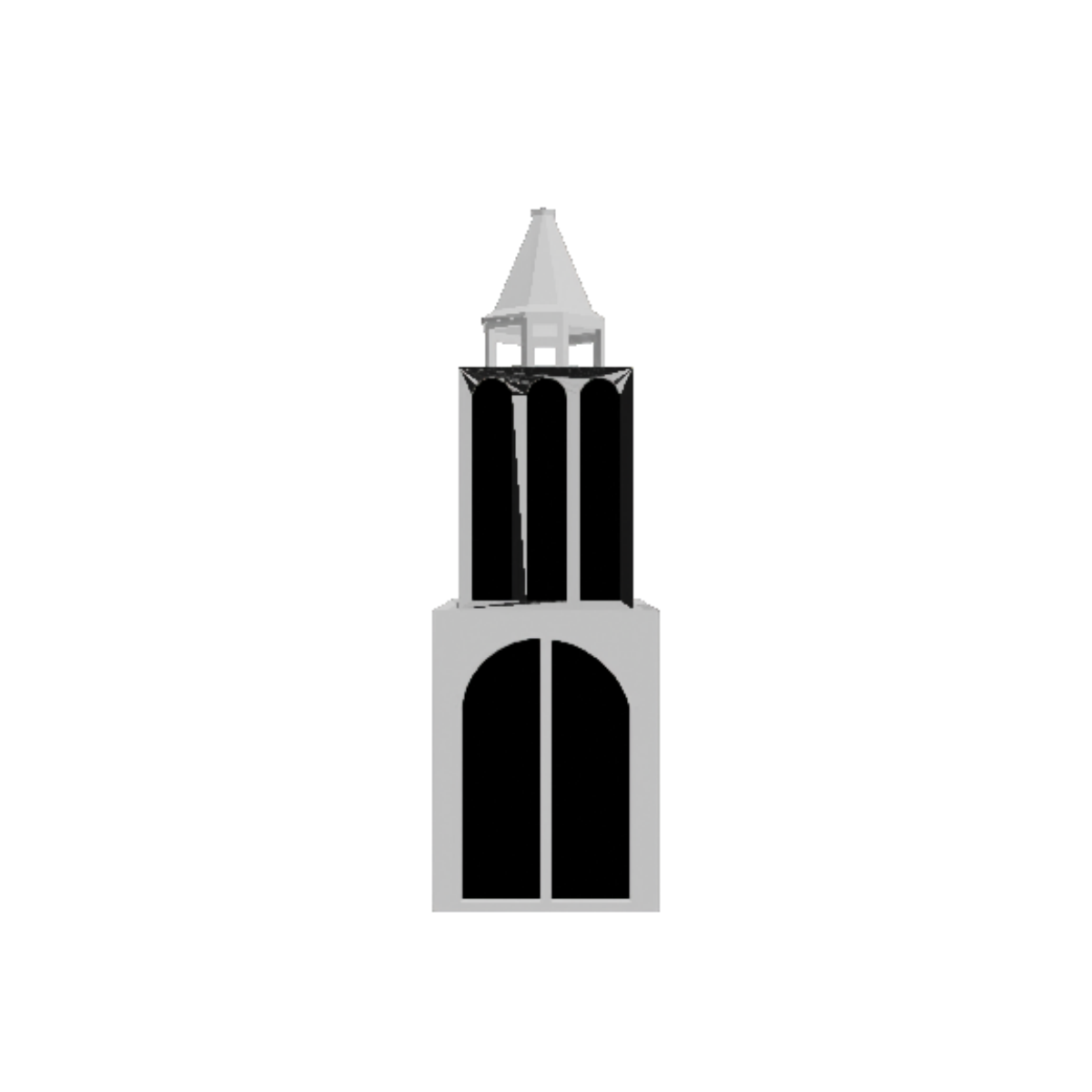}
			\end{subfigure} &
			\begin{subfigure}[c]{0.104\textwidth}  
				\centering
				\includegraphics[width=\linewidth]{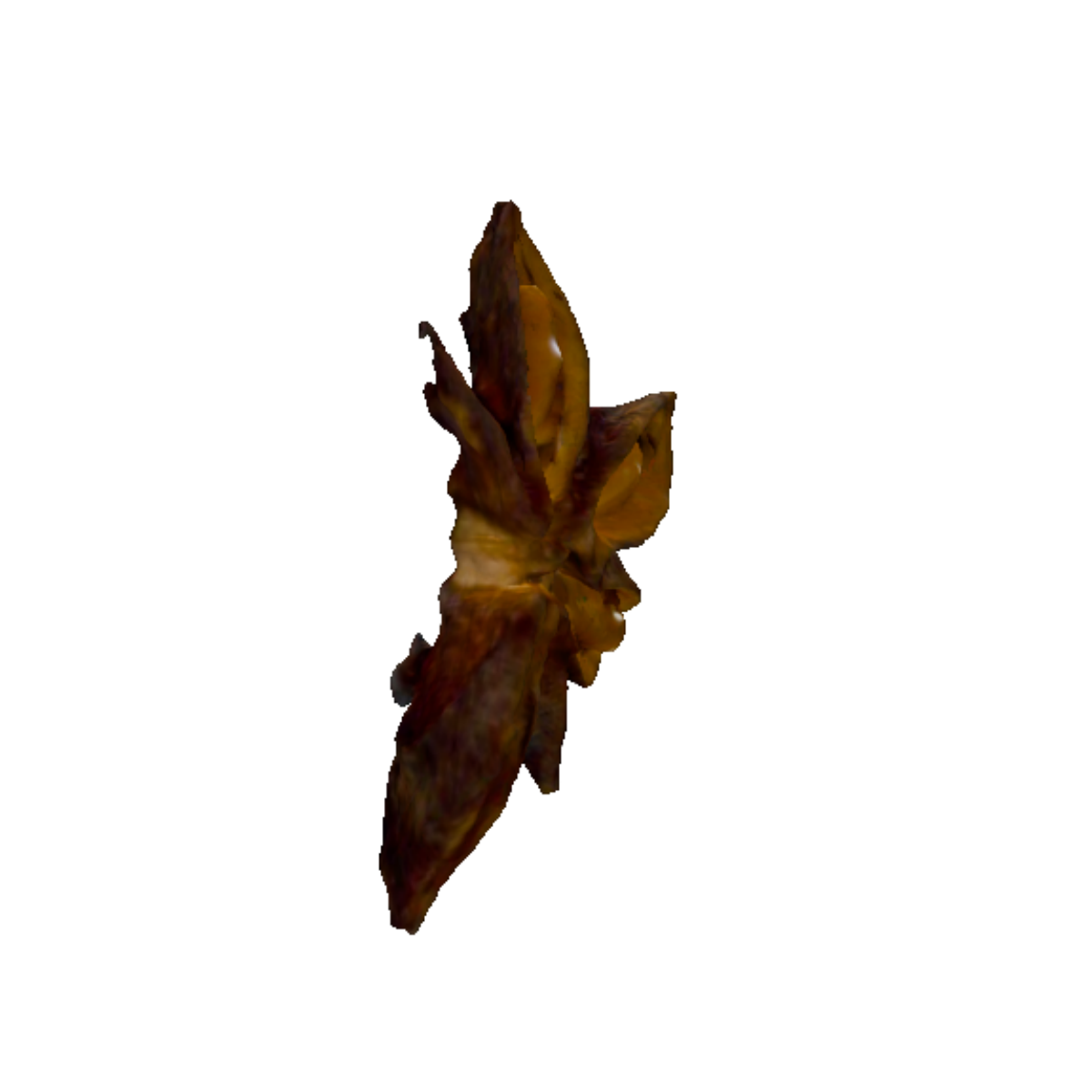}
			\end{subfigure} &
			\begin{subfigure}[c]{0.104\textwidth}  
				\centering
				\includegraphics[width=\linewidth]{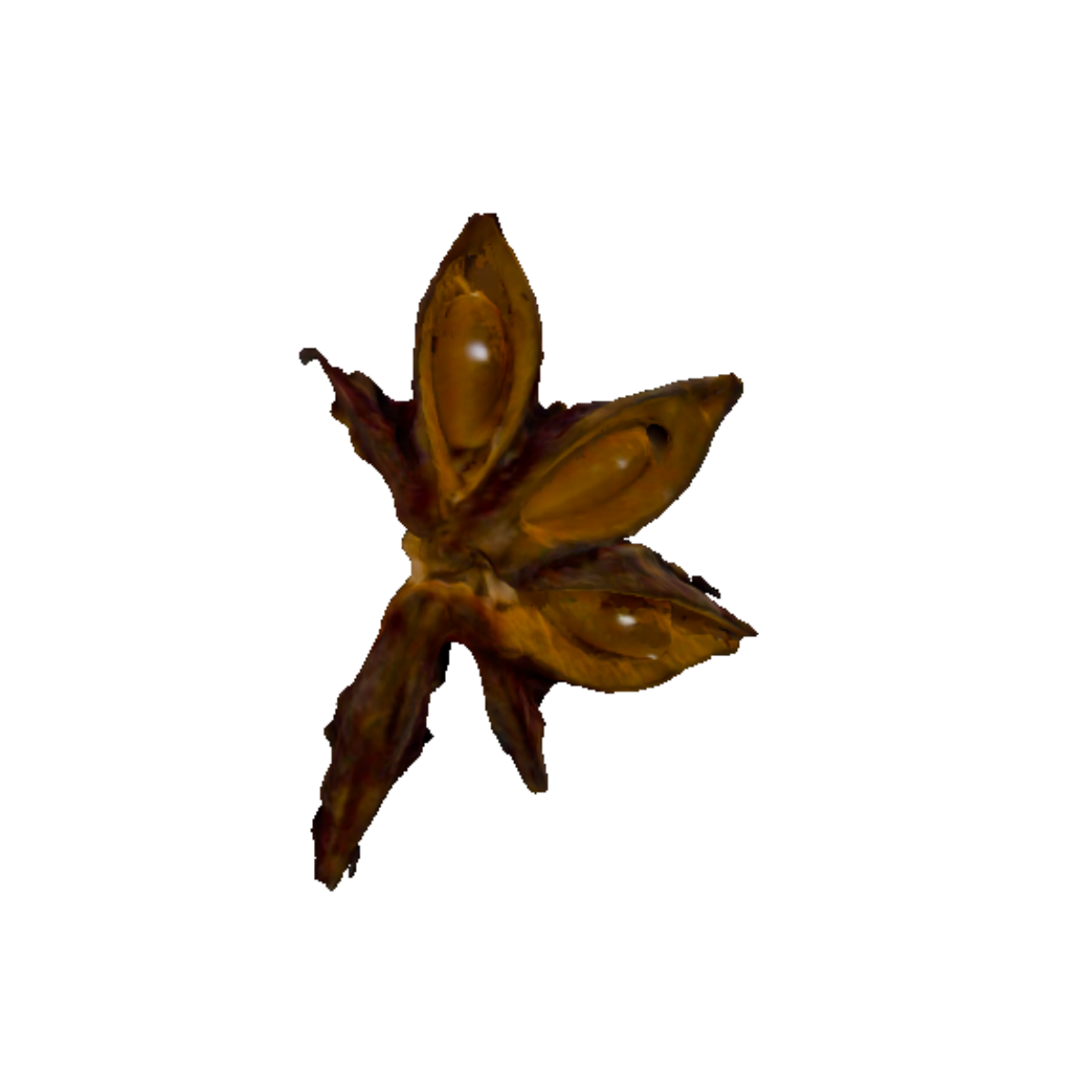}
			\end{subfigure}&
			\begin{subfigure}[c]{0.104\textwidth}  
				\centering
				\includegraphics[width=\linewidth]{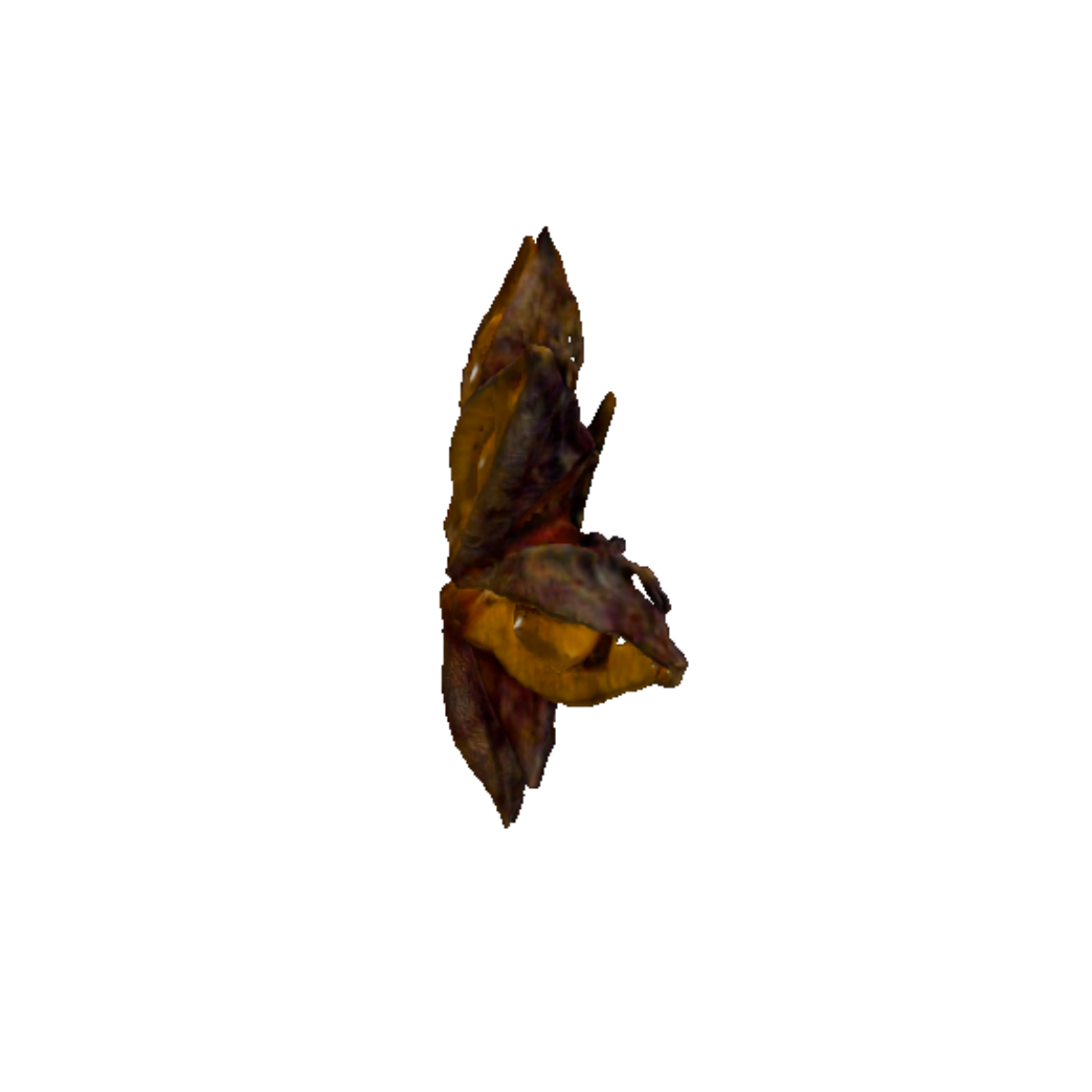}
			\end{subfigure} \\
			SV3D &
			\begin{subfigure}[c]{0.104\textwidth}  
				\centering
				\includegraphics[width=\linewidth]{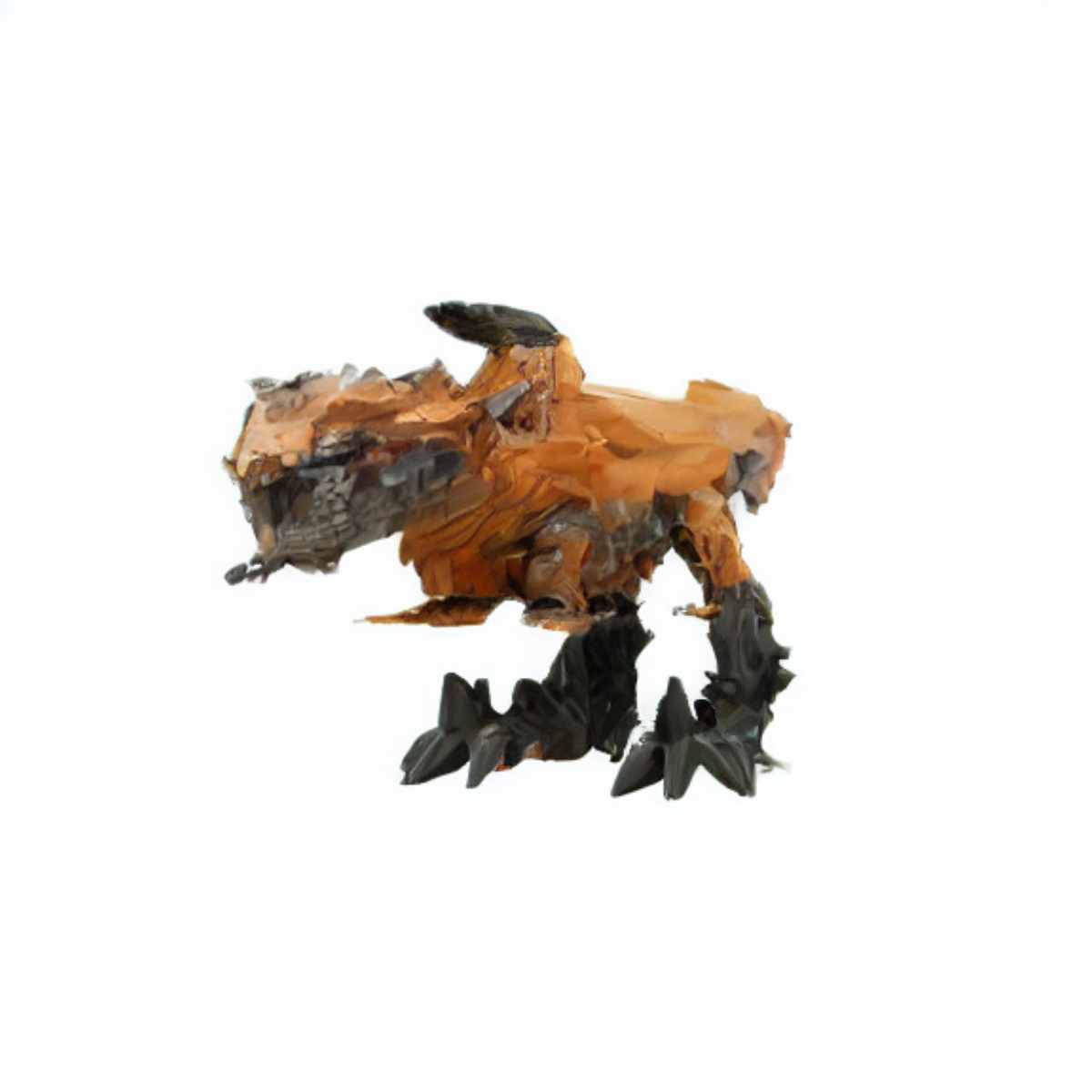}
			\end{subfigure} &
			\begin{subfigure}[c]{0.104\textwidth}  
				\centering
				\includegraphics[width=\linewidth]{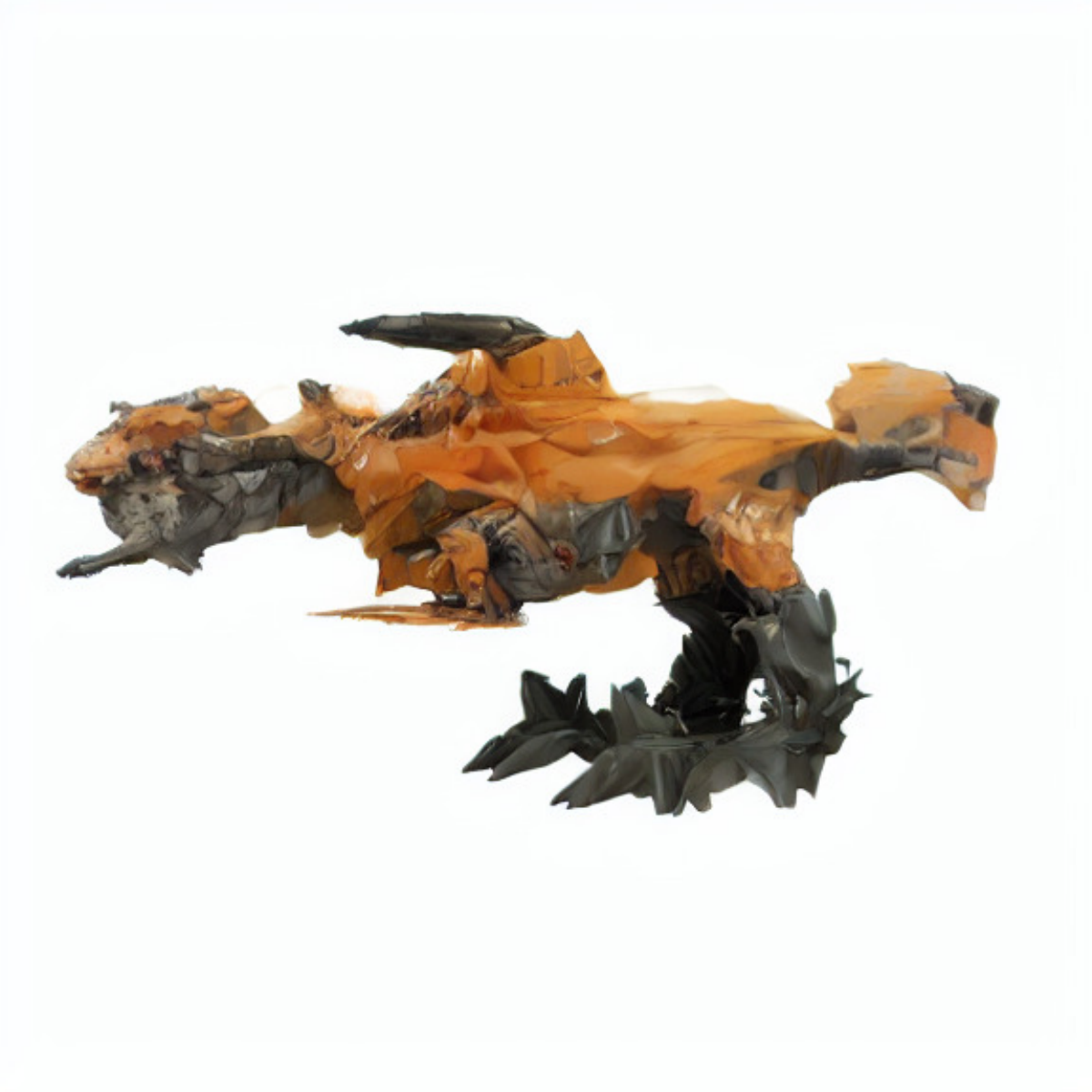}
			\end{subfigure} &
			\begin{subfigure}[c]{0.104\textwidth}  
				\centering
				\includegraphics[width=\linewidth]{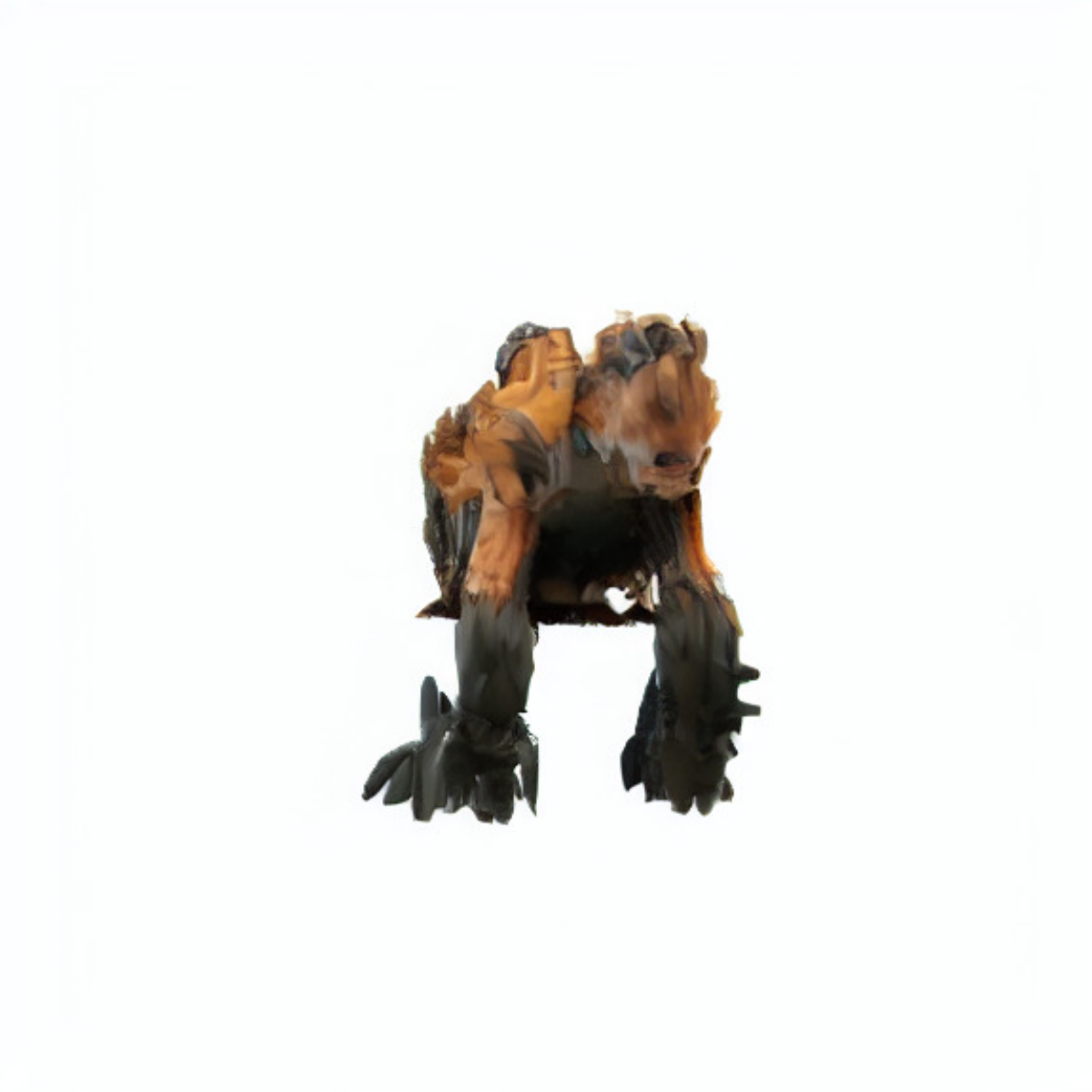}
			\end{subfigure} &
			\begin{subfigure}[c]{0.104\textwidth}  
				\centering
				\includegraphics[width=\linewidth]{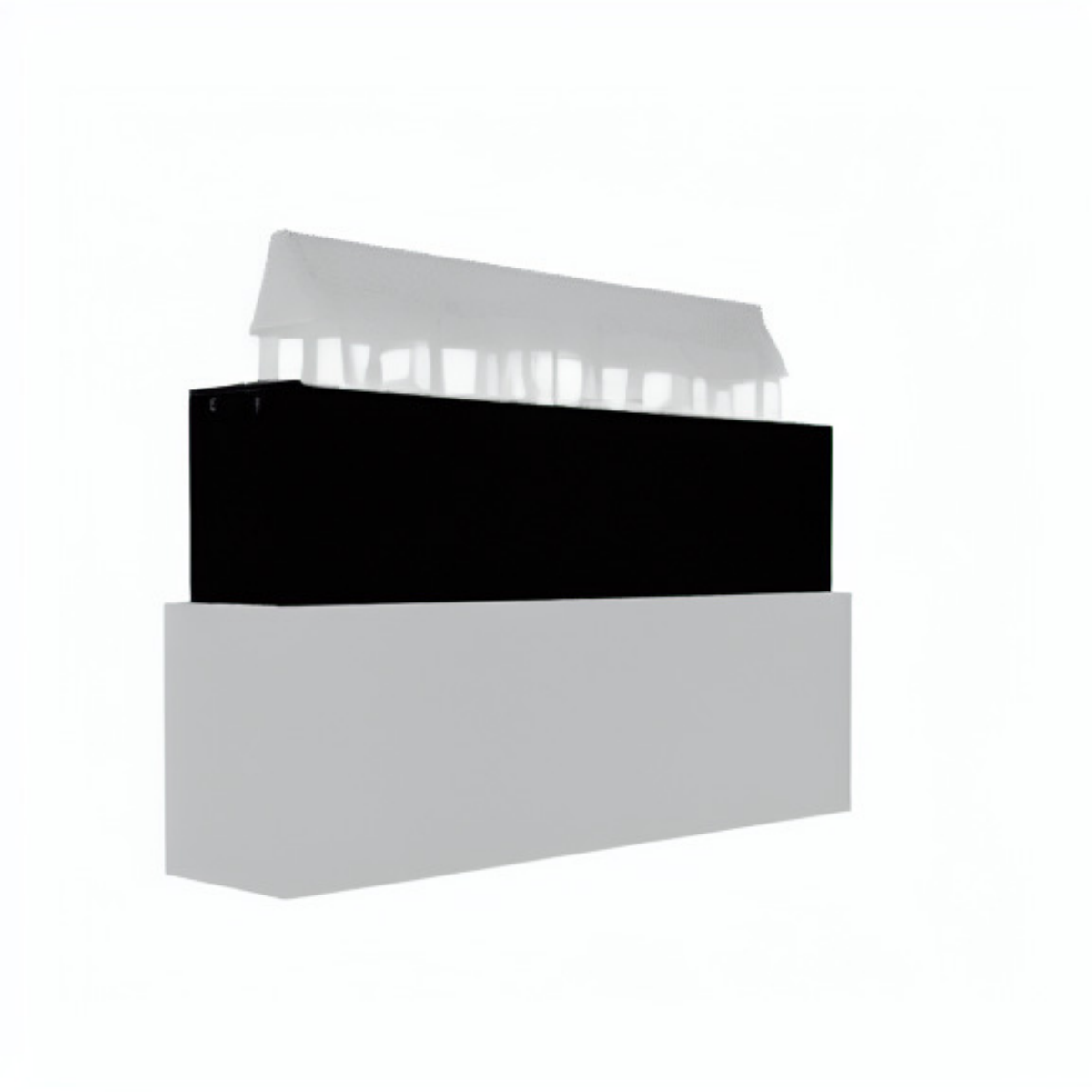}
			\end{subfigure} &
			\begin{subfigure}[c]{0.104\textwidth}  
				\centering
				\includegraphics[width=\linewidth]{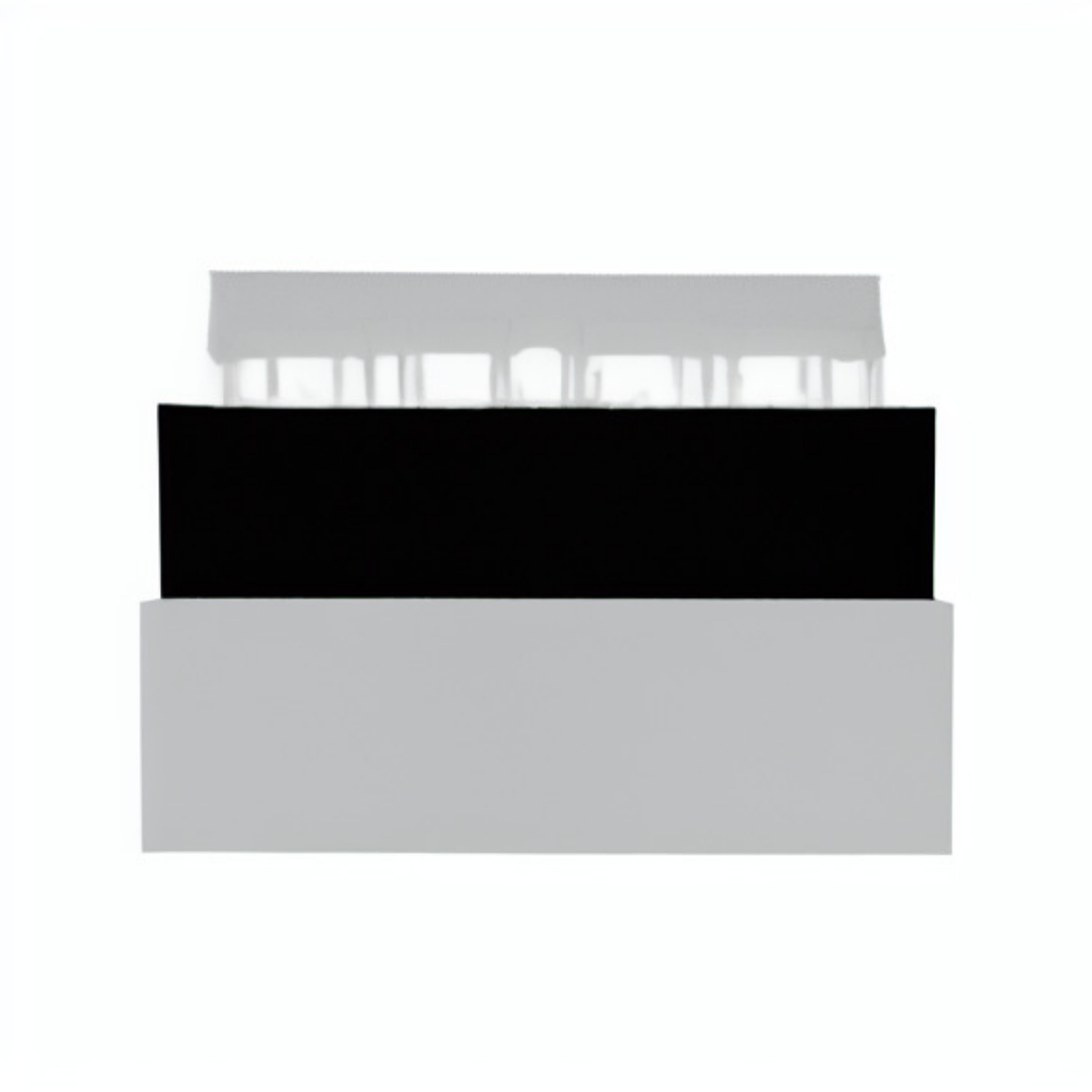}
			\end{subfigure} &
			\begin{subfigure}[c]{0.104\textwidth} 
				\centering
				\includegraphics[width=\linewidth]{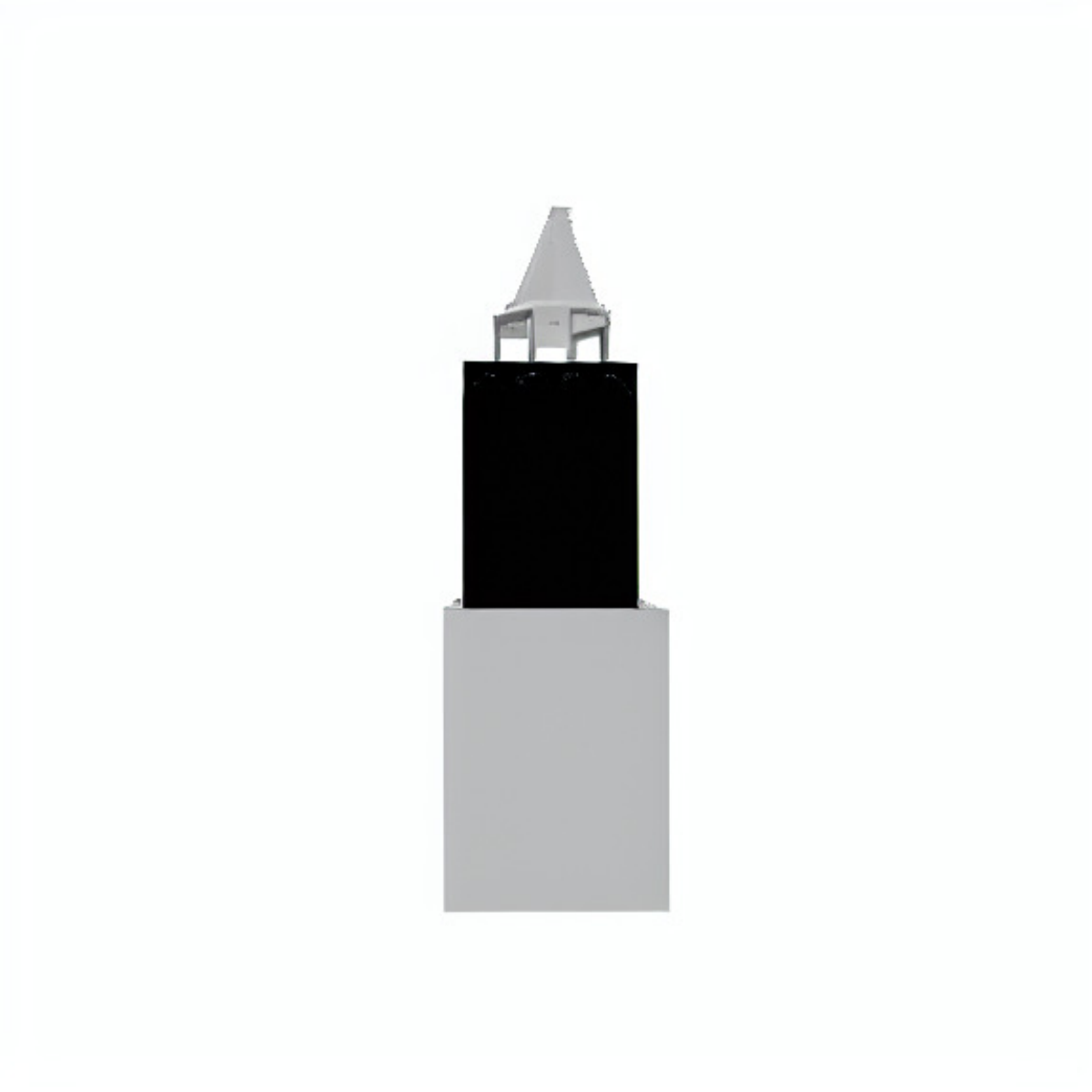}
			\end{subfigure} &
			\begin{subfigure}[c]{0.104\textwidth}  
				\centering
				\includegraphics[width=\linewidth]{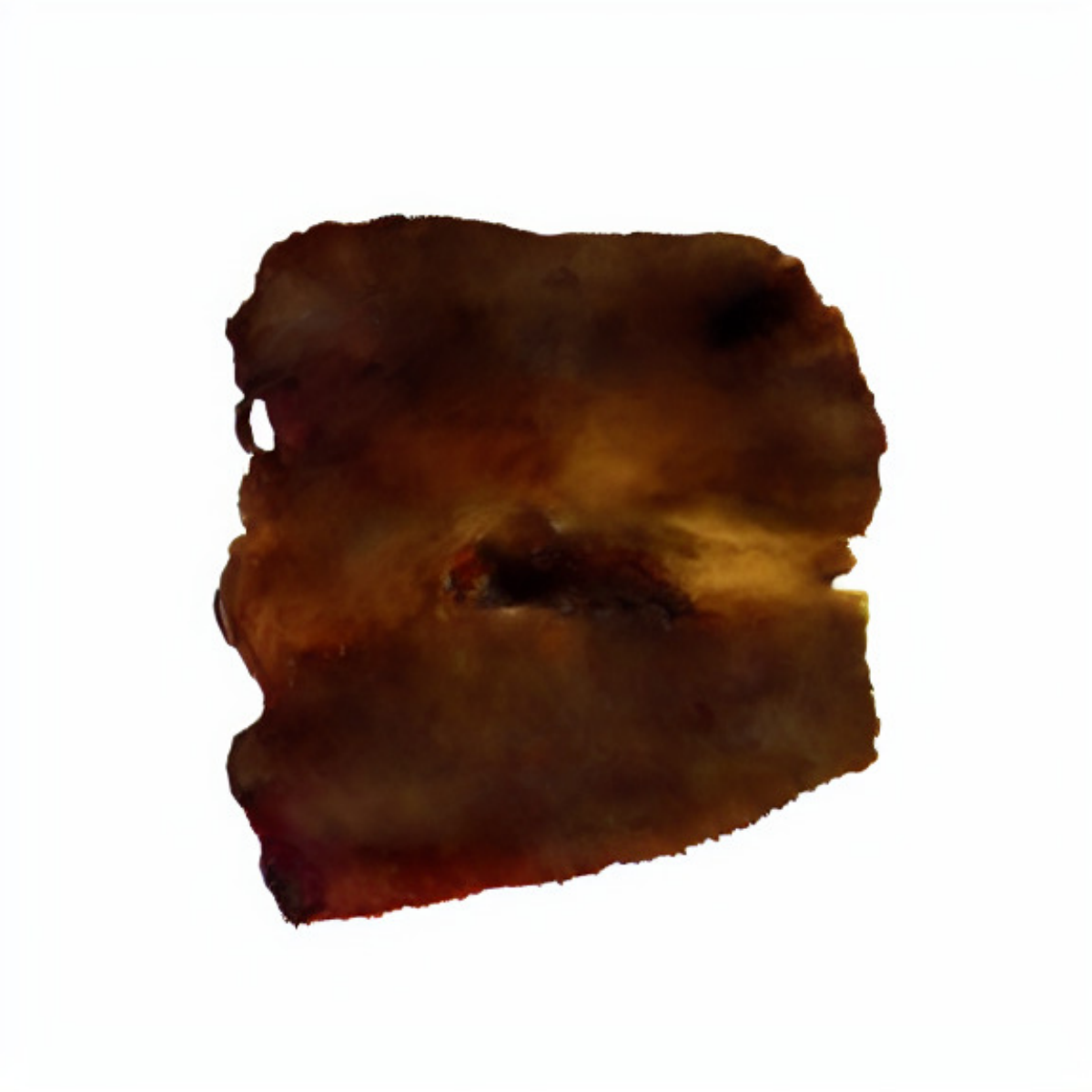}
			\end{subfigure} &
			\begin{subfigure}[c]{0.104\textwidth}  
				\centering
				\includegraphics[width=\linewidth]{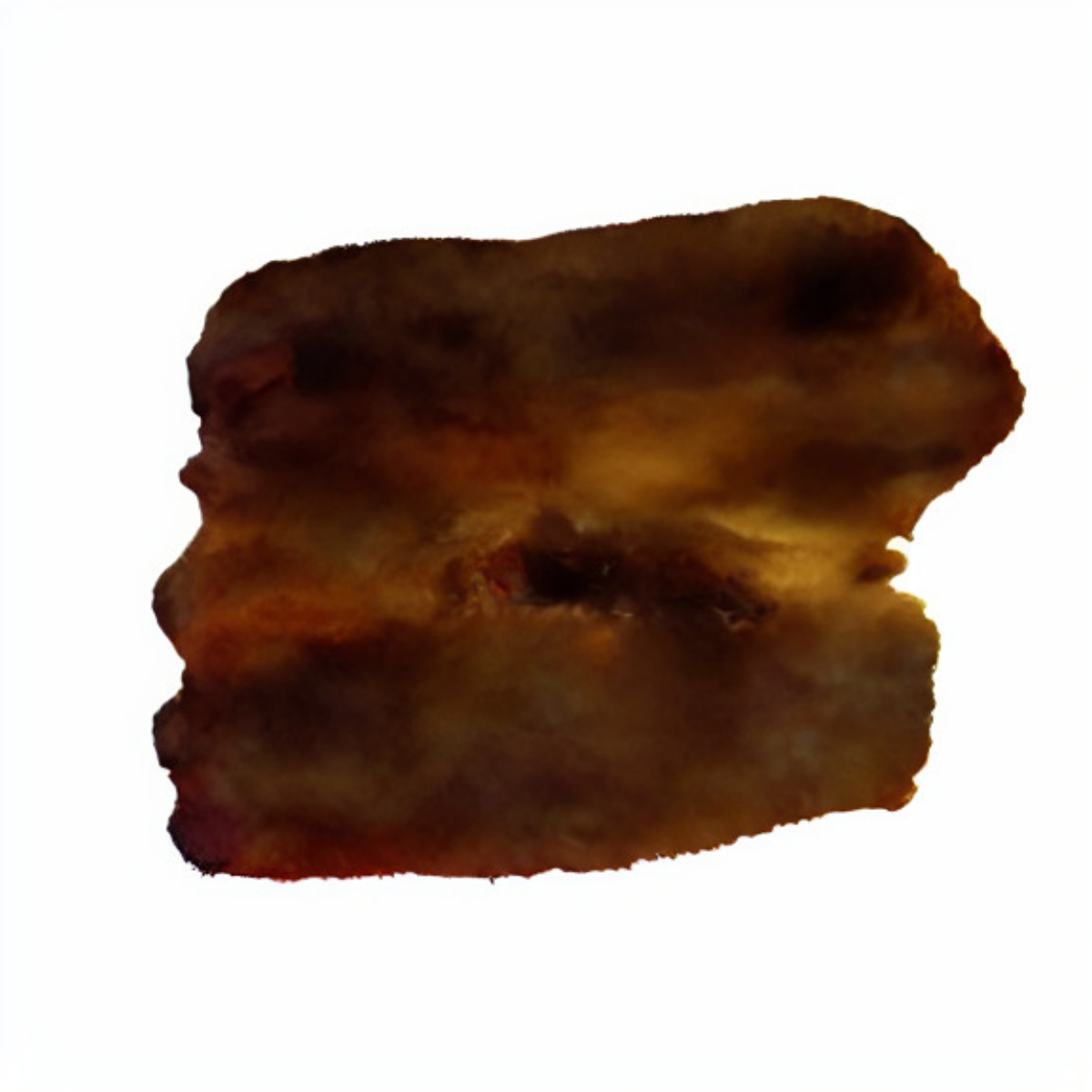}
			\end{subfigure}&
			\begin{subfigure}[c]{0.104\textwidth}  
				\centering
				\includegraphics[width=\linewidth]{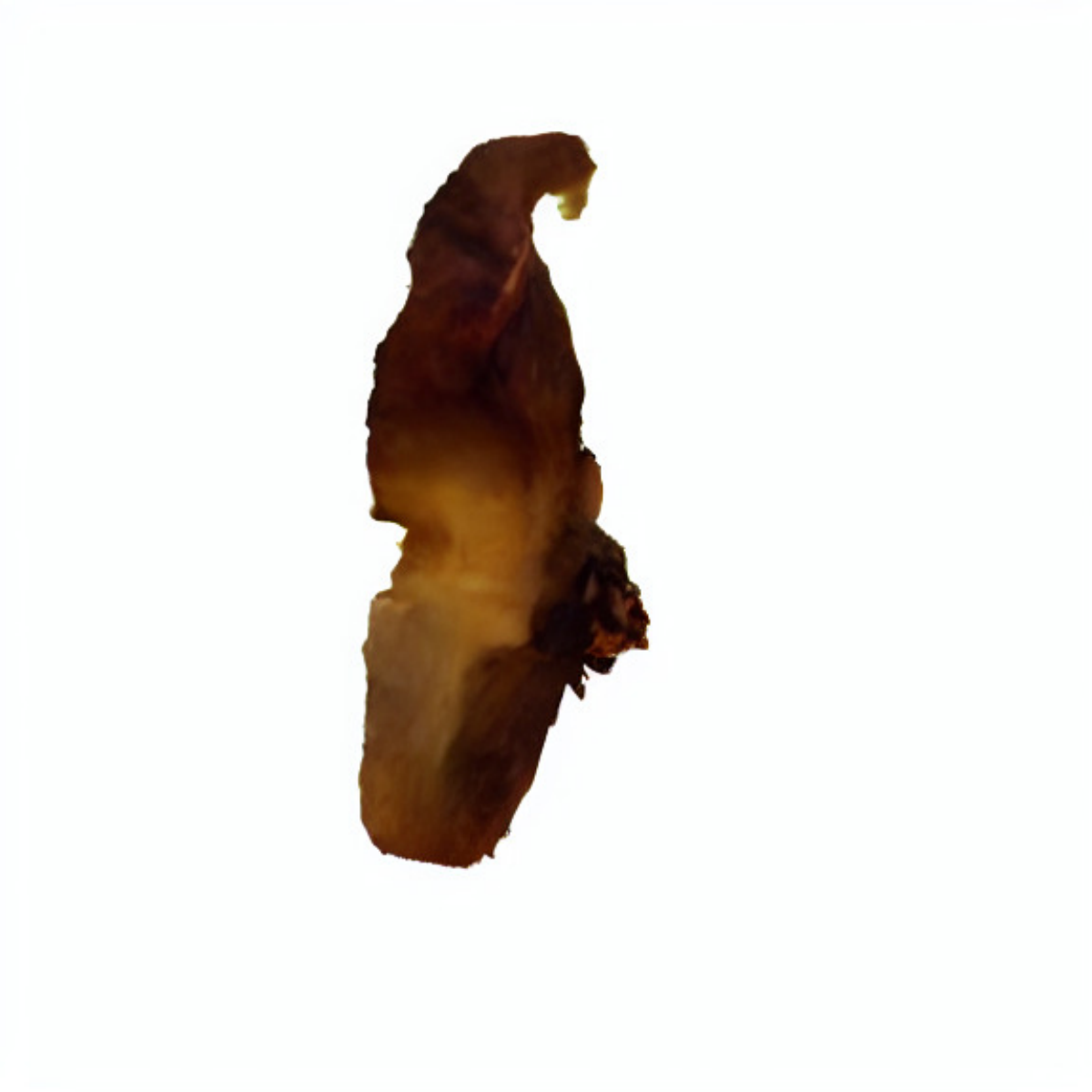}
			\end{subfigure} \\
			\makecell{SV3D\\\textbf{with}\\\textbf{EDN}} &
			\begin{subfigure}[c]{0.104\textwidth}  
				\centering
				\includegraphics[width=\linewidth]{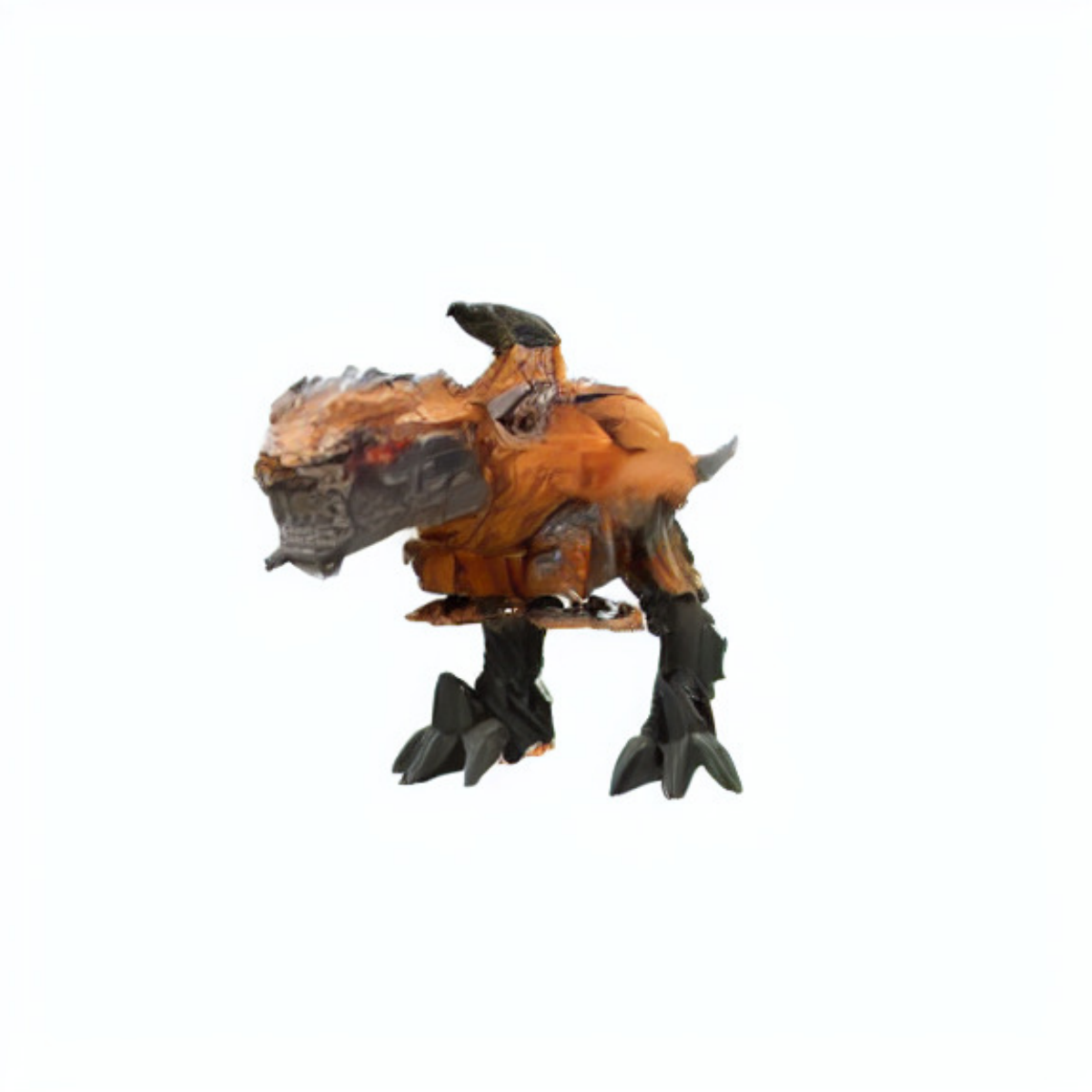}
			\end{subfigure} &
			\begin{subfigure}[c]{0.104\textwidth}  
				\centering
				\includegraphics[width=\linewidth]{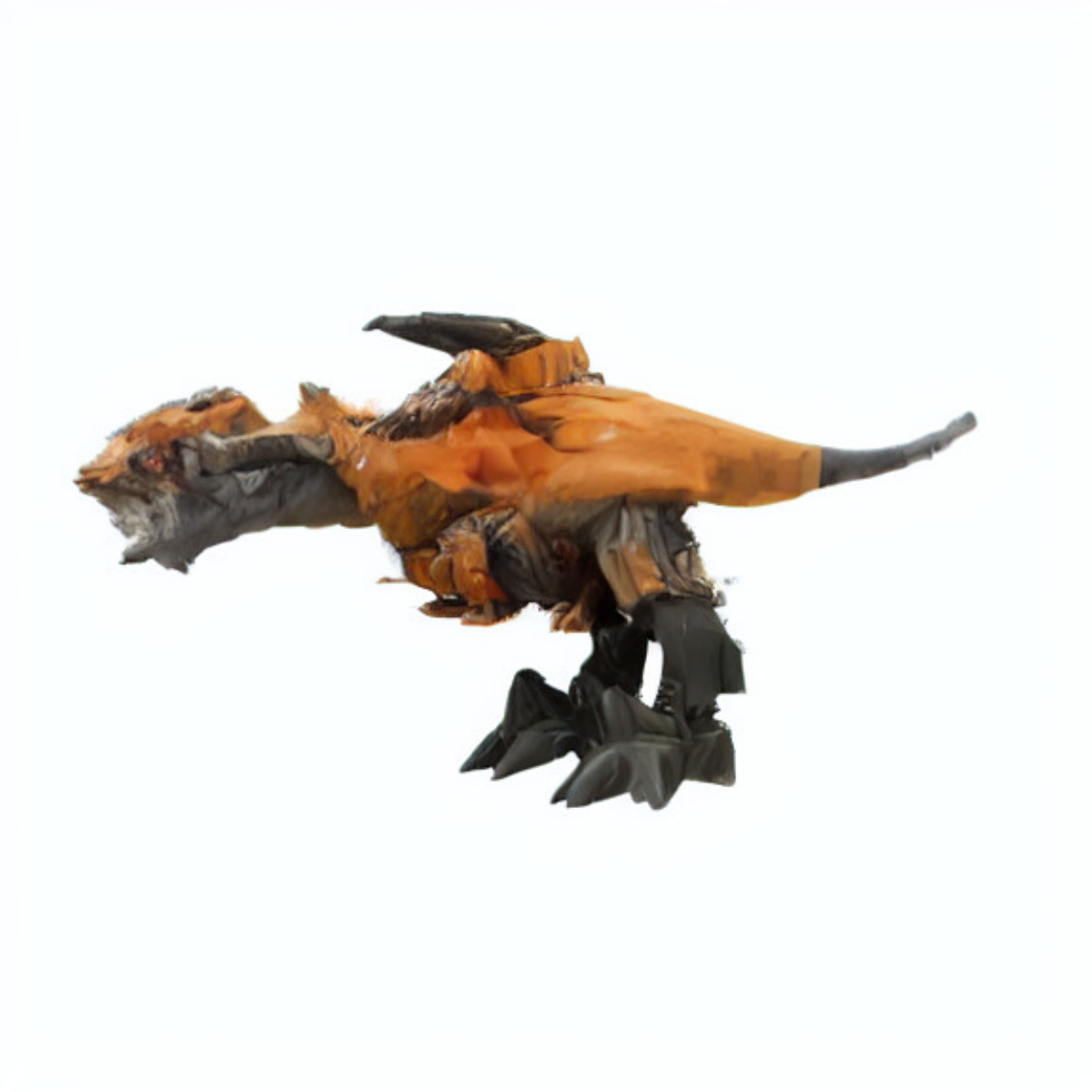}
			\end{subfigure} &
			\begin{subfigure}[c]{0.104\textwidth}  
				\centering
				\includegraphics[width=\linewidth]{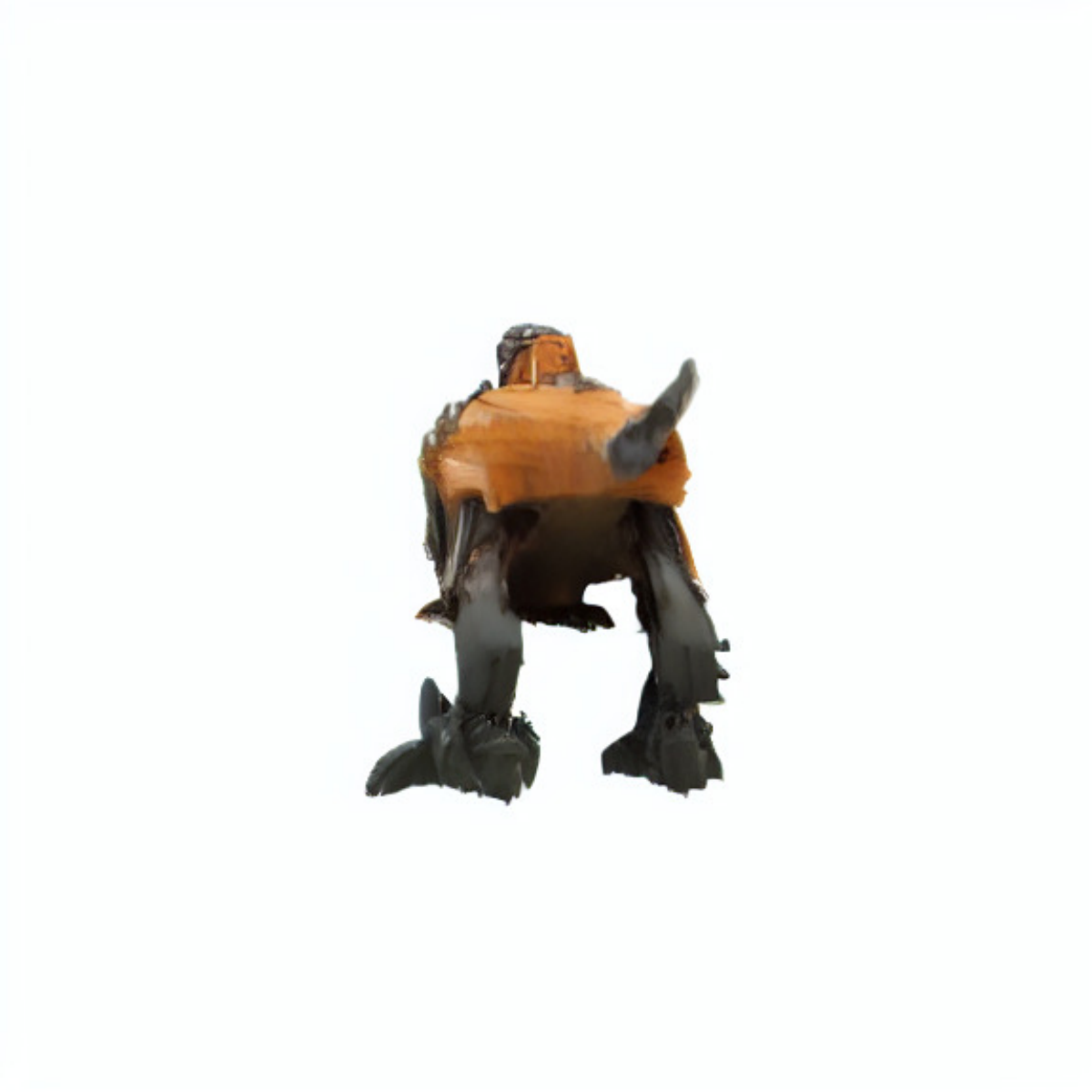}
			\end{subfigure} &
			\begin{subfigure}[c]{0.104\textwidth}  
				\centering
				\includegraphics[width=\linewidth]{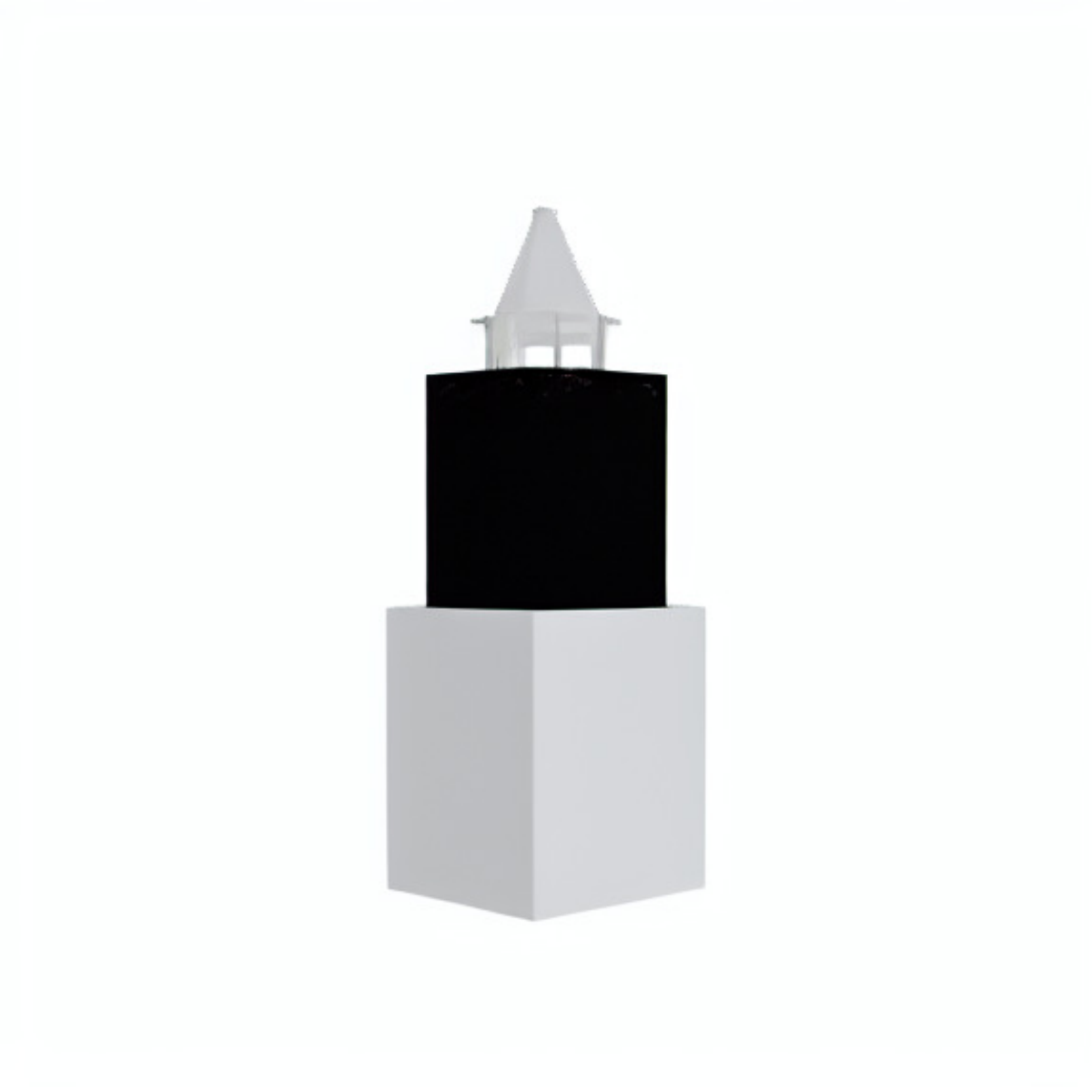}
			\end{subfigure} &
			\begin{subfigure}[c]{0.104\textwidth}  
				\centering
				\includegraphics[width=\linewidth]{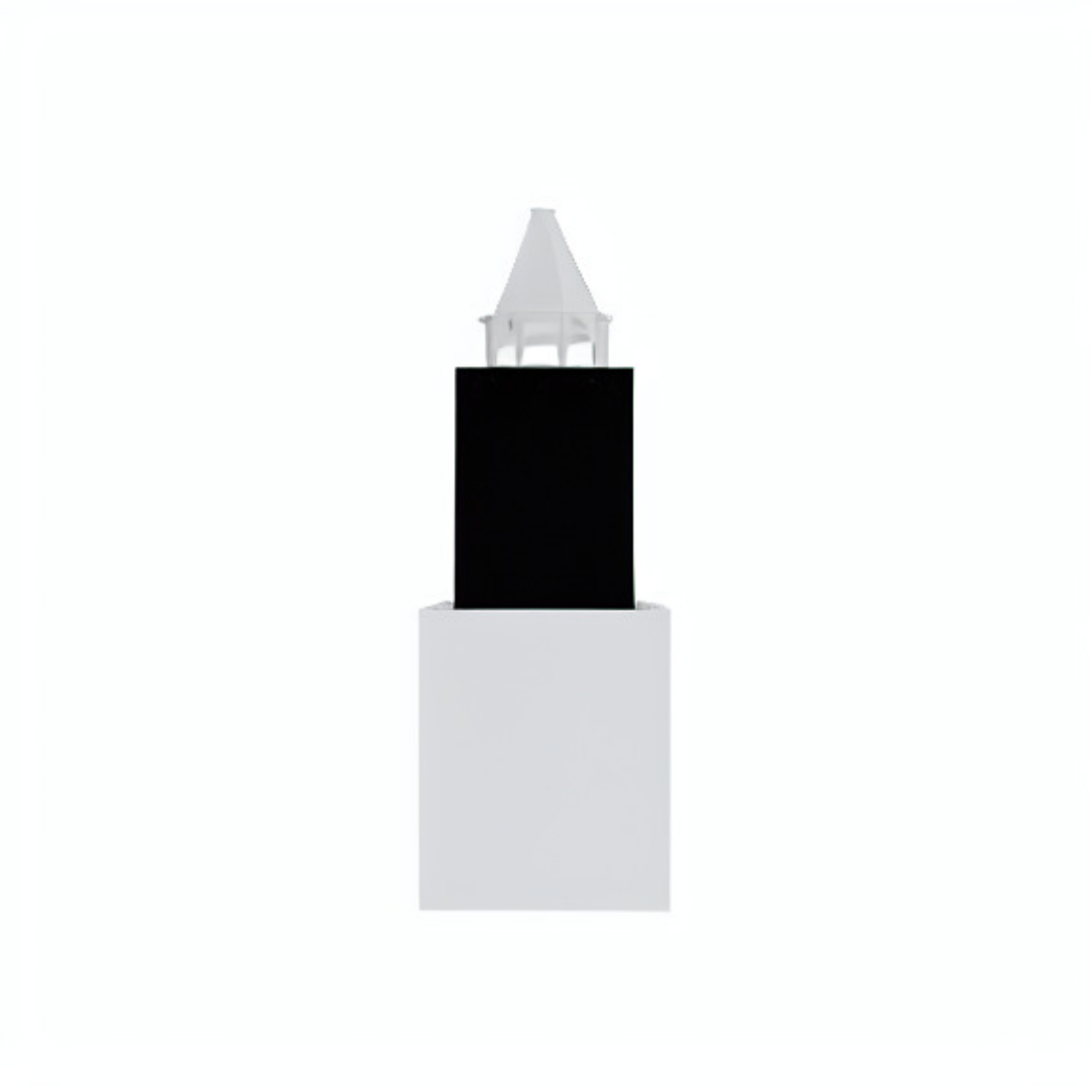}
			\end{subfigure} &
			\begin{subfigure}[c]{0.104\textwidth} 
				\centering
				\includegraphics[width=\linewidth]{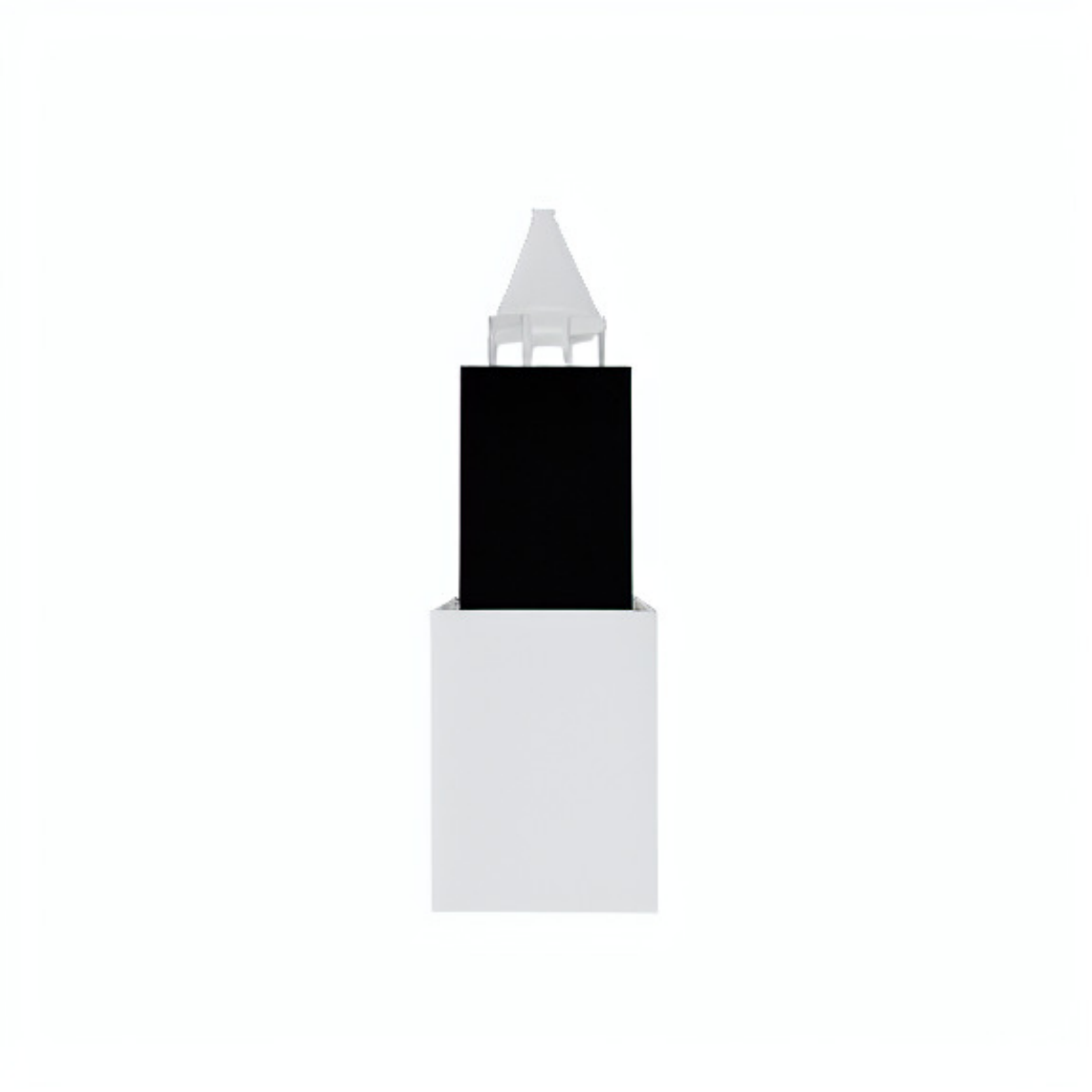}
			\end{subfigure} &
			\begin{subfigure}[c]{0.104\textwidth}  
				\centering
				\includegraphics[width=\linewidth]{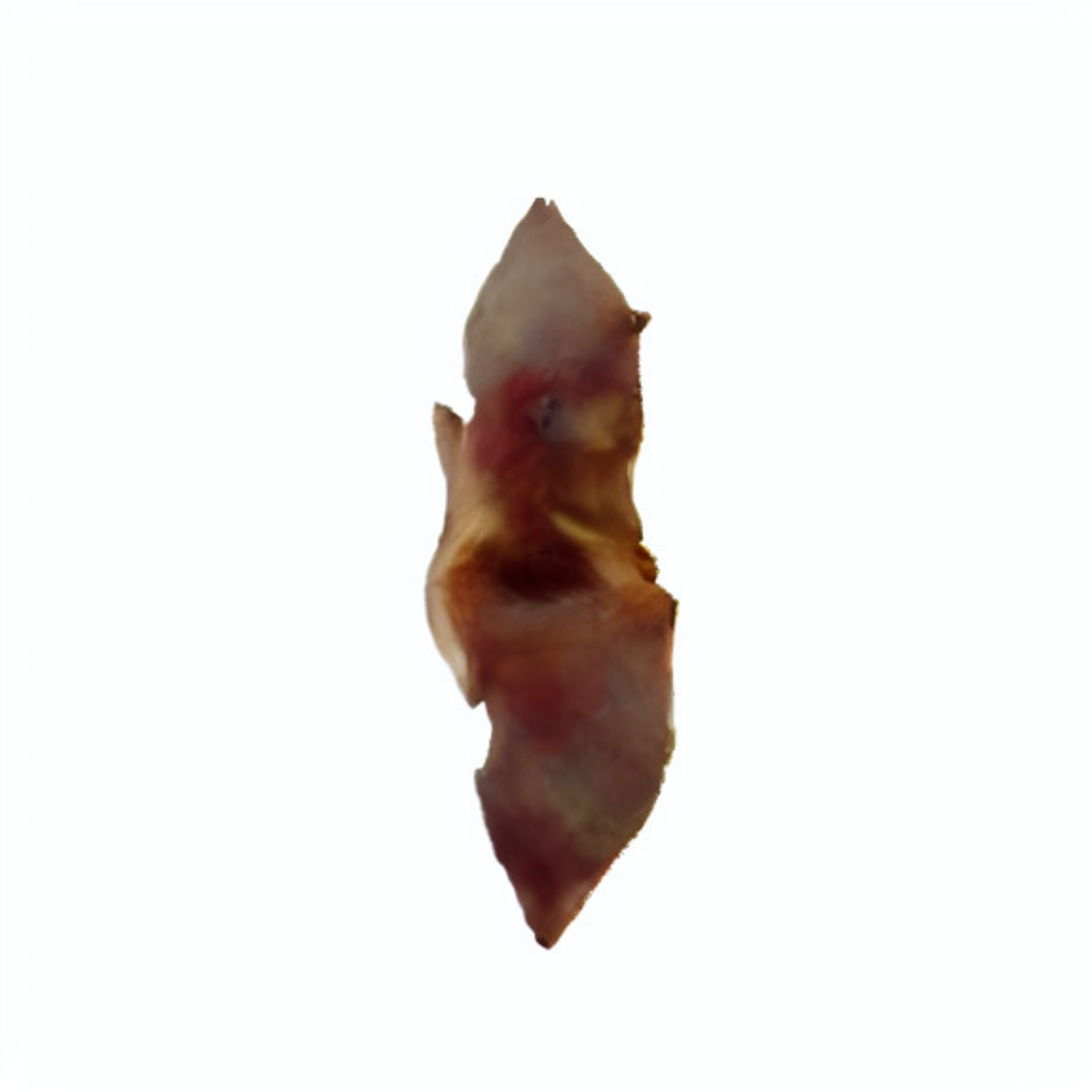}
			\end{subfigure} &
			\begin{subfigure}[c]{0.104\textwidth}  
				\centering
				\includegraphics[width=\linewidth]{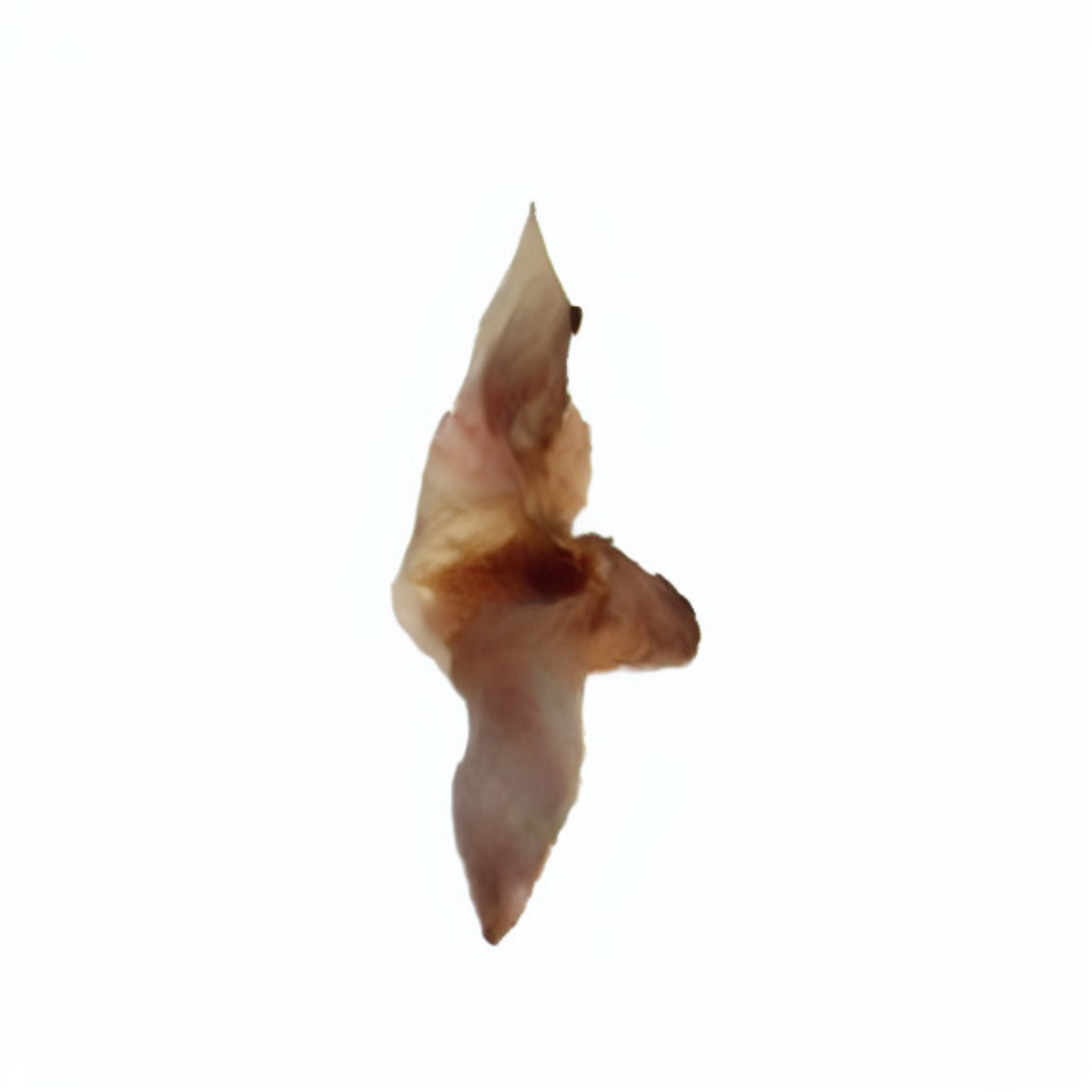}
			\end{subfigure}&
			\begin{subfigure}[c]{0.104\textwidth}  
				\centering
				\includegraphics[width=\linewidth]{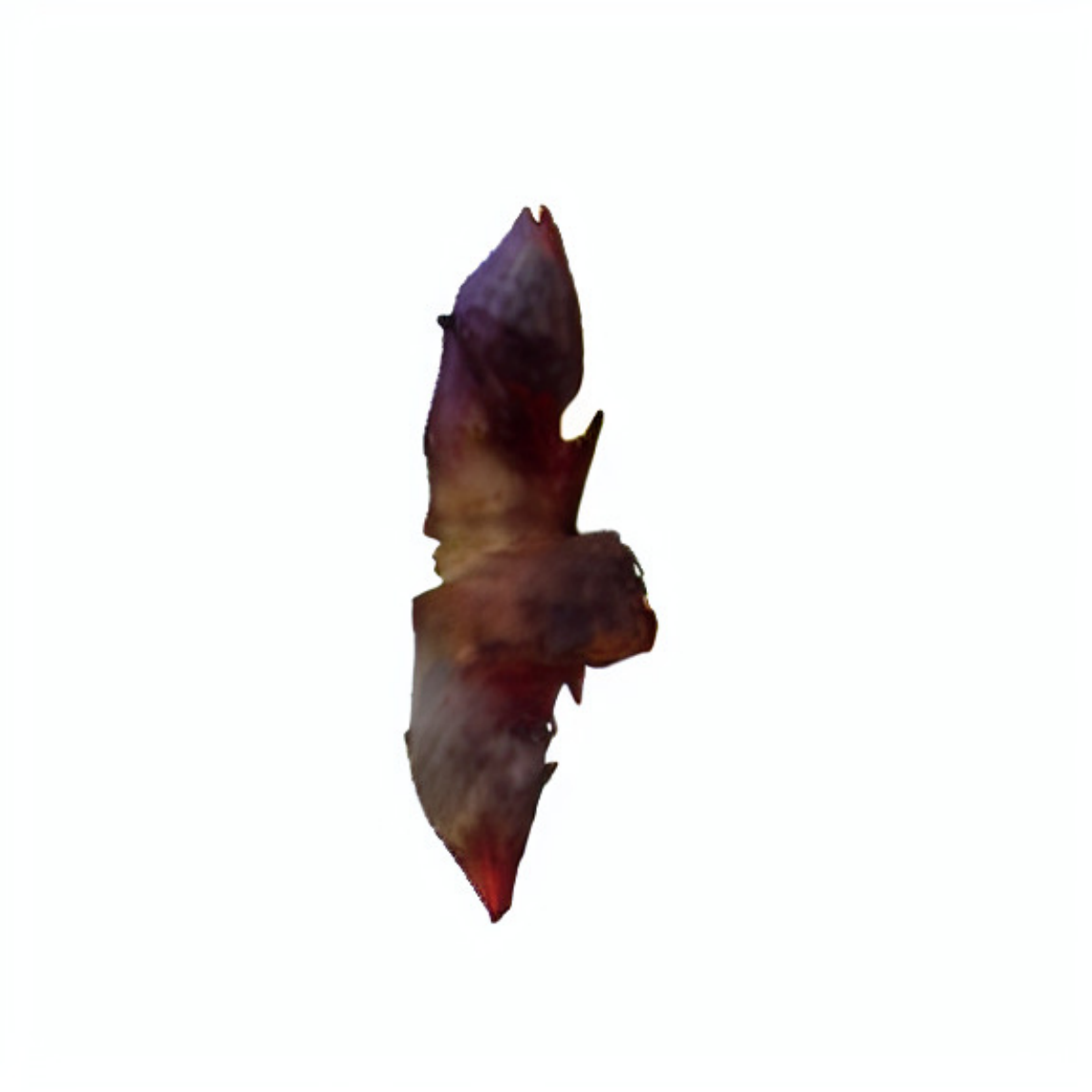}
			\end{subfigure} \\
			\makecell{Vivid\\1-to-3} &
			\begin{subfigure}[c]{0.104\textwidth}  
				\centering
				\includegraphics[width=\linewidth]{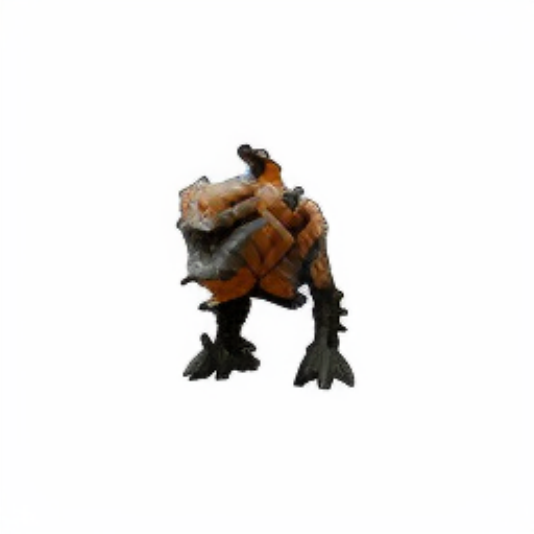}
			\end{subfigure} &
			\begin{subfigure}[c]{0.104\textwidth}  
				\centering
				\includegraphics[width=\linewidth]{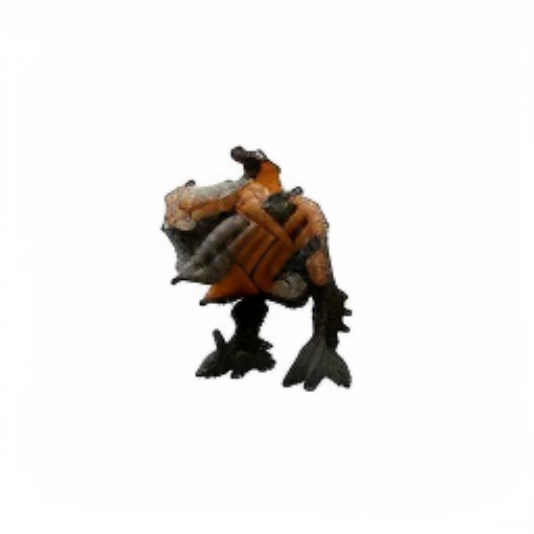}
			\end{subfigure} &
			\begin{subfigure}[c]{0.104\textwidth}  
				\centering
				\includegraphics[width=\linewidth]{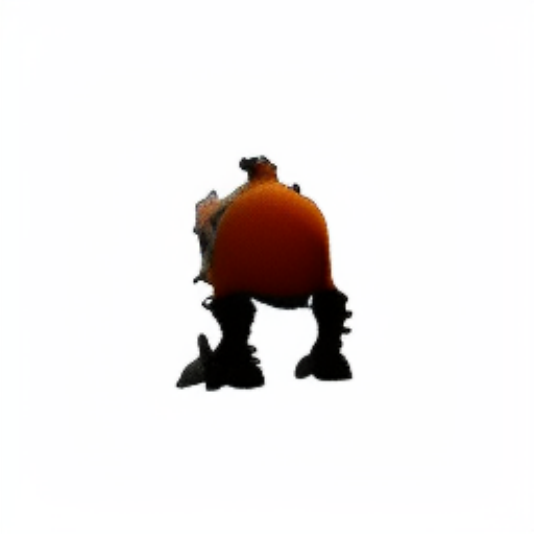}
			\end{subfigure} &
			\begin{subfigure}[c]{0.104\textwidth}  
				\centering
				\includegraphics[width=\linewidth]{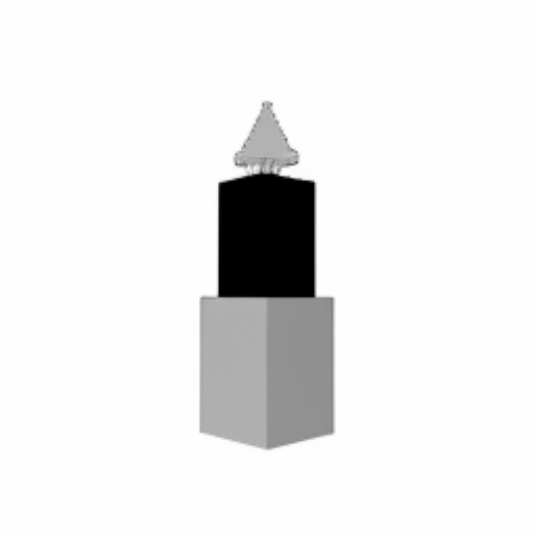}
			\end{subfigure} &
			\begin{subfigure}[c]{0.104\textwidth}  
				\centering
				\includegraphics[width=\linewidth]{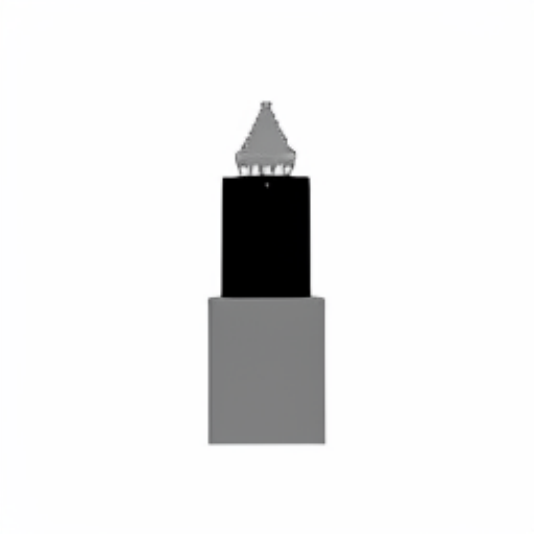}
			\end{subfigure} &
			\begin{subfigure}[c]{0.104\textwidth} 
				\centering
				\includegraphics[width=\linewidth]{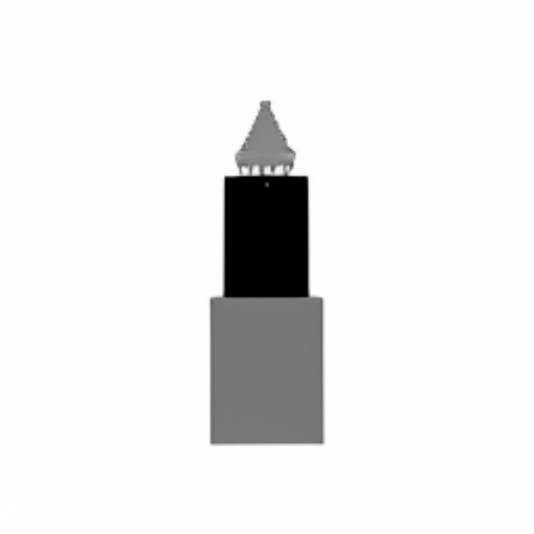}
			\end{subfigure} &
			\begin{subfigure}[c]{0.104\textwidth}  
				\centering
				\includegraphics[width=\linewidth]{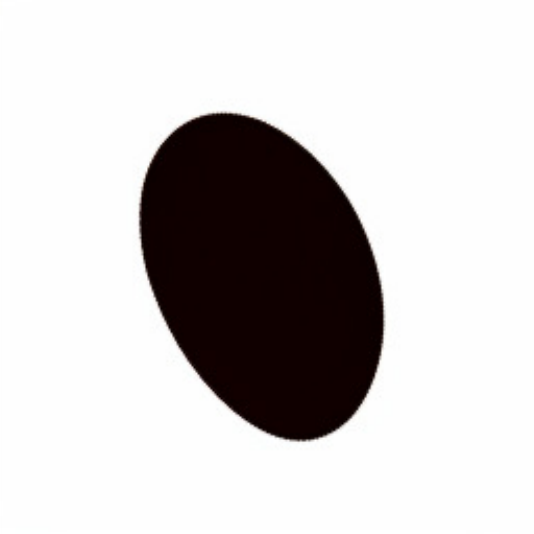}
			\end{subfigure} &
			\begin{subfigure}[c]{0.104\textwidth}  
				\centering
				\includegraphics[width=\linewidth]{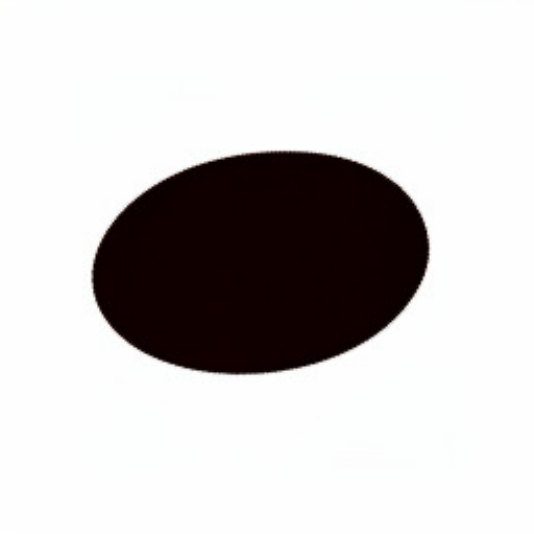}
			\end{subfigure}&
			\begin{subfigure}[c]{0.104\textwidth}  
				\centering
				\includegraphics[width=\linewidth]{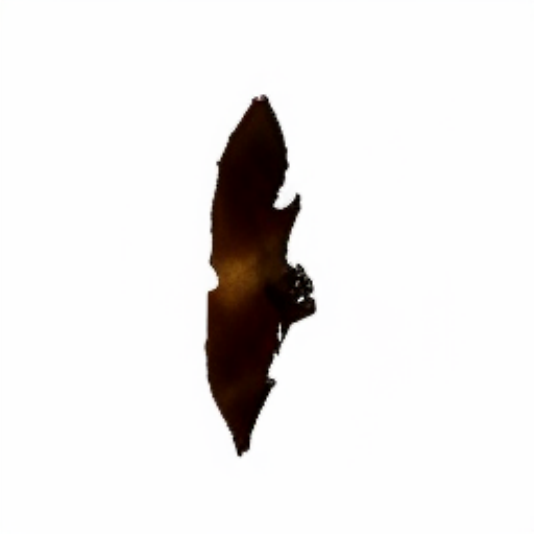}
			\end{subfigure} \\
			\makecell{Zero\\1-to-3\\XL} &
			\begin{subfigure}[c]{0.104\textwidth}  
				\centering
				\includegraphics[width=\linewidth]{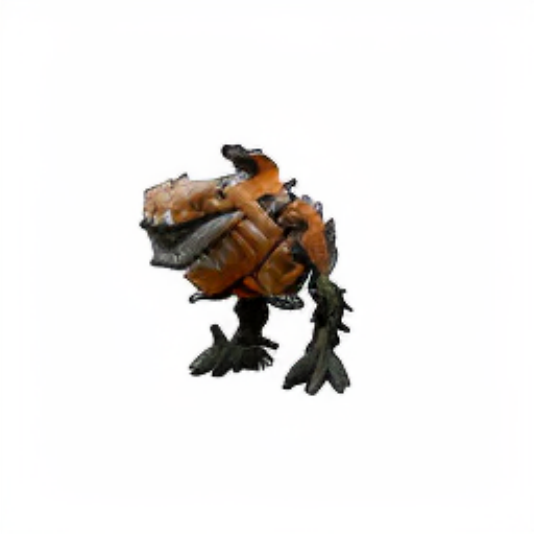}
			\end{subfigure} &
			\begin{subfigure}[c]{0.104\textwidth}  
				\centering
				\includegraphics[width=\linewidth]{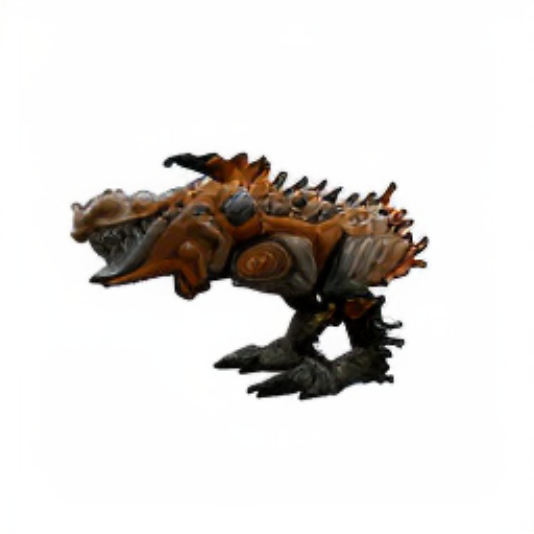}
			\end{subfigure} &
			\begin{subfigure}[c]{0.104\textwidth}  
				\centering
				\includegraphics[width=\linewidth]{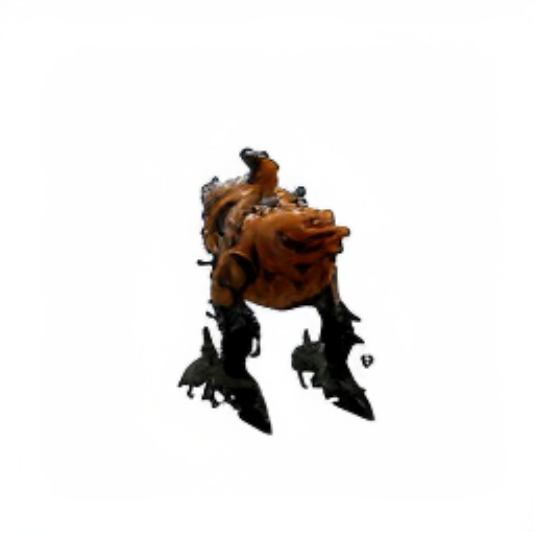}
			\end{subfigure} &
			\begin{subfigure}[c]{0.104\textwidth}  
				\centering
				\includegraphics[width=\linewidth]{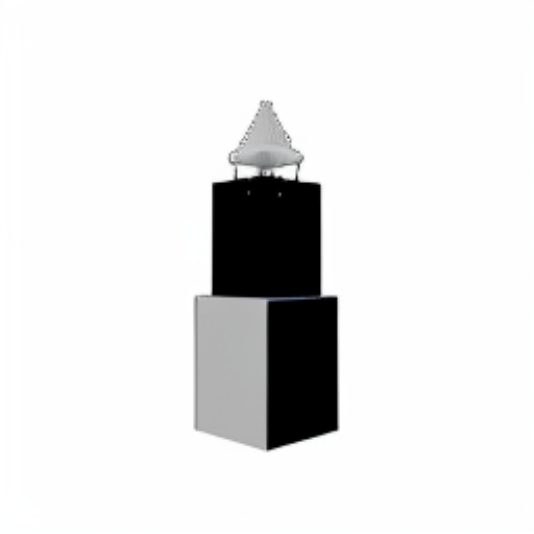}
			\end{subfigure} &
			\begin{subfigure}[c]{0.104\textwidth}  
				\centering
				\includegraphics[width=\linewidth]{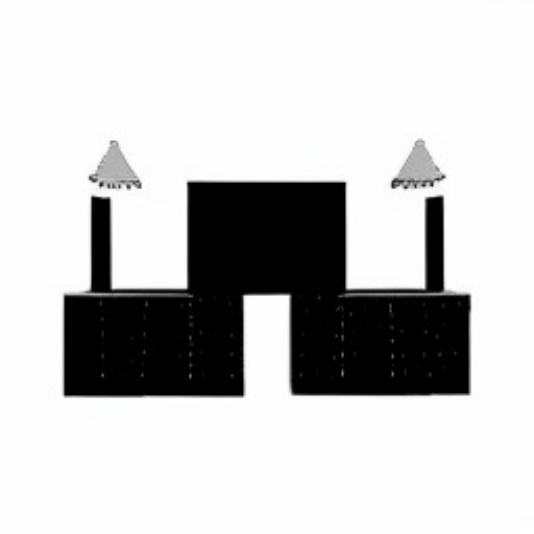}
			\end{subfigure} &
			\begin{subfigure}[c]{0.104\textwidth} 
				\centering
				\includegraphics[width=\linewidth]{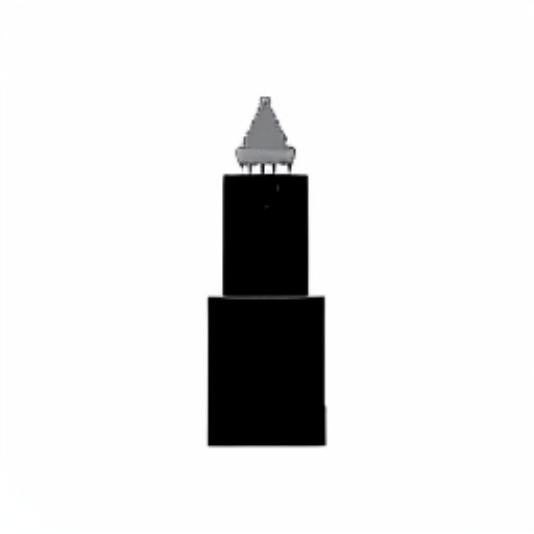}
			\end{subfigure} &
			\begin{subfigure}[c]{0.104\textwidth}  
				\centering
				\includegraphics[width=\linewidth]{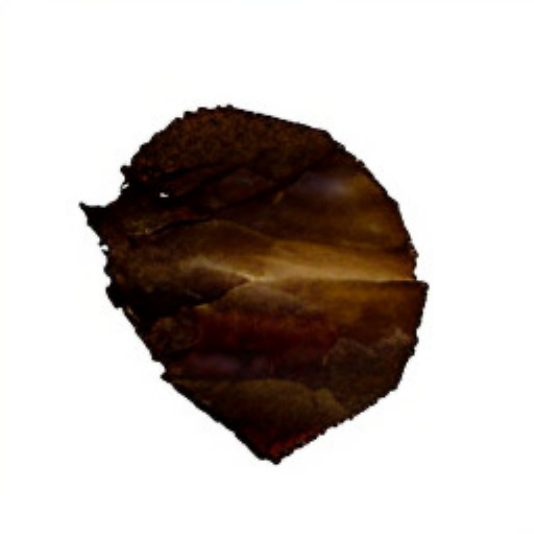}
			\end{subfigure} &
			\begin{subfigure}[c]{0.104\textwidth}  
				\centering
				\includegraphics[width=\linewidth]{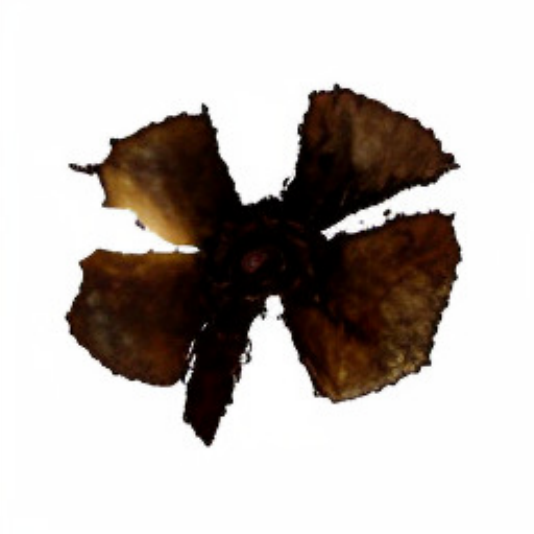}
			\end{subfigure}&
			\begin{subfigure}[c]{0.104\textwidth}  
				\centering
				\includegraphics[width=\linewidth]{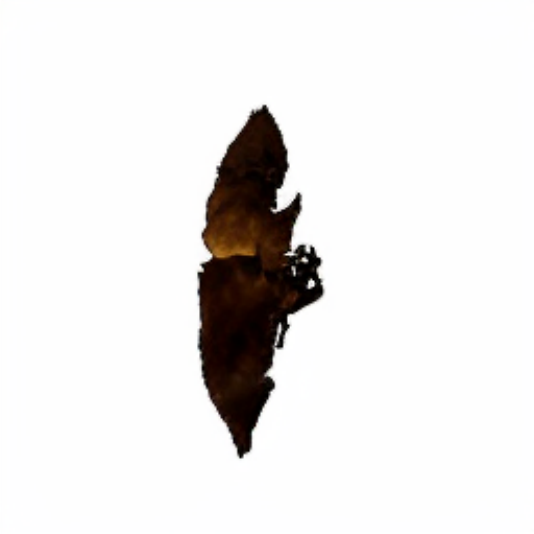}
			\end{subfigure} \\
			Mv-Adapter &
			\begin{subfigure}[c]{0.104\textwidth}  
				\centering
				\includegraphics[width=\linewidth]{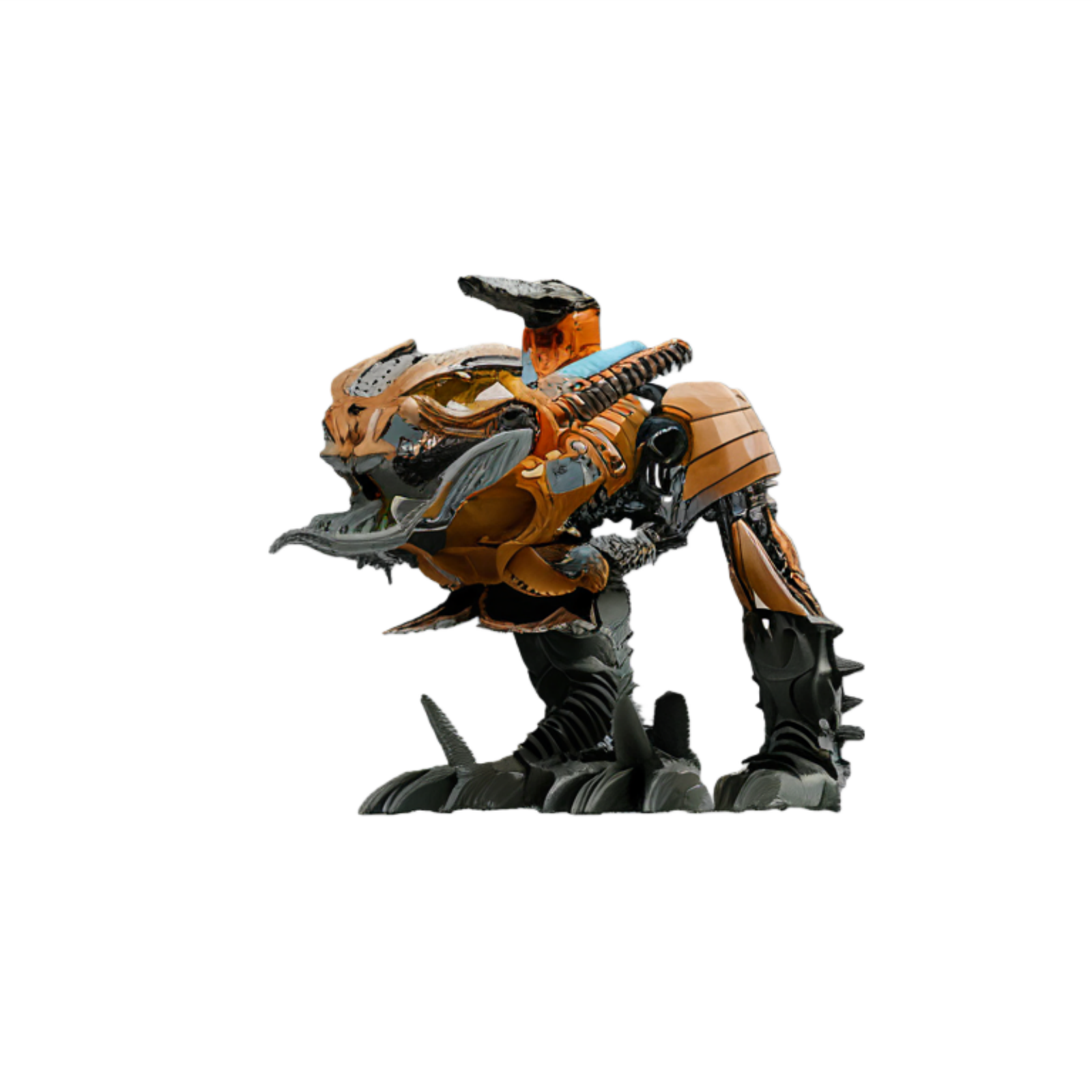}
			\end{subfigure} &
			\begin{subfigure}[c]{0.104\textwidth}  
				\centering
				\includegraphics[width=\linewidth]{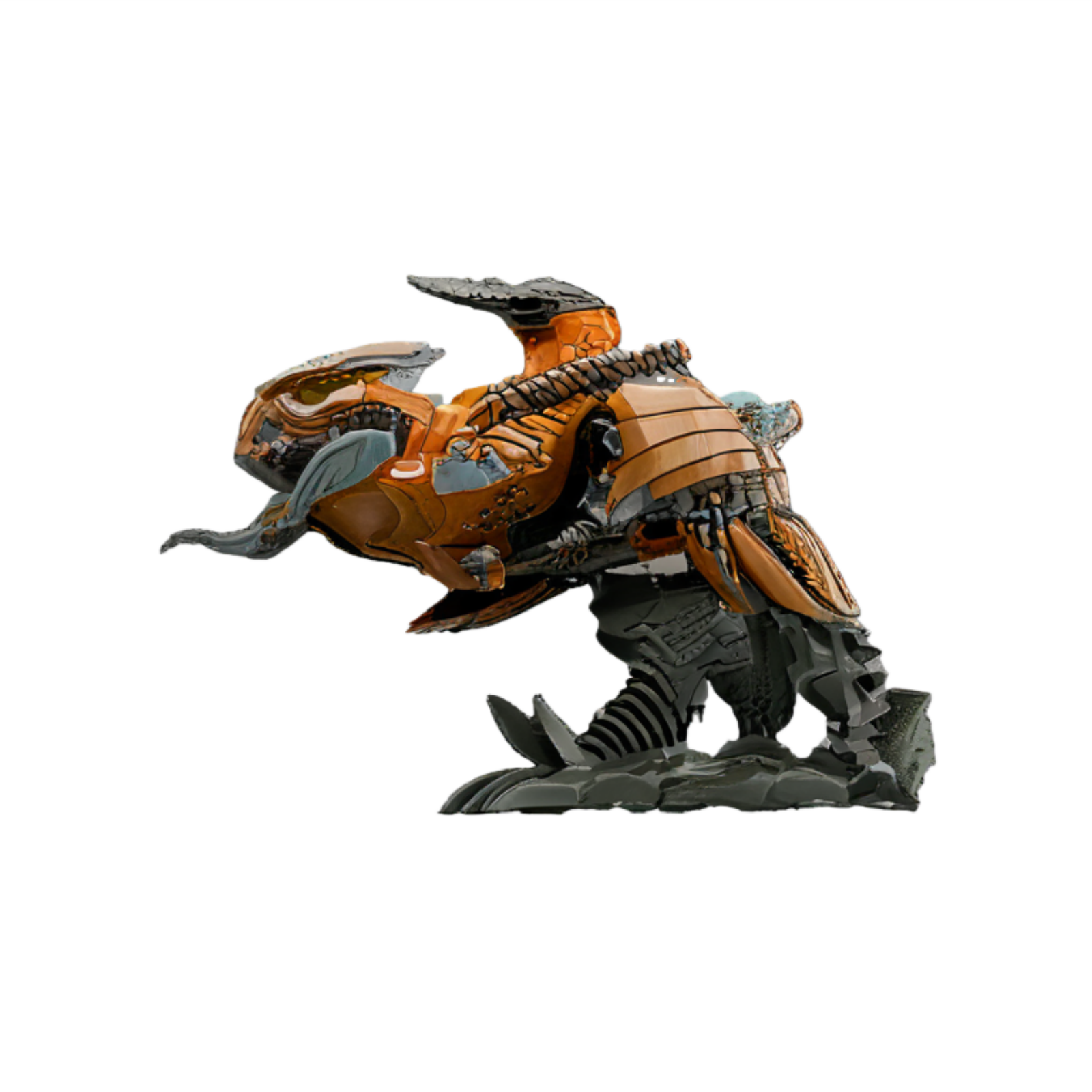}
			\end{subfigure} &
			\begin{subfigure}[c]{0.104\textwidth}  
				\centering
				\includegraphics[width=\linewidth]{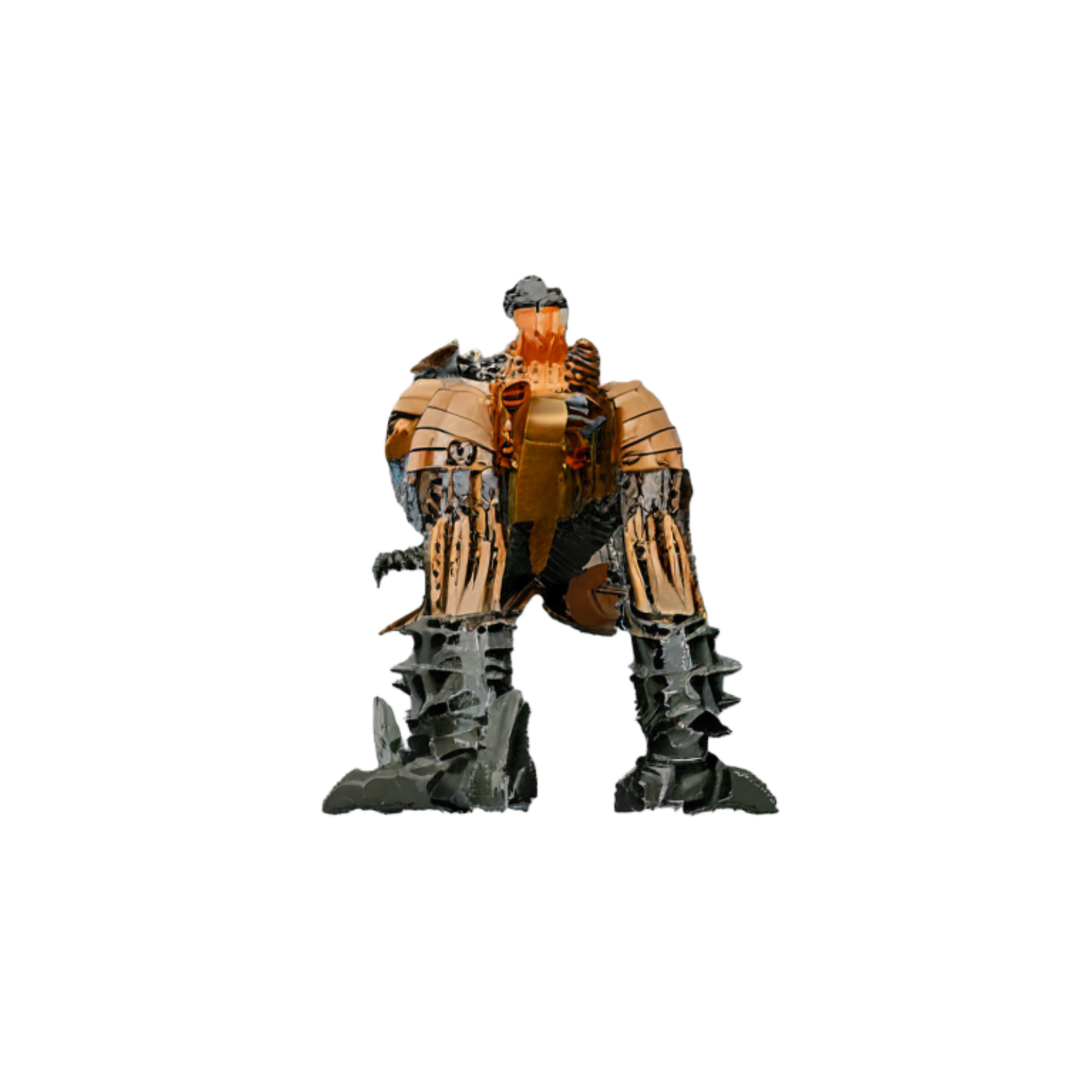}
			\end{subfigure} &
			\begin{subfigure}[c]{0.104\textwidth}  
				\centering
				\includegraphics[width=\linewidth]{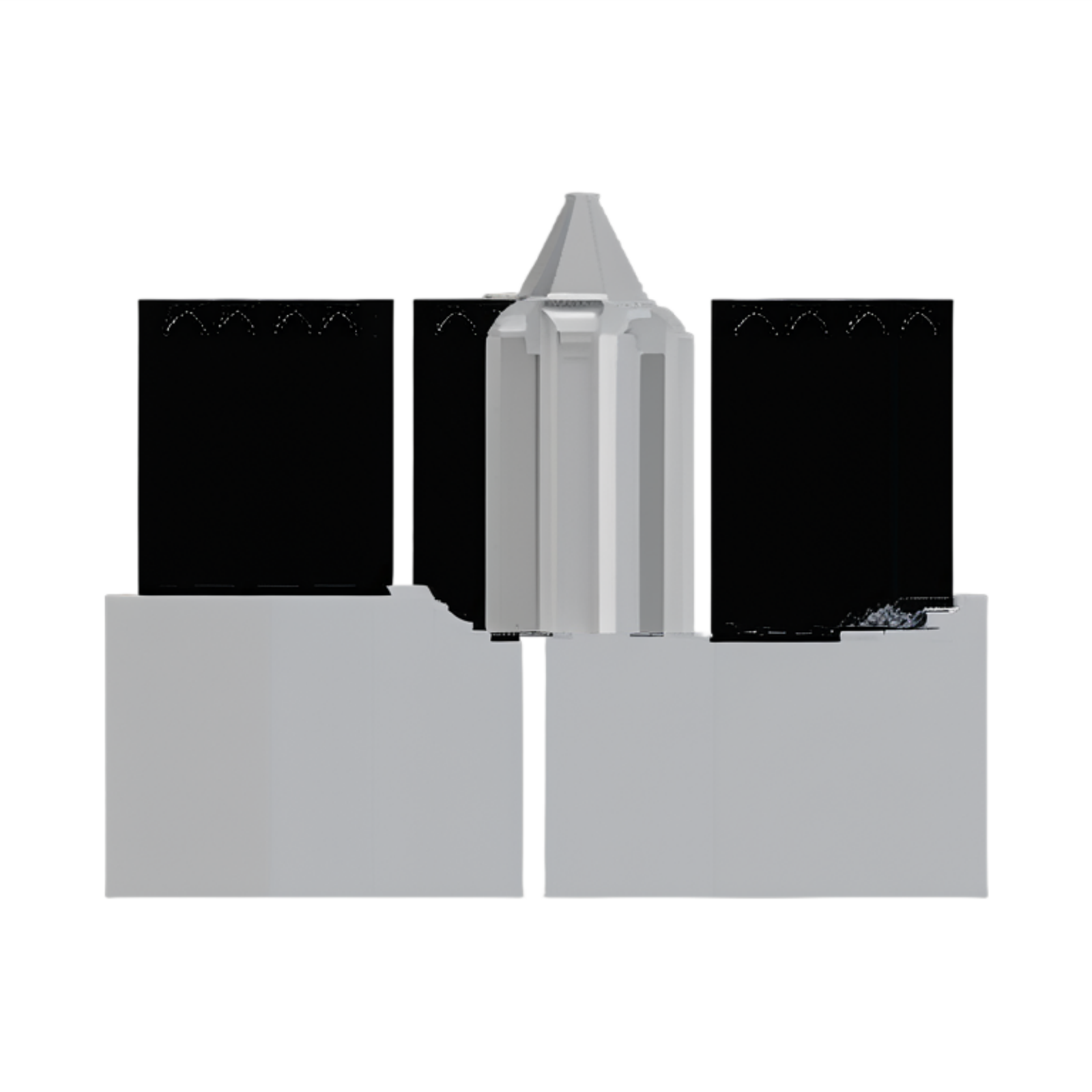}
			\end{subfigure} &
			\begin{subfigure}[c]{0.104\textwidth}  
				\centering
				\includegraphics[width=\linewidth]{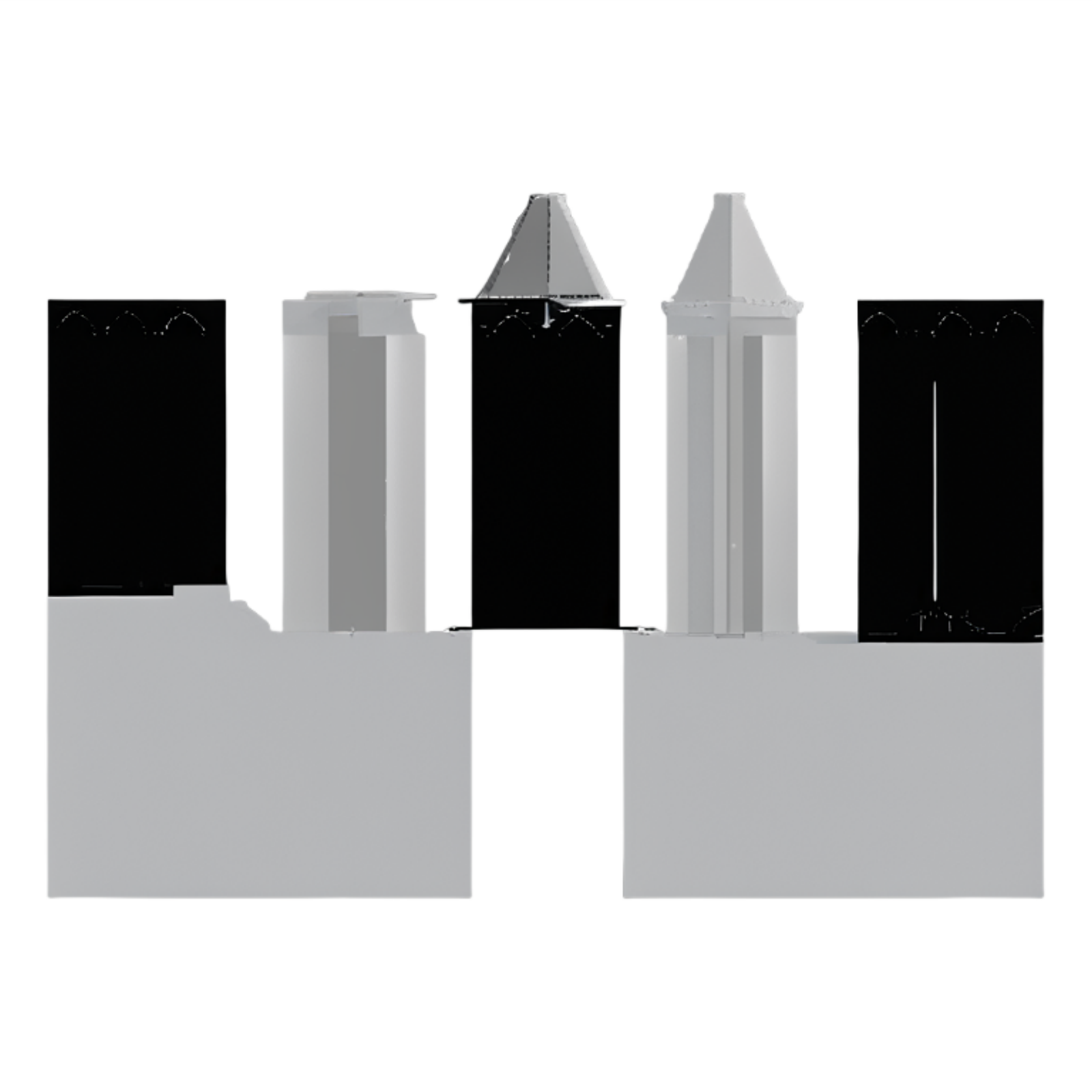}
			\end{subfigure} &
			\begin{subfigure}[c]{0.104\textwidth} 
				\centering
				\includegraphics[width=\linewidth]{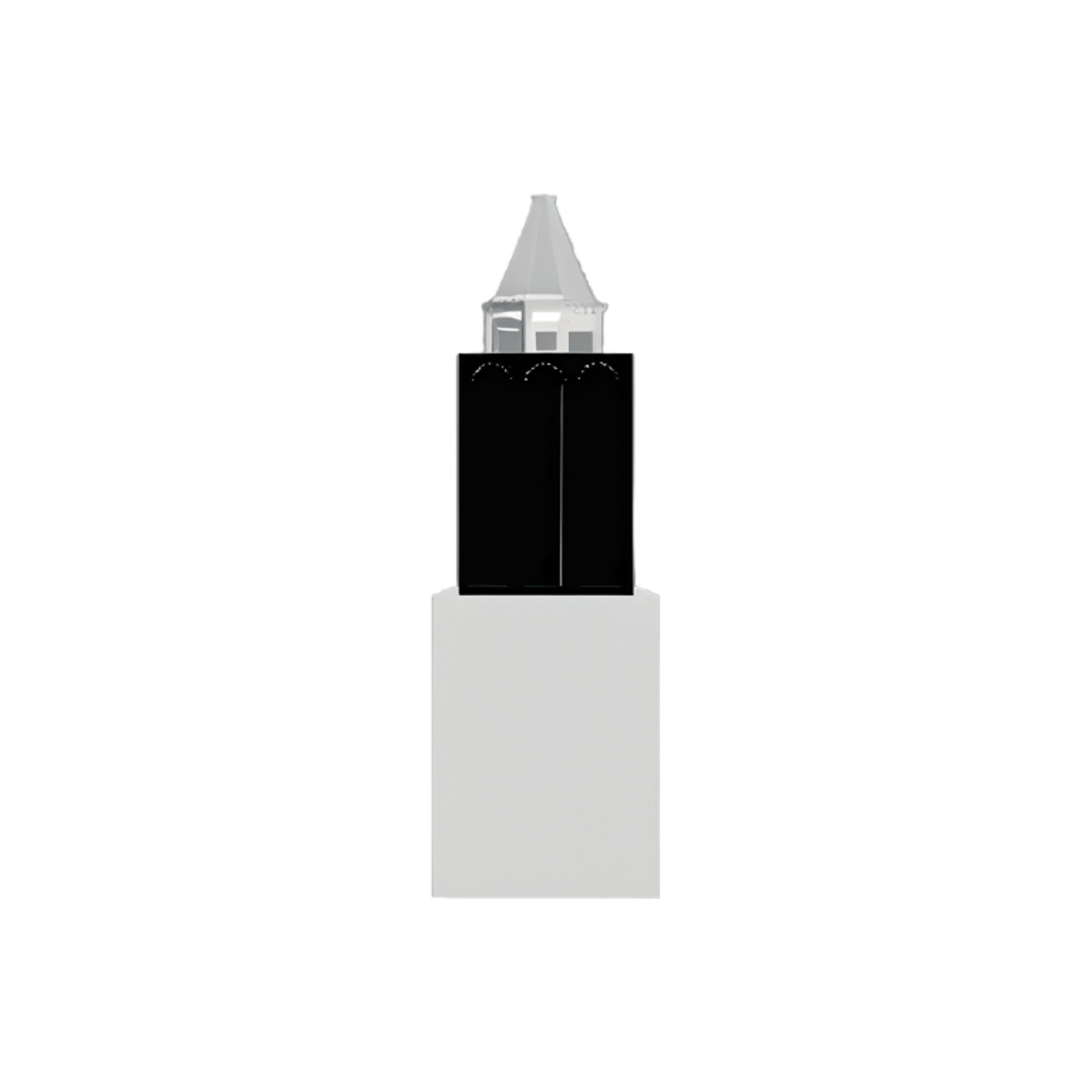}
			\end{subfigure} &
			\begin{subfigure}[c]{0.104\textwidth}  
				\centering
				\includegraphics[width=\linewidth]{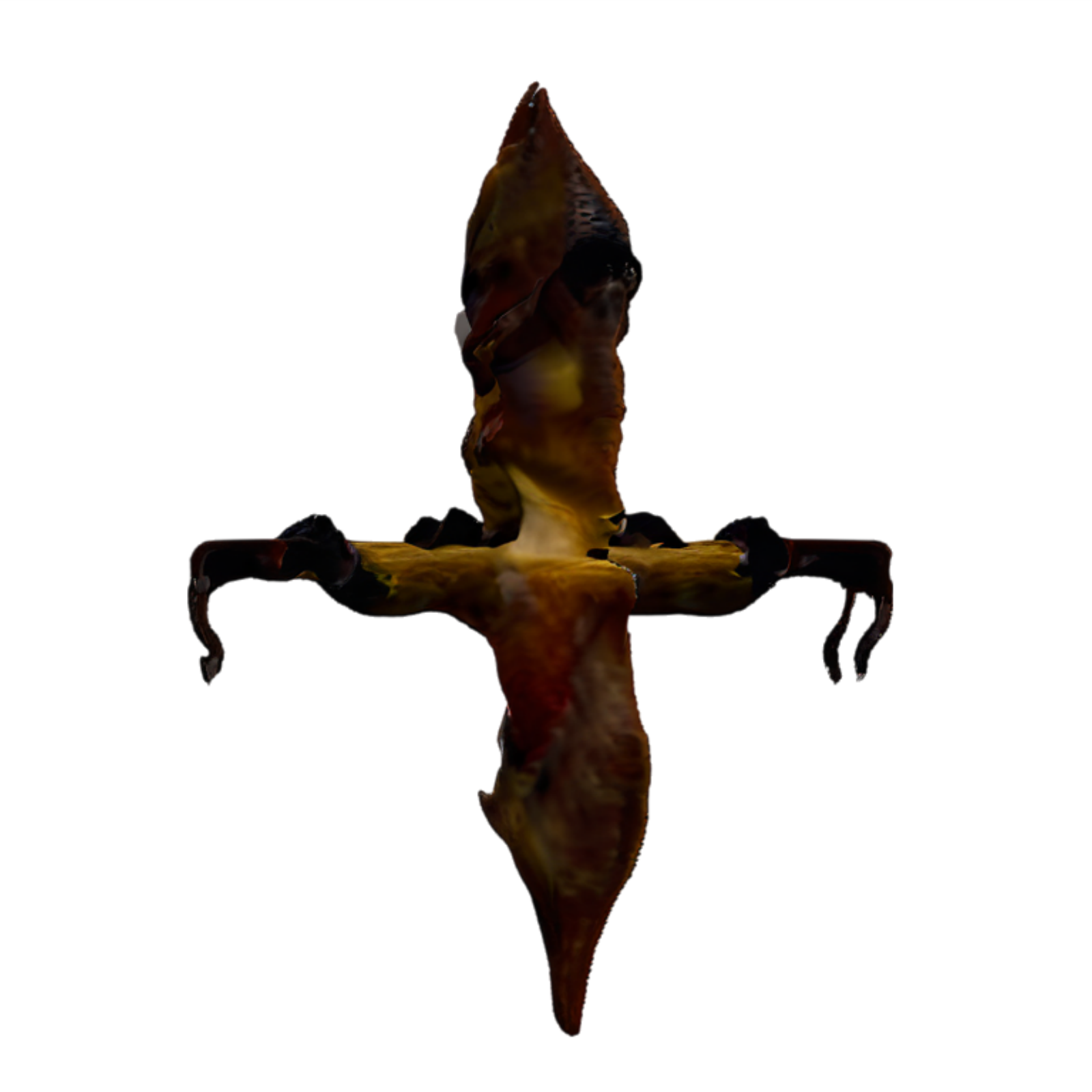}
			\end{subfigure} &
			\begin{subfigure}[c]{0.104\textwidth}  
				\centering
				\includegraphics[width=\linewidth]{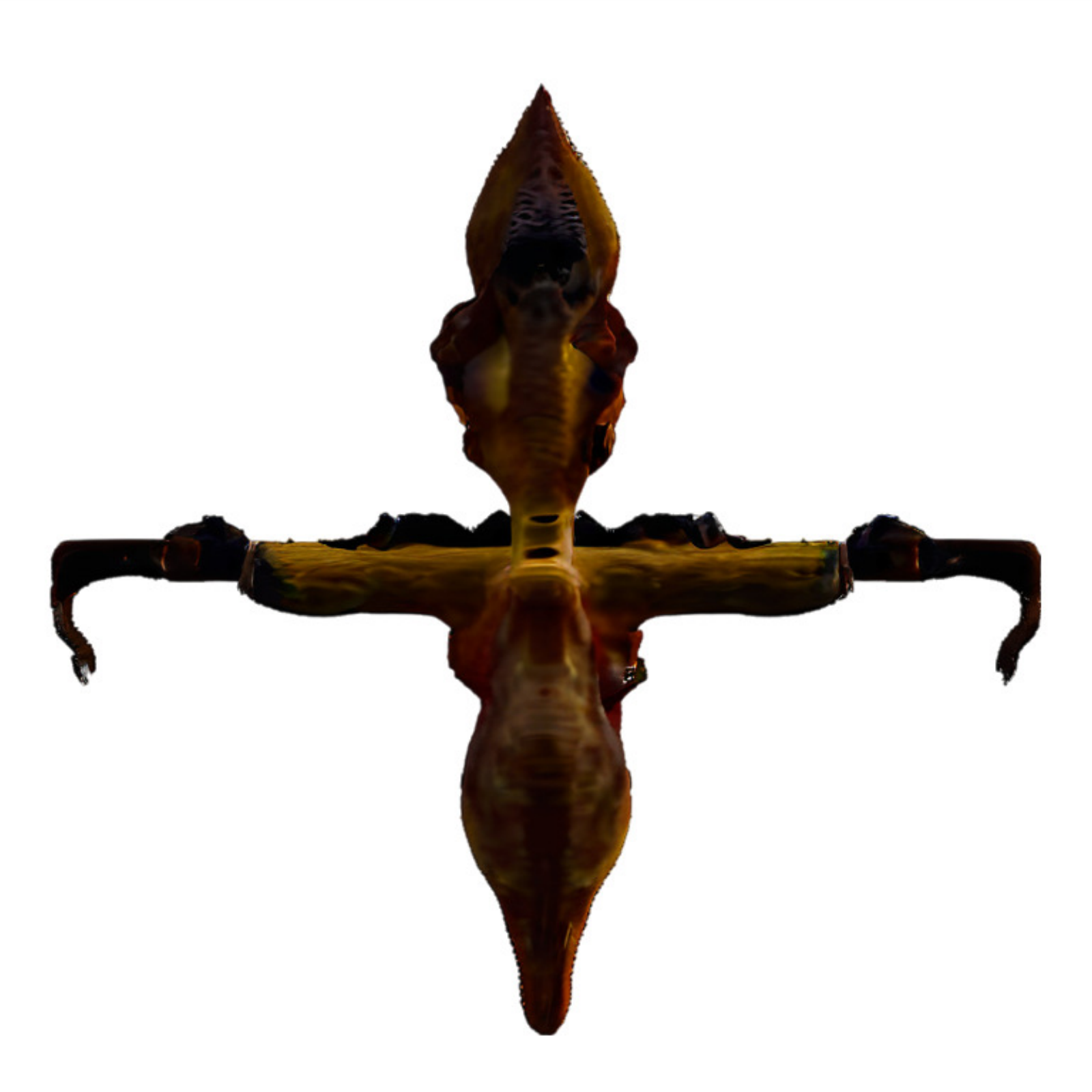}
			\end{subfigure}&
			\begin{subfigure}[c]{0.104\textwidth}  
				\centering
				\includegraphics[width=\linewidth]{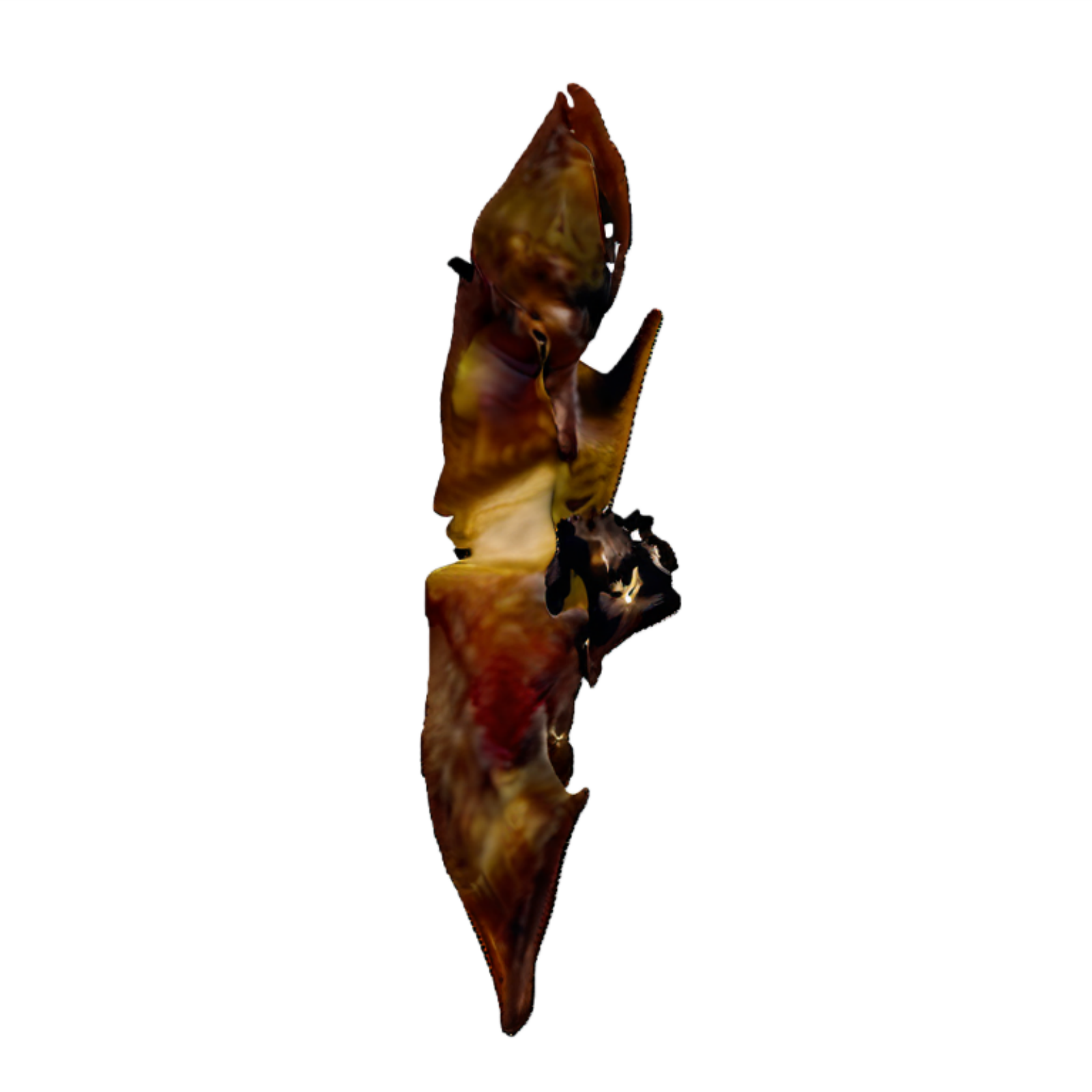}
			\end{subfigure} \\
			\makecell{Mv-Adapter\\\textbf{with}\\\textbf{EDN}} &
			\begin{subfigure}[c]{0.104\textwidth}  
				\centering
				\includegraphics[width=\linewidth]{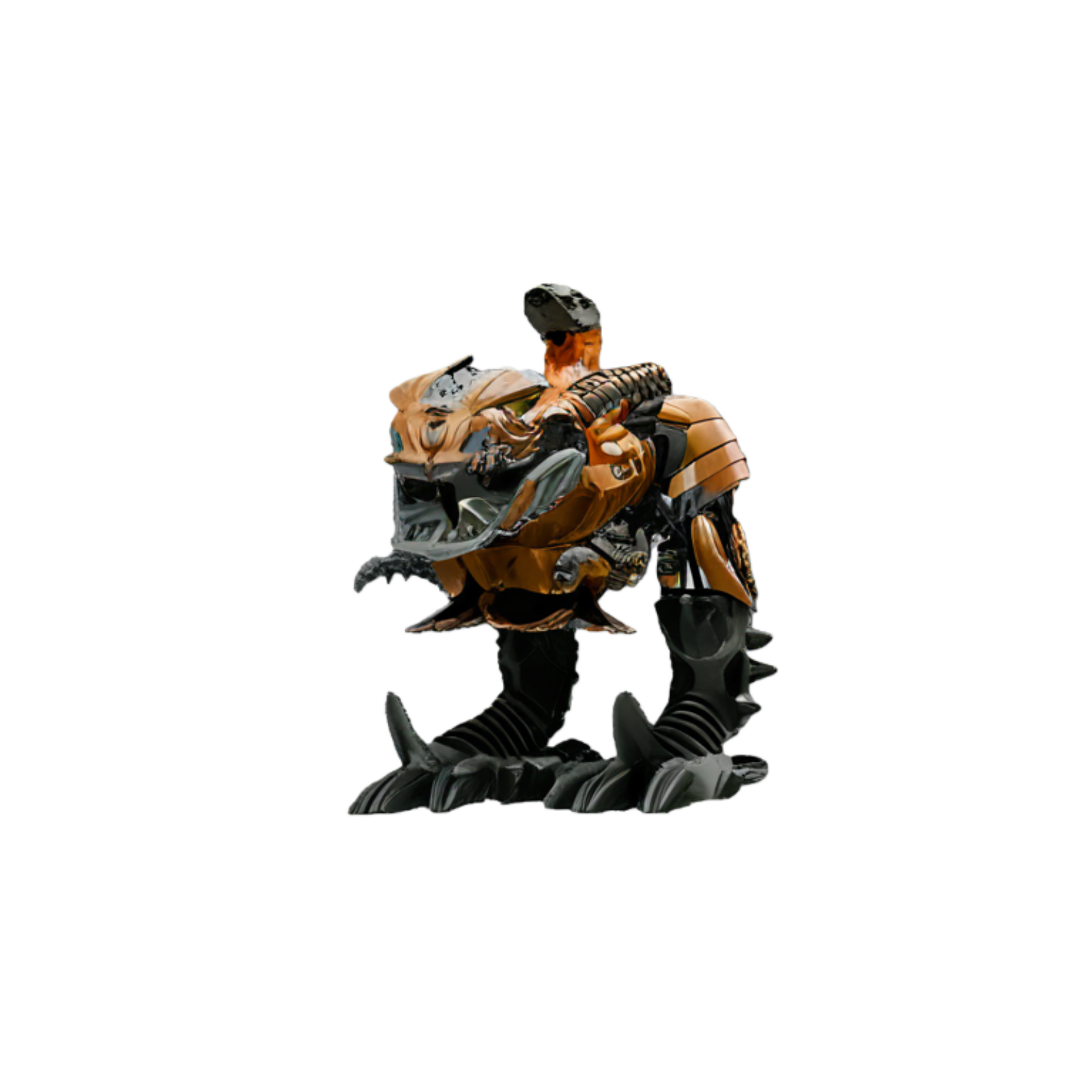}
			\end{subfigure} &
			\begin{subfigure}[c]{0.104\textwidth}  
				\centering
				\includegraphics[width=\linewidth]{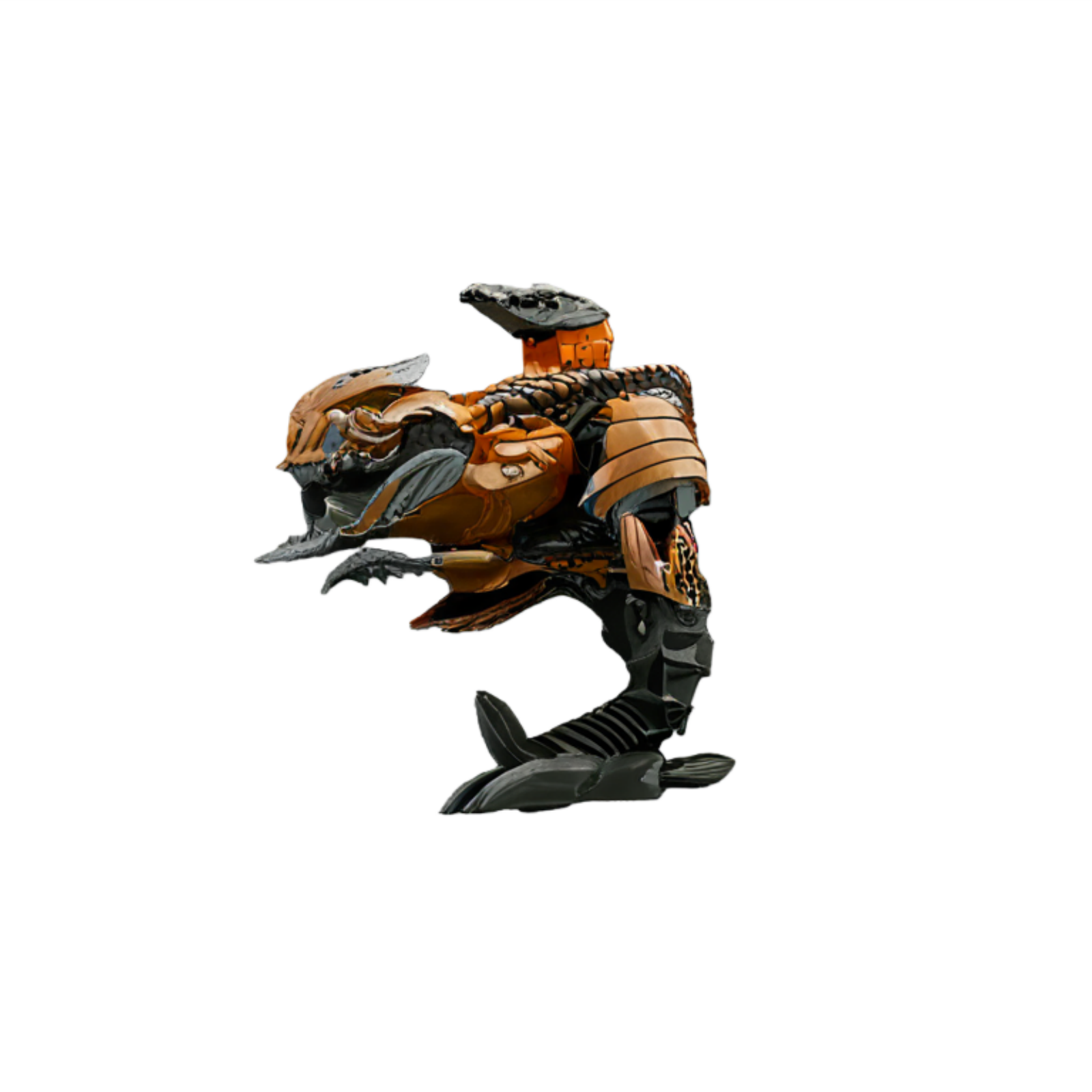}
			\end{subfigure} &
			\begin{subfigure}[c]{0.104\textwidth}  
				\centering
				\includegraphics[width=\linewidth]{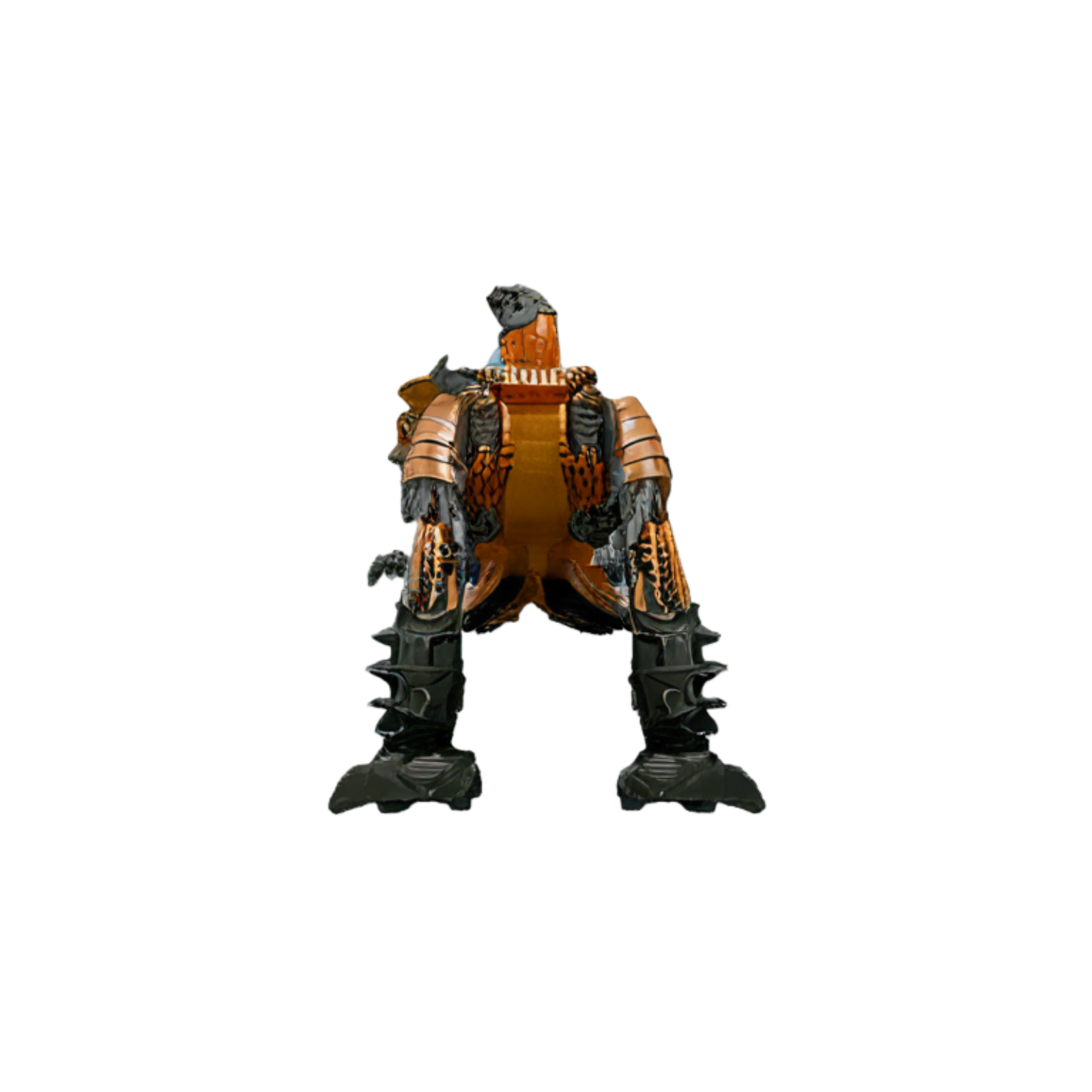}
			\end{subfigure} &
			\begin{subfigure}[c]{0.104\textwidth}  
				\centering
				\includegraphics[width=\linewidth]{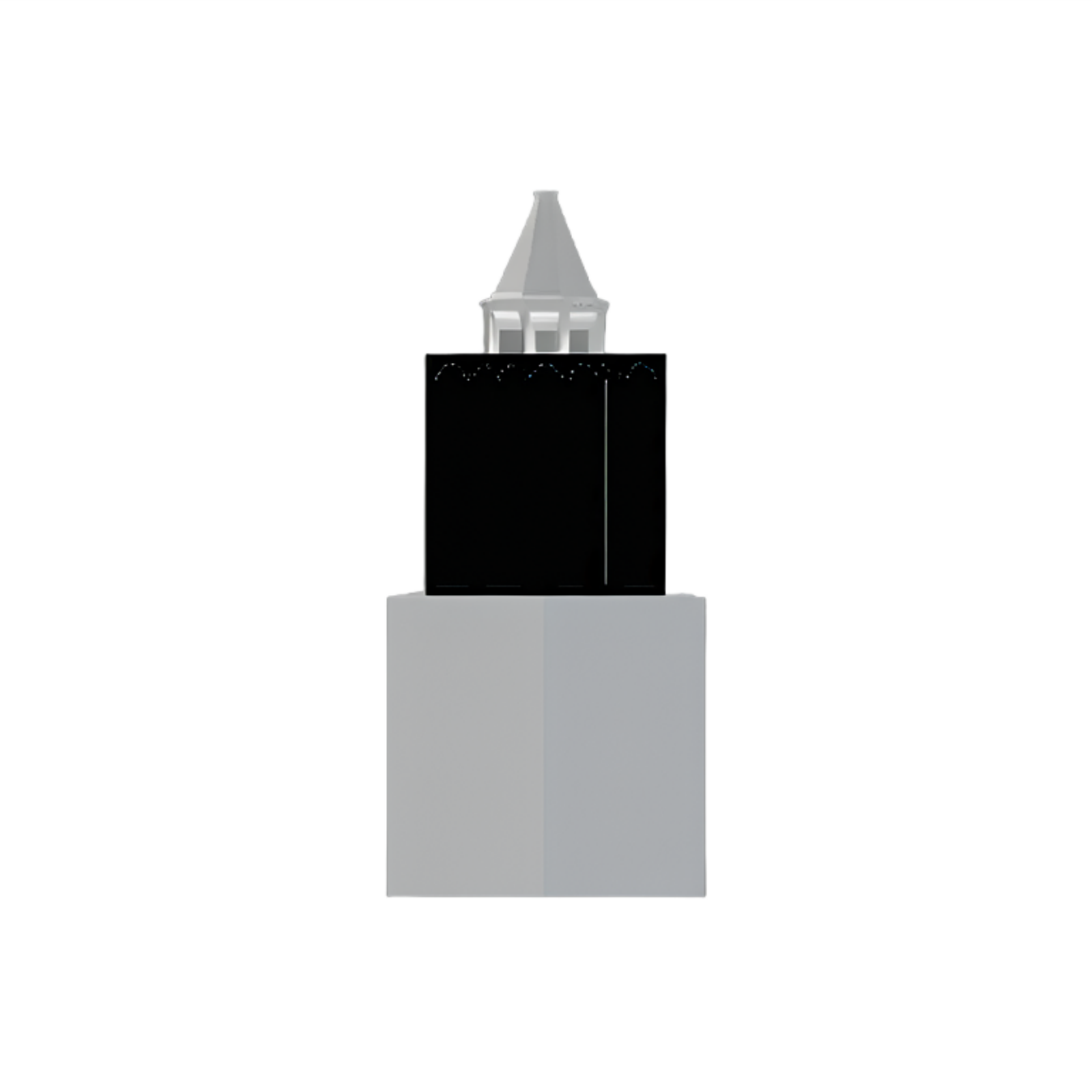}
			\end{subfigure} &
			\begin{subfigure}[c]{0.104\textwidth}  
				\centering
				\includegraphics[width=\linewidth]{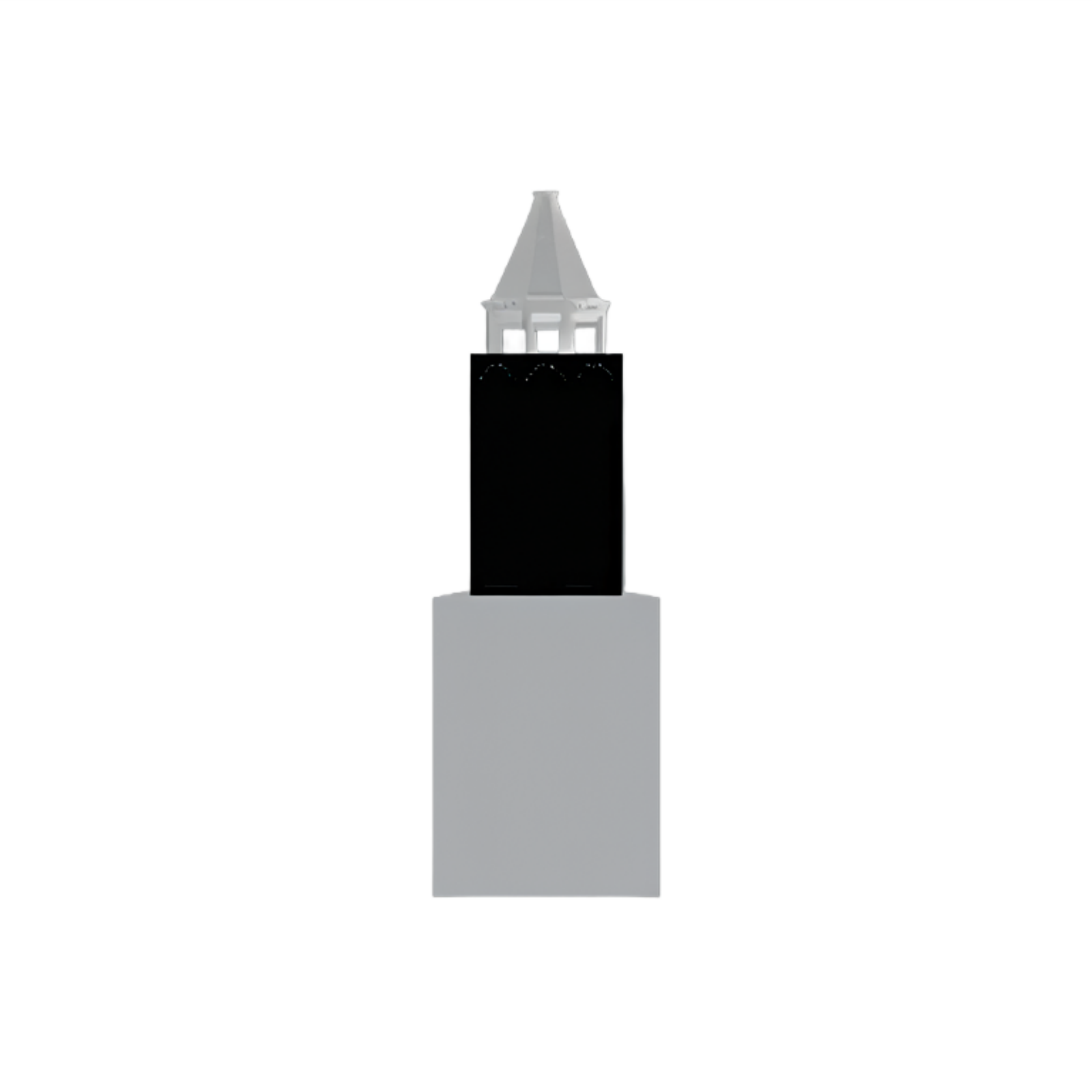}
			\end{subfigure} &
			\begin{subfigure}[c]{0.104\textwidth} 
				\centering
				\includegraphics[width=\linewidth]{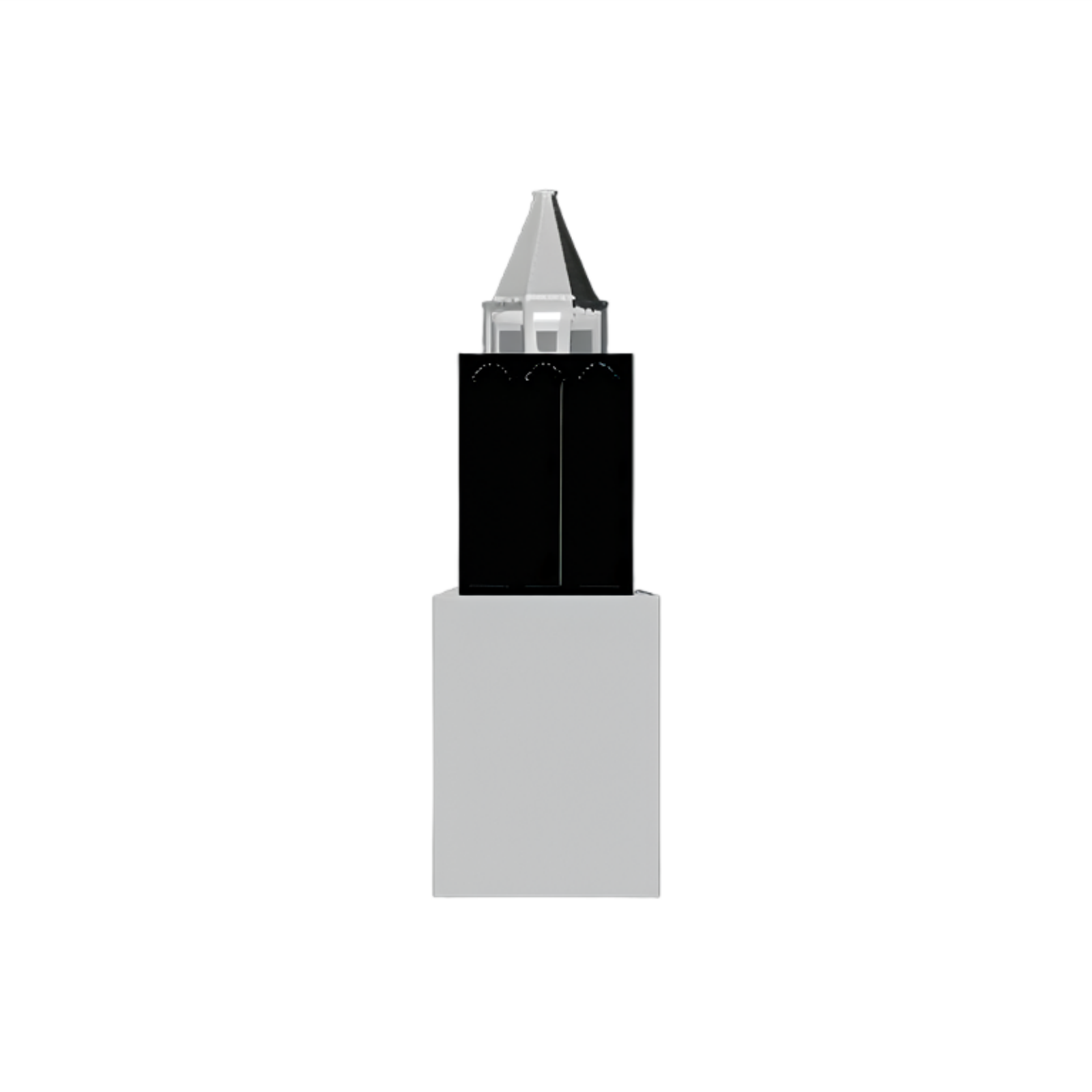}
			\end{subfigure} &
			\begin{subfigure}[c]{0.104\textwidth}  
				\centering
				\includegraphics[width=\linewidth]{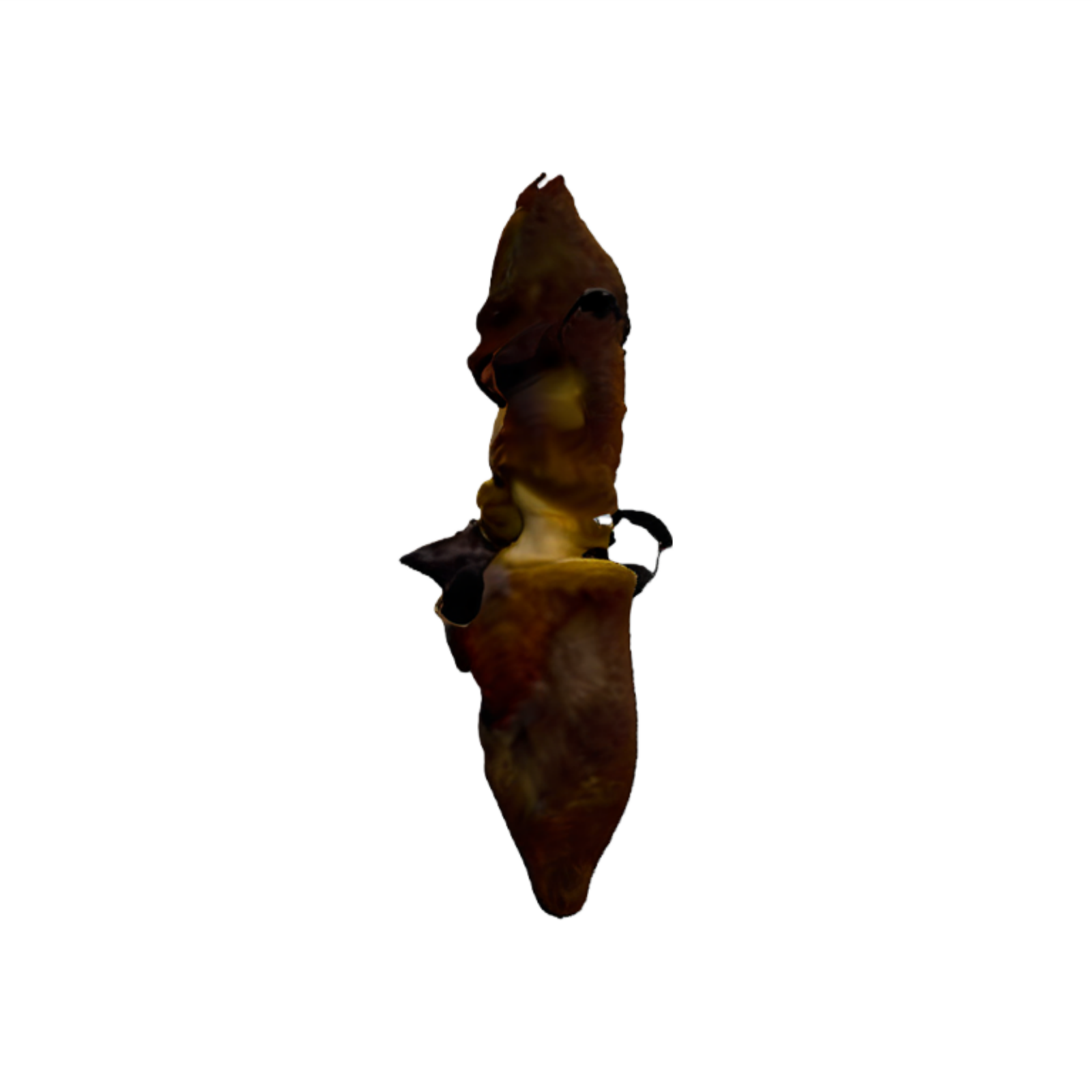}
			\end{subfigure} &
			\begin{subfigure}[c]{0.104\textwidth}  
				\centering
				\includegraphics[width=\linewidth]{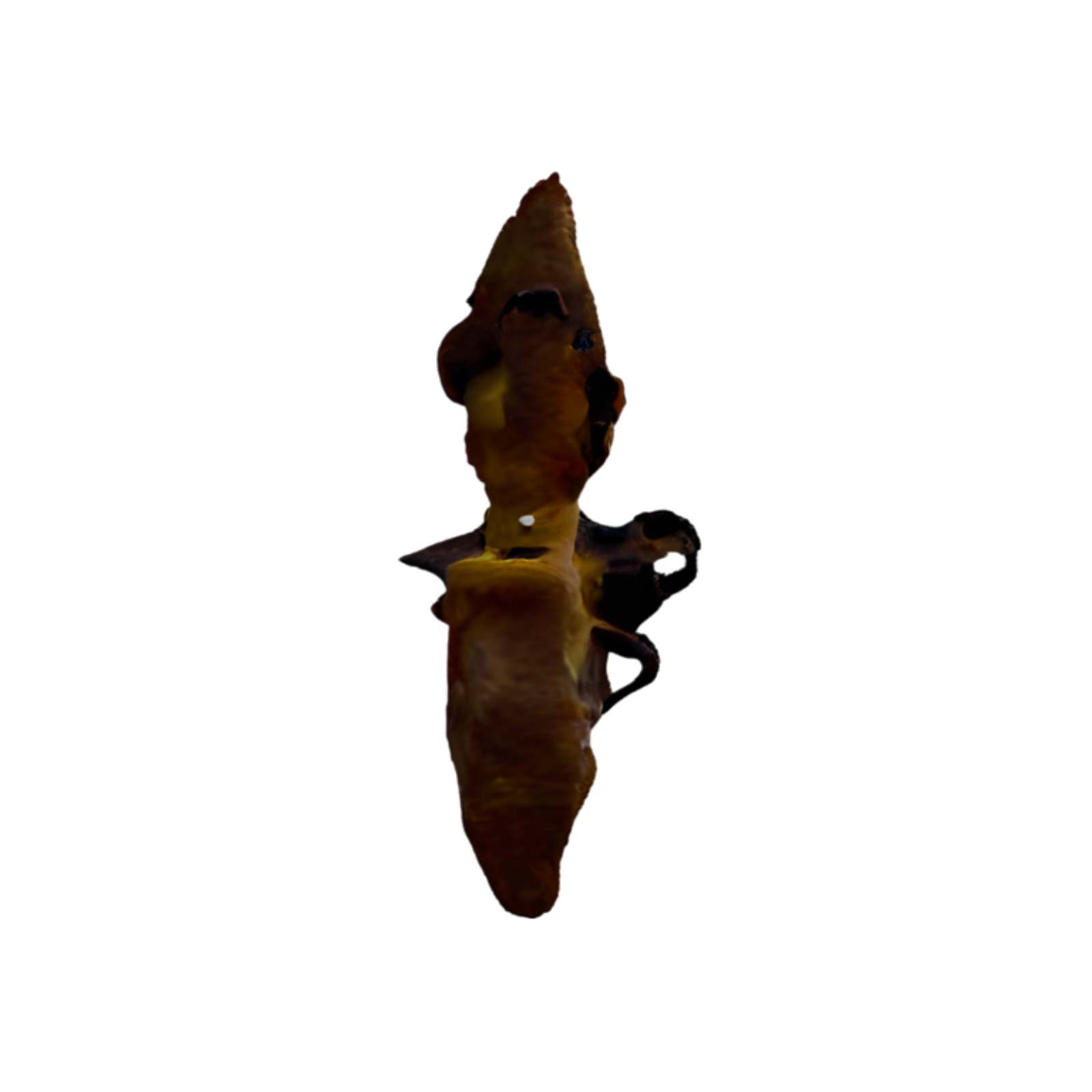}
			\end{subfigure}&
			\begin{subfigure}[c]{0.104\textwidth}  
				\centering
				\includegraphics[width=\linewidth]{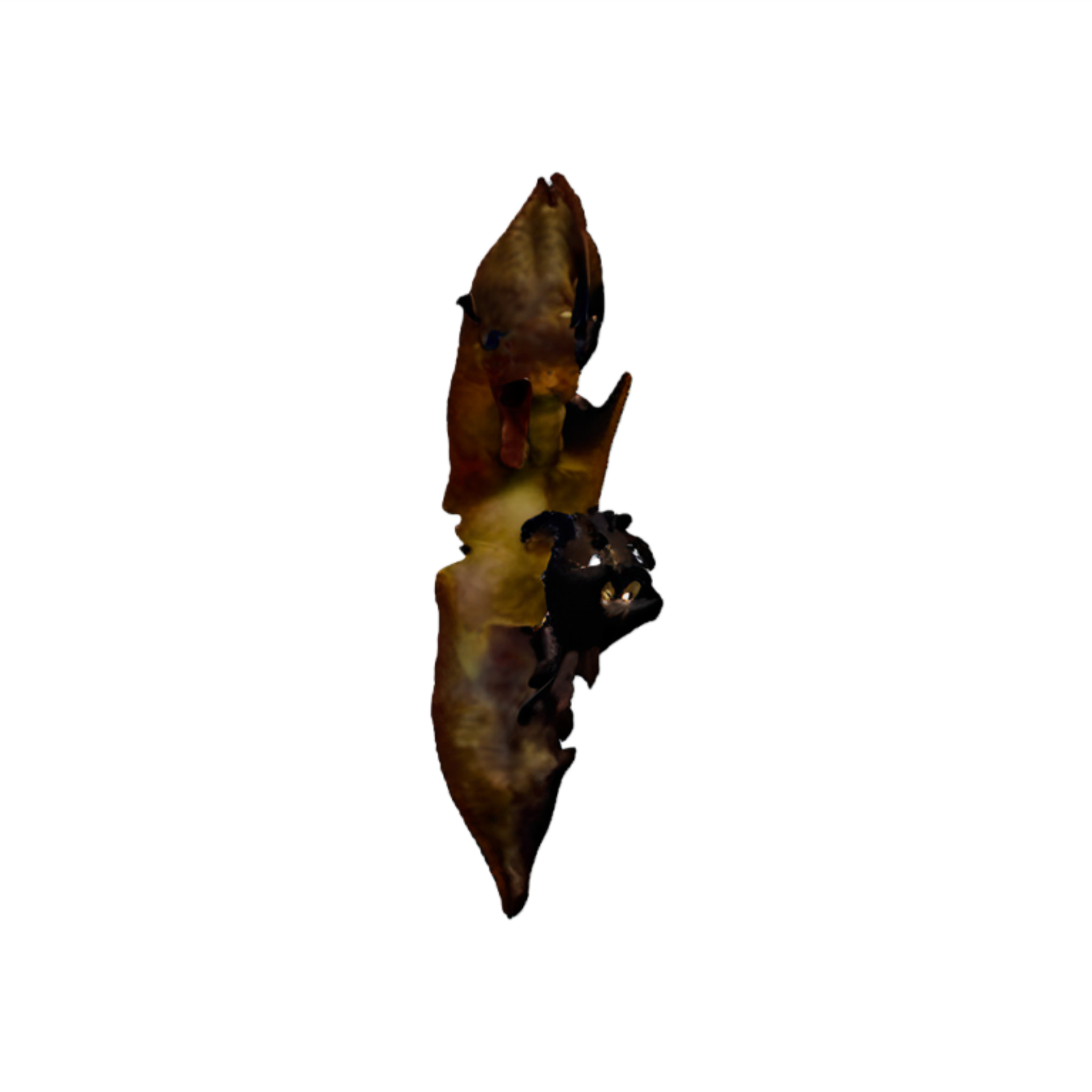}
			\end{subfigure}
		\end{tabular}
		\caption{Visual results of different novel view synthesis models on static orbits.}
		\label{qualitative_results_under_static_orbit}
	\end{figure*}
